\documentclass[10pt]{article} %
\usepackage[preprint]{tmlr}

\usepackage[utf8]{inputenc}  %
\usepackage[T1]{fontenc}  %

\usepackage{xspace}
\makeatletter
\DeclareRobustCommand\onedot{\futurelet\@let@token\@onedot}
\def\@onedot{\ifx\@let@token.\else.\null\fi\xspace}
\def\eg{\emph{e.g}\onedot}
\def\ie{\emph{i.e}\onedot}
\def\cf{\emph{c.f}\onedot}
\makeatother

\usepackage{amsmath}
\usepackage{amssymb}
\usepackage{amsfonts}
\newcommand*{\JnF}{\mathcal{J}\&\mathcal{F}}

\usepackage{pifont}

\usepackage{placeins}  %

\usepackage{verbatim} %
\usepackage[table,dvipsnames]{xcolor} %
\usepackage{soul} %

\usepackage{hyperref}
\hypersetup{
    colorlinks,
    linkcolor={MidnightBlue},
    citecolor={MidnightBlue},
    urlcolor={MidnightBlue},
    pdfauthor={The DINO Team},
    pdftitle={DINOv3},
}

\usepackage[capitalize,nameinlink]{cleveref}
\creflabelformat{equation}{#2\textup{#1}#3}
\crefname{figure}{Fig.}{Figs.}
\Crefname{figure}{Figure}{Figures}
\crefname{section}{Sec.}{Secs.}
\Crefname{section}{Section}{Sections}
\crefname{equation}{Eq.}{Eqs.}
\Crefname{equation}{Equation}{Equations}
\crefname{table}{Tab.}{Tabs.}
\Crefname{table}{Table}{Tables}
\crefname{appendix}{App.}{Apps.}
\Crefname{appendix}{Appendix}{Appendices}

\usepackage{graphicx}
\usepackage{epstopdf}
\usepackage{caption} %
\usepackage{subcaption} %
\usepackage{wrapfig}

\usepackage{booktabs}
\usepackage{multirow}
\usepackage{array}
\usepackage[export]{adjustbox} %
\usepackage[ruled,linesnumbered]{algorithm2e} %

\usepackage{tikz}
\usetikzlibrary{positioning}
\usepackage{pgfplots}
\pgfplotsset{compat=1.18}
\usepgfplotslibrary{statistics}
\usetikzlibrary{patterns}
\usetikzlibrary{external}

\usepackage{pifont}

\usepackage[weather,misc]{ifsym}
\newcommand{\orangefire}{{\color{orange}\Fire}}
\newcommand{\bluesnow}{{\color{cyan}\Snow}}

\definecolor{mint}{RGB}{173,235,179}
\definecolor{dinocolor}{RGB}{70,130,180}
\definecolor{frozencolor}{RGB}{239,100,97}
\definecolor{othercolor}{RGB}{211,211,211}

\newcommand{\omitme}[1]{}

\newcommand{\Gramname}{Gram Anchoring\xspace}
\newcommand{\gramname}{Gram anchoring\xspace}

\newcommand{\LVDdataset}{LVD-1689M\xspace}
\newcommand{\SATdataset}{SAT-493M\xspace}

\newcommand{\resultsTableHeaderSuper}{\raisebox{0.5pt}{\footnotesize\hspace{-1pt}\textit{Supervised backbones}}}
\newcommand{\resultsTableHeaderWeakly}{\raisebox{0.5pt}{\footnotesize\hspace{-1pt}\textit{Weakly-supervised backbones}}}
\newcommand{\resultsTableHeaderAgg}{\raisebox{0.5pt}{\footnotesize\hspace{-1pt}\textit{Agglomerative backbones}}}
\newcommand{\resultsTableHeaderSelf}{\raisebox{0.5pt}{\footnotesize\hspace{-1pt}\textit{Self-supervised backbones}}}

\tikzset{external/only named=true}  %

\title{\centering DINOv3}
\author{
    \centering
    Oriane Siméoni$^*$\hfill
    Huy\,V.\,Vo$^*$\hfill
    Maximilian Seitzer$^*$\hfill
    Federico Baldassarre$^*$\hfill
    Maxime Oquab$^*$\\[0.25em]
    Cijo Jose\hspace{1em}
    Vasil Khalidov\hspace{1em}
    Marc Szafraniec\hspace{1em}
    Seungeun Yi\hspace{1em}
    Michaël Ramamonjisoa\\[0.25em]
    Francisco Massa\hspace{2.25em}
    Daniel Haziza\hspace{2.25em}
    Luca Wehrstedt\hspace{2.25em}
    Jianyuan Wang \\[0.25em]
    Timothée Darcet\hspace{1.1em}
    Théo Moutakanni\hspace{1.1em}
    Leonel Sentana\hspace{1.1em}
    Claire Roberts\\[0.25em]
    Andrea Vedaldi\hspace{1.25em}
    Jamie Tolan\hspace{1.25em}
    John Brandt$^1$\hspace{1.25em}
    Camille Couprie\\[0.25em]
    Julien Mairal$^2$\hspace{0.7em}
    Hervé Jégou\hspace{1.1em}
    Patrick Labatut\hspace{0.9em}
    Piotr Bojanowski \\
    \normalfont
    \vspace{0.75em}
    \addr{Meta AI Research} $\;\;\;\;\;\;\;\;$ \addr{$^1$WRI} $\;\;\;\;\;\;\;\;$ \addr{$^2$Inria} \\
    \vspace{0.75em}
    \small{$^*$corresponding authors: \texttt{\{osimeoni,huyvvo,seitzer,baldassarre,qas\}@meta.com}}
}
\date{}

\begin{document}

\maketitle

\begin{abstract}

Self-supervised learning holds the promise of eliminating the need for manual data annotation, enabling models to scale effortlessly to massive datasets and larger architectures. 
By not being tailored to specific tasks or domains, this training paradigm has the potential to learn visual representations from diverse sources, ranging from natural to aerial images---using a single algorithm. 
This technical report introduces DINOv3, a major milestone toward realizing this vision by leveraging simple yet effective strategies. 
First, we leverage the benefit of scaling both dataset and model size by careful data preparation, design, and optimization. 
Second, we introduce a new method called \gramname, which effectively addresses the known yet unsolved issue of dense feature maps degrading during long training schedules. 
Finally, we apply post-hoc strategies that further enhance our models' flexibility with respect to resolution, model size, and alignment with text.
As a result, we present a versatile vision foundation model that outperforms the specialized state of the art across a broad range of settings, without fine-tuning.
DINOv3 produces high-quality dense features that achieve outstanding performance on various vision tasks, significantly surpassing previous self- and weakly-supervised foundation models.
We also share the DINOv3 suite of vision models, designed to advance the state of the art on a wide spectrum of tasks and data by providing scalable solutions for diverse resource constraints and deployment scenarios. 

\end{abstract}

\section{Introduction}
Foundation models have become a central building block in modern computer vision, enabling broad generalization across tasks and domains through a single, reusable model.
Self-supervised learning (SSL) is a powerful approach for training such models, by learning directly from raw pixel data and leveraging the natural co-occurrences of patterns in images. 
Unlike weakly and fully supervised pretraining methods~\citep{radford2021learning,dehghani2023scaling,bolya2025perception} which require images paired with high-quality metadata, SSL unlocks training on massive, raw image collections. 
This is particularly effective for training large-scale visual encoders thanks to the availability of virtually unlimited training data.
DINOv2~\citep{oquab2024dinov2} exemplifies these strengths, achieving impressive results in image understanding tasks~\citep{wang2025vggt} and enabling pre-training for complex domains such as histopathology~\citep{chen2024uni}.
Models trained with SSL exhibit additional desirable properties: they are robust to input distribution shifts, provide strong global and local features, and generate rich embeddings that facilitate physical scene understanding. 
Since SSL models are not trained for any specific downstream task, they produce versatile and robust generalist features. 
For instance, DINOv2 models deliver strong performance across diverse tasks and domains without requiring task-specific finetuning, allowing a single frozen backbone to serve multiple purposes.
Importantly, self-supervised learning is especially suitable to train on the vast amount of available observational data in domains like histopathology~\citep{vorontsov2024foundation}, biology~\citep{kim2025self}, medical imaging~\citep{perez2025exploring}, remote sensing~\citep{cong2022satmae,tolan2024very}, astronomy~\citep{parker2024astroclip}, or high-energy particle physics~\citep{dillon2022symmetries}. 
These domain often lack metadata and have already been shown to benefit from foundation models like DINOv2.  
Finally, SSL, requiring no human intervention, is well-suited for lifelong learning amid the growing volume of web data.

\begin{figure}[t]
    \centering
    \begin{minipage}[b]{0.35\textwidth}
        \input{figures/introduction/evolution_in_time}        

        \vspace{1.1em}
        \input{figures/introduction/comparison_us_bestintown}
        \vspace{-0.2em}
    \end{minipage}%
    \hfill
    \begin{minipage}[b]{0.6\textwidth}
        \hfill
        \begin{tikzpicture}[every node/.style={inner sep=0,outer sep=0}, node distance=0.02\textwidth and 0.02\textwidth]
          \node (img1) {\includegraphics[width=0.49\textwidth]{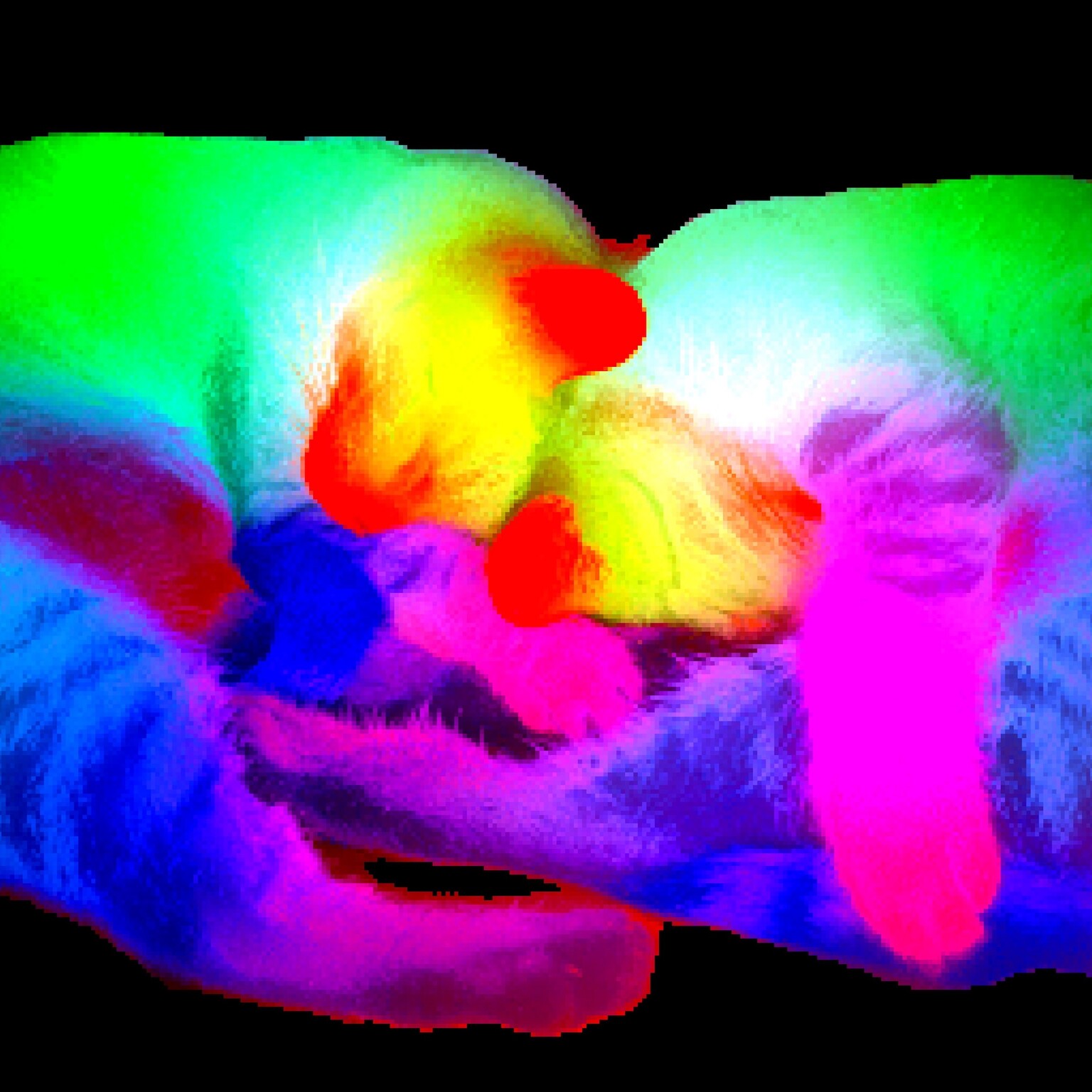}};
          \node[right=of img1] (img2) {\includegraphics[width=0.49\textwidth]{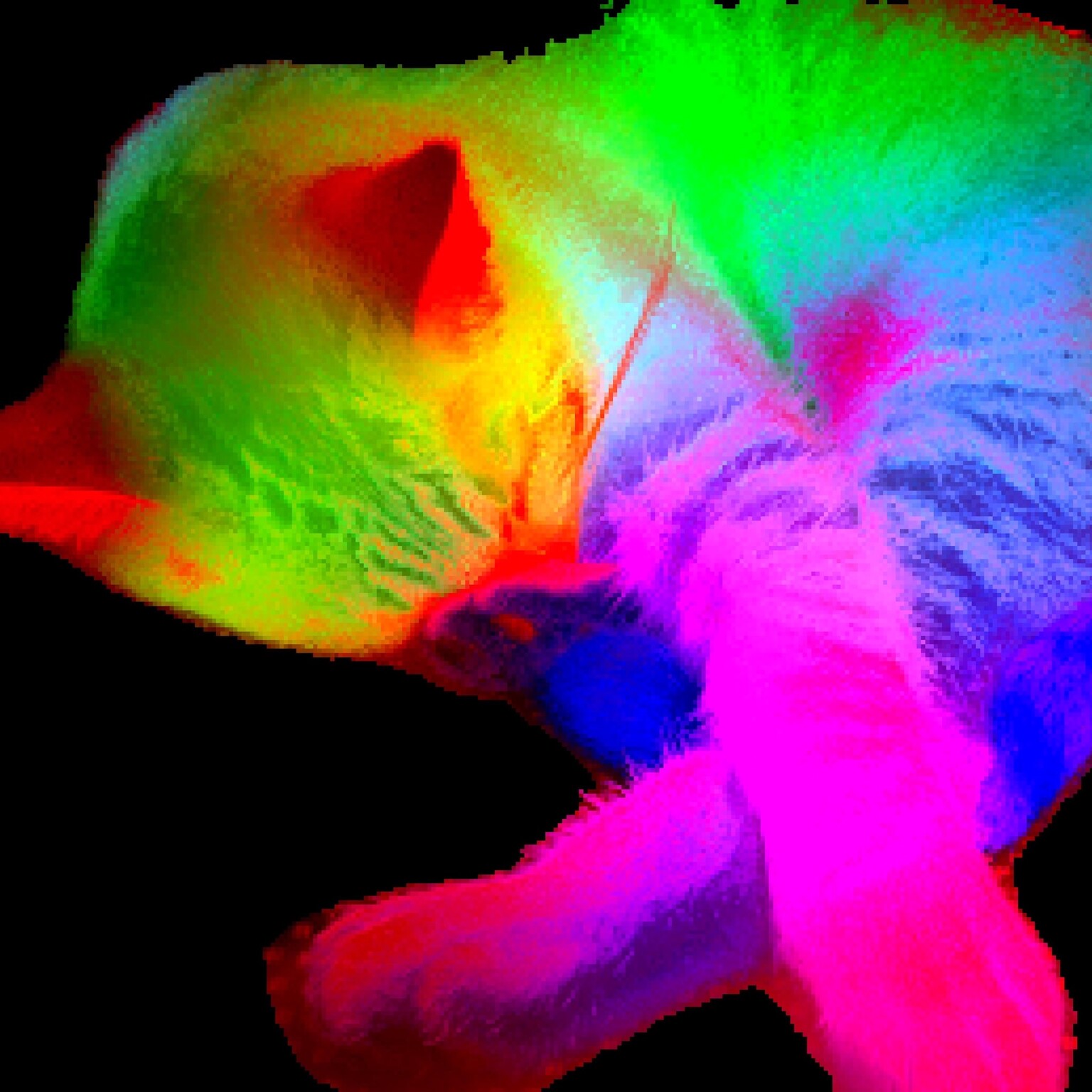}};
          \node[below=of img1] (img3) {\includegraphics[width=0.49\textwidth]{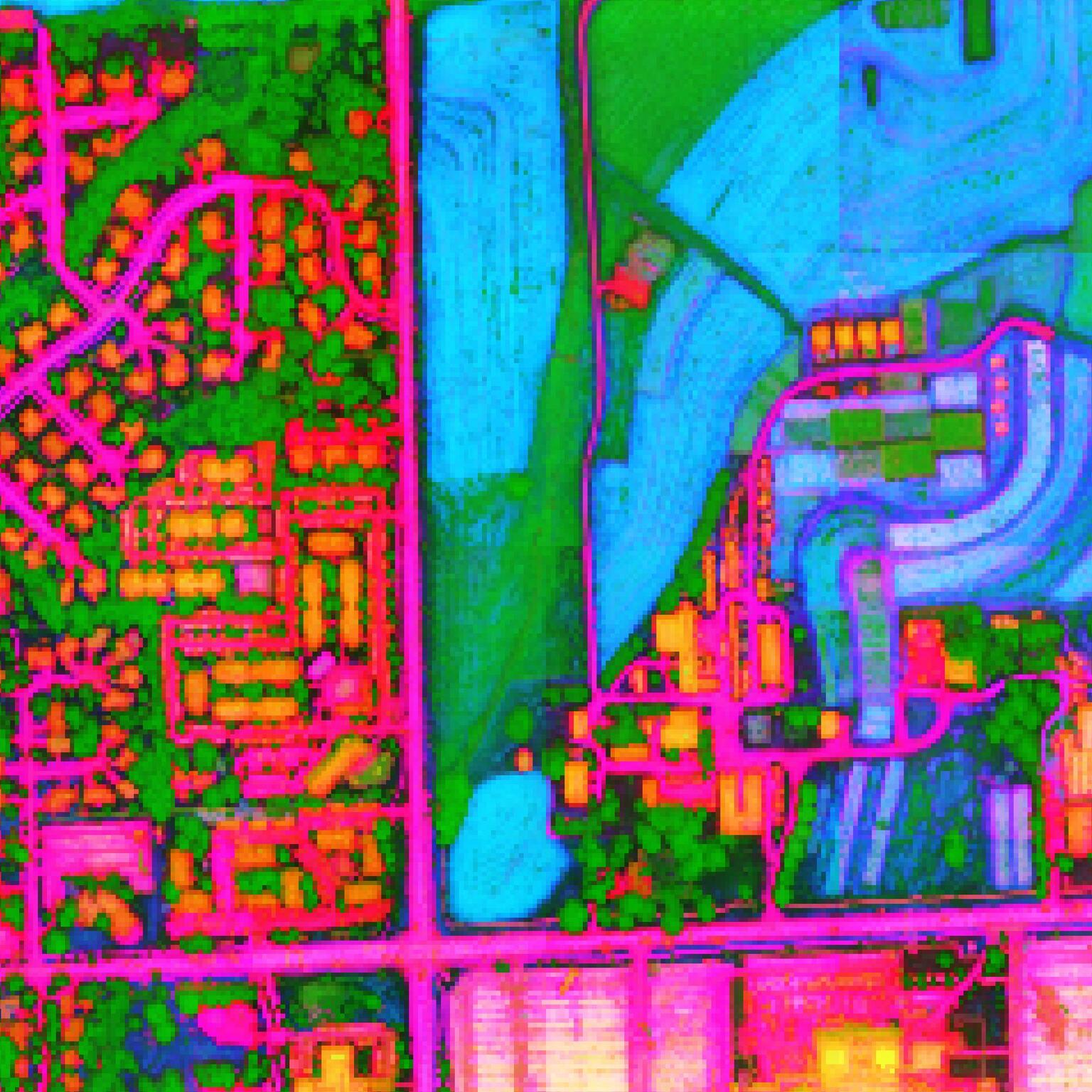}};
          \node[right=of img3] (img4) {\includegraphics[width=0.49\textwidth]{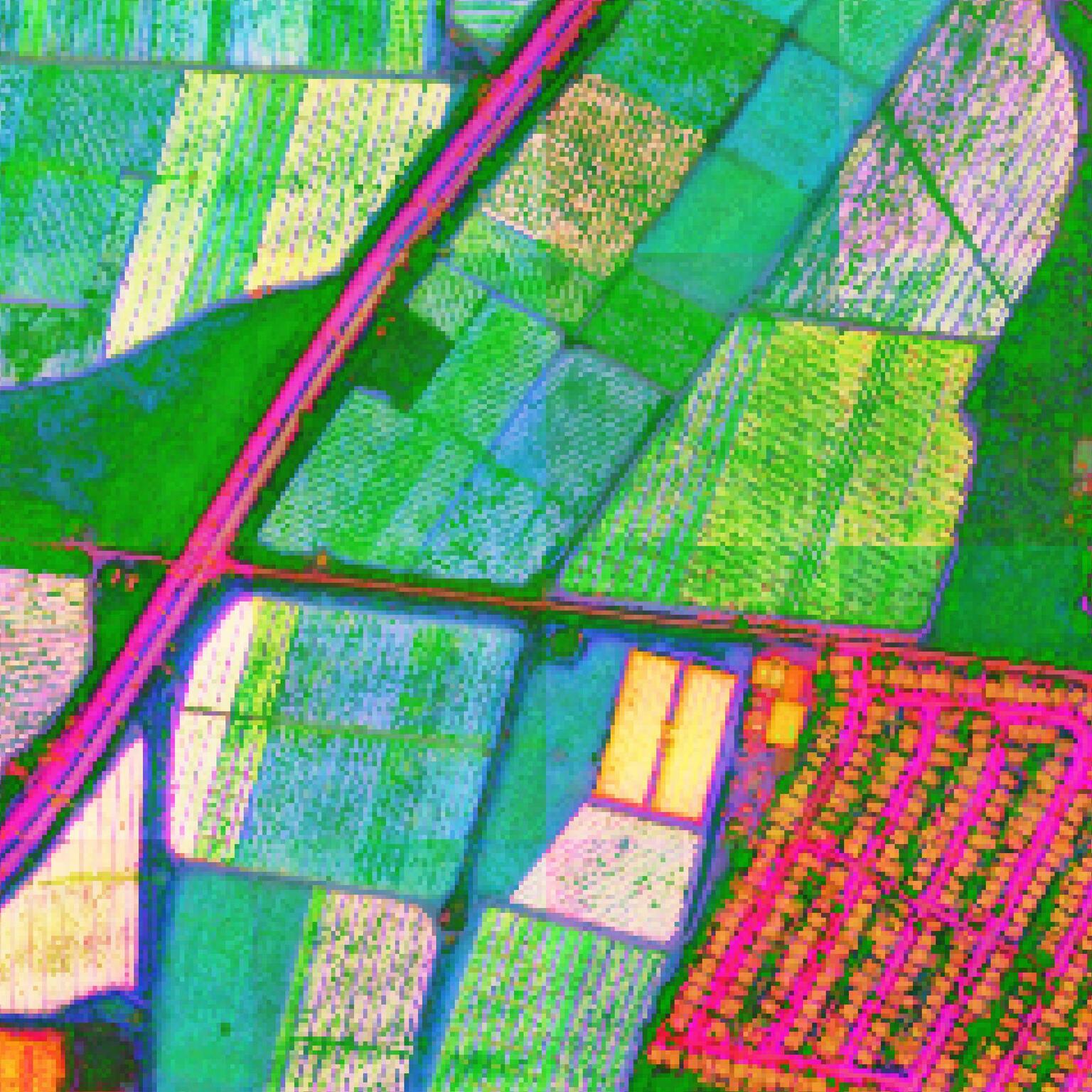}};
          \node[font=\bfseries,overlay] at ([xshift=-0.636\textwidth,yshift=-0.75\baselineskip]img1.north west) {(a)};
          \node[font=\bfseries,overlay,text=white] at ([xshift=0.35cm,yshift=-0.75\baselineskip]img1.north west) {(c)};
          \node[font=\bfseries,overlay] at ([xshift=-0.636\textwidth,yshift=-0.75\baselineskip]img3.north west) {(b)};
          \node[font=\bfseries,overlay,text=white] at ([xshift=0.35cm,yshift=-0.75\baselineskip]img3.north west) {(d)};
        \end{tikzpicture}
        
    \end{minipage}
    \caption{(a) Evolution of linear probing results on ImageNet1k (IN1k) over the years, comparing fully- (SL), weakly- (WSL) and self-supervised learning (SSL) methods. 
    Despite coming into the picture later, SSL has quickly progressed and now reached the Imagenet accuracy plateau of recent years. 
    On the other hand, we demonstrate that SSL offers the unique promise of high-quality dense features.
    With DINOv3, we markedly improve over weakly-supervised models on dense tasks, as shown by the relative performance of the best-in-class WSL models to DINOv3 (b).
    We also produce PCA maps of features obtained from high resolution images with DINOv3 trained on natural (c) and aerial images (d).}
    \label{fig:pcas}
\end{figure}

In practice, the promise of SSL, namely producing arbitrarily large and powerful models by leveraging large amounts of unconstrained data, remains challenging at scale. 
While model instabilities and collapse are mitigated by the heuristics proposed by \citet{oquab2024dinov2}, more problems emerge from scaling further. 
First, it is unclear how to collect useful data from unlabeled collections. 
Second, in usual training practice, employing cosine schedules implies knowing the optimization horizon a priori, which is difficult when training on large image corpora. 
Third, the performance of the features gradually decreases after early training, confirmed by visual inspection of the patch similarity maps. 
This phenomenon appears in longer training runs with models above ViT-Large size (300M parameters), reducing the usefulness of scaling DINOv2.

Addressing the problems above leads to this work, \emph{DINOv3}, which advances SSL training at scale.  
We demonstrate that a \emph{single frozen SSL backbone} can serve as a universal visual encoder that achieves state-of-the-art performance on challenging downstream tasks, outperforming supervised and metadata-reliant pre-training strategies. 
Our research is guided by the following objectives: 
(1) training a foundational model versatile across tasks and domains, 
(2) improving the shortcomings of existing SSL models on dense features, 
(3) disseminating a family of models that can be used off-the-shelf.  
We discuss the three aims in the following.

\begin{figure}[t]
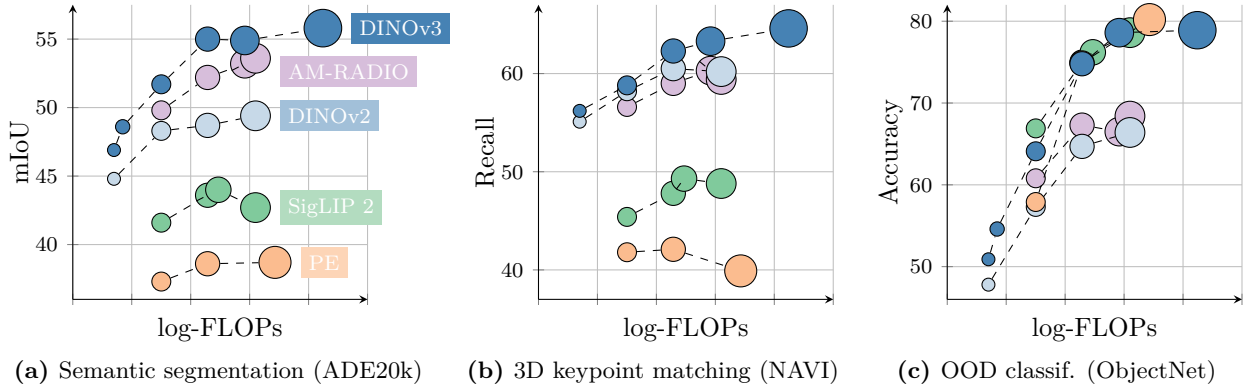

    \begin{subfigure}[b]{0.35\textwidth}
        \centering
        \input{figures/introduction/family_models/family_models_ade20k}
        \vspace{-15pt}
        \caption{Semantic segmentation (ADE20k)}
   \end{subfigure}%
   \hfill%
    \begin{subfigure}[b]{0.30\textwidth}
        \centering
        \input{figures/introduction/family_models/family_models_probe3d}
        \vspace{-15pt}
        \caption{3D keypoint matching (NAVI)}
   \end{subfigure}%
   \hfill%
    \begin{subfigure}[b]{0.30\textwidth}
        \centering
        \input{figures/introduction/family_models/family_models_objectnet}
        \vspace{-15pt}
        \caption{OOD classif. (ObjectNet)}
   \end{subfigure}
   \caption{Performance of the DINOv3 family of models, compared to other families of self- or weakly-supervised models, on different benchmarks. DINOv3 significantly surpasses others on dense benchmarks, including models that leverage mask annotation priors such as AM-RADIO~\citep{heinrich2025radiov25}.}
   \label{fig:intro:family}
\end{figure}

\paragraph{Strong \& Versatile Foundational Models}  
DINOv3 aims to offer a high level of versatility along two axes, which is enabled by the scaling of the model size and training data. 
First, a key desirable property for SSL models is to achieve excellent performance while being kept frozen, ideally reaching similar state-of-the-art results as specialized models. 
In that case, a single forward pass can deliver cutting-edge results across multiple tasks, leading to substantial computational savings---an essential advantage for practical applications, particularly on edge devices. 
We show the wide breadth of tasks that DINOv3 can successfully be applied to in \cref{sec:results}.
Second, a scalable SSL training pipeline that does not depend on metadata unlocks numerous scientific applications. 
By pre-training on a diverse set of images, whether web images or observational data, SSL models generalize across a large set of domains and tasks. 
As illustrated in \cref{fig:pcas}(d), the PCA of DINOv3 features extracted from a high-resolution aerial image clearly allows to separates roads, houses, and greenery, highlighting the model’s feature quality. 

\begin{figure}[t]
\begin{minipage}{\linewidth}
    \includegraphics[height=0.248\linewidth, width=0.248\linewidth]{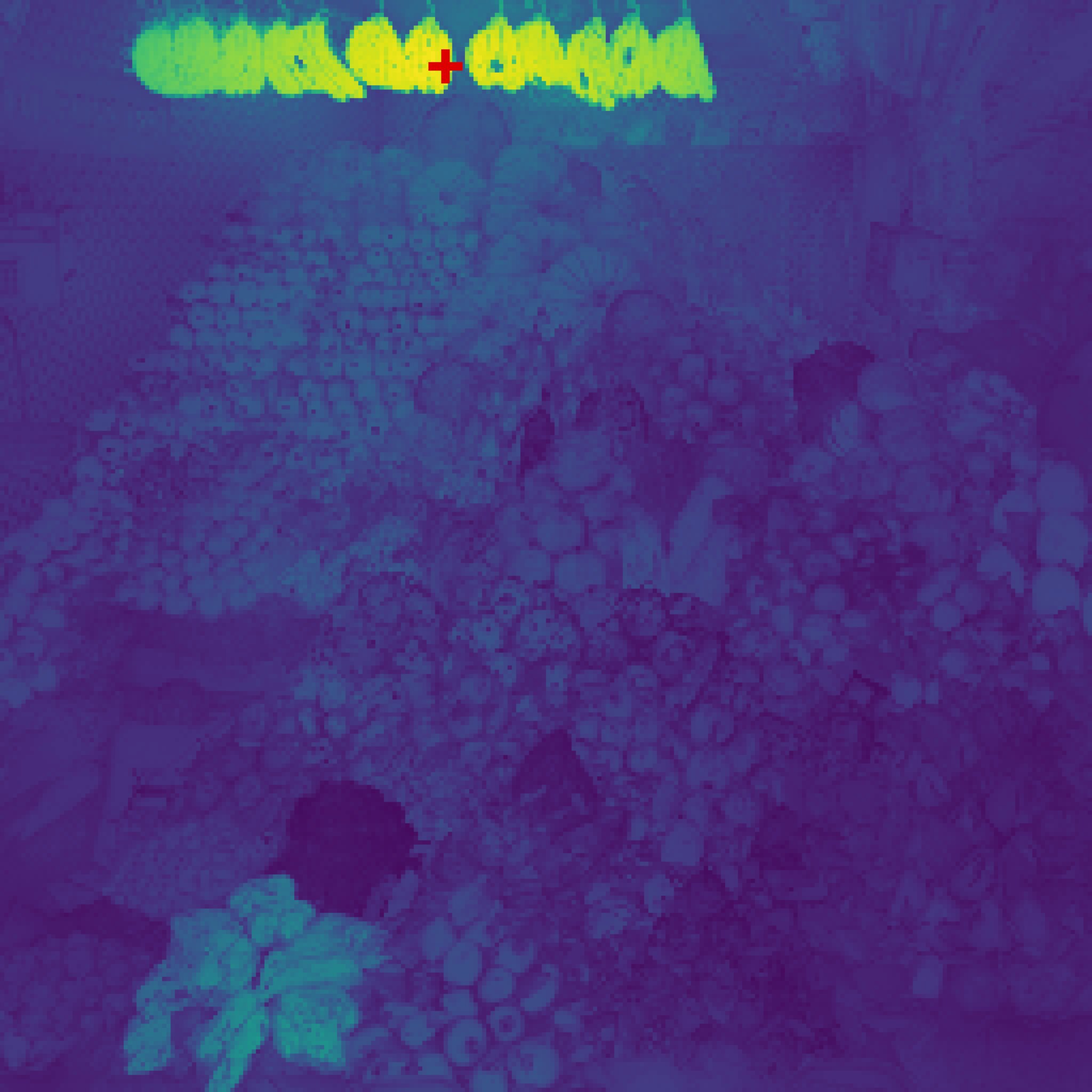}\hfill%
    \includegraphics[height=0.248\linewidth, width=0.248\linewidth]{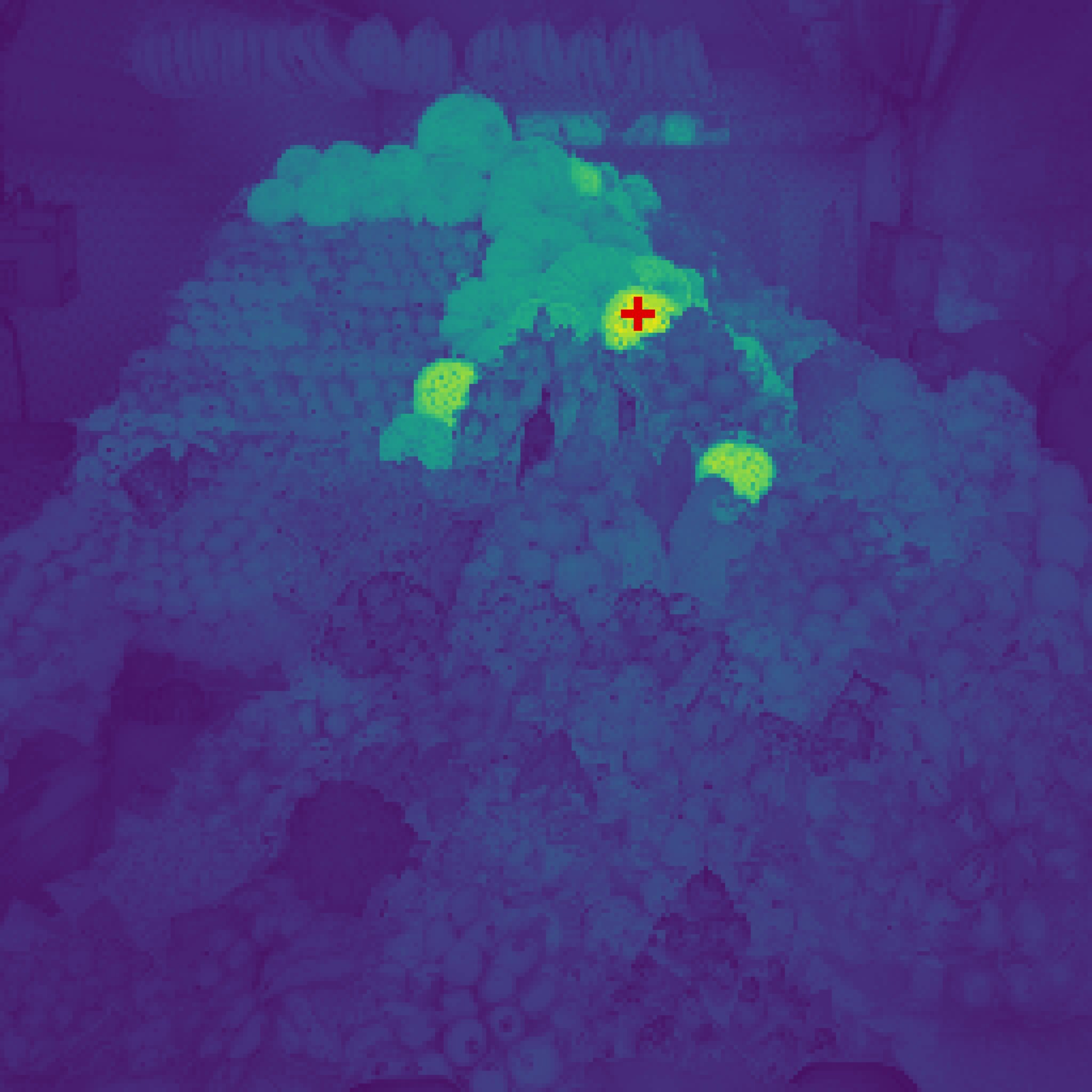}\hfill%
    \includegraphics[height=0.248\linewidth, width=0.248\linewidth]{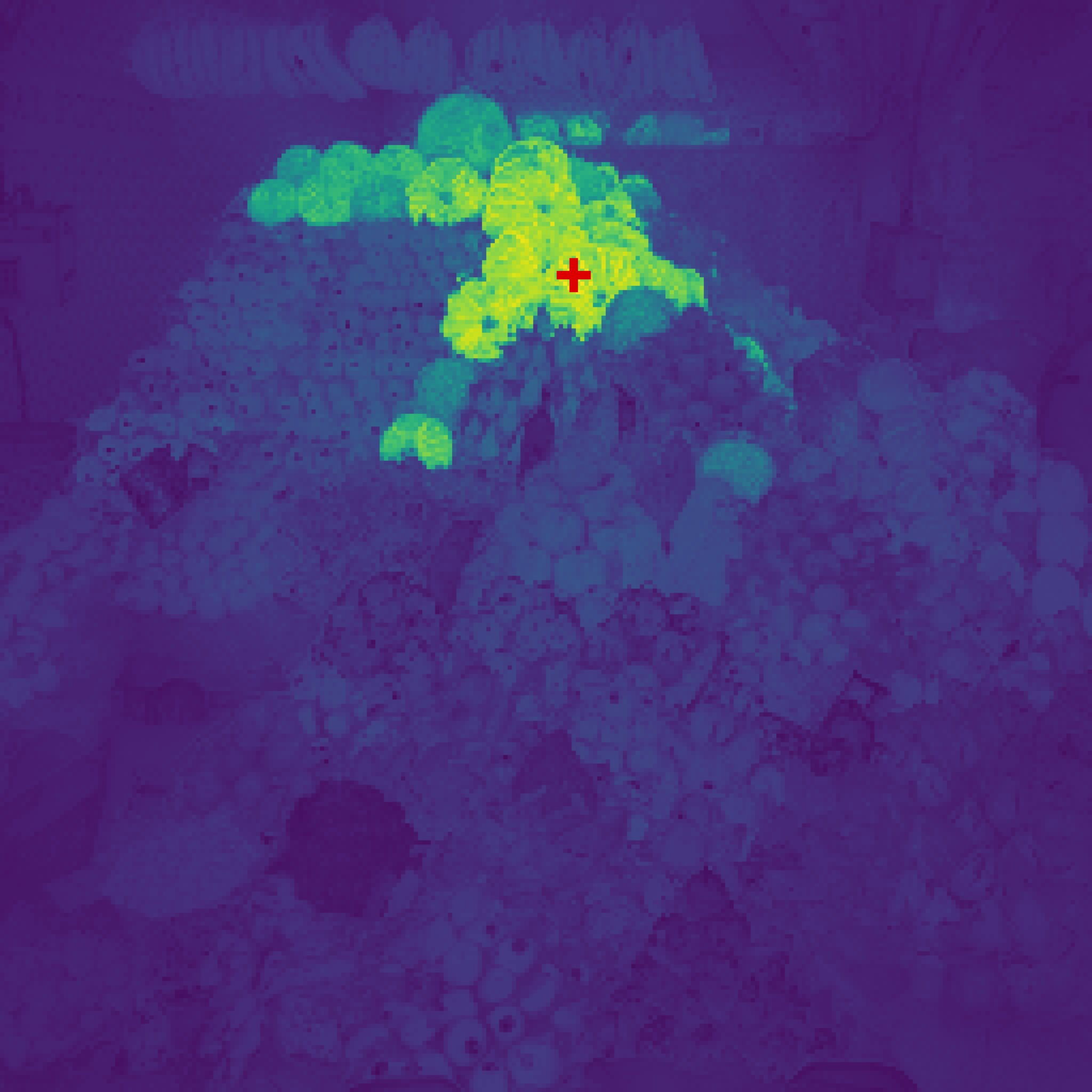}\hfill%
    \includegraphics[height=0.248\linewidth, width=0.248\linewidth] {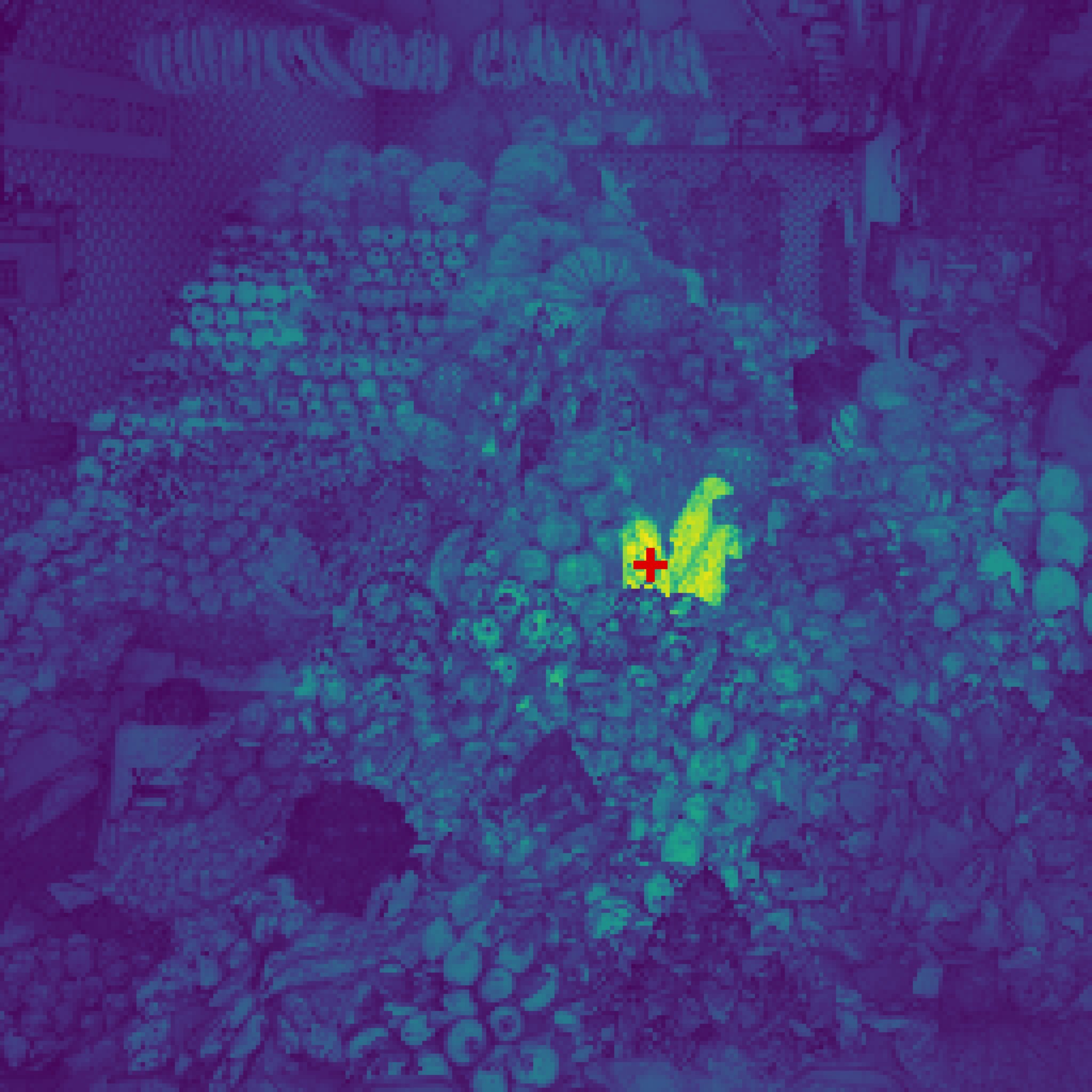}
\end{minipage}\\
\begin{minipage}{0.248\linewidth}
    \includegraphics[height=\linewidth, width=\linewidth]{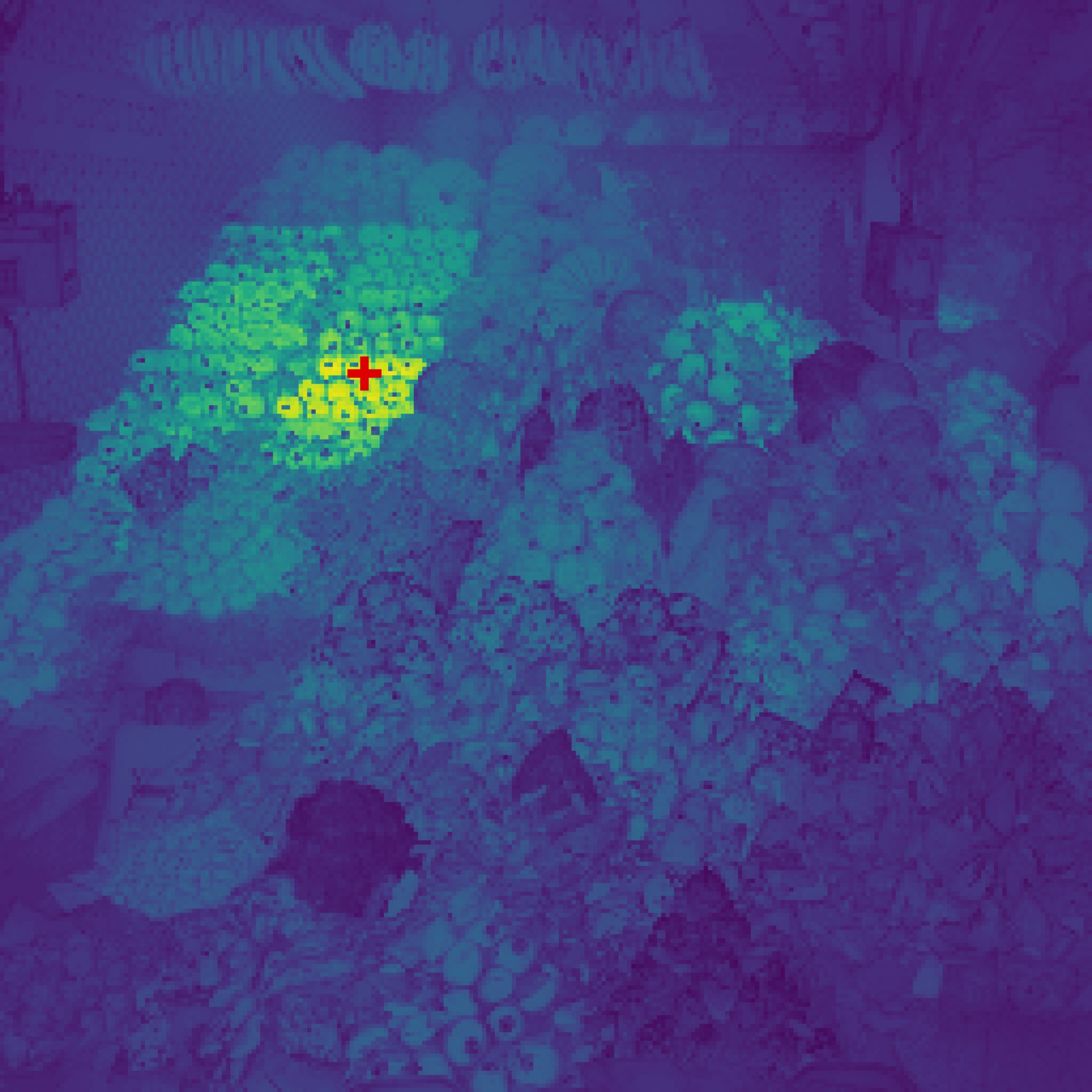}\\
    \includegraphics[height=\linewidth, width=\linewidth]{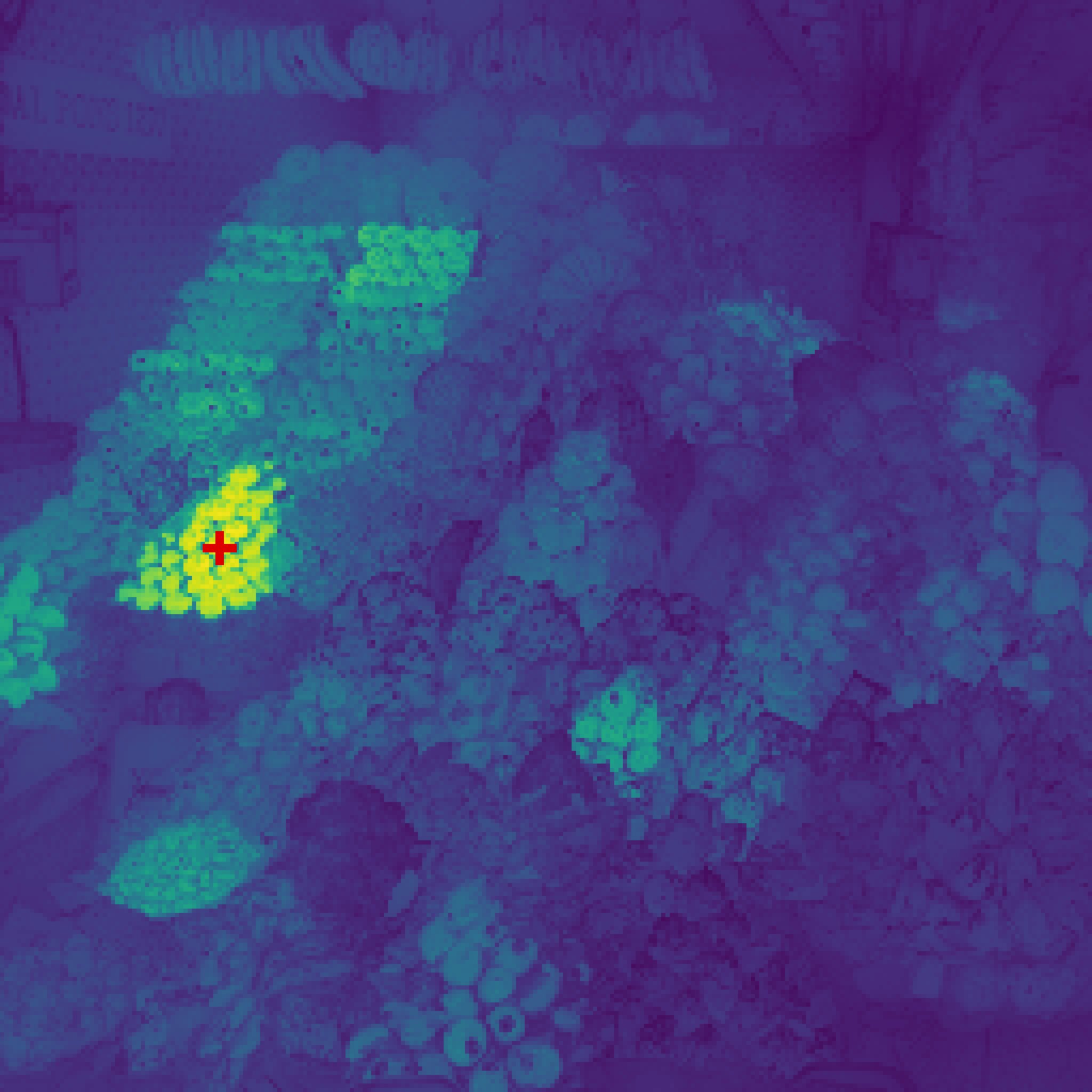}
\end{minipage}\hfill%
\begin{minipage}{0.499\linewidth}%
    \includegraphics[height=0.998\linewidth, width=1\linewidth]{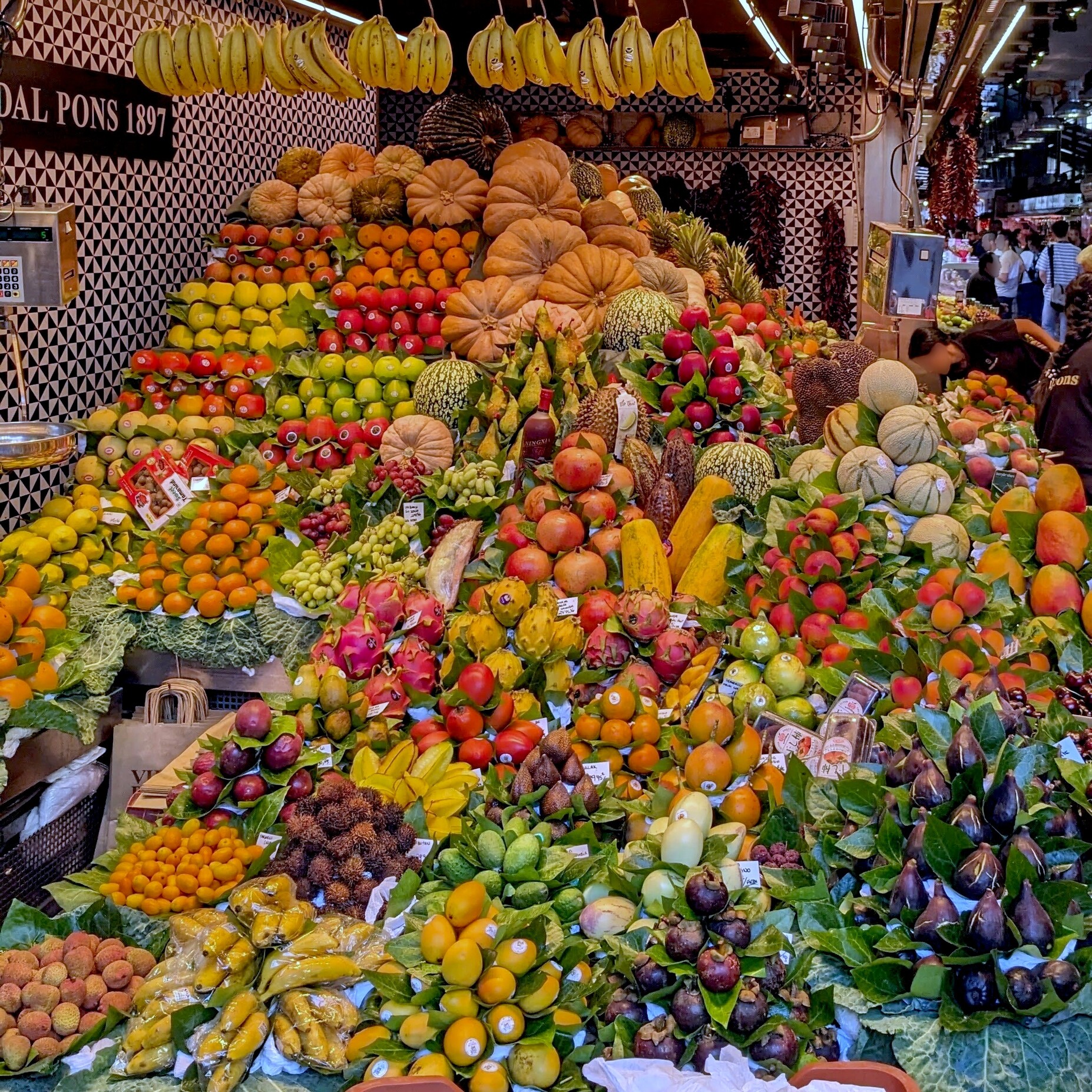}%
\end{minipage}\hfill%
\begin{minipage}{0.248\linewidth}
    \includegraphics[height=\linewidth, width=\linewidth]{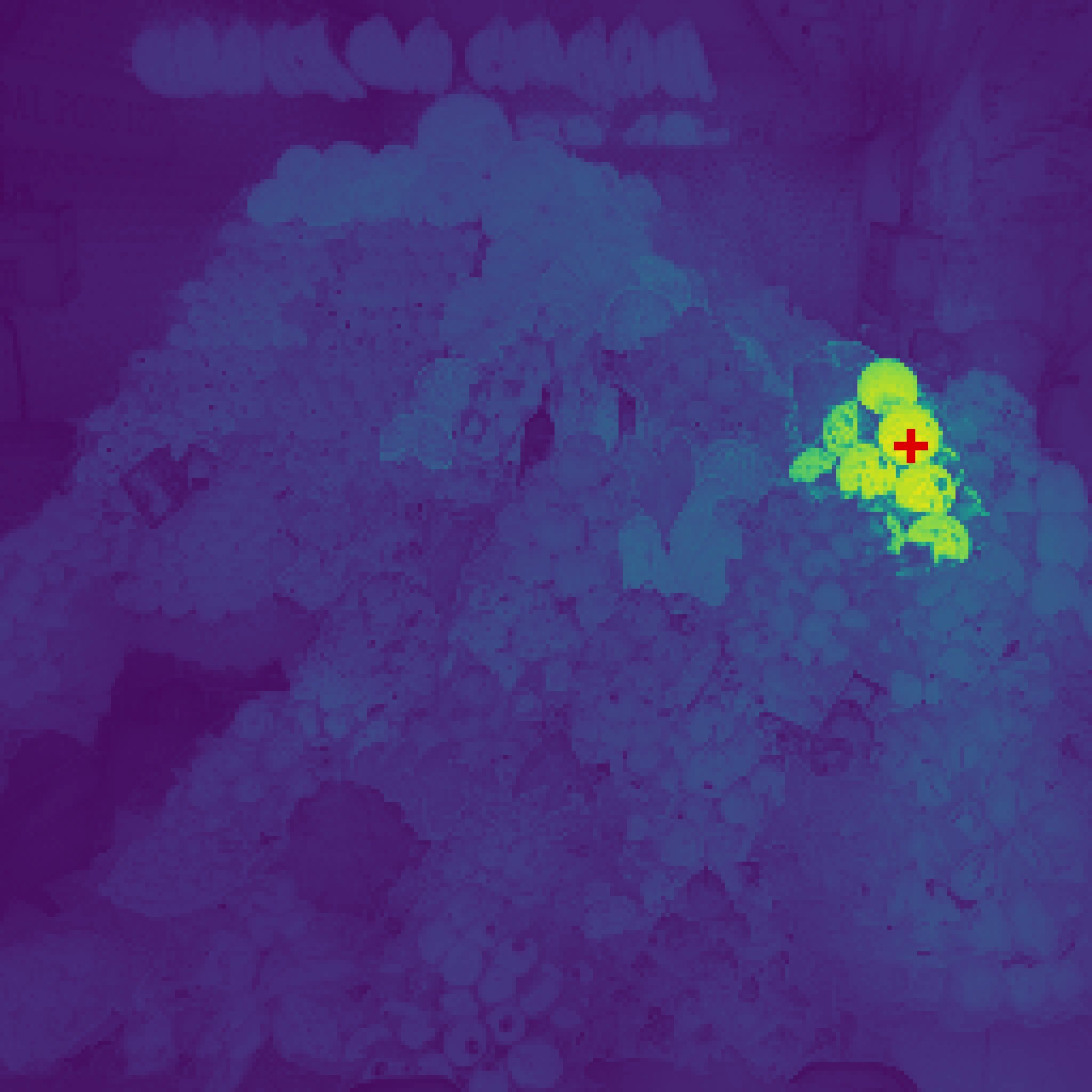}\\
    \includegraphics[height=\linewidth, width=\linewidth]{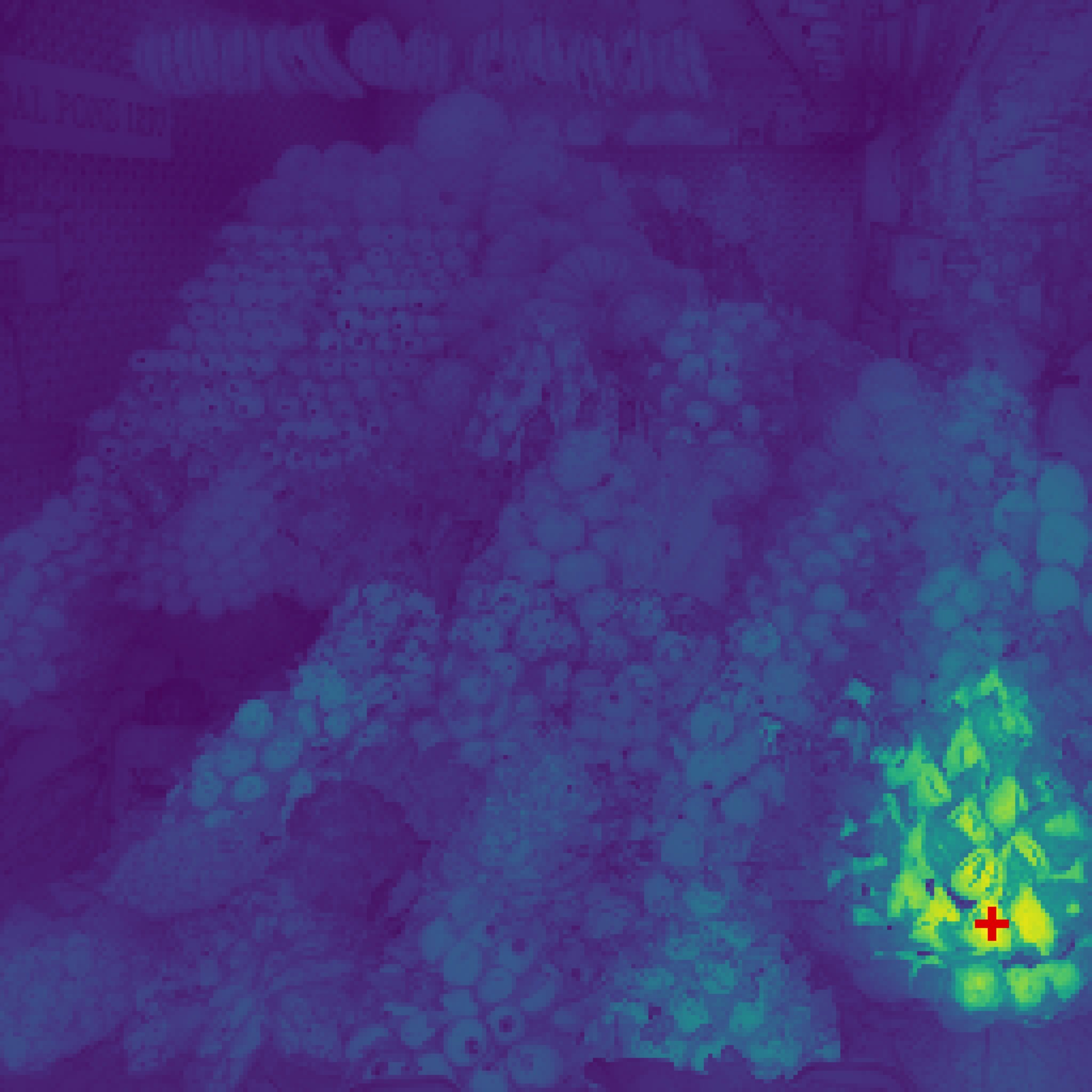}
\end{minipage}     
\caption{High-resolution dense features. We visualize the cosine similarity maps obtained with DINOv3 output features between the patches marked with a red cross and all other patches. Input image at 4096×4096. \emph{Please zoom in}, do you agree with DINOv3?}
    \label{fig:intro:dense-quality}
\end{figure}

\paragraph[Superior Feature Maps Through Gram]{Superior Feature Maps Through \Gramname}
Another key feature of DINOv3 is a significant improvement of its dense feature maps. 
The DINOv3 SSL training strategy aims at producing models excelling at high-level semantic tasks while producing excellent feature maps amenable to solving geometric tasks such as depth estimation, or 3D matching.
In particular, the models should produce dense features that can be used off-the-shelf or with little post-processing. 
The compromise between dense and global representation is especially difficult to optimize when training with vast amounts of images, since the objective of high-level understanding can conflict with the quality of the dense feature maps. 
These contradictory objectives lead to a collapse of dense features with large models and long training schedules.
Our new \gramname strategy effectively mitigates this collapse (see \cref{sec:gram}).
As a result, DINOv3 obtains significantly better dense feature maps than DINOv2, staying clean even at high resolutions (see \cref{fig:intro:dense-quality}). 

\paragraph{The DINOv3 Family of Models} 
Solving the degradation of dense feature map with \gramname unlocks the power of scaling.
As a consequence, training a much larger model with SSL leads to significant performance improvements. 
In this work, we successfully train a DINO model with 7B parameters. 
Since such a large model requires significant resources to run, we apply distillation to compress its knowledge into smaller variants. 
As a result, we present the \emph{DINOv3 family of vision models}, a comprehensive suite designed to address a wide spectrum of computer vision challenges. 
This model family aims to advance the state of the art by offering scalable solutions adaptable to diverse resource constraints and deployment scenarios.
The distillation process produces model variants at multiple scales, including Vision Transformer (ViT) Small, Base, and Large, as well as ConvNeXt-based architectures.
Notably, the efficient and widely adopted ViT-L model achieves performance close to that of the original 7B teacher across a variety of tasks. 
Overall, the DINOv3 family demonstrates strong performance on a broad range of benchmarks, matching or exceeding the accuracy of competing models on global tasks, while significantly outperforming them on dense prediction tasks, as visible in \cref{fig:intro:family}.

\paragraph{Overview of Contributions}
In this work, we introduce multiple contributions to address the challenge of scaling SSL towards a large frontier model.
We build upon recent advances in automatic data curation~\citep{vo2024automatic} to obtain a large ``background'' training dataset that we carefully mix with a bit of specialized data (ImageNet-1k). 
This allows leveraging large amounts of unconstrained data to improve the model performance. 
This contribution \textbf{(i)} around data scaling will be described in \cref{sec:data}.

We increase our main model size to 7B parameters by defining a custom variant of the ViT architecture.
We include modern position embeddings (axial RoPE) and develop a regularization technique to avoid positional artifacts. 
Departing from the multiple cosine schedules in DINOv2, we train with constant hyperparameter schedules for 1M iterations. 
This allows producing models with stronger performance. 
This contribution \textbf{(ii)} on model architecture and training will be described in \cref{sec:algo}.

With the above techniques, we are able to train a model following the DINOv2 algorithm at scale. 
However, as mentioned previously, scale leads to a degradation of dense features. 
To address this, we propose a core improvement of the pipeline with a \gramname training phase. 
This cleans the noise in the feature maps, leading to impressive similarity maps, and drastically improving the performance on both parametric and non-parametric dense tasks. 
This contribution \textbf{(iii)} on Gram training will be described in \cref{sec:gram}.

Following previous practice, the last steps of our pipeline consist of a high-resolution post-training phase and distillation into a series of high-performance models of various sizes.
For the latter, we develop a novel and efficient single-teacher multiple-students distillation procedure. 
This contribution \textbf{(iv)} transfers the power of our 7B frontier model to a family of smaller practical models for common usage, that we describe in \cref{sec:distillation}.

As measured in our thorough benchmarking, results in \cref{sec:results} show that our approach defines a new standard in dense tasks and performs comparably to CLIP derivatives on global tasks. 
In particular, \textit{with a frozen vision backbone}, we achieve state-of-the-art performance on longstanding computer vision problems such as object detection (COCO detection, mAP 66.1) and image segmentation (ADE20k, mIoU 63.0), outperforming specialized fine-tuned pipelines. 
Moreover, we provide evidence of the generality of our approach across domains by applying the DINOv3 algorithm to satellite imagery, in \cref{sec:satellite}, surpassing all prior approaches. 

\section{Related Work}

\paragraph{Self-Supervised Learning}
Learning without annotations requires an artificial learning task that provides supervision in lieu for training.
The art and challenge of SSL lies in carefully designing these so-called pre-text tasks in order to learn powerful representations for downstream tasks.
The language domain, by its discrete nature, offers straightforward ways to set up such tasks, which led to many successful unsupervised pre-training approaches for text data.
Examples include word embeddings~\citep{mikolov2013distributed,bojanowski2017enriching}, sentence representations~\citep{devlin2018bert,liu2019roberta}, and plain language models~\citep{mikolov2010recurrent,zaremba2014recurrent}.
In contrast, computer vision presents greater challenges due to the continuous nature of the signal. 
Early attempts mimicking language approaches extracted supervisory signals from parts of an image to predict other parts, \eg by predicting relative patch position~\citep{doersch2015unsupervised}, patch re-ordering~\citep{noroozi2016jigsaw, misra2020self}, or inpainting~\citep{pathakCVPR16context}.
Other tasks involve re-colorizing images~\citep{zhang2016colorful} or predicting image transformations~\citep{gidaris2018unsupervised}.

Among these tasks, \emph{inpainting-based} approaches have gathered significant interest thanks to the flexibility of the patch-based ViT architecture~\citep{he2021masked, bao2021beit, el2021large}.
The objective is to reconstruct corrupted regions of an image, which can be viewed as a form of denoising auto-encoding and is conceptually related to the masked token prediction task in BERT pretraining~\citep{devlin2018bert}.
Notably, \citet{he2021masked} demonstrated that pixel-based masked auto-encoders~(MAE) can be used as strong initializations for finetuning on downstream tasks.
In the following, \citet{baevski2022data2vec,baevski2023efficient,assran2023self} showed that predicting a \emph{learned latent space} instead of the pixel space leads to more powerful, higher-level features---a learning paradigm called JEPA: ``Joint-Embedding Predictive Architecture''~\citep{lecun2022path}.
Recently, JEPAs have also been extended to video training~\citep{bardes2024revisiting,assran2025vjepa2}.

A second line of work, closer to ours, leverages \emph{discriminative signals between images} to learn visual representations. 
This family of methods traces its origins to early deep learning research~\citep{hadsell2006dimensionality}, but gained popularity with the introduction of instance classification techniques~\citep{dosovitskiy2016discriminative, bojanowski2017unsupervised, wu2018unsupervised}.
Subsequent advancements introduced contrastive objectives and information-theoretic criteria~\citep{henaff2019data,he2020momentum, chen2020exploring, chen2020simple,grill2020bootstrap, bardes2021vicreg}, as well as self clustering-based strategies~\citep{caron2018deep,asano2019self,caron2020unsupervised,caron2021emerging}.
More recent approaches, such as iBOT~\citep{zhou2021ibot}, combine these discriminative losses with masked reconstruction objectives.
All of these methods show the ability to learn strong features and achieve high performance on standard benchmarks like ImageNet~\citep{russakovsky2015imagenet}.
However, most face challenges scaling to larger model sizes~\citep{chen2021empirical}.

\paragraph{Vision Foundation Models}
The deep learning revolution began with the AlexNet breakthrough~\citep{krizhevsky2012imagenet}, a deep convolutional neural network that outperformed all previous methods on the ImageNet challenge~\citep{deng2009imagenet,russakovsky2015imagenet}. 
Already early on, features learned end-to-end on the large manually-labeled ImageNet dataset were found to be highly effective for a wide range of transfer learning tasks~\citep{oquab2014learning}.
Early work on vision \emph{foundation models} then focused on architecture development, including VGG~\citep{simonyan2015very}, GoogleNet~\citep{szegedy2015going}, and ResNets~\citep{he2016deep}. %

Given the effectiveness of \emph{scaling}, subsequent works explored training larger models on big datasets.
\citet{sun2017revisiting} expanded supervised training data with the proprietary JFT dataset containing 300 million labeled images, showing impressive results.
JFT also enabled significant performance gains for \citet{kolesnikov2020big}.
In parallel, scaling was explored using a combination of supervised and unsupervised data. 
For instance, an ImageNet-supervised model can be used to produce pseudo-labels for unsupervised data, which then serve to train larger networks~\citep{yalniz2019billion}.
Subsequently, the availability of large supervised datasets such as JFT also facilitated the adaptation of the transformer architecture to computer vision~\citep{dosovitskiy2020image}.
In particular, achieving performance comparable to that of the original vision transformer (ViT) without access to JFT requires substantial effort~\citep{touvron2020training, touvron2022deit}. 
Due to the learning capacity of ViTs, scaling efforts were further extended by \citet{zhai2022scaling}, culminating in the very large ViT-22B encoder~\citep{dehghani2023scaling}.

Given the complexity of manually labeling large datasets, \emph{weakly-supervised training}---where annotations are derived from metadata associated with images---provides an effective alternative to supervised training.
Early on, \citet{joulin2016learning} demonstrated that a network can be pre-trained by simply predicting all words in the image caption as targets. 
This initial approach was further refined by leveraging sentence structures~\citep{li2017learning}, incorporating other types of metadata and involve curation~\citep{mahajan2018exploring}, and scaling~\citep{singh2022revisiting}.
However, weakly-supervised algorithms only reached their full potential with the introduction of contrastive losses and the joint-training of caption representations, as exemplified by Align~\citep{jia2021scaling} and CLIP~\citep{radford2021learning}.

This highly successful approach inspired numerous \emph{open-source reproductions and scaling efforts}. 
OpenCLIP~\citep{cherti2023reproducible} was the first open-source effort to replicate CLIP by training on the LAION dataset~\citep{schuhmann2021laion}; following works leverage pre-trained backbones by fine-tuning them in a CLIP-style manner~\citep{sun2023eva,sun2024eva}.
Recognizing that data collection is a critical factor in the success of CLIP training, MetaCLIP~\citep{xu2024demystifying} precisely follows the original CLIP procedure to reproduce its results, whereas \citet{fang2024data} use supervised datasets to curate pretraining data.
Other works focus on improving the training loss, \eg using a sigmoid loss in SigLIP~\citep{zhai2023sigmoid}, or leveraging a pre-trained image encoder~\citep{zhai2022lit}.
Ultimately though, the most critical components for obtaining cutting-edge foundation models are abundant high-quality data and substantial compute resources.
In this vein, SigLIP 2~\citep{tschannen2025siglip} and Perception Encoder (PE)~\citep{bolya2025perception} achieve impressive results after training on more than $40$B image-text pairs.
The largest PE model is trained on $86$B billion samples with a global batch size of $131$K.
Finally, a range of more complex and natively multimodal approaches have been proposed; these include contrastive captioning~\citep{yu2022coca}, masked modeling in the latent space~\citep{bao2021beit, wang2022image, fang2022eva, wang2023onepeace}, and auto-regressive training~\citep{fini2024multimodal}.\looseness-1

In contrast, relatively little work has focused on \emph{scaling unsupervised image pretraining}.
Early efforts include \citet{caron2019unsupervised} and \citet{goyal2019scaling} utilizing the YFCC dataset~\citep{thomee2016yfcc100m}.
Further progress has been achieved by focusing on larger datasets and models~\citep{goyal2021self, goyal2022vision}, as well as initial attempts at data curation for SSL~\citep{tian2021divide}.
Careful tuning of the training algorithms, larger architectures, and more extensive training data lead to the impressive results of DINOv2~\citep{oquab2024dinov2}; for the first time, an SSL model matched or surpassed open-source CLIP variants on a range of tasks. 
This direction has recently been further pushed by~\citet{fan2025scaling} by scaling to larger models without data curation, or by \citet{venkataramanan2025franca} using open datasets and improved training recipes.

\paragraph{Dense Transformer Features}
\begin{figure}[t]
    \begin{subfigure}{0.19\textwidth}
        \centering
        \includegraphics[width=\linewidth]{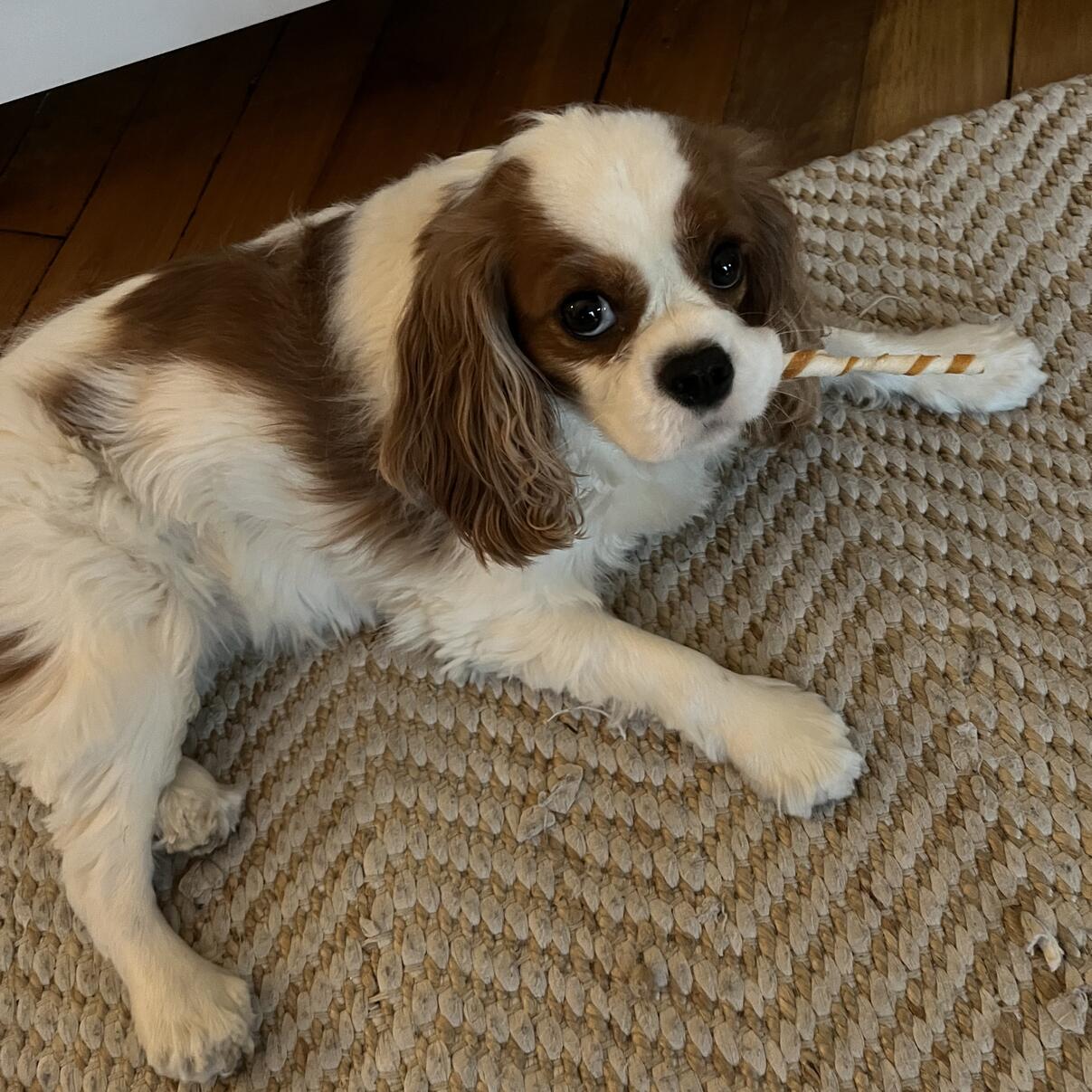}
        \caption*{Input}
    \end{subfigure}\hfill
    \begin{subfigure}{0.19\textwidth}
        \centering
        \includegraphics[width=\linewidth]{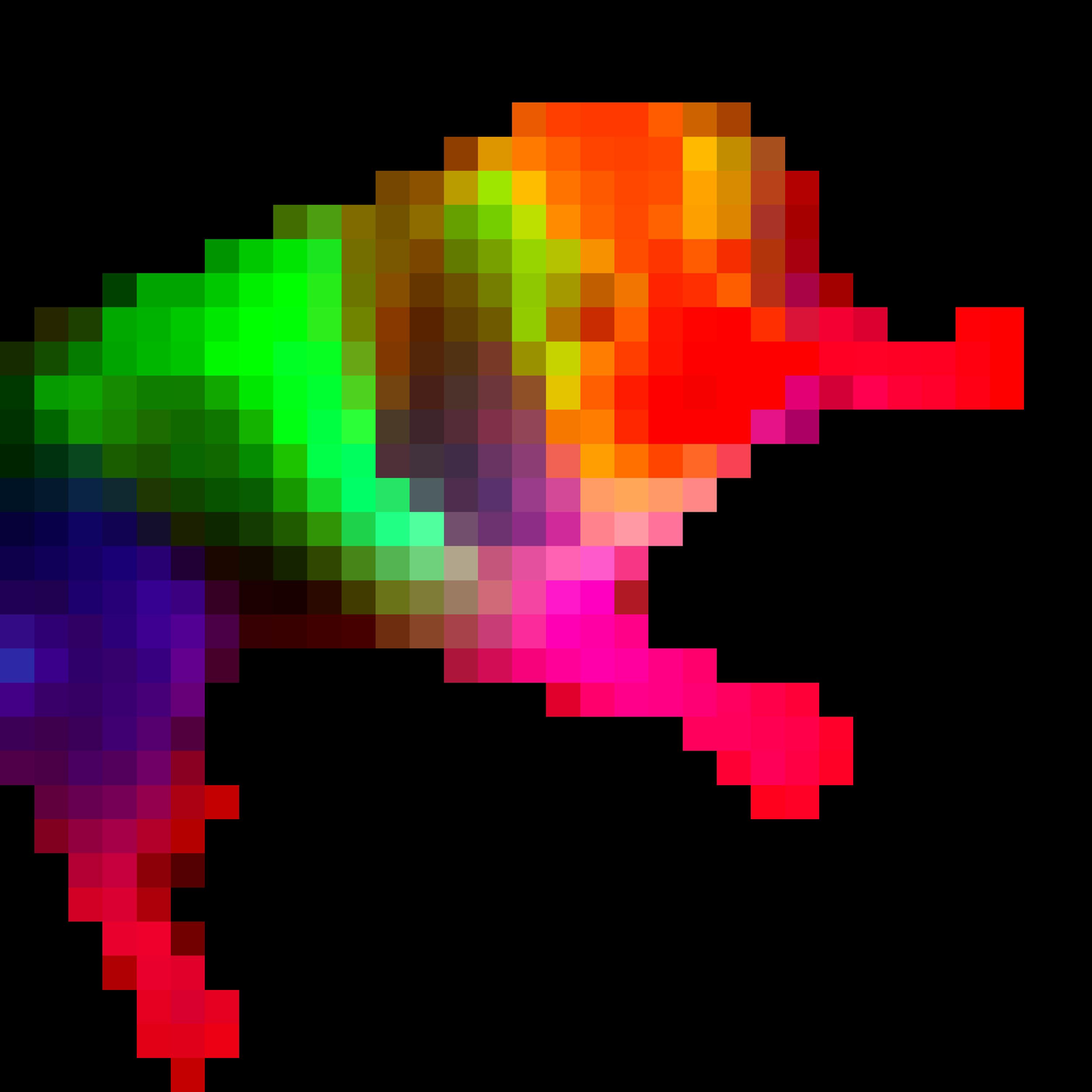}
        \caption*{$512 \times 512$}
    \end{subfigure}\hfill
    \begin{subfigure}{0.19\textwidth}
        \centering
        \includegraphics[width=\linewidth]{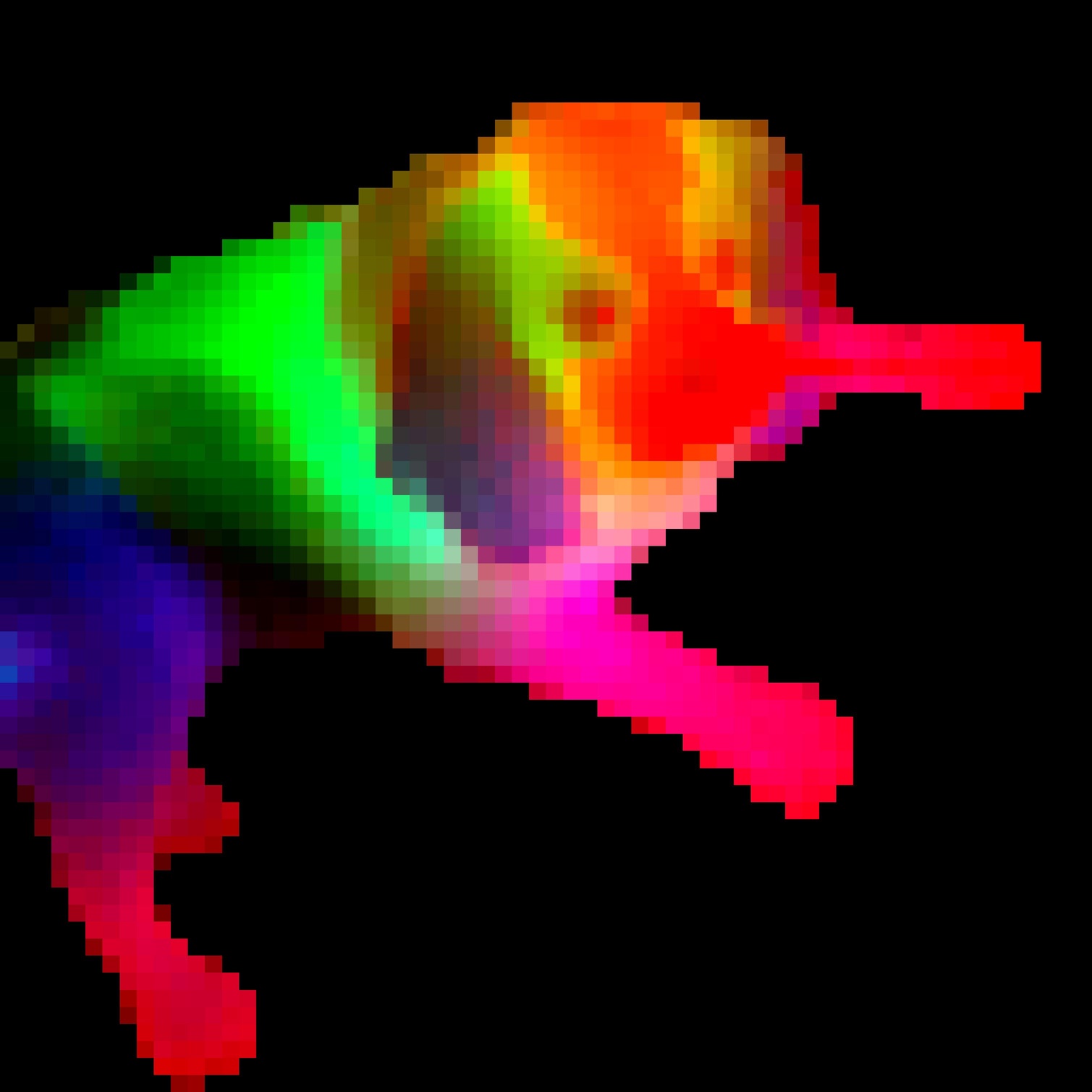}
        \caption*{$1024 \times 1024$}
    \end{subfigure}\hfill
    \begin{subfigure}{0.19\textwidth}
        \centering
        \includegraphics[width=\linewidth]{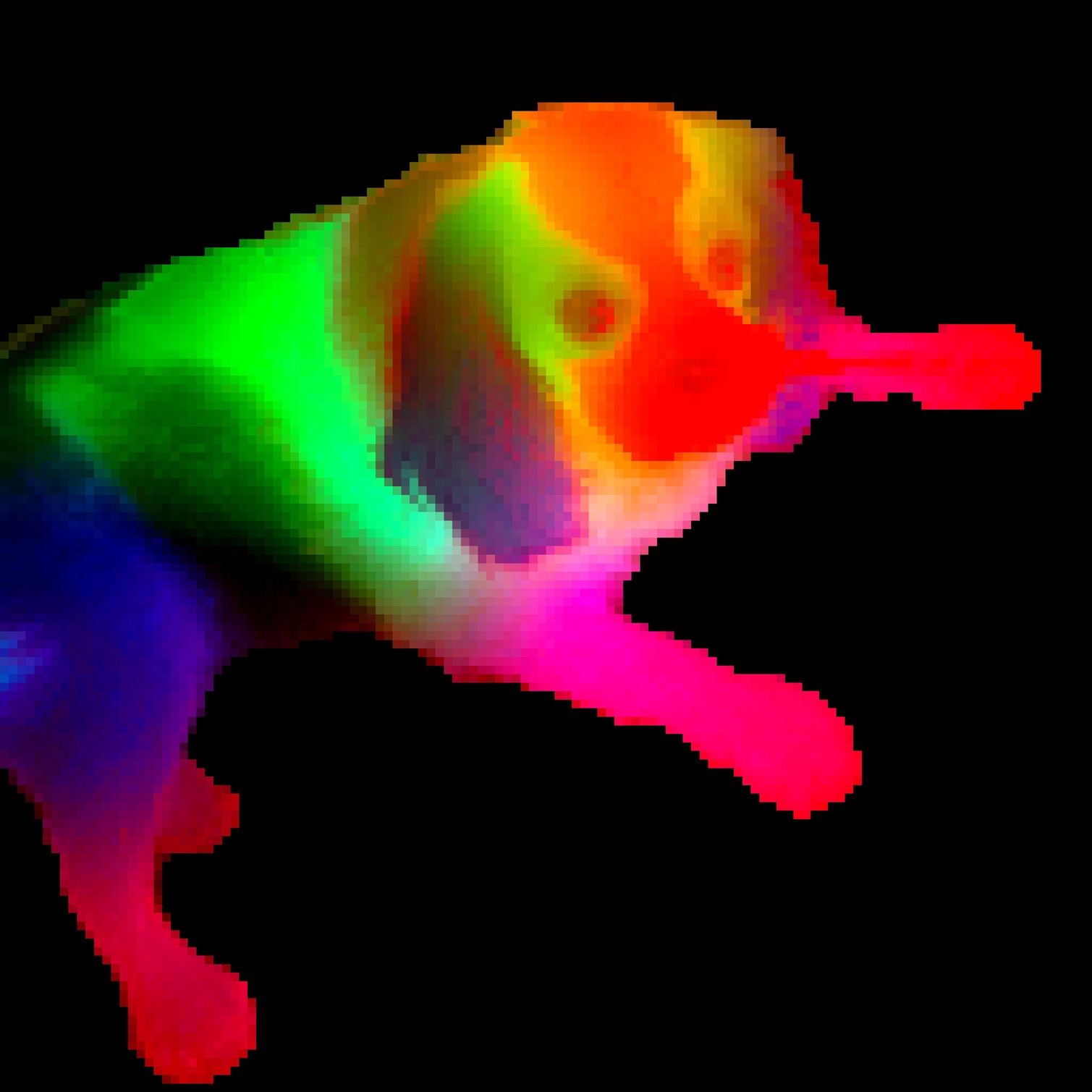}
        \caption*{$2048 \times 2048$}
    \end{subfigure}\hfill
    \begin{subfigure}{0.19\textwidth}
        \centering
        \includegraphics[width=\linewidth]{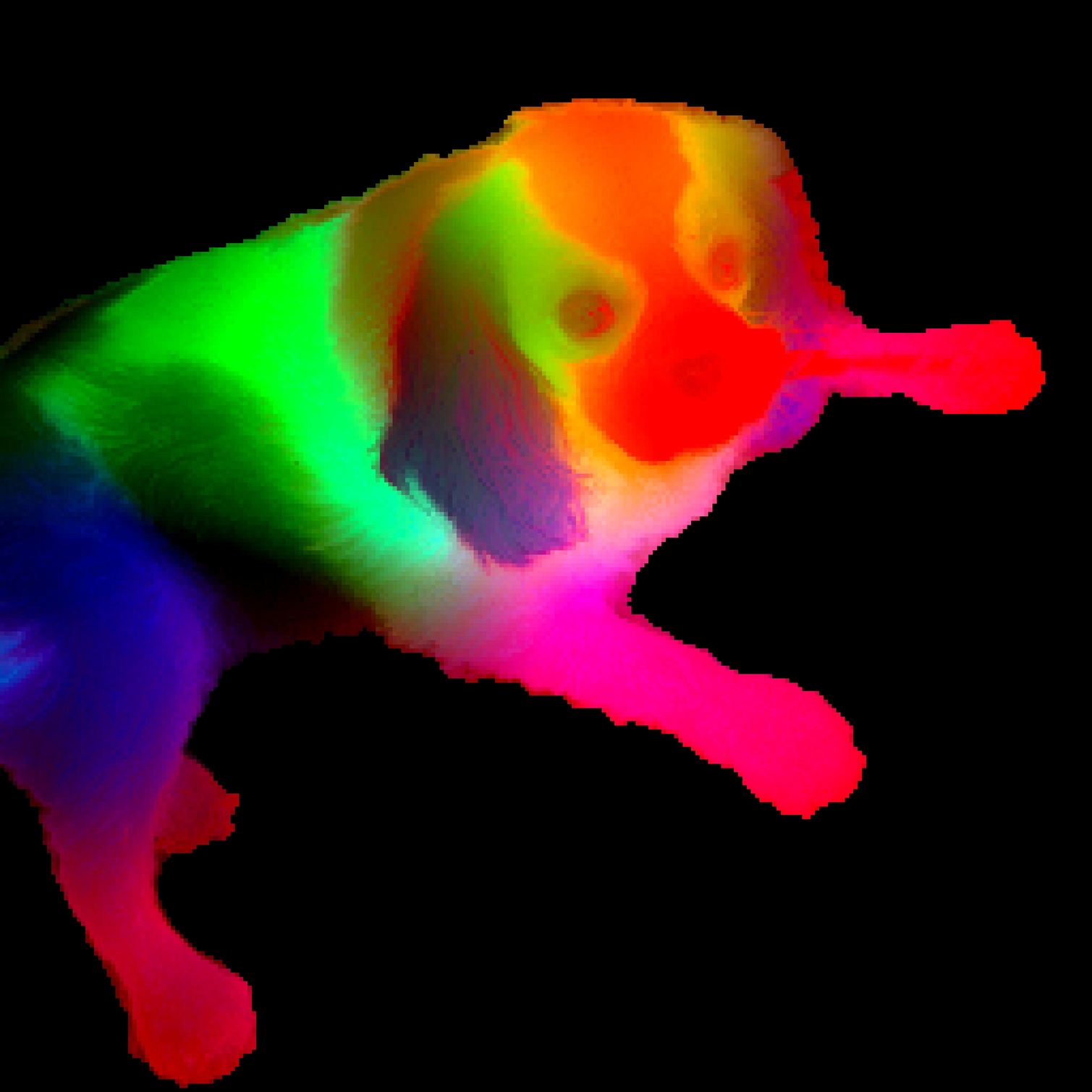}
        \caption*{$4096 \times 4096$}
    \end{subfigure}%
    \caption{
        DINOv3 at very high resolution. 
        We visualize dense features of DINOv3 by mapping the first three components of a PCA computed over the feature space to RGB.
        To focus the PCA on the subject, we mask the feature maps via background subtraction.
        With increasing resolution, DINOv3 produces crisp features that stay semantically meaningful.
        We visualize more PCAs in \cref{sec:results-qualitative}.
    }
    \label{fig:visualization-extreme-resolutions}
\end{figure}

A broad range of modern vision applications consume \emph{dense features} of pre-trained transformers, including multi-modal models~\citep{liu2023llava,beyer2024paligemma}, generative models~\citep{yu2024repa,yao2025reconstruction}, 3D understanding~\citep{wang2025vggt}, video understanding~\citep{lin2023video,wang2024internvideo2}, and robotics~\citep{driess2023palme,kim24openvla}.
On top of that, traditional vision tasks such as detection, segmentation, or depth estimation require accurate local descriptors.
To enhance the quality of SSL-trained local descriptors, a substantial body of work focuses on developing \emph{local SSL losses}.
Examples include leveraging spatio-temporal consistency in videos, \eg using point track loops as training signal~\citep{jabri2020space}, exploiting the spatial alignment between different crops of the same image~\citep{pinheiro2020unsupervised,bardes2022vicregl}, or enforcing consistency between neighboring patches~\citep{yun2022patch}.
\citet{darcet2025cluster} show that predicting clustered local patches leads to improved dense representations.
DetCon~\citep{henaff2021efficient} and ORL~\citep{xie2021unsupervised} perform contrastive learning on region proposals but assume that such proposals exist \emph{a priori}; this assumption is relaxed by approaches such as ODIN~\citep{henaff2022object} and SlotCon~\citep{wen2022slotcon}.
Without changing the training objective, \citet{darcet2024vision} show that adding register tokens to the input sequence greatly improves dense feature maps, and recent works find this can be done without model training~\citep{jiang2025vision, chen2025vision}.

A recent trend are distillation-based, ``\emph{agglomerative}'' methods that combine information from multiple image encoders with varying in global and local feature quality, trained using different levels of supervision~\citep{ranzinger2024radio, bolya2025perception}: AM-RADIO~\citep{ranzinger2024radio} combines the strengths of the fully-supervised SAM~\citep{kirillov2023segment}, the weakly-supervised CLIP, and the self-supervised DINOv2 into a unified backbone.
The Perception Encoder~\citep{bolya2025perception} similarly distills SAM(v2) into a specialized dense variant called PEspatial. 
They use an objective enforcing cosine similarity between student and teacher patches to be high, where their teacher is trained with mask annotations. 
Similar losses were shown to be effective in the context of style transfer, by reducing the inconsistency between the Gram matrices of feature dimensions~\citep{gatys2016image,johnson2016perceptual,yoo2024sagagan}.
In this work, we adopt a Gram objective to regularize cosine similarity between student and teacher patches, favoring them being close. 
In our case, we use earlier iterations of the SSL model itself as the teacher, demonstrating that early-stage SSL models effectively guides SSL training for both global and dense tasks. 

Other works focus on post-hoc improvements to the local features of SSL-trained models.
For example, \citet{ziegler2022self} fine-tune a pre-trained model with a dense clustering objective; similarly, \citet{salehi2023time} fine-tune by aligning patch features temporally, in both cases enhance the quality of local features.
Closer to us, \citet{pariza2025near} propose a patch-sorting based objective to encourage the student and teacher to produce features with consistent neighbor ordering. 
Without finetuning, STEGO~\citep{hamilton2022unsupervised} learns a non-linear projection on top of frozen SSL features to form compact clusters and amplify correlation patterns. 
Alternatively, \citet{simoncini2024no} augment self-supervised features by concatenating gradients from different self-supervised objectives to frozen SSL features. 
Recently, \citet{wysoczanska2023clipdino} show that noisy feature maps are significantly improved through a weighted average of patches.

Related, but not specific to SSL, some recent works generate high-resolution feature maps from ViT feature maps~\citep{fu2024featup}, which are often low-resolution due to patchification of images. %
In contrast with this body of work, our models natively deliver high-quality dense feature maps that remain stable and consistent across resolutions, as shown in \cref{fig:visualization-extreme-resolutions}.

\section{Training at Scale Without Supervision}

DINOv3 is a next-generation model designed to produce the most robust and flexible visual representations to date by pushing the boundaries of self-supervised learning. 
We draw inspiration from the success of large language models (LLMs), for which scaling-up the model capacity leads to outstanding \emph{emerging properties}.
By leveraging models and training datasets that are an order of magnitude larger, we seek to unlock the full potential of SSL and drive a similar paradigm shift for computer vision, unencumbered by the limitations inherent to traditional supervised or task-specific approaches. 
In particular, SSL produces rich, high-quality visual features that are not biased toward any specific supervision or task, thereby providing a versatile foundation for a wide range of downstream applications. 
While previous attempts at scaling SSL models have been hindered by issues of instability, this section describes how we harness the benefits of scaling with careful data preparation, design, and optimization. 
We first describe the dataset creation procedure (\cref{sec:data}), then present the self-supervised SSL recipe used for this first training phase of DINOv3 (\cref{sec:algo}).
This includes the choice of architecture, loss functions, and optimization techniques. 
The second training phase, focusing on dense features, will be described in \cref{sec:gram}.

\subsection{Data Preparation}
\label{sec:data}
Data scaling is one of the driving factors behind the success of large foundation models~\citep{touvron2023llama,radford2021learning,xu2024demystifying,oquab2024dinov2}.
However, increasing naively the size of the training data does not necessarily translate into higher model quality and better performance on downstream benchmarks~\citep{goyal2021self,oquab2024dinov2,vo2024automatic}: 
Successful data scaling efforts typically involve careful data curation pipelines.
These algorithms may have different objectives: either focusing on improving data \textit{diversity} and \textit{balance}, or data \textit{usefulness}---its relevance to common practical applications.
For the development of DINOv3, we combine two complementary approaches to improve both the generalizability and performance of the model, striking a balance between the two objectives.

\begin{table}[t]
    \small
    \centering
    \caption{
        Influence of training data on features quality shown via performance on downstream tasks. 
        We compare datasets curated with \emph{clustering}~\citep{vo2024automatic} and \emph{retrieval}~\citep{oquab2024dinov2}
        to \emph{raw} data and
        to our data mixture. 
        This ablation study is run for a shorter schedule of 200k iterations.
    }
    \label{table:data_ablation}
    \begin{tabular}{@{} l c ccccc @{}}
        \toprule 
        Dataset && IN1k k-NN & IN1k Linear & ObjectNet & iNaturalist 2021 & Paris Retrieval \\
        \midrule
        Raw && 80.1 & 84.8 & 70.3 & 70.1 & 63.3 \\
        Clustering  && 79.4 & 85.4 & 72.3 & 81.3 & 85.2 \\
        Retrieval && 84.0 & 86.7 & 70.7 & 86.0 & 82.7 \\
        \midrule
        \LVDdataset (ours) && \bf 84.6 & \bf 87.2 & \bf 72.8 & \bf 87.0 & \bf 85.9 \\
        \bottomrule   
    \end{tabular}
\end{table}

\paragraph{Data Collection and Curation}
We build our large-scale pre-training dataset by leveraging a large data pool of web images collected from public posts on Instagram. These images already went through platform-level content moderation to help prevent harmful contents and we obtain an initial data pool of approximately 17 billions of images.
Using this raw data pool, we create three dataset \emph{parts}.
We construct the first part by applying the automatic curation method based on hierarchical $k$-means from~\citet{vo2024automatic}.  
We employ DINOv2 as image embeddings, and use 5 levels of clustering with the number of clusters from the lowest to highest levels being $200$M, $8$M, $800$k, $100$k, and $25$k respectively. 
After building the hierarchy of clusters, we apply the balanced sampling algorithm proposed in~\citet{vo2024automatic}.
This results in a curated subset of 1,689 million images (named \LVDdataset) that guarantees a balanced coverage of all visual concepts appearing on the web.
For the second part, we adopt a retrieval-based curation system similar to the procedure proposed by~\citet{oquab2024dinov2}.
We retrieve images from the data pool that are similar to those from selected seed datasets, creating a dataset that covers visual concepts relevant for downstream tasks.
For the third part, we use raw publicly available computer vision datasets including ImageNet1k~\citep{deng2009imagenet}, ImageNet22k~\citep{russakovsky2015imagenet}, and Mapillary Street-level Sequences~\citep{Warburg_CVPR_2020}.
This final part allows us to optimize our model's performance, following \citet{oquab2024dinov2}. 

\paragraph{Data Sampling}
During pre-training, we use a sampler to mix different data parts together.
There are several different options for mixing the above data components.
One is to train with \textit{homogeneous} batches of data that come from a single, randomly selected component in each iteration.
Alternatively, we can optimize the model on \textit{heterogeneous} batches that are assembled by data from all components, selected using certain ratios.
Inspired by \citet{charton2024emergent}, who observed that it is beneficial to have homogeneous batches consisting of very high quality data from a small dataset, we randomly sample in each iteration either a homogeneous batch from ImageNet1k alone or a heterogeneous batch mixing data from all other components.
In our training, homogeneous batches from ImageNet1k account for 10\% of training.

\paragraph{Data Ablation}
To assess the impact of our data curation technique, we perform an ablation study to compare our data mix against datasets curated with clustering or retrieval-based methods alone, and the raw data pool.
To this end, we train a model on each dataset and compare their performance on standard downstream tasks.
For efficiency, we use a shorter schedule of 200k iterations instead of 1M iterations.
In \cref{table:data_ablation}, it can be seen that no single curation technique works best across all benchmarks, and that our full pipeline allows us to obtain the best of both worlds.

\subsection{Large-Scale Training with Self-Supervision}
\label{sec:algo}
While models trained with SSL have demonstrated interesting properties~\citep{chen2020big,caron2021emerging}, most SSL algorithms have not been scaled-up to larger models sizes. %
This is either due to issues with training stability~\citep{darcet2025cluster}, or overly simplistic solutions that fail to capture the full complexity of the visual world. %
When trained at scale~\citep{goyal2022vision}, models trained with SSL do not necessarily show impressive performance. %
One notable exception is DINOv2, a model with 1.1 billion parameters trained on curated data, matching the performance of weakly-supervised models like CLIP~\citep{radford2021learning}. 
A recent effort to scale DINOv2 to 7 billion parameters~\citep{fan2025scaling} demonstrates promising results on global tasks, but with disappointing results on dense prediction.
Here, we aim to scale up the model and data, and obtain even more powerful visual representations with both improved global and local properties.

\begin{table}[t]
    \centering
    \small
    \caption{
      Comparison of the teacher architectures used in DINOv2 and DINOv3 models.
      We keep the model $40$ blocks deep, and increase the embedding dimension to $4096$.
      Importantly, we use a patch size of $16$ pixels, changing the effective sequence length for a given resolution.
    }
    \begin{tabular}{l cc}
        \toprule
        Teacher model & DINOv2 & DINOv3 \\
        \midrule
        Backbone & ViT-giant & ViT-7B \\
        \#Params & 1.1B & 6.7B \\
        \#Blocks & 40 & 40 \\
        Patch Size & 14 & 16 \\
        Pos.~Embeddings & Learnable & RoPE \\
        Registers & 4 & 4 \\
        Embed.~Dim. & 1536 & 4096 \\
        FFN Type & SwiGLU & SwiGLU \\
        FFN Hidden Dim. & 4096 & 8192 \\
        Attn.~Heads & 24 & 32 \\
        Attn.~Heads Dim. & 64 & 128 \\
        \midrule
        DINO Head MLP & 4096-4096-256 & 8192-8192-512 \\
        DINO Prototypes & 128k & 256k \\
        iBOT Head MLP & 4096-4096-256 & 8192-8192-384 \\
        iBOT Prototypes & 128k & 96k \\
        \bottomrule
    \end{tabular}
    \label{tab:models_comparison}
    \vspace{-0.5em}
\end{table}

\paragraph{Learning Objective} We train the model with a discriminative self-supervised strategy which is a mix of several self-supervised objectives with both global and local loss terms. 
Following DINOv2~\citep{oquab2024dinov2},
we use an image-level objective~\citep{caron2021emerging} $\mathcal{L_{\mathrm{DINO}}}$, and balance it with a patch-level latent reconstruction objective~\citep{zhou2021ibot} $\mathcal{L_{\mathrm{iBOT}}}$. 
We also replace the centering from DINO with the Sinkhorn-Knopp from SwAV~\citep{caron2020unsupervised} in both objectives.
Each objective is computed using the output of a dedicated head on top of the backbone network, allowing for some specialization of features before the computation of the losses. 
Additionally, we use a dedicated layer normalization applied to the backbone outputs of the local and global crops. 
Empirically, we found this change to stabilize ImageNet kNN-classification late in training (+0.2 accuracy) and improve dense performance (\eg +1 mIoU on ADE20k segmentation, -0.02 RMSE on NYUv2 depth estimation).
In addition, a Koleo regularizer $\mathcal{L}_{\mathrm{Koleo}}$ is added to encourage the features within a batch to spread uniformly in the space~\citep{sablayrolles2018spreading}. 
We use a distributed implementation of Koleo in which the loss is applied in small batches of $16$ samples---possibly across GPUs.
Our initial training phase is carried by optimizing the following loss:
\begin{equation}
    \mathcal{L_{\mathrm{Pre}}} = \mathcal{L_{\mathrm{DINO}}} + \mathcal{L}_{\mathrm{iBOT}} + 0.1 * \mathcal{L}_{\mathrm{DKoleo}.}
\end{equation}

\paragraph{Updated Model Architecture}
For the model scaling aspect of this work, we increase the size of the model to 7B parameters, and provide in \cref{tab:models_comparison} a comparison of the corresponding hyperparameters with the 1.1B parameter model trained in the DINOv2 work. We also employ a custom variant of RoPE: our base implementation assigns coordinates in a normalized $[-1,1]$ box to each patch, then applies a bias in the multi-head attention operation depending on the relative position of two patches. In order to improve the robustness of the model to resolutions, scales and aspect ratios, we employ \textit{RoPE-box jittering}. The coordinate box $[-1,1]$ is randomly scaled to $[-s,s]$, where $s \in [0.5, 2]$.
Together, these changes enable DINOv3 to better learn detailed and robust visual features, improving its performance and scalability.

\paragraph{Optimization}
Training large models on very large datasets represents a complicated experimental workflow.
Because the interplay between model capacity and training data complexity is hard to assess \emph{a priori}, it is impossible to guess the right optimization horizon.
To overcome this, we get rid of all parameter scheduling, and train with constant learning rate, weight decay, and teacher EMA momentum.
This has two main benefits.
First, we can continue training as long as downstream performance continues to improve. %
Second, the number of optimization hyperparameters is reduced, making it easier to choose them properly.
For the training to start properly, we still use a linear warmup for learning rate and teacher temperature. %
Following common practices, we use AdamW~\citep{loshchilov2017decoupled}, and set the total batch size to $4096$ images split across $256$ GPUs.
We train our models using the multi-crop strategy~\citep{caron2020unsupervised}, taking $2$ global crops and $8$ local crops per image.
We use square images with a side length of $256$/$112$ pixels for global/local crops, which, along with the change in patch size, results in the same effective sequence length per image as in DINOv2 and a total sequence length of $3.7$M tokens per batch.
Additional hyperparameters can be found in \cref{app:implem-details} and in the code release.

\begin{figure}[t]
    \begin{minipage}[b]{.6\linewidth}
        \centering
        \small
        \renewcommand{\arraystretch}{0.2}
        \setlength{\tabcolsep}{0.3pt}
        \begin{tabular}{ccccccc}
           \rotatebox[origin=c]{-90}{\includegraphics[width=0.32\linewidth]{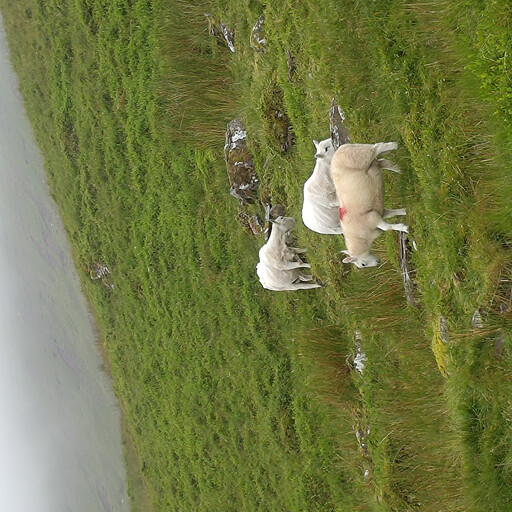}} & 
          \includegraphics[width=0.32\linewidth]{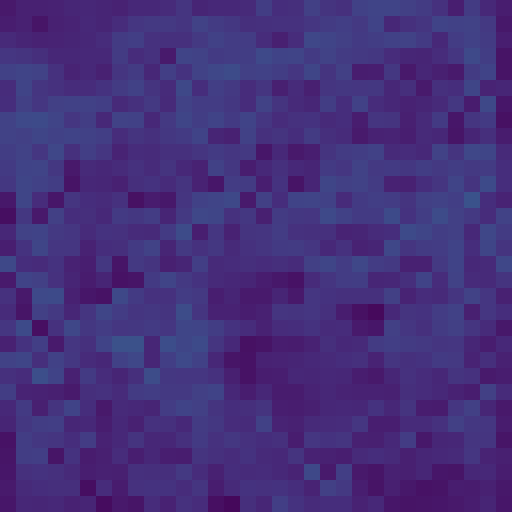} & 
          \includegraphics[width=0.32\linewidth]{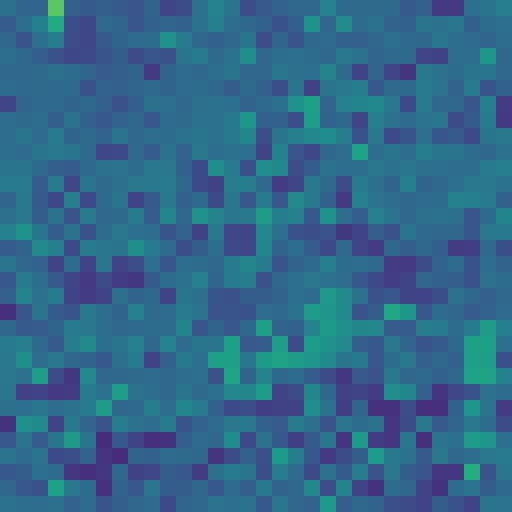} 
          \\
           \rotatebox[origin=c]{-90}{\includegraphics[width=0.32\linewidth]{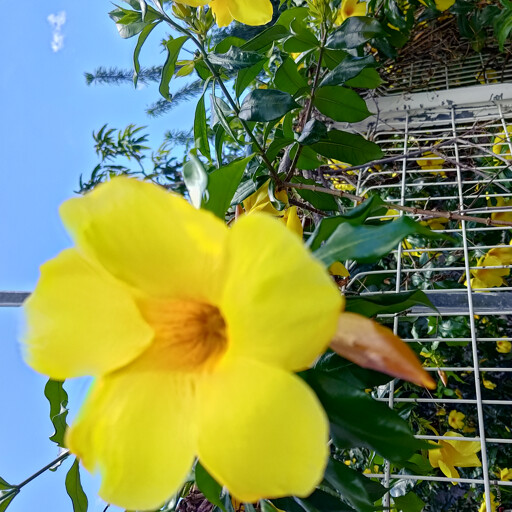}} & 
          \includegraphics[width=0.32\linewidth]{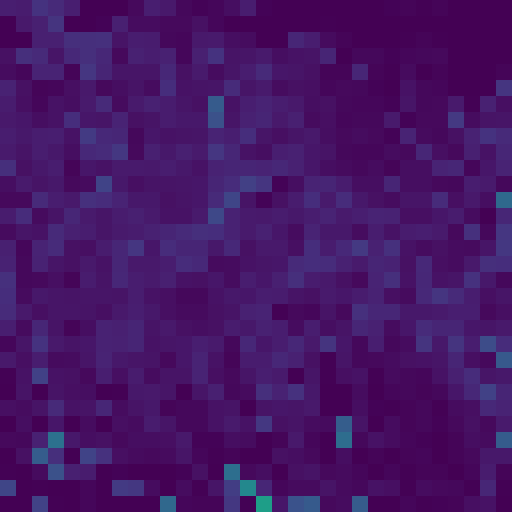} & 
          \includegraphics[width=0.32\linewidth]{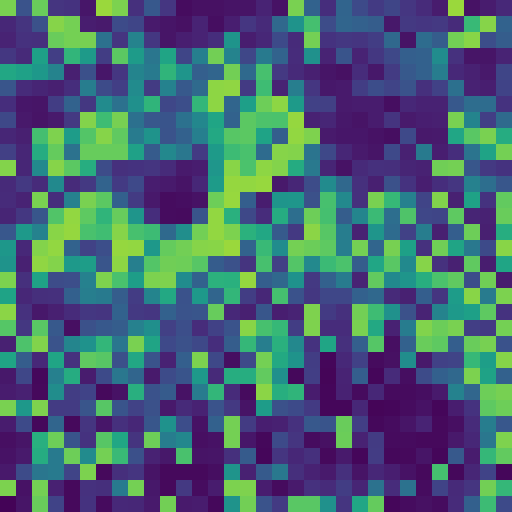} 
          \\[0.5em]
          Image & 200k & 1M
        \end{tabular}
        
        \subcaption{Cosine similarities between the CLS and output patches.}
        \label{fig:evolution-cls}
    \end{minipage}%
    \hfill
    \begin{minipage}[b]{.4\linewidth}
        \centering
        \refstepcounter{subfigure}  %
        \label{fig:evolution-cls-g}
        \includegraphics{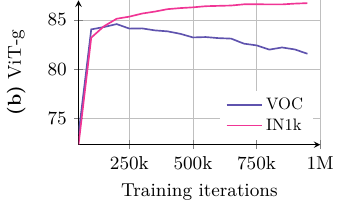}%

        \vspace{0.35em}
        \refstepcounter{subfigure}  %
        \label{fig:evolution-cls-7B}
        \includegraphics{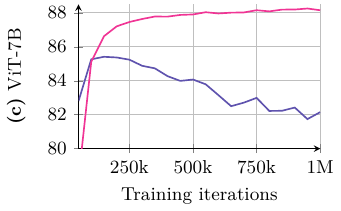}%
        \vspace{0.35em}
    \end{minipage}%
    
    \caption{Evolution of the cosine similarities (a) and of the accuracy on ImageNet1k linear (IN1k) and segmentation on VOC for ViT-g (b) and ViT-7B (c). We observe that the segmentation performance is maximal when the cosine similarities between the patch tokens and the class tokens are low. As training progresses, these similarities increase and the performance on dense tasks decreases.}
    \label{fig:gram:evolution}
\end{figure}

\section[Gram: A Regularization for Dense Features]{\Gramname: A Regularization for Dense Features}
\label{sec:gram}

To fully leverage the benefits of large-scale training, we aim to train the 7B model for an extended duration, with the notion that it could potentially train indefinitely. As expected, prolonged training leads to improvements on global benchmarks. However, as training progresses, the performance degrades on dense tasks (\cref{fig:evolution-cls-g,fig:evolution-cls-7B}). 
This phenomenon, which is due to the emergence of patch-level inconsistencies in feature representations, undermines the interest behind extended training.\footnote{%
We also observed different types of outliers appearing with continued training; we provide a discussion in \cref{app:sec:artifacts}.
} 
In this section, we first analyze the loss of patch-level consistency, then propose a new objective to mitigate it, called \emph{\gramname}. 
We finally discuss the impact of our approach on both training stability and model performance.

\subsection{Loss of Patch-Level Consistency Over Training}

\begin{figure}[t]
    \centering
    \small
    \renewcommand{\arraystretch}{0.2}
    \setlength{\tabcolsep}{0.3pt}
    \begin{tabular}{@{}ccccccc@{}}
       \rotatebox[origin=c]{-90}{\includegraphics[width=0.165\linewidth]{images/evolution_cosine_u1/P_20240709_135144.jpg_croped.lr.jpg}} & 
      \includegraphics[width=0.165\linewidth]{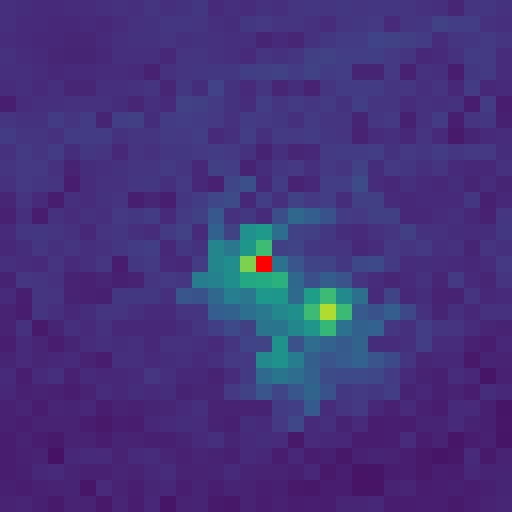} & 
      \includegraphics[width=0.165\linewidth]{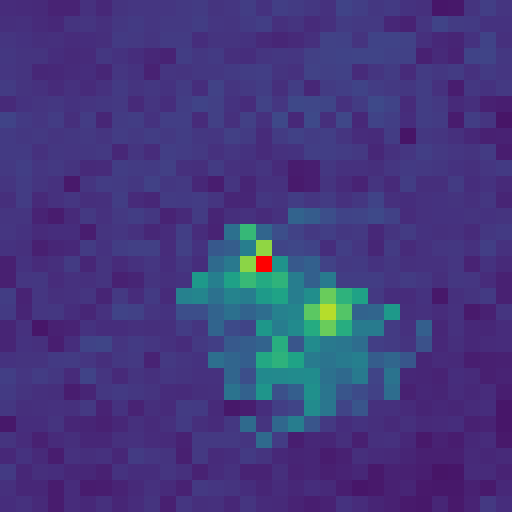} & 
      \includegraphics[width=0.165\linewidth]{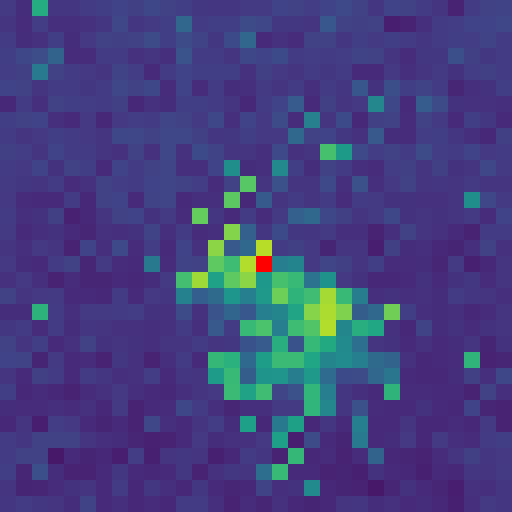} & 
      \includegraphics[width=0.165\linewidth]{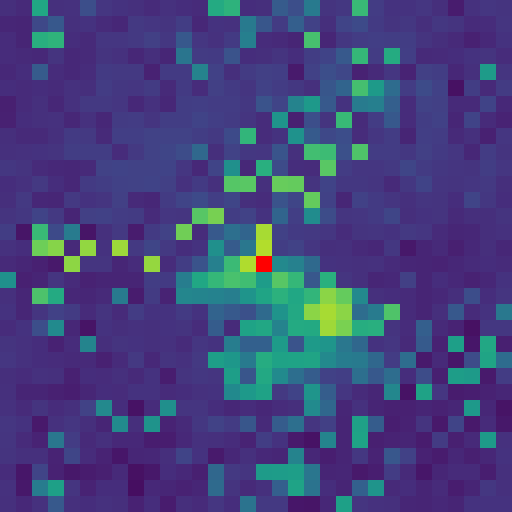} & 
      \includegraphics[width=0.165\linewidth]{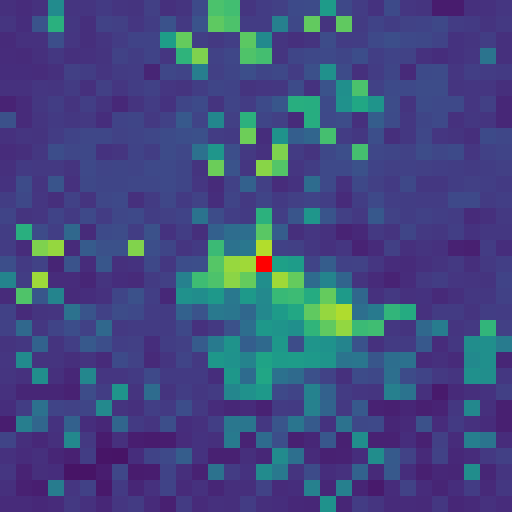} 
      \\
       \rotatebox[origin=c]{-90}{\includegraphics[width=0.165\linewidth]{images/evolution_cosine_u1/P_20250105_134132.jpg_croped.lr.jpg}} & 
      \includegraphics[width=0.165\linewidth]{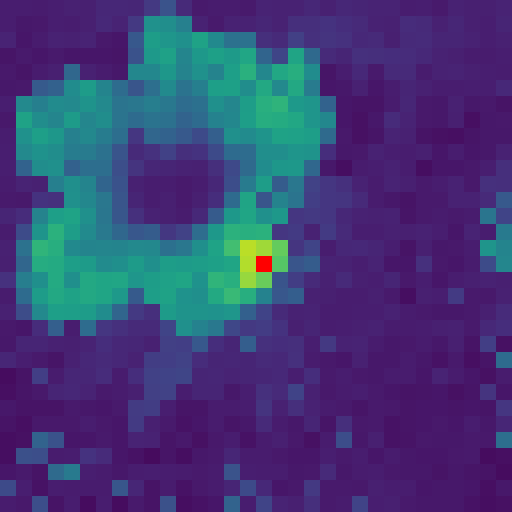} & 
      \includegraphics[width=0.165\linewidth]{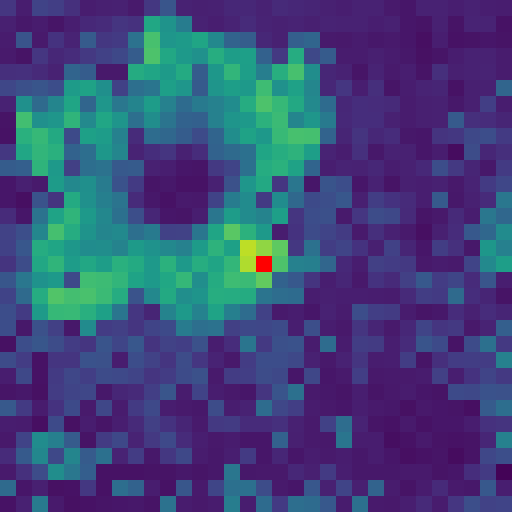} & 
      \includegraphics[width=0.165\linewidth]{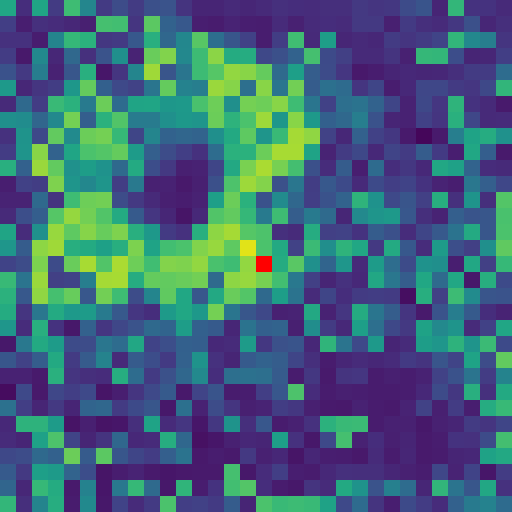} & 
      \includegraphics[width=0.165\linewidth]{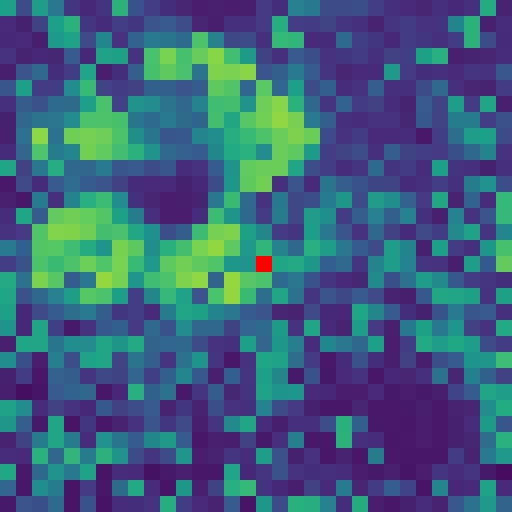} & 
      \includegraphics[width=0.165\linewidth]{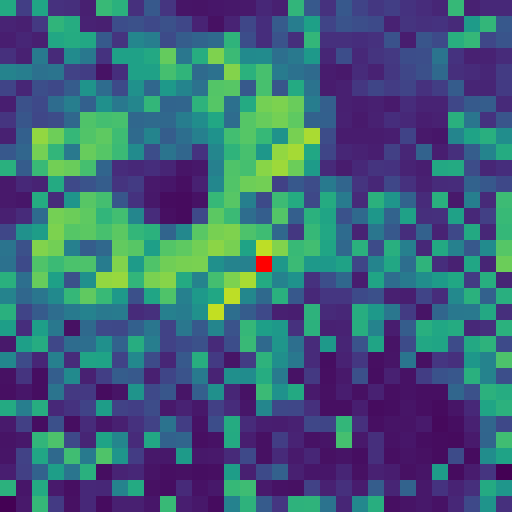} 
      \\[0.5em]
      Image & 200k & 400k & 600k & 800k & 1M \\
    \end{tabular}
    \caption{
     Evolution of the cosine similarity between the patch noted in red and all other patches.
     As training progresses, the features produced by the model become less localized and the similarity maps become noisier.
    }
    \label{fig:evolution-cosines}
\end{figure}

During extended training, we observe consistent improvements in global metrics but a notable decline in performance on dense prediction tasks. 
This behavior was previously observed, to a lesser extent, during the training of DINOv2, and also discussed in the scaling effort of \citet{fan2025scaling}.
However, to the best of our knowledge, it remains unresolved to date.
We illustrate the phenomenon in \cref{fig:evolution-cls-g,fig:evolution-cls-7B}, which present the performance of the model across iterations on both image classification and segmentation tasks. 
For classification, we train a linear classifier on ImageNet-1k using the CLS token and report top-1 accuracy. 
For segmentation, we train a linear layer on patch features extracted from Pascal VOC and report mean Intersection over Union (mIoU). 
We observe that both for the ViT-g and the ViT-7B, the classification accuracy monotonically improves throughout training. 
However, segmentation performance declines in both cases after approximately 200k iterations, falling below its early levels in the case of the ViT-7B.

To better understand this degradation, we analyze the quality of patch features by visualizing cosine similarities between patches. \cref{fig:evolution-cosines} shows the cosine similarity maps between the backbone’s output patch features and a reference patch (highlighted in red). At 200k iterations, the similarity maps are smooth and well-localized, indicating consistent patch-level representations. However, by 600k iterations and beyond, the maps degrade substantially,  with an increasing number of irrelevant patches with high similarity to the reference patch. 
This loss of patch-level consistency correlates with the drop in dense task performance.

These patch-level irregularities differ from the high-norm patch outliers described in \citet{darcet2024vision}. 
Specifically, with the integration of register tokens, patch norms remain stable throughout training. However, we notice that the cosine similarity between the CLS token and the patch outputs gradually increases during training. This is expected, yet it means that the locality of the patch features diminishes. We visualize this phenomenon in \cref{fig:evolution-cls}, which depicts the cosine maps at 200k and 1M iterations. 
In order to mitigate the drop on dense tasks, we propose a new objective specifically designed to regularize the patch features and ensure a good patch-level consistency, while preserving high global performance.

\subsection[Gram Objective]{\Gramname Objective}

\begin{figure}[t]
    \centering
    \begin{subfigure}{0.32\linewidth} 
        \centering
        \includegraphics{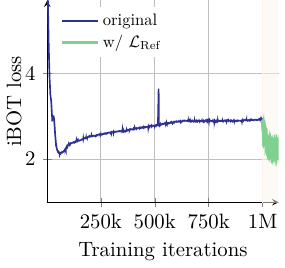}
        \caption{iBOT loss}
    \end{subfigure}
    \hfill
    \begin{subfigure}{0.32\linewidth} 
        \centering
        \includegraphics{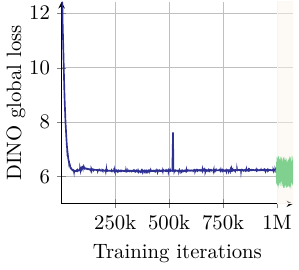}
        \caption{DINO global loss}
    \end{subfigure}
    \hfill
    \begin{subfigure}{0.32\linewidth} 
        \centering
        \includegraphics{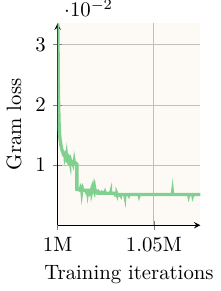}
        \caption{Gram loss}
    \end{subfigure}
    \caption{Evolution trough the training iterations of the patch-level iBOT loss, the global loss DINO (applied to the global crops) and the newly introduced Gram loss. We \colorbox{White!90!Tan}{highlight} the iterations of the refinement step $\mathcal{L}_\mathrm{Ref}$ which uses the Gram objective. }
    \label{fig:losses}
\end{figure}

Throughout our experiments, we have identified a relative independence between learning strong discriminative features and maintaining local consistency, as observed in the lack of correlation between global and dense performance. 
While combining the global DINO loss with the local iBOT loss has begun to address this issue, we observe that the balance is unstable, with global representation dominating as training progresses.
Building on this insight, we propose a novel solution that explicitly leverages this independence. 

We introduce a new objective which mitigates the degradation of patch-level consistency by enforcing the quality of the patch-level consistency, without impacting the features themselves.
This new loss function %
operates on the Gram matrix: the matrix of all pairwise dot products of patch features in an image.
We want to push the Gram matrix of the student towards that of an earlier model, referred to as the \emph{Gram teacher}.
We select the Gram teacher by taking an early iteration of the teacher network, which exhibits superior dense properties. 
By operating on the Gram matrix rather than the feature themselves, the local features are free to move, provided the structure of similarities remains the same.
Suppose we have an image composed of $P$ patches, and a network that operates in dimension $d$.
Let us denote by $\mathbf{X}_S$ (respectively $\mathbf{X}_{G}$) the $P \times d$ matrix of $\mathbf{L}_2$-normalized local features of the student (respectively the Gram teacher).
We define the loss $\mathcal{L}_{\text{Gram}}$ as follows: 
\begin{equation}
  \mathcal{L}_{\text{Gram}} = \left \| \mathbf{X}_{S} \cdot \mathbf{X}_{S}^\top - \mathbf{X}_{G} \cdot \mathbf{X}_{G}^\top  \right \|_\text{F}^2.
  \label{eq:gram-loss}
\end{equation}

We only compute this loss on the global crops.
Even though it can be applied early on during the training, for efficiency, we start only after $1$M iterations.
Interestingly, we observe that the late application of $\mathcal{L}_{\text{Gram}}$ still manages to ``repair'' very degraded local features.
In order to further improve performance, we update the Gram teacher every 10k iterations at which the Gram teacher becomes identical to the main EMA teacher.  
We call this second step of training the \emph{refinement step}, which optimizes the objective $\mathcal{L}_\mathrm{Ref}$, with 
\begin{equation}
    \mathcal{L}_\mathrm{Ref} = w_{\mathrm{D}}\mathcal{L_{\mathrm{DINO}}} + \mathcal{L}_{\mathrm{iBOT}} + w_{\mathrm{DK}}\mathcal{L}_{\mathrm{DKoleo}} + w_{\mathrm{Gram}}\mathcal{L}_{\mathrm{Gram}}.
\end{equation}
We visualize the evolution of different losses in \cref{fig:losses} and observe that applying the Gram objective significantly influences the iBOT loss, causing it to decrease more rapidly. 
This suggests that the stability introduced by the stable Gram teacher positively impacts the iBOT objective. In contrast, the Gram objective does not have a significant effect on the DINO losses. 
This observation implies that the Gram and iBOT objectives impact the features in a similar way, whereas the DINO losses affect them differently.

\begin{figure}[t]
    \centering
    \begin{subfigure}{0.32\linewidth} 
        \centering
        \includegraphics{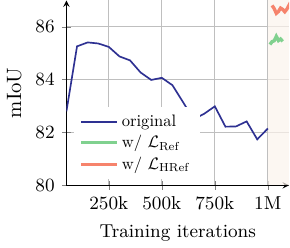}
        \caption{VOC}
    \end{subfigure}
    \hfill
    \begin{subfigure}{0.32\linewidth} 
      \includegraphics{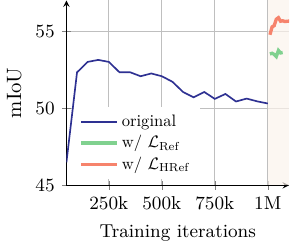}
        \caption{ADE20k}
    \end{subfigure}
    \hfill
    \begin{subfigure}{0.32\linewidth} 
      \includegraphics{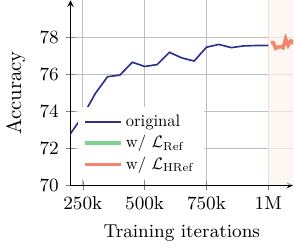}
        \caption{ObjectNet}
    \end{subfigure}    
    \caption{
        Evolution of the results on different benchmarks after applying our proposed \emph{\gramname} method.
        We visualize results when continuing the original training with our refinement step, noted `$\mathcal{L}_\mathrm{Ref}$'.
        We also plot results obtained when using higher-resolution features for the Gram objective as introduced in following \cref{sec:high-res-feats} and noted `$\mathcal{L}_\mathrm{HRef}$'.
        We \colorbox{White!90!Tan}{highlight} the iterations which use the Gram objective. 
    }
    \label{fig:evolution-gram}
\end{figure}

Regarding performance, we observe the impact of the new loss is almost immediate. 
As shown in \cref{fig:evolution-gram}, incorporating \gramname leads to significant improvements on dense tasks within the first 10k iterations. 
We also see notable gains on the ADE20k benchmark following the Gram teacher updates. 
Additionally, longer training further benefits performance on the ObjectNet benchmark and other global benchmarks show mild impact from the new loss.

\subsection{Leveraging Higher-Resolution Features}
\label{sec:high-res-feats}

\begin{figure}[t]
    \begin{subfigure}{.495\linewidth}    
        \centering
        \renewcommand{\arraystretch}{0.2}
        \setlength{\tabcolsep}{0.5pt}
        \begin{tabular}{cccc}
        \includegraphics[height=0.23\textwidth]{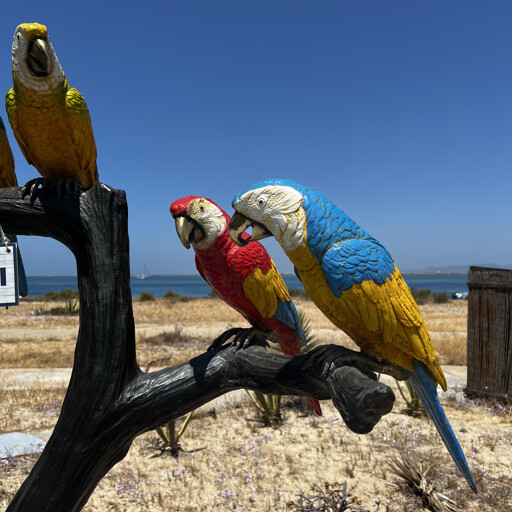} &
            \includegraphics[height=0.23\textwidth]{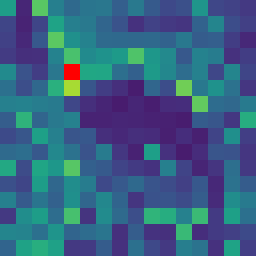} & 
            \includegraphics[height=0.23\textwidth]{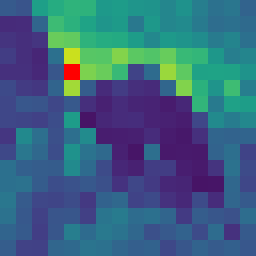} & 
            \includegraphics[height=0.23\textwidth]{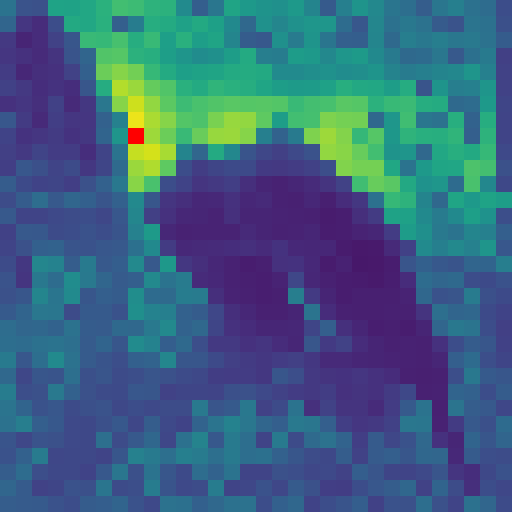} \\
            \rotatebox[origin=c]{-90}{\includegraphics[height=0.23\textwidth]{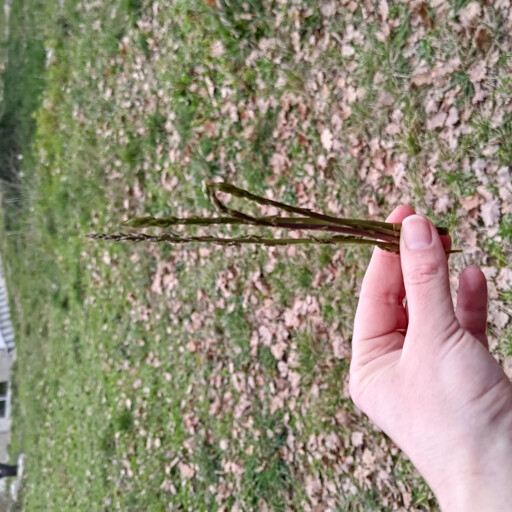}} &
            \includegraphics[height=0.23\textwidth]{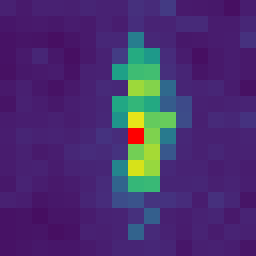} & 
            \includegraphics[height=0.23\textwidth]{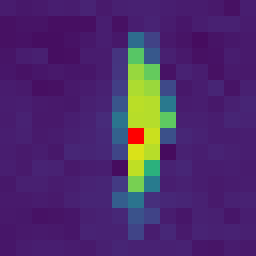} & 
            \includegraphics[height=0.23\textwidth]{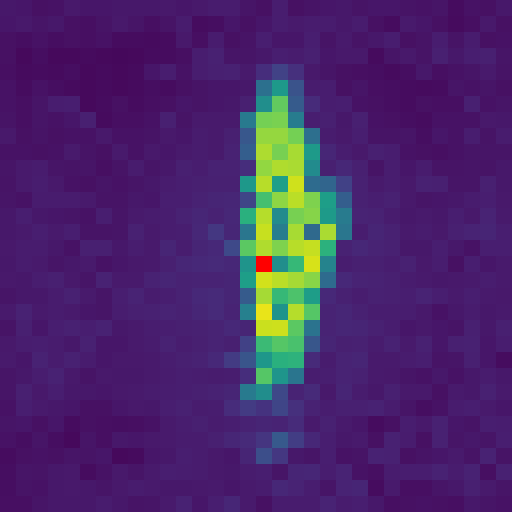} \\\\
            Input &  $256$ &  Downsamp. &  $512$
        \end{tabular}
        \caption{Gram matrices at different input resolutions.}
        \label{fig:gram-resolutions}
    \end{subfigure}   
    \hfill 
    \begin{subfigure}[b]{.495\linewidth}
        \centering
        \small
        \setlength{\tabcolsep}{5pt}
        \begin{tabular}{@{}lcc  ccc@{}}
          \toprule
           & Teacher & Res.  &   IN1k   & ADE  & NYU \\
          Method & Iteration &&  Linear & mIoU & RMSE \\
          \midrule
          Baseline & --- & --- &          \bf 88.2 & 50.3 & 0.307 \\
          GRAM &          200k & $\times$1 & 88.0  & 53.6 & 0.285 \\
          GRAM &          200k & $\times$2 & 88.0 & \bf 55.7 & \bf 0.281 \\
          GRAM &          100k & $\times$2 & 87.9 & \bf 55.7 & 0.284 \\
          GRAM & \phantom{00}1M& $\times$2 & 88.1 & 54.9 & 0.290 \\
          \bottomrule
        \end{tabular}
        \vspace{1.5em}
        \caption{
          Ablation of Gram teachers and resolutions.
        }
        \label{tab:gram-teacher-res}
    \end{subfigure}
    \caption{Quantitative and qualitative study of the impact of high-resolution Gram. We show (a) the improved cosine maps after down-sampling the high-resolution maps into smaller ones, and (b) the quantitative improvements brought by varying the training iteration and the resolution of the Gram teacher.
    }
    \label{fig:gram}
\end{figure}

Recent work shows that a weighted average of patch features can yield stronger local representations by smoothing outlier patches and enhancing patch-level consistency~\citep{wysoczanska2023clipdino}. 
On the other hand, feeding higher-resolution images into the backbone produces finer and more detailed feature maps. 
We leverage the benefits of both observations to compute high-quality features for Gram teacher.
Specifically, we first input images at twice the normal resolution into the Gram teacher, then $2\times$ down-sample the resulting feature maps with the bicubic interpolation to achieve the desired smooth feature maps that match the size of the student output.
\cref{fig:gram-resolutions} visualizes the Gram matrices of patch features obtained with images at resolutions 256 and 512, as well as those obtained after $2\times$ down-sampling features from the 512-resolution (denoted as ‘downsamp.’). We observe that the superior patch-level consistency in the higher-resolution features is preserved through down-sampling, resulting in smoother and more coherent patch-level representations. As a side note, our model can seamlessly process images at varying resolutions without requiring adaptation, thanks to the adoption of Rotary Positional Embeddings (RoPE) introduced by \citet{su2024roformer}.

We compute the Gram matrix of the down-sampled features and use it to replace $\mathbf{X}_{G}$ in the objective $\mathcal{L_{\mathrm{Gram}}}$. We note the new resulting refinement objective as $\mathcal{L}_\mathrm{HRef}$. 
This approach enables the Gram objective to effectively distill the improved patch consistency of smoothed high-resolution features into the student model.
As shown in \cref{fig:evolution-gram} and \cref{tab:gram-teacher-res}, this distillation translates into better predictions on dense tasks, yielding additional gains on top of the benefit brought by $\mathcal{L}_\mathrm{Ref}$ (+2 mIoU on ADE20k).
We also ablate the choice of Gram teacher in \cref{tab:gram-teacher-res}. 
Interestingly, choosing the Gram teacher from 100k or 200k does not significantly impact the results,
but using a much later Gram teacher (1M iterations) is detrimental because the patch-level consistency of such a teacher is inferior. 

\begin{figure}[t]
    \centering
    \renewcommand{\arraystretch}{0.2}
    \setlength{\tabcolsep}{0.2pt}
    \begin{tabular}{@{}cccccc@{}}
    \raisebox{40pt}{\rotatebox[origin=c]{90}{original}} & 
    \rotatebox[origin=c]{-90}{\includegraphics[width=0.19\textwidth]{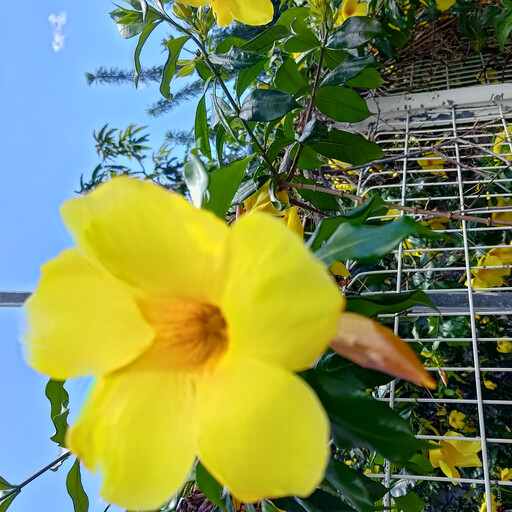}} & 
    \includegraphics[width=0.19\textwidth]{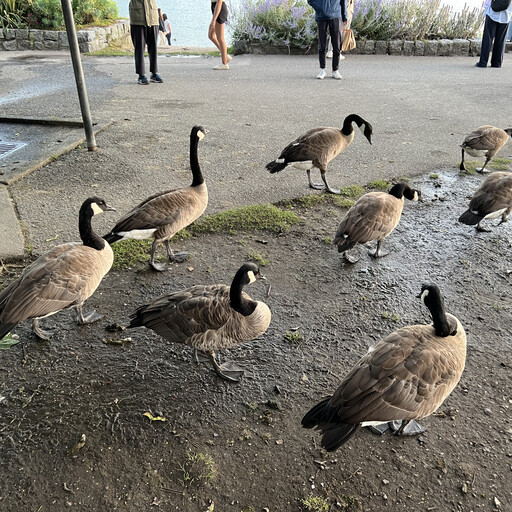} & 
    \includegraphics[width=0.19\textwidth]{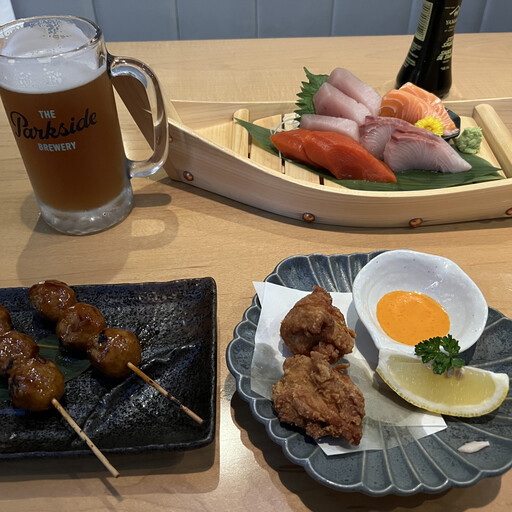} & 
    \includegraphics[width=0.19\textwidth]{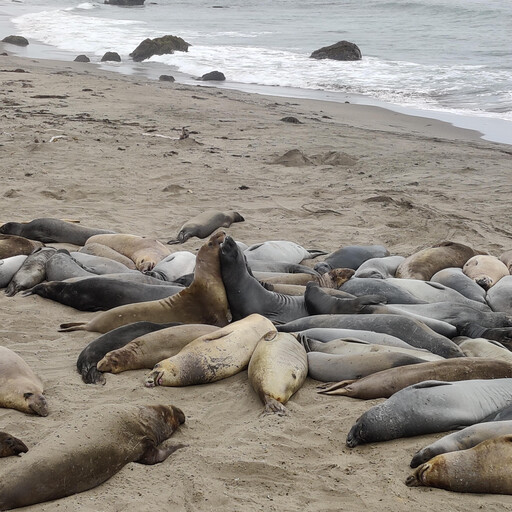} & 
    \rotatebox[origin=c]{-90}{\includegraphics[width=0.19\textwidth]{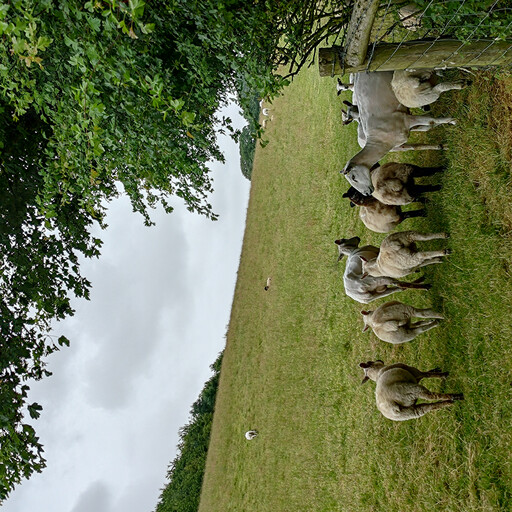}}
    
    \\
    \raisebox{40pt}{\rotatebox[origin=c]{90}{wo/ $\mathcal{L}_\mathrm{HRef}$}} & 
    \includegraphics[width=0.19\textwidth]{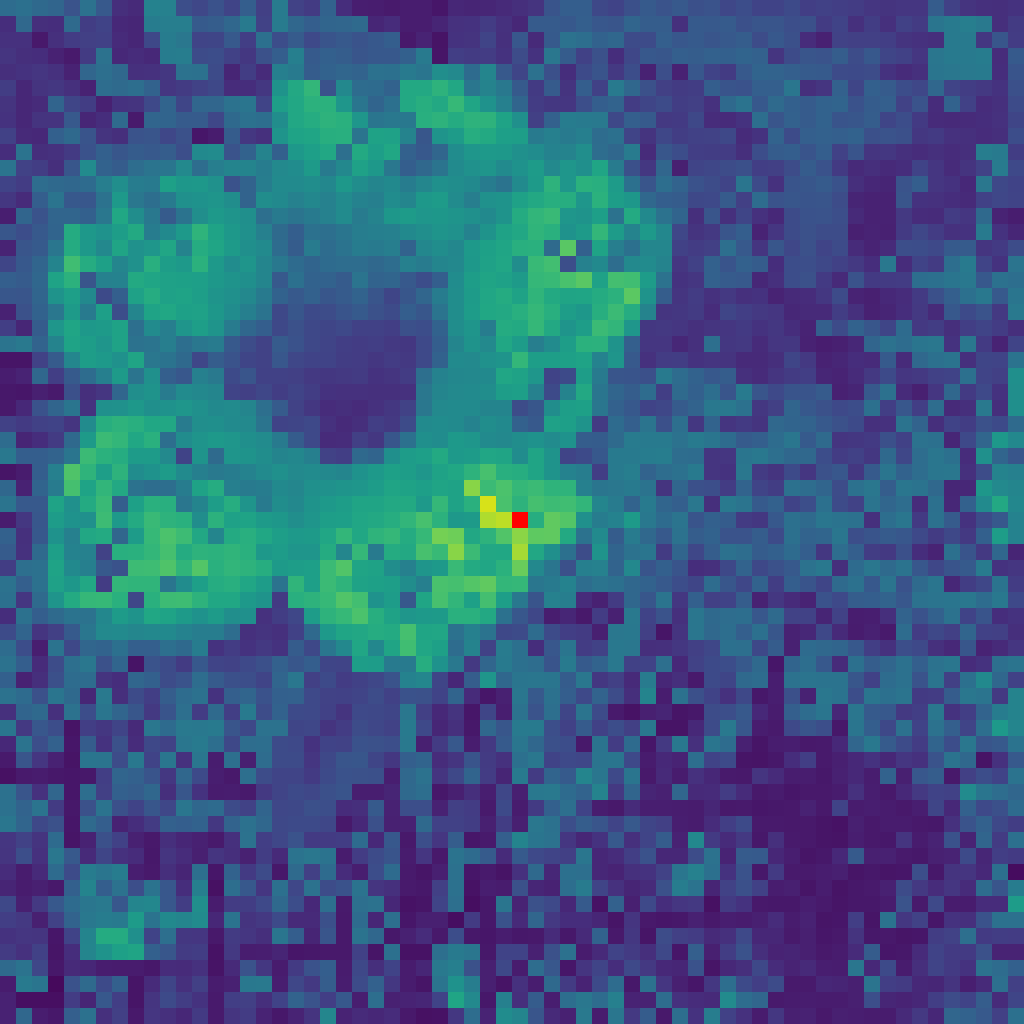} &
    \includegraphics[width=0.19\textwidth]{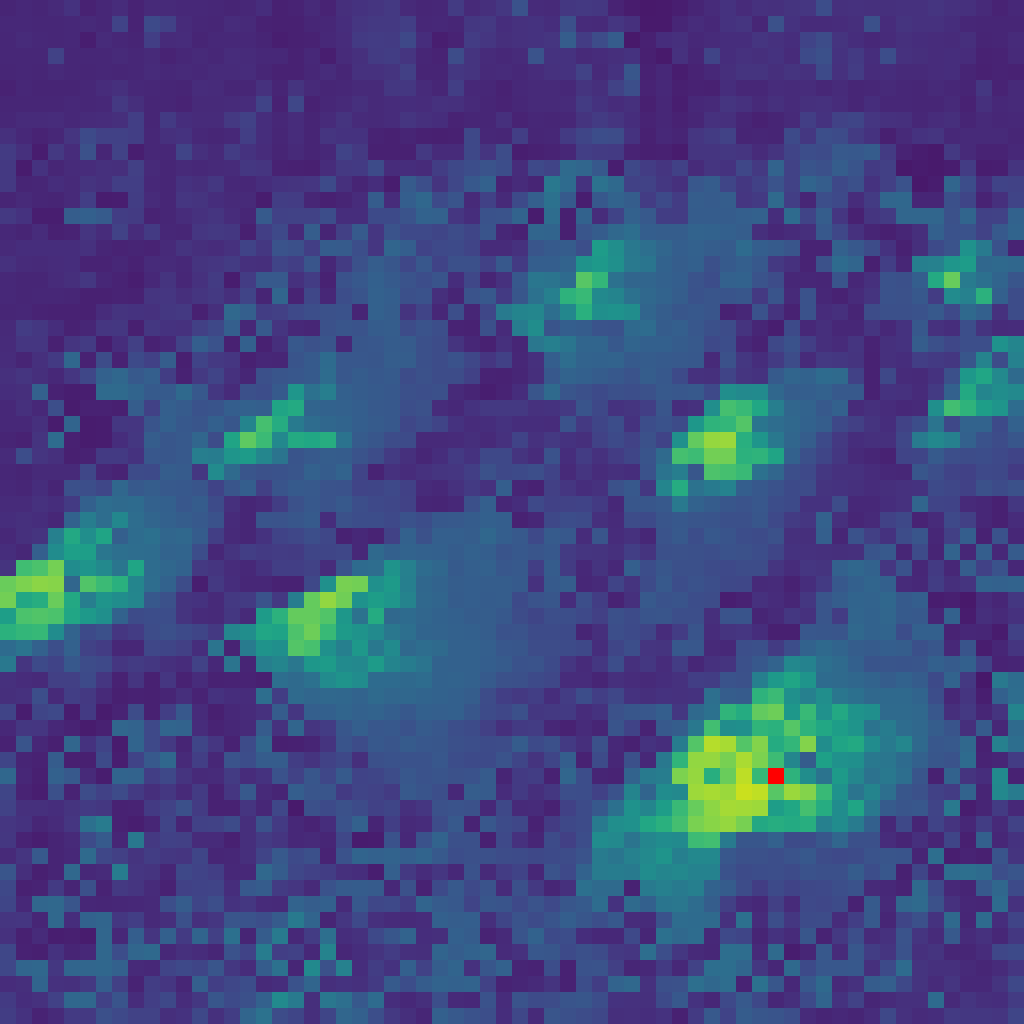} & 
    \includegraphics[width=0.19\textwidth]{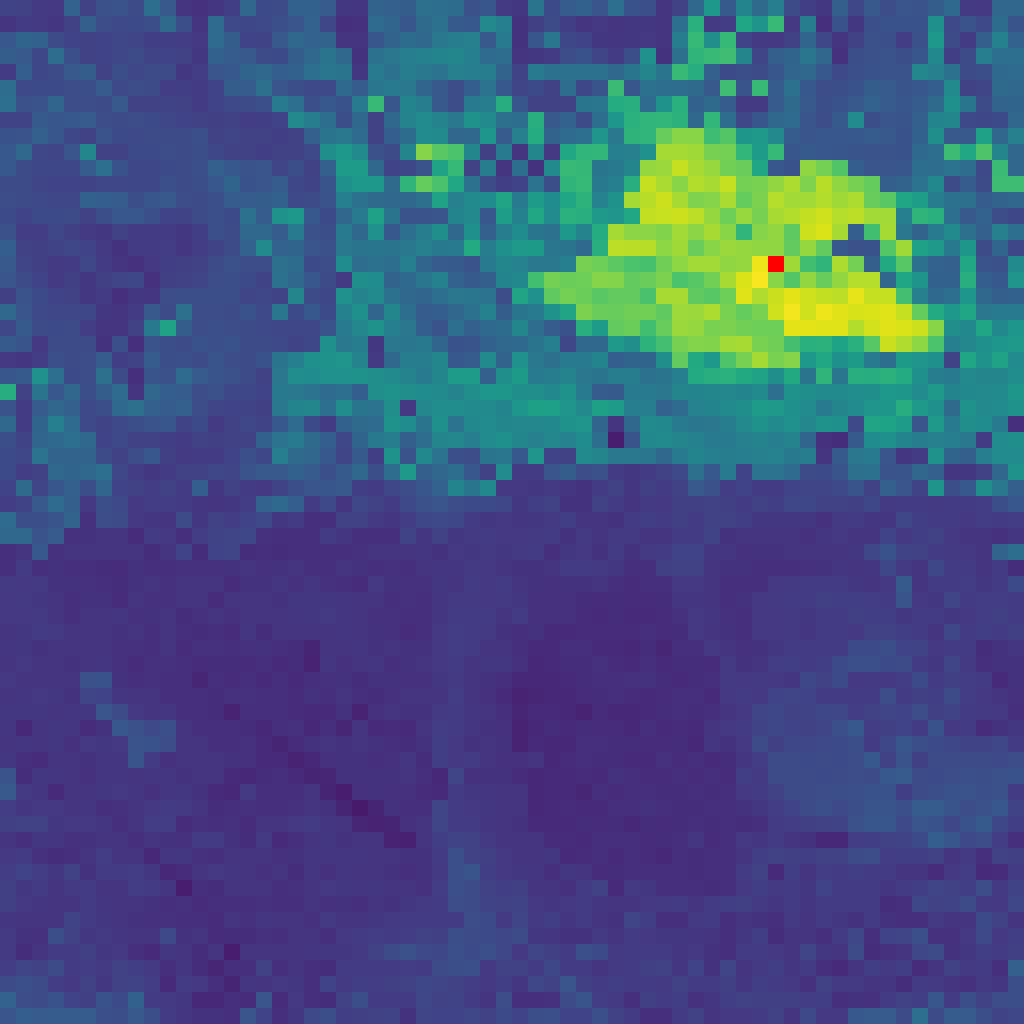} &  
    \includegraphics[width=0.19\textwidth]{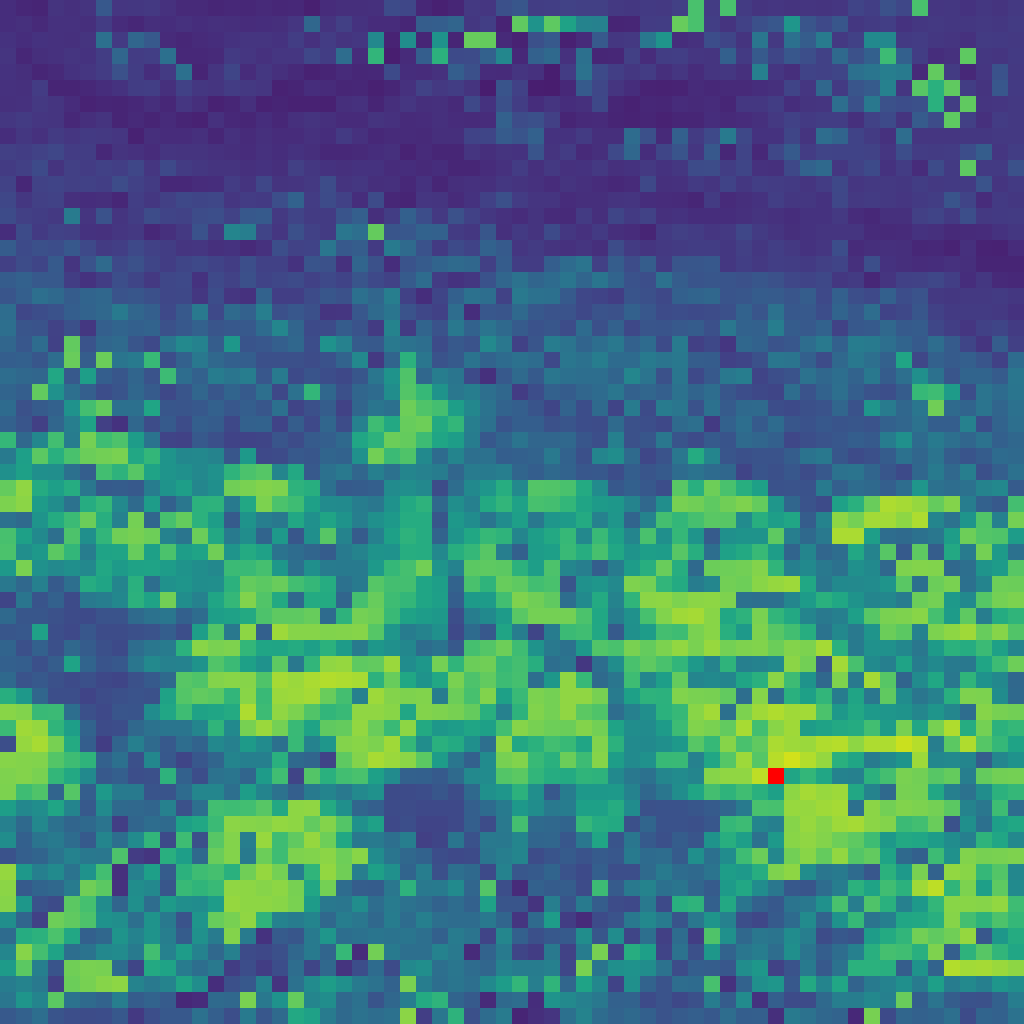} & 
    \includegraphics[width=0.19\textwidth]{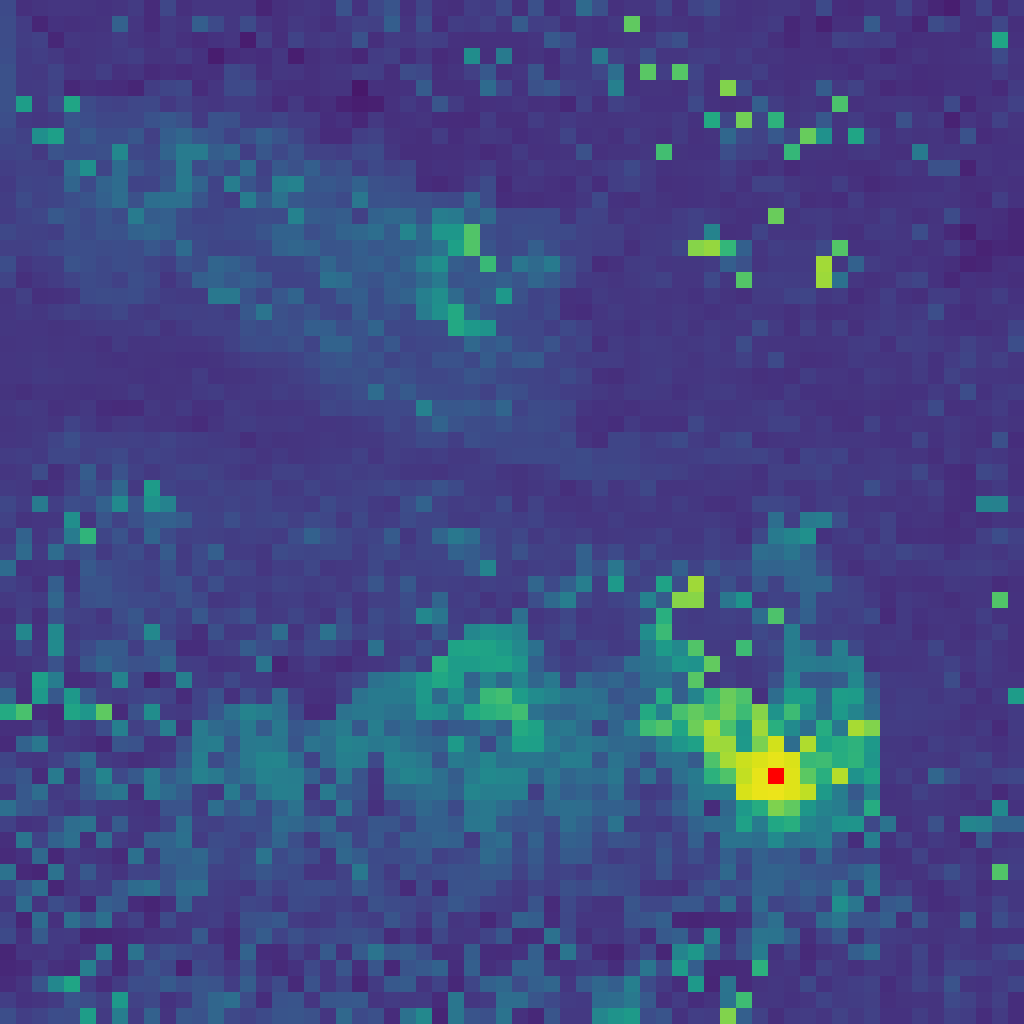}
    \\
    \raisebox{40pt}{\rotatebox[origin=c]{90}{w/ $\mathcal{L}_\mathrm{HRef}$}} & 
    \includegraphics[width=0.19\textwidth]{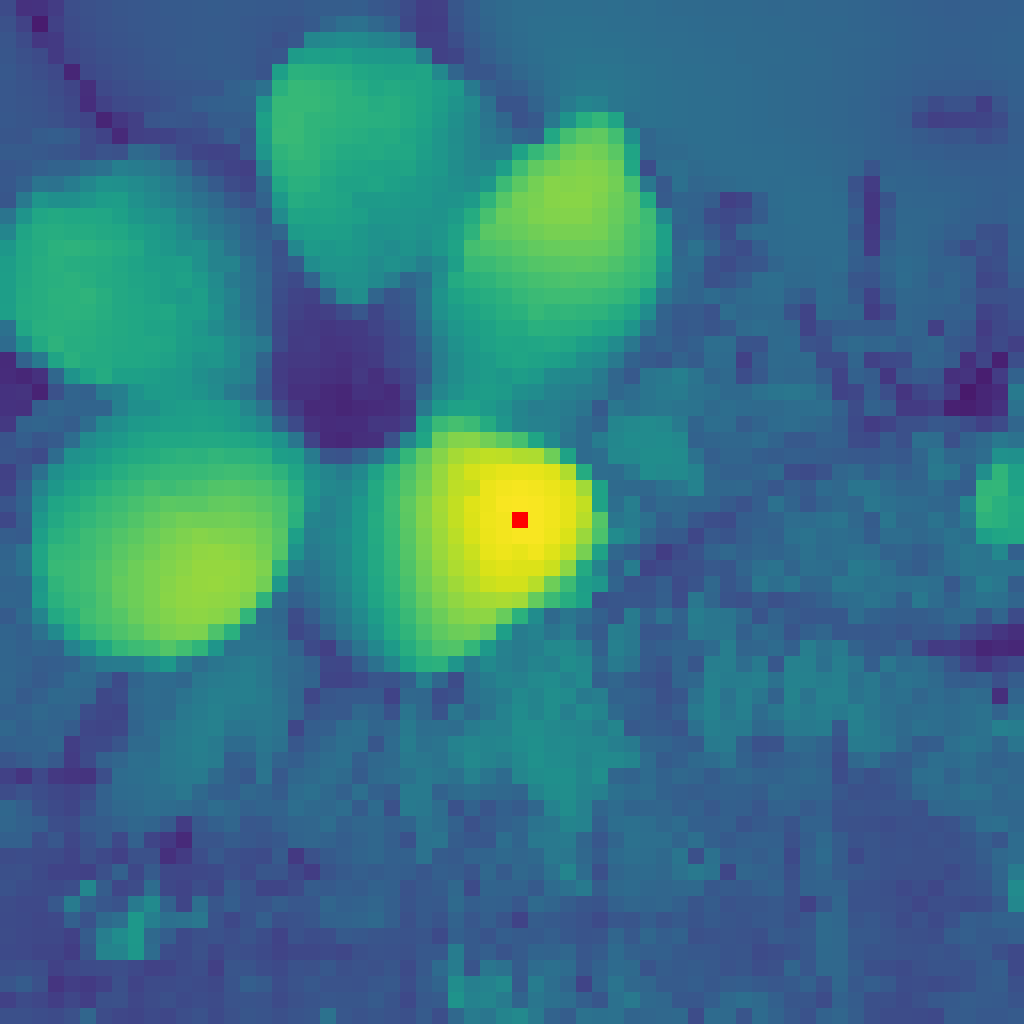} &
    \includegraphics[width=0.19\textwidth]{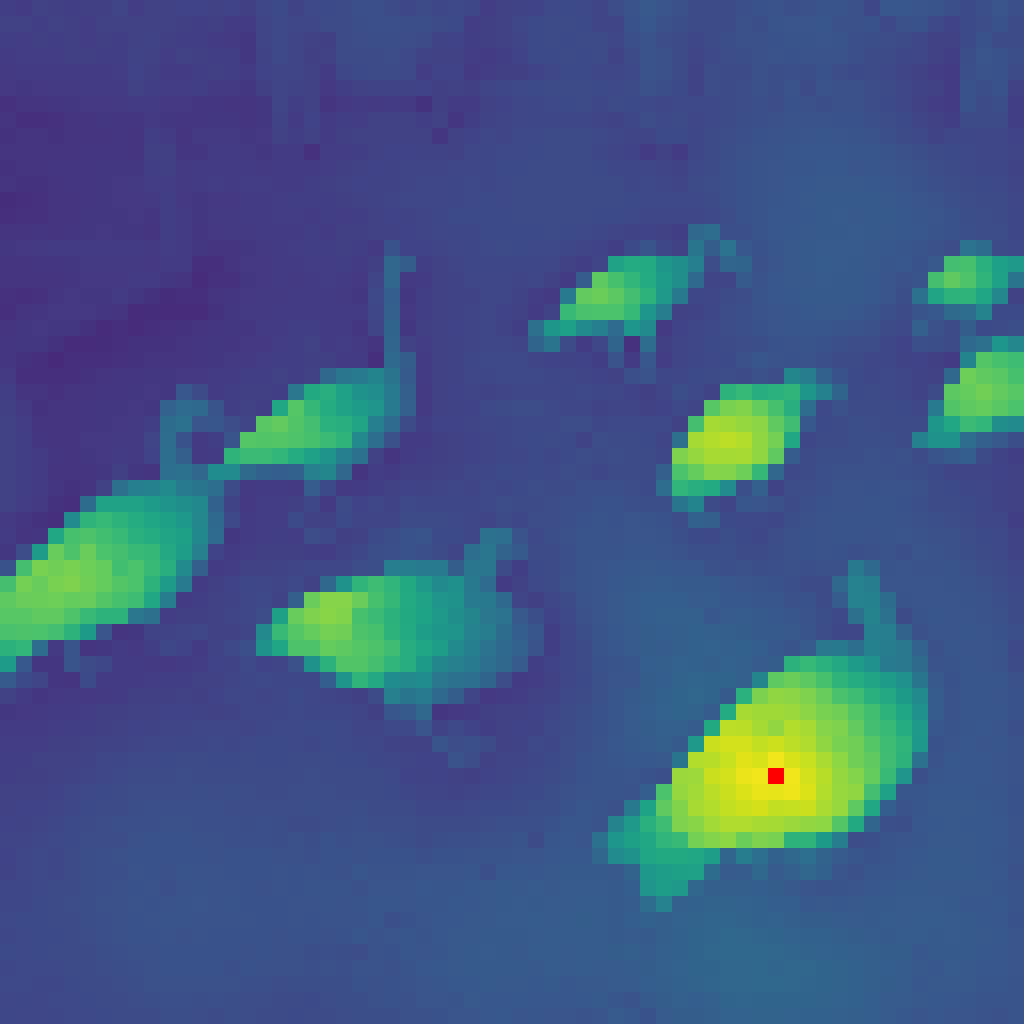} & 
    \includegraphics[width=0.19\textwidth]{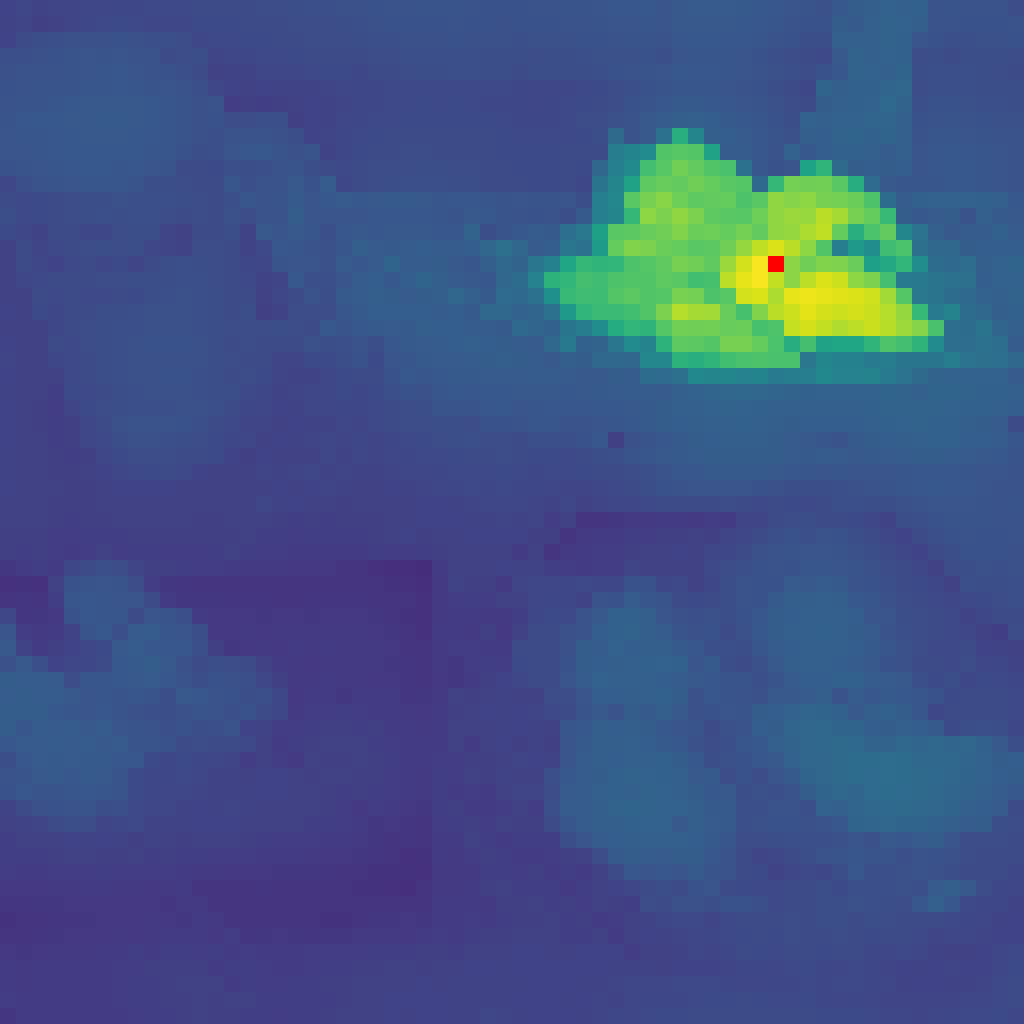} &  
    \includegraphics[width=0.19\textwidth]{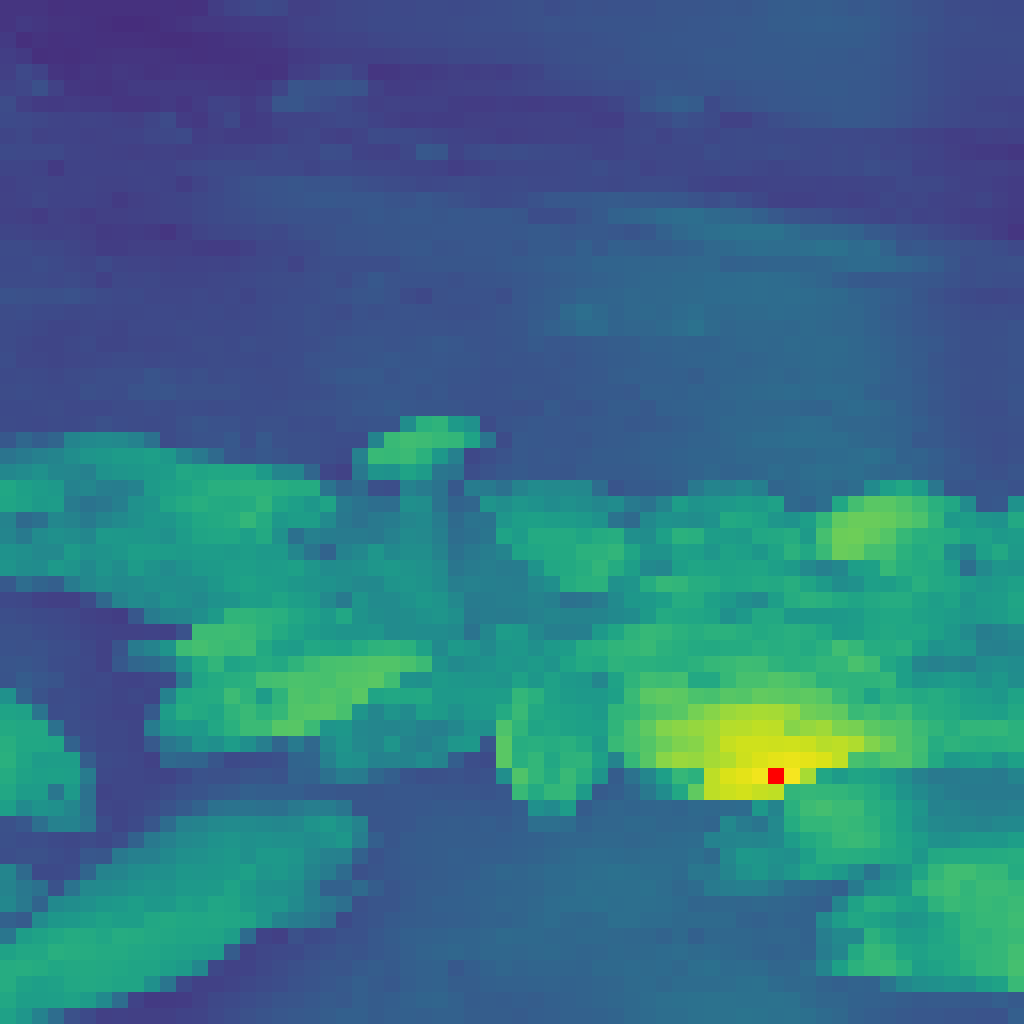} &   
    \includegraphics[width=0.19\textwidth]{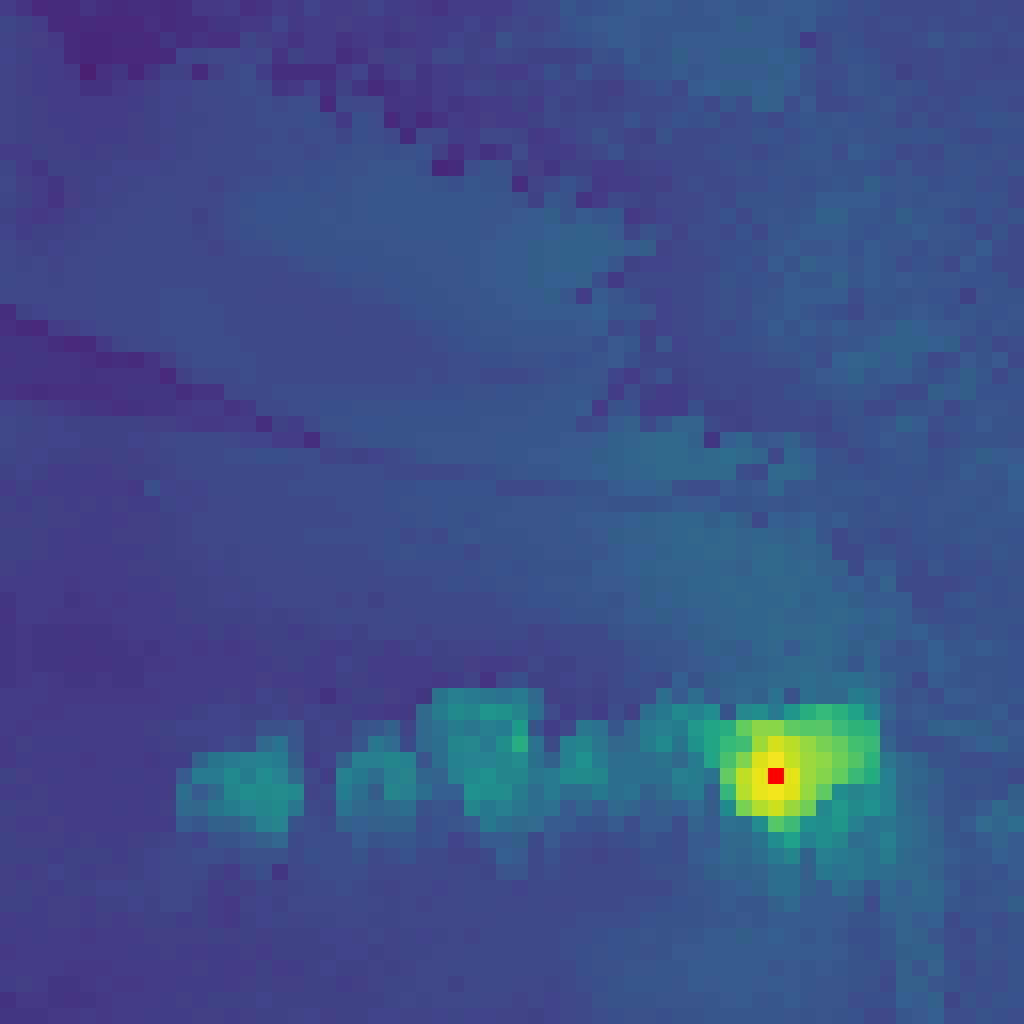}
    \end{tabular}
    \caption{Qualitative effect of \gramname. We visualize cosine maps before and after using the refinement objective $\mathcal{L}_\mathrm{HRef}$. The input resolution of the images is $1024 \times 1024$ pixels.}
    \label{fig:gram-matrices-comp}
\end{figure}

Finally, we qualitatively illustrate the effect of Gram anchoring to patch-level consistency in \cref{fig:gram-matrices-comp} which visualizes the Gram matrices patch features obtained with the initial training and high-resolution \gramname refinement.
We observe great improvements in feature correlations that our high-resolution refinement procedure brings about.

\section{Post-Training}

This section presents \emph{post-training} stages. 
This includes a high-resolution adaptation phase enabling effective inference at different input resolutions (\cref{sec:algo:hrft}), model distillation producing quality and efficient smaller-sized models (\cref{sec:distillation}), and text alignment adding zero-shot capabilities to DINOv3 (\cref{sec:alignementtxt}).

\subsection{Resolution Scaling}
\label{sec:algo:hrft}

We train our model at a relatively small resolution of $256$, which gives us a good trade-off between speed and effectiveness. 
For a patch size of $16$, this setup leads to the same input sequence length as DINOv2, which was trained with resolution $224$ and patch size $14$.
However, many contemporary computer vision applications require processing images at significantly higher resolutions, often $512 \times 512$ pixels or greater, to capture intricate spatial information. 
The inference image resolution is also not fixed in practice and varies depending on specific use cases.
To address this, we extend our training regime with a high-resolution adaptation step~\citep{touvron2019fixing}. 
To ensure high performance across a range of resolutions, we utilize \emph{mixed resolutions}, sampling differently-sized pairs of global and local crops per mini-batch.
Specifically, we consider global crop sizes from $\{512, 768\}$ and local crop sizes from $\{112, 168, 224, 336\}$ and train the model for 10k additional iterations.

Similar to the main training, a key component of this high-resolution adaptation phase is the addition of \gramname, using the 7B teacher as Gram teacher.
We found this component to be essential: without it, the model performance on dense prediction tasks degrades significantly. 
The \gramname encourages the model to maintain consistent and robust feature correlations across spatial locations, which is crucial when dealing with the increased complexity of high-resolution inputs. 

\begin{figure}[t]
    \centering
    \begin{subfigure}{0.245\textwidth}
        \includegraphics{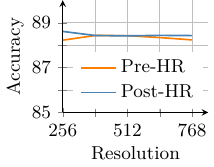}
            \caption{IN1k linear}
        \label{fig:exp:hrft-resolution-in}
    \end{subfigure}%
    \hfill
    \begin{subfigure}{0.245\textwidth}
        \includegraphics{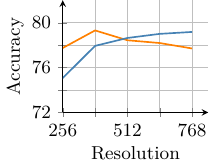}
        \caption{ObjectNet}
        \label{fig:exp:hrft-resolution-objectnet}
    \end{subfigure}%
    \hfill
    \begin{subfigure}{0.245\textwidth}
        \includegraphics{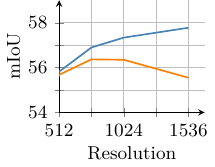}
        \caption{ADE20k}
        \label{fig:exp:hrft-resolution-ade20k}
    \end{subfigure}%
    \hfill
    \begin{subfigure}{0.245\textwidth}
        \includegraphics{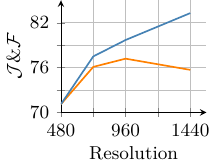}
        \caption{DAVIS}
        \label{fig:exp:hrft-resolution-davis}
    \end{subfigure}
    \caption{Effect of high resolution adaptation. 
    Results before (`Pre-HR') and after (`Post-HR') resolution scaling (\cref{sec:algo:hrft}) on (a) linear classification on ImageNet, (b) applied OOD to ObjectNet, (c) linear semantic segmentation on ADE20k, and (d) segmentation tracking on DAVIS at different evaluation resolutions.}
    \label{fig:exp:hrft-resolution}
\end{figure}

Empirically, we observe that this relatively brief but targeted high-resolution step substantially enhances the overall model's quality and allows it to generalize across a wide range of input sizes, as shown visually in \cref{fig:visualization-extreme-resolutions}. 
In \cref{fig:exp:hrft-resolution}, we compare our 7B model before and after adaptation. 
We find that resolution scaling leads to a small gain on ImageNet classification (a) with relatively stable performance w.r.t.~resolution. 
However, in ObjectNet OOD transfer (b), we observe that the performance tends to degrade slightly for lower resolutions, while improving for higher resolutions. 
This is largely compensated by the improvement in the quality of local features at high resolution, shown by the positive trend in segmentation on ADE20k (c) and tracking on DAVIS (d).
Adaptation leads to local features that \emph{improve with image size}, leveraging the richer spatial information available at larger resolutions and effectively enabling high-resolution inference.
Interestingly, the adapted model supports resolutions way beyond the maximum training resolution of 768---we visually observe stable feature maps at resolutions above 4k (\cf \cref{fig:visualization-extreme-resolutions}).

\definecolor{preHRcolor}{named}{orange}
\definecolor{postHRcolor}{named}{dinocolor}

\subsection{Model Distillation}
\label{sec:distillation}
\paragraph{A Family of Models for Multiple Use-Cases}
We perform knowledge distillation of the ViT-7B model into smaller Vision Transformer variants (ViT-S, ViT-B, and ViT-L), which are highly valued by the community for their improved manageability and efficiency. 
Our distillation approach uses the same training objective as in the first training phase, ensuring consistency in learning signals. 
However, instead of relying on an exponential moving average (EMA) of model weights, we use the 7B model directly as the teacher to guide the smaller student models. In this case, the teacher model is fixed. 
We do not observe patch-level consistency issues and therefore do not apply the \gramname technique.
This strategy enables the distilled models to inherit the rich representational power of the large teacher while being more practical for deployment and experimentation.

Our ViT-7B model is distilled into a series of ViT models with sizes covering a broad range of compute budgets, and allowing proper comparison with concurrent models. They include the standard ViT-S (21M params), B (86M), L (0.3B), along with a 
custom ViT-S+ (29M) and a
custom ViT-H+ (0.8B) model
to close the performance gap with the self-distilled 7B teacher model. Indeed, we observe in DINOv2 that smaller student models can reach a performance on par with their teacher as the distillation.
As a result, the distilled models deliver frontier-level performance for a fraction of the inference compute as we see in \cref{tab:family:distillation}. 
We train the models for 1M iterations then perform 250k iterations of learning-rate cooldown following a cosine schedule before applying the high-resolution phase described in \cref{sec:algo:hrft} above without \gramname.

\begin{figure}[t]
    \centering
    \includegraphics{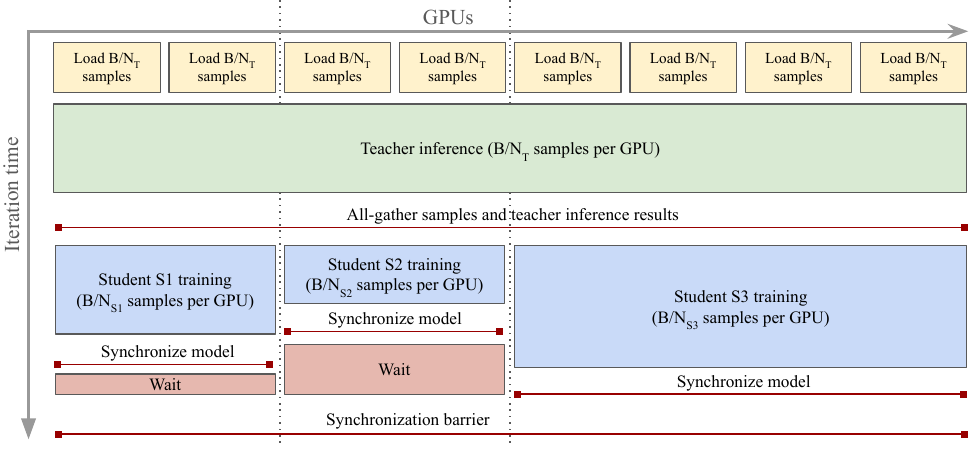}
    \caption{
        Multi-student distillation procedure. 
        In this diagram, we distill 3 students in parallel: we first share teacher inference across all T nodes to save compute, and gather inputs and results on all GPUs. 
        Then, smaller groups perform student training. We adjust the size of these groups such that the training step has the same duration across all students $Si$, minimizing idle time waiting at the synchronization barrier.
    }
    \label{fig:distillation_diagram}
\end{figure}

\paragraph{Efficient Multi-Student Distillation}
As the inference cost for a large teacher can be orders of magnitude higher than for students (see \cref{tab:family-models-flops}), we design a parallel distillation pipeline that allows training multiple students at the same time and sharing the teacher inference across all nodes involved in the training (see \cref{fig:distillation_diagram} for a diagram). Let $C_T$ and $C_S$ be respectively the cost of running the teacher inference and the student training on a single sample, in single-teacher/single-student distillation with batch-size $B$ where each of the $N$ GPUs processes a $B / N$ slice of the data, the teacher inference costs $B/N \times C_T$ and the student training costs $B/N \times C_S$ per GPU. In multi-student distillation, we proceed as follows. Each student $Si$ is assigned a set of $N_{Si}$ GPUs for training, and all $N_T = \sum{N_{Si}}$ GPUs are part of the global inference group. At each iteration, we first run the teacher inference on the global group for a $B/N_T \times C_T$ compute cost per GPU. We then run an \textit{all-gather} collective operation to share the input data and inference result with all compute nodes. Finally, each student group separately performs student training for a $B/N_{Si} \times C_{Si}$ cost.

The above calculations shows that adding an additional student to the distillation pipeline will (1) reduce the per-GPU compute at each iteration, thus globally improving distillation speed, and (2) increase the overall compute only by the training cost of the new student, since the total teacher inference cost is now fixed. The implementation only requires setting up GPU process groups carefully, adapting data-loaders and teacher inference to ensure inputs and outputs are synchronized across groups using NCCL collectives. As the groups are synchronized at each iteration, in order to maximize speed, we adapt the number of GPUs for each student such that their iteration times are roughly the same.
With this procedure, we seamlessly train multiple students, and produce a whole family of distilled models from our flagship 7B model. 

\subsection{Aligning DINOv3 with Text}
\label{sec:alignementtxt}
Open-vocabulary image-text alignment has received significant interest and enthusiasm from the research community, thanks to its potential to enable flexible and scalable multimodal understanding. 
A large body of work has focused on improving the quality of CLIP \citep{radford2021learning}, which originally learned only a global alignment between image and text representations. 
While CLIP has demonstrated impressive zero-shot capabilities, its focus on global features limits its ability to capture fine-grained, localized correspondences. 
More recent works~\citep{zhai2022lit} have shown that effective image-text alignment can be achieved with pre-trained self-supervised visual backbones.  
This makes it possible to leverage these powerful models in multi-modal settings, facilitating richer and more precise text-to-image associations that extend beyond global semantics while also reducing computational costs, since the visual encoding is already learned.

We align a text encoder with our DINOv3 model by adopting the training strategy previously proposed in \citet{jose2025dinov2}.  
This approach follows the LiT training paradigm~\citep{zhai2022lit}, training a text representation from scratch to match images to their captions with a contrastive objective, while keeping the vision encoder frozen. 
To allow for some flexibility on the vision side, two transformer layers are introduced on top of the frozen visual backbone. 
A key enhancement of this method is the concatenation of the mean-pooled patch embeddings with the output CLS token before matching to the text embeddings.
This enables aligning both global and local visual features to text, leading to improved performance on dense prediction tasks without requiring additional heuristics or tricks. 
Furthermore, we use to the same data curation protocol as established in \citet{jose2025dinov2} to ensure consistency and comparability.

\section{Results}
\label{sec:results}

In this section, we evaluate our flagship DINOv3 7B model on a variety of computer vision tasks.
Throughout our experiments, unless otherwise specified, \emph{we keep DINOv3 frozen} and solely use its representations.
We demonstrate that with DINOv3, finetuning is not necessary to obtain strong performance.
This section is organized as follows. We first probe the quality of DINOv3's dense (\cref{sec:results-dense-features}) and global (\cref{sec:results-global-features}) image representations using lightweight evaluation protocols and compare it to the strongest available vision encoders. 
We show that DINOv3 learns exceptional dense features while offering robust and versatile global image representations.
Then, we consider DINOv3 as a basis for developing more complex computer vision systems (\cref{sec:results-system-level}).
We show with little effort on top of DINOv3, we are able to achieve results competitive with or exceeding the state of the art in tasks as diverse as object detection, semantic segmentation, 3D view estimation, or relative monocular depth estimation.

\subsection{DINOv3 provides Exceptional Dense Features}
\label{sec:results-dense-features}
We first investigate the raw quality of DINOv3's dense representations using a diverse set of lightweight evaluations.
In all cases, we utilize the frozen patch features of the last layer, and evaluate them using 
(1) qualitative visualizations (\cref{sec:results-qualitative}), 
(2) dense linear probing (\cref{sec:results-dense-linear-probing}: segmentation, depth estimation),
(3) non-parametric approaches (\cref{sec:results-correspondence-estimation}: 3D correspondence estimation, \cref{sec:results-object-discovery}: object discovery, \cref{sec:results-tracking}: tracking), 
and (4) lightweight attentive probing (\cref{sec:results-video-classification}: video classification).

\paragraph{Baselines}
We compare the dense features of DINOv3 with those of the strongest publicly available image encoders, both weakly- and self-supervised ones. 
We consider the weakly-supervised encoders Perception Encoder (PE) Core~\citep{bolya2025perception} and SigLIP 2~\citep{tschannen2025siglip}, which use CLIP-style image-text contrastive learning.
We also compare to the strongest self-supervised methods: DINOv3's predecessor DINOv2~\citep{oquab2024dinov2} with registers~\citep{darcet2024vision}, Web-DINO~\citep{fan2025scaling}, a recent scaling effort of DINO, and Franca~\citep{venkataramanan2025franca}, the best open-data SSL model.   %
Finally, we compare to the agglomerative models AM-RADIOv2.5~\citep{heinrich2025radiov25}, distilled from DINOv2, CLIP~\citep{radford2021learning}, DFN~\citep{fang2024data}, and Segment Anything (SAM)~\citep{kirillov2023segment}, and to PEspatial, distilling SAM 2~\citep{ravi2025sam} into PEcore.
For each baseline, we report the performance of the strongest model available and specify the architecture in the tables.

\subsubsection{Qualitative Analysis}
\label{sec:results-qualitative}
\begin{figure}[t]
    \centering
    \begin{subfigure}{0.195\textwidth}
        \centering
        \includegraphics[width=\linewidth]{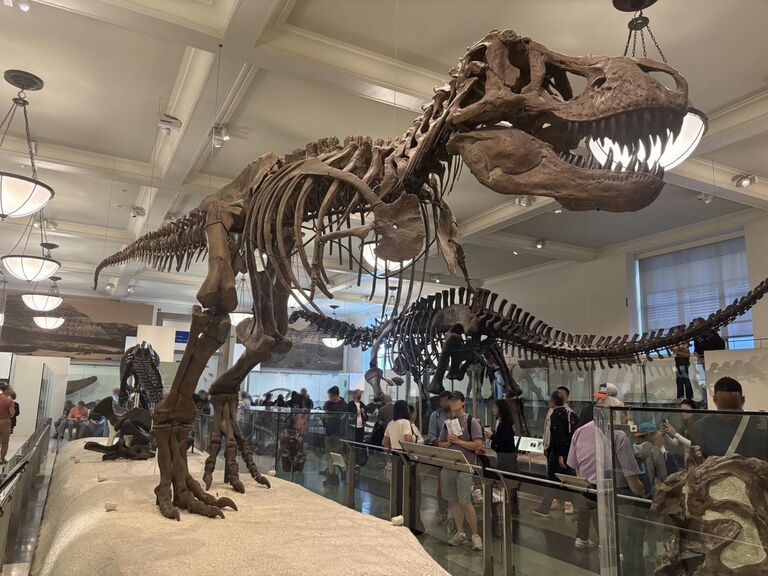}
    \end{subfigure}\hfill
    \begin{subfigure}{0.195\textwidth}
        \centering
        \includegraphics[width=\linewidth]{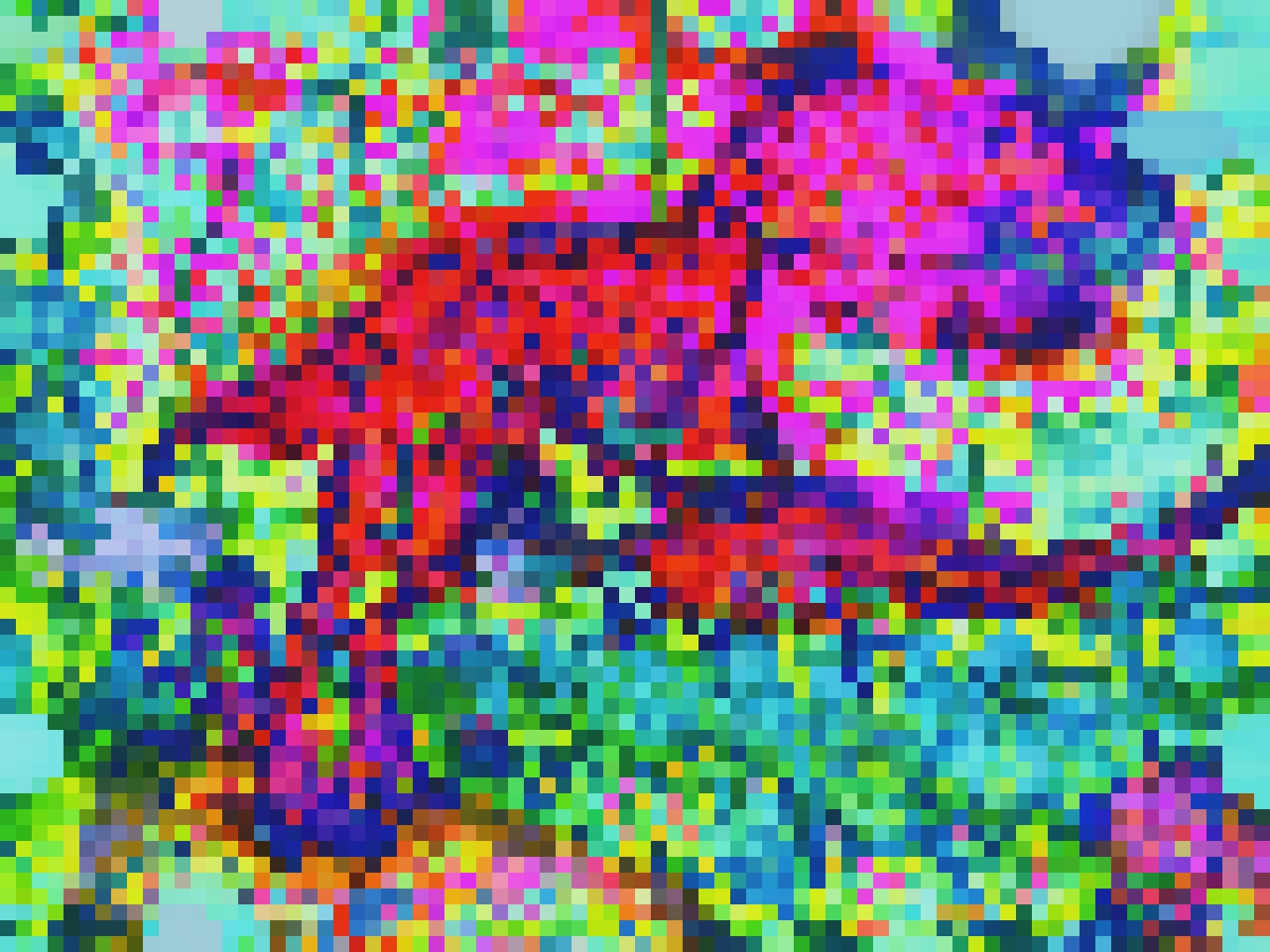}
    \end{subfigure}\hfill
    \begin{subfigure}{0.195\textwidth}
        \centering
        \includegraphics[width=\linewidth]{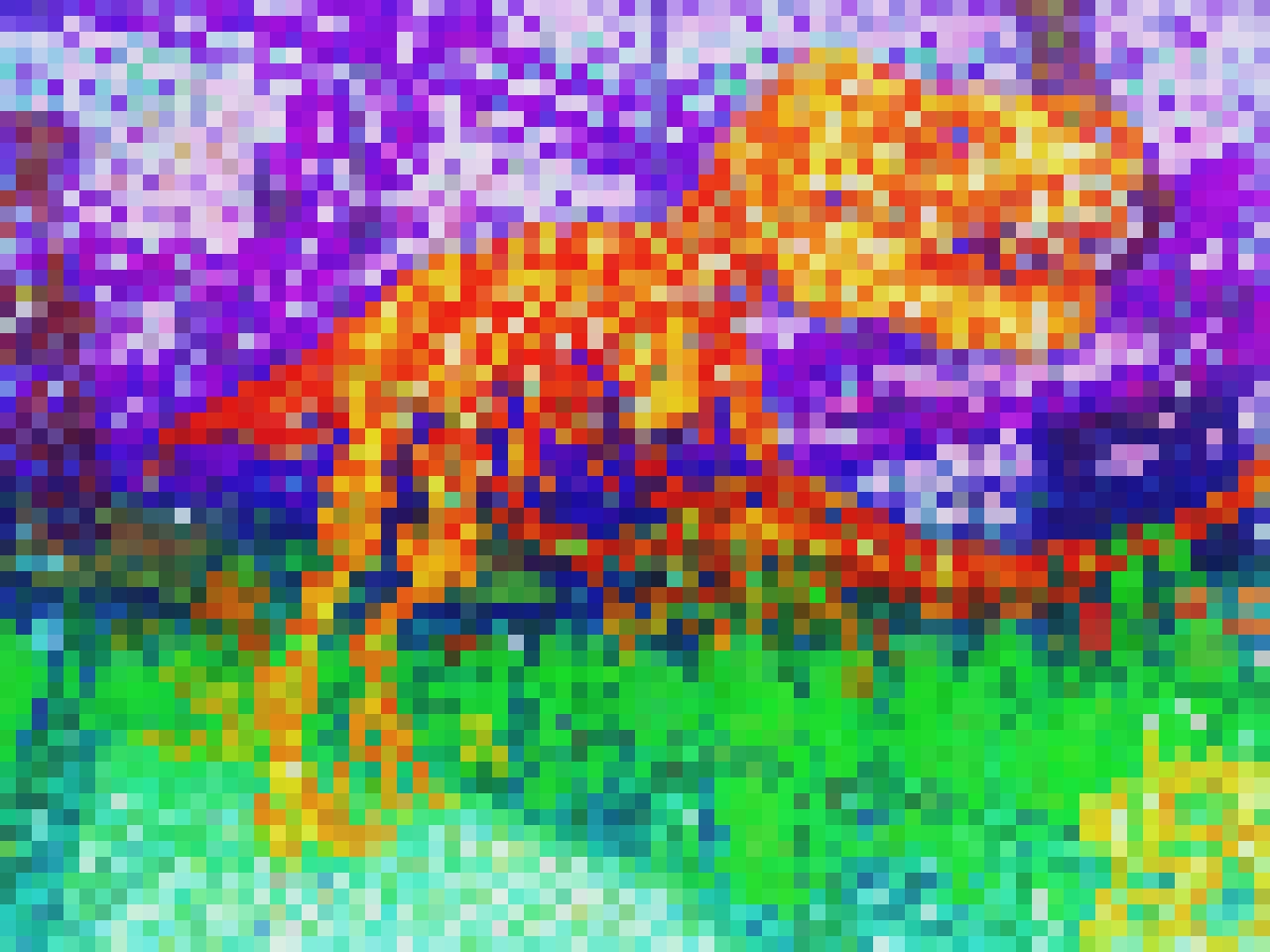}
    \end{subfigure}\hfill
    \begin{subfigure}{0.195\textwidth}
        \centering
        \includegraphics[width=\linewidth]{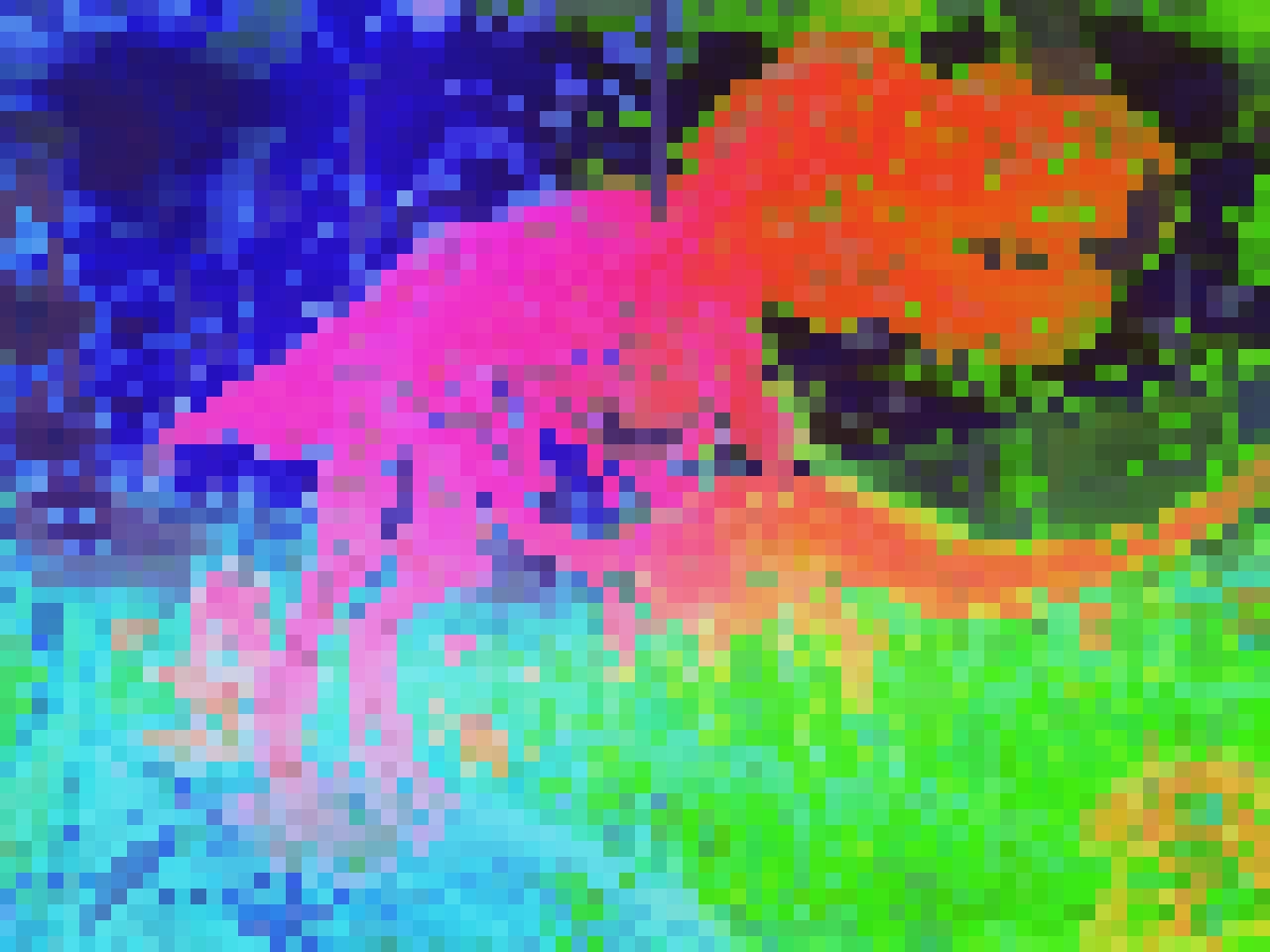}
    \end{subfigure}\hfill
    \begin{subfigure}{0.195\textwidth}
        \centering
        \includegraphics[width=\linewidth]{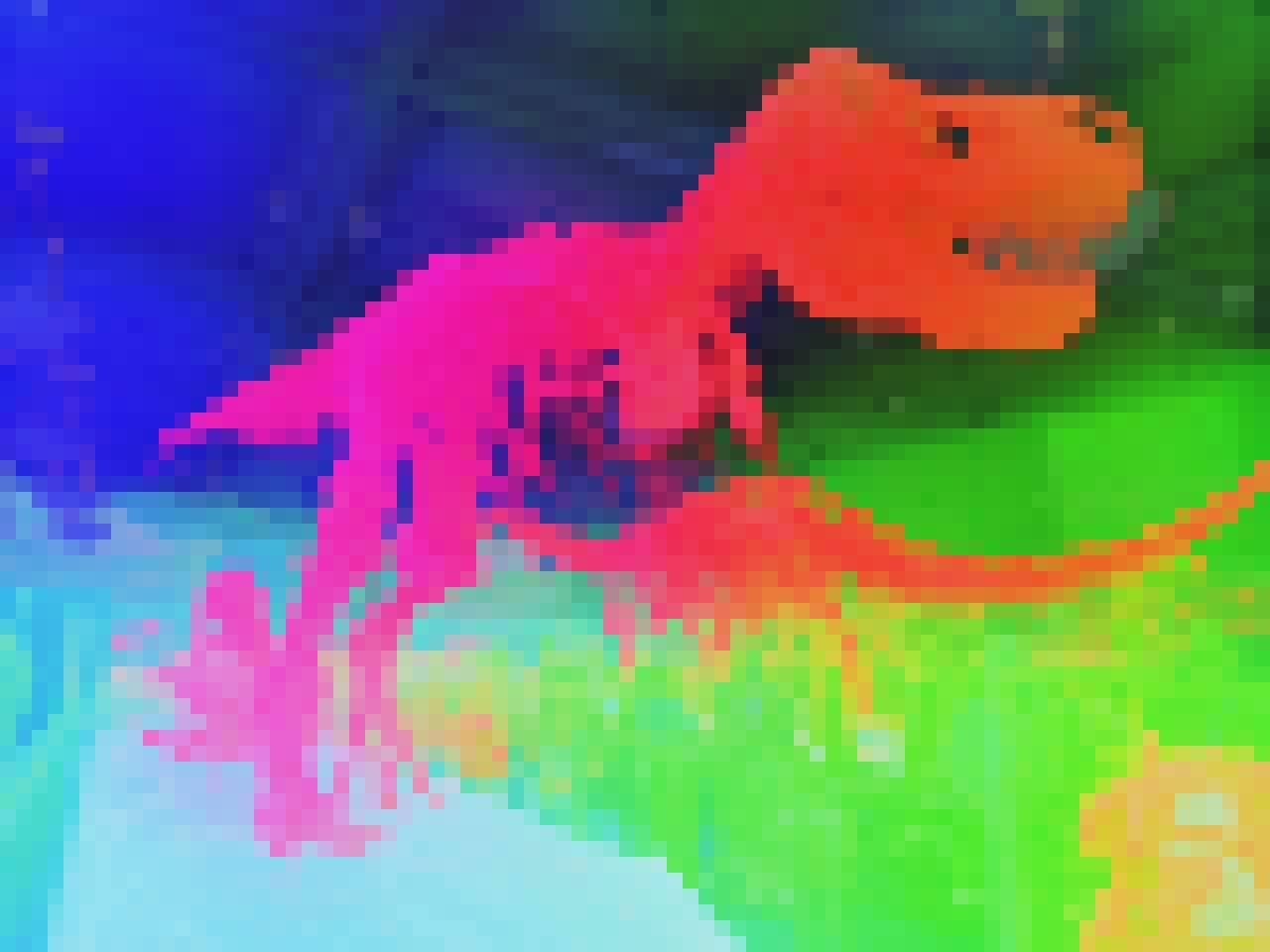}
    \end{subfigure}\vspace{0.25em}
    \begin{subfigure}{0.195\textwidth}
        \centering
        \includegraphics[width=\linewidth]{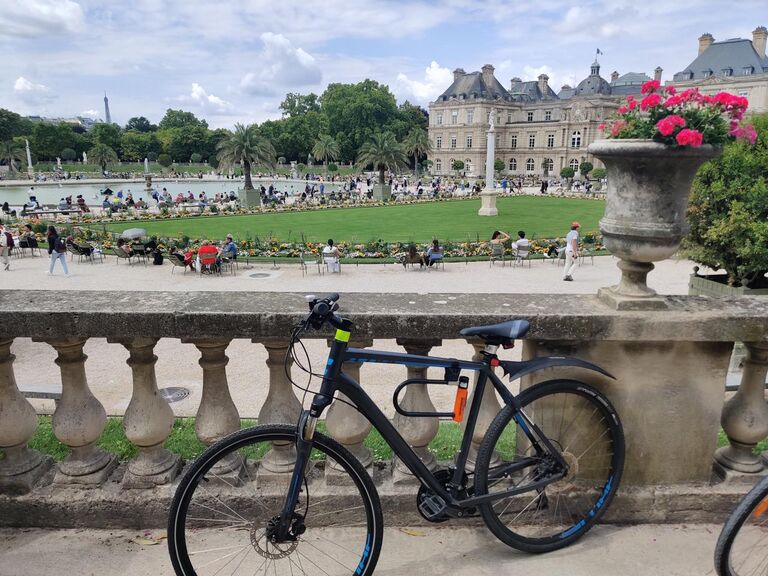}
    \end{subfigure}\hfill
    \begin{subfigure}{0.195\textwidth}
        \centering
        \includegraphics[width=\linewidth]{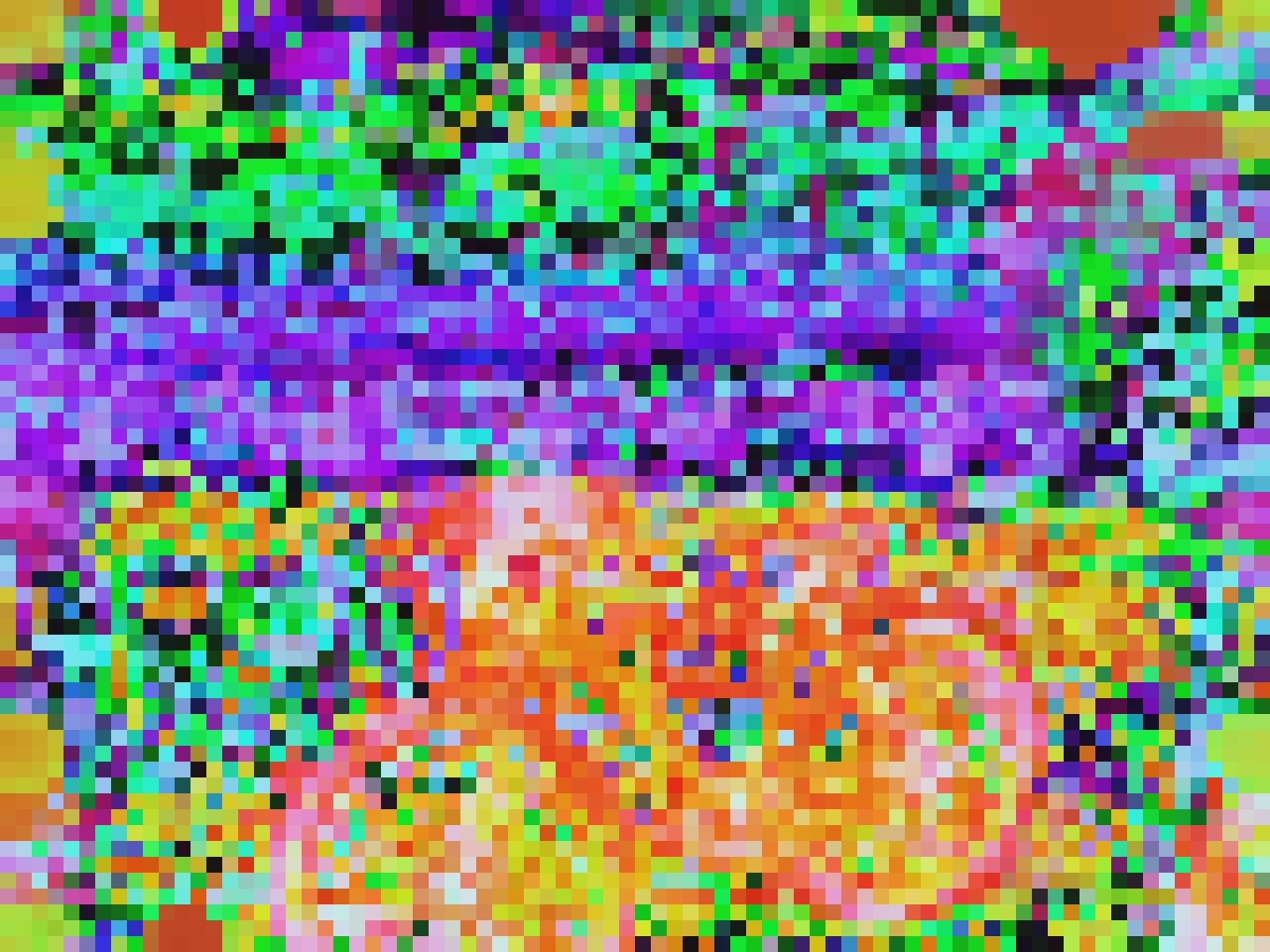}
    \end{subfigure}\hfill
    \begin{subfigure}{0.195\textwidth}
        \centering
        \includegraphics[width=\linewidth]{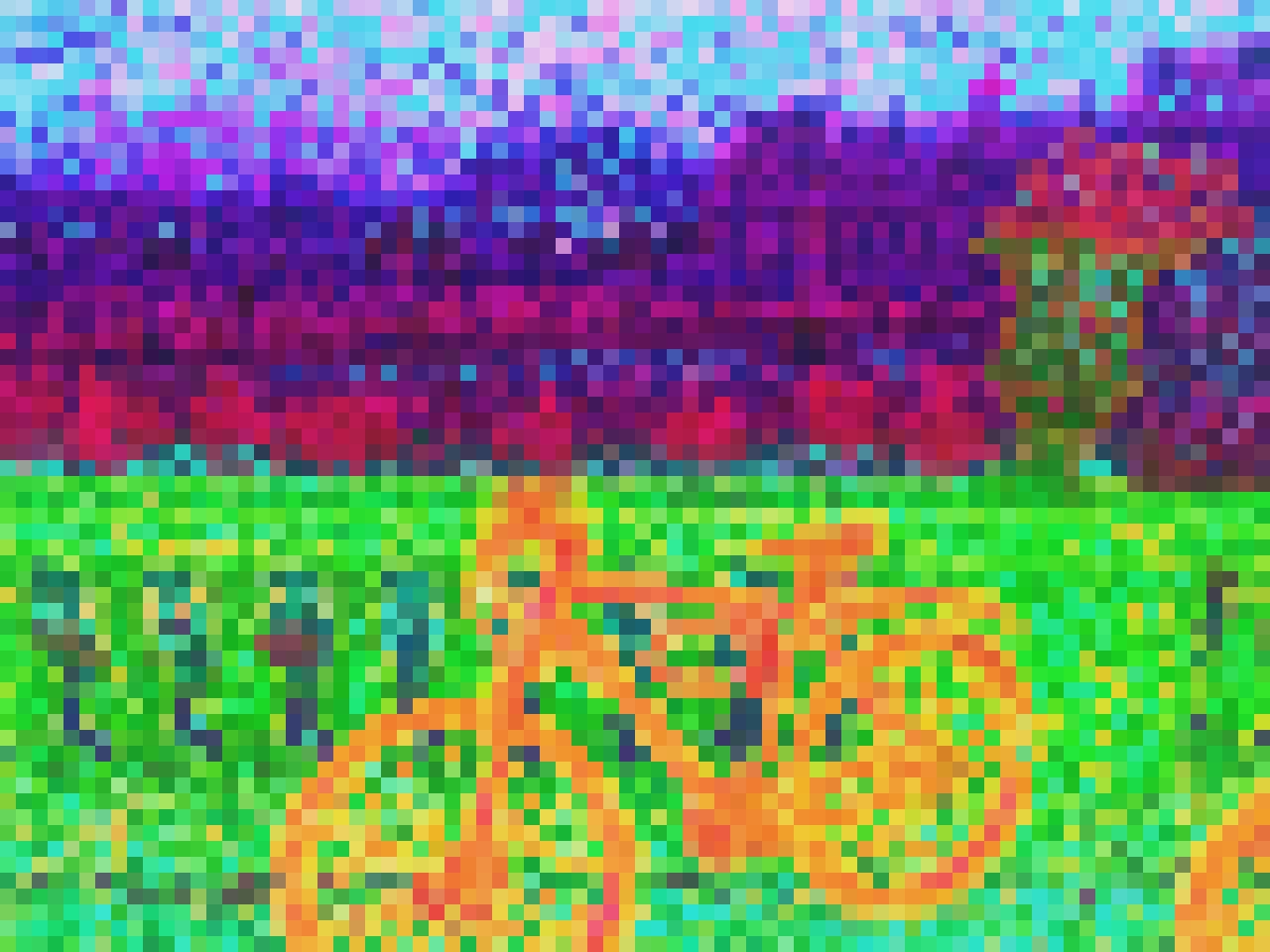}
    \end{subfigure}\hfill
    \begin{subfigure}{0.195\textwidth}
        \centering
        \includegraphics[width=\linewidth]{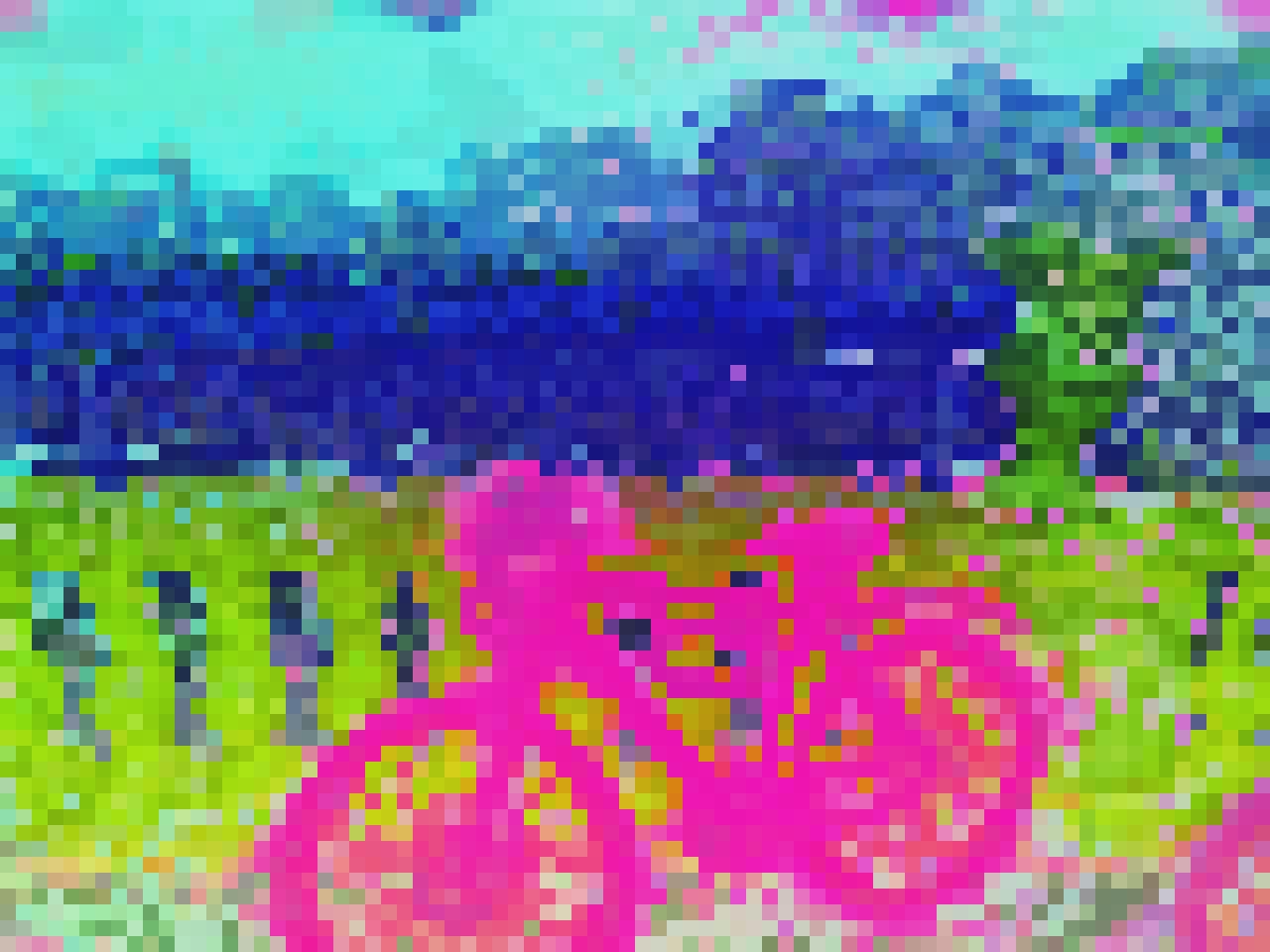}
    \end{subfigure}\hfill
    \begin{subfigure}{0.195\textwidth}
        \centering
        \includegraphics[width=\linewidth]{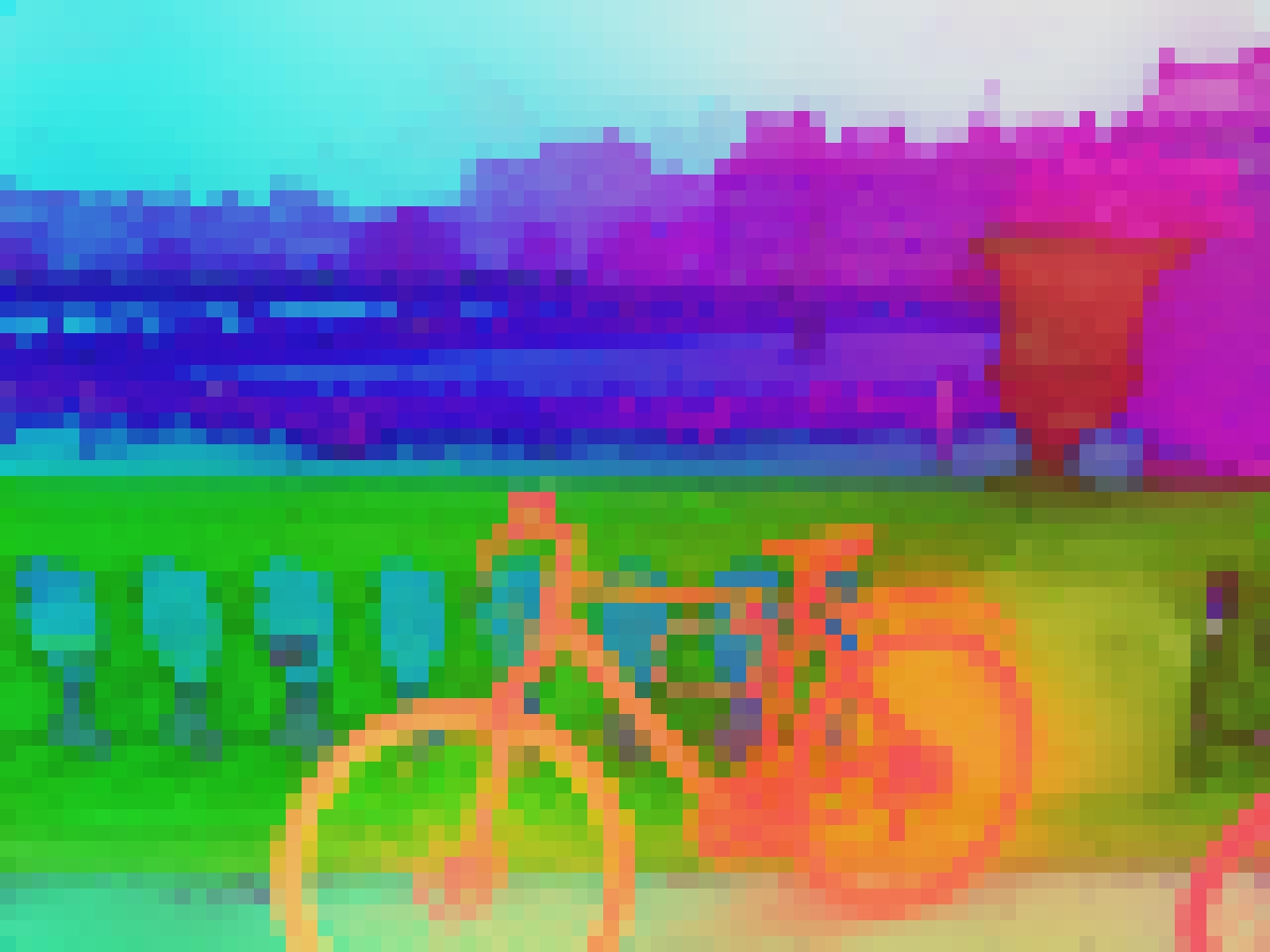}
    \end{subfigure}\vspace{0.25em}
    \begin{subfigure}{0.195\textwidth}
        \centering
        \includegraphics[width=\linewidth]{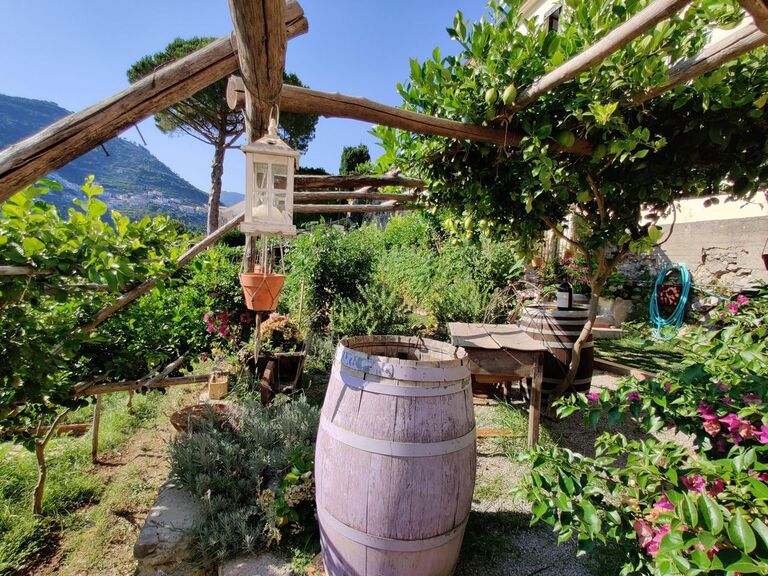}
    \end{subfigure}\hfill
    \begin{subfigure}{0.195\textwidth}
        \centering
        \includegraphics[width=\linewidth]{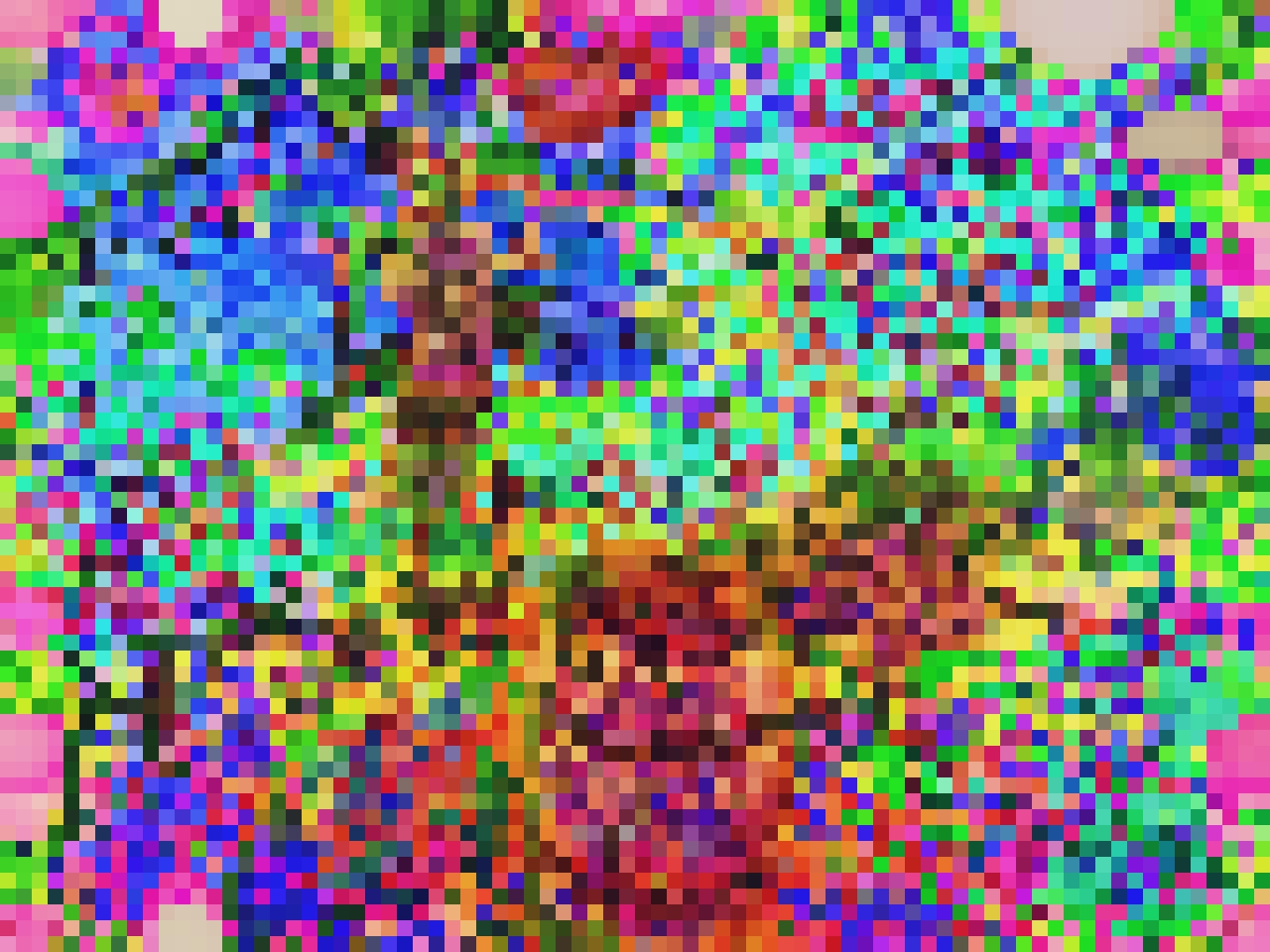}
    \end{subfigure}\hfill
    \begin{subfigure}{0.195\textwidth}
        \centering
        \includegraphics[width=\linewidth]{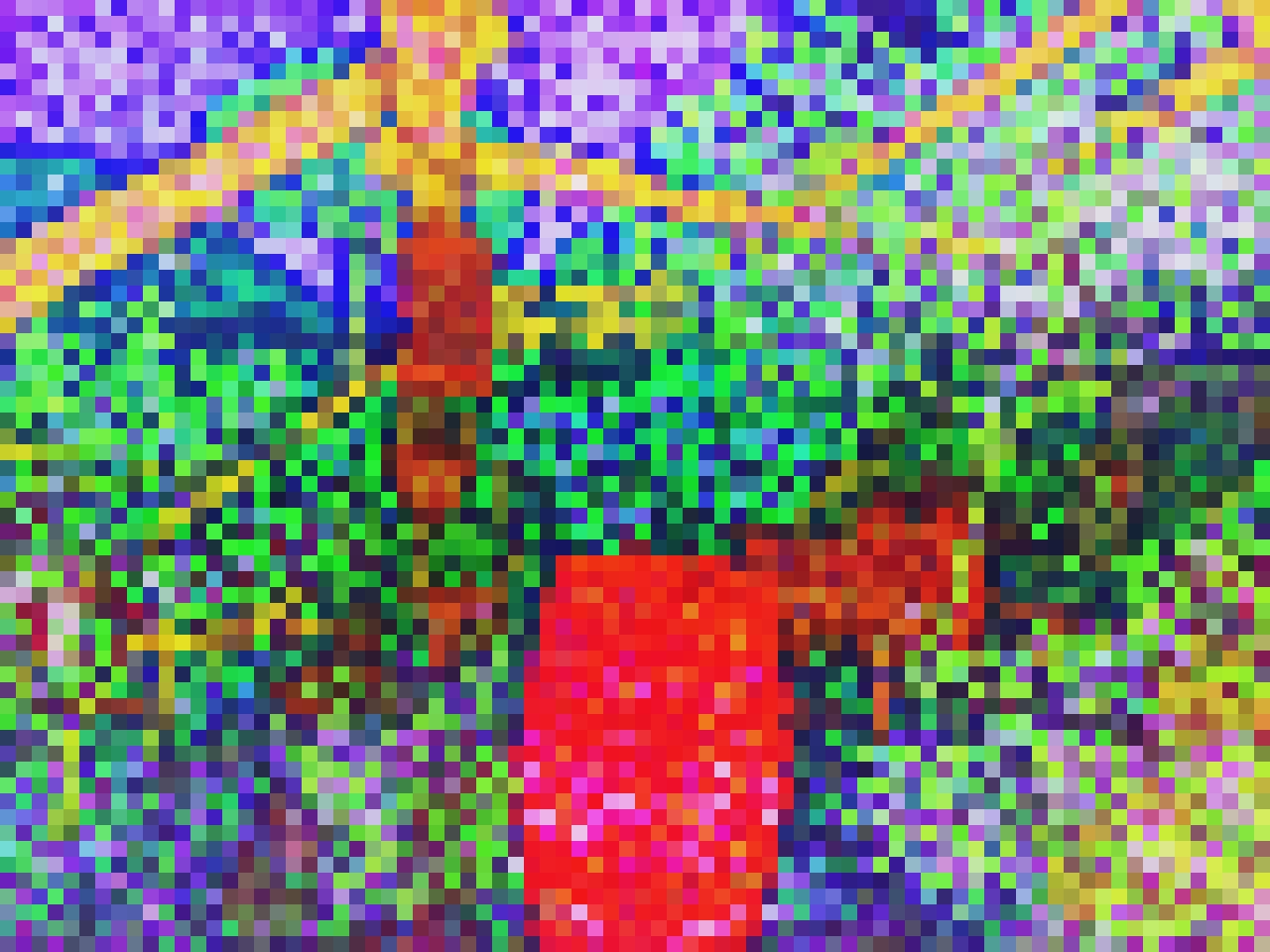}
    \end{subfigure}\hfill
    \begin{subfigure}{0.195\textwidth}
        \centering
        \includegraphics[width=\linewidth]{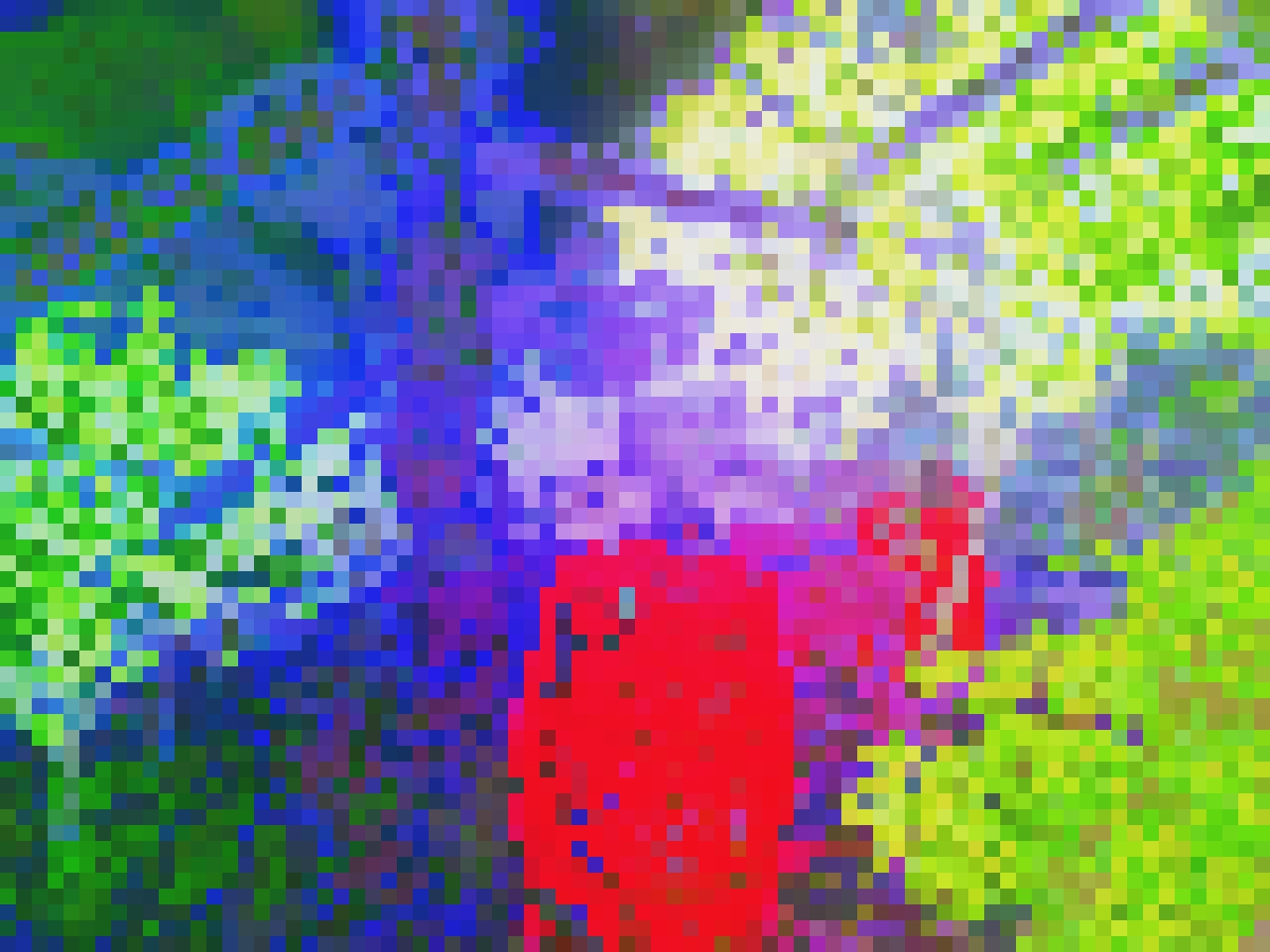}
    \end{subfigure}\hfill
    \begin{subfigure}{0.195\textwidth}
        \centering
        \includegraphics[width=\linewidth]{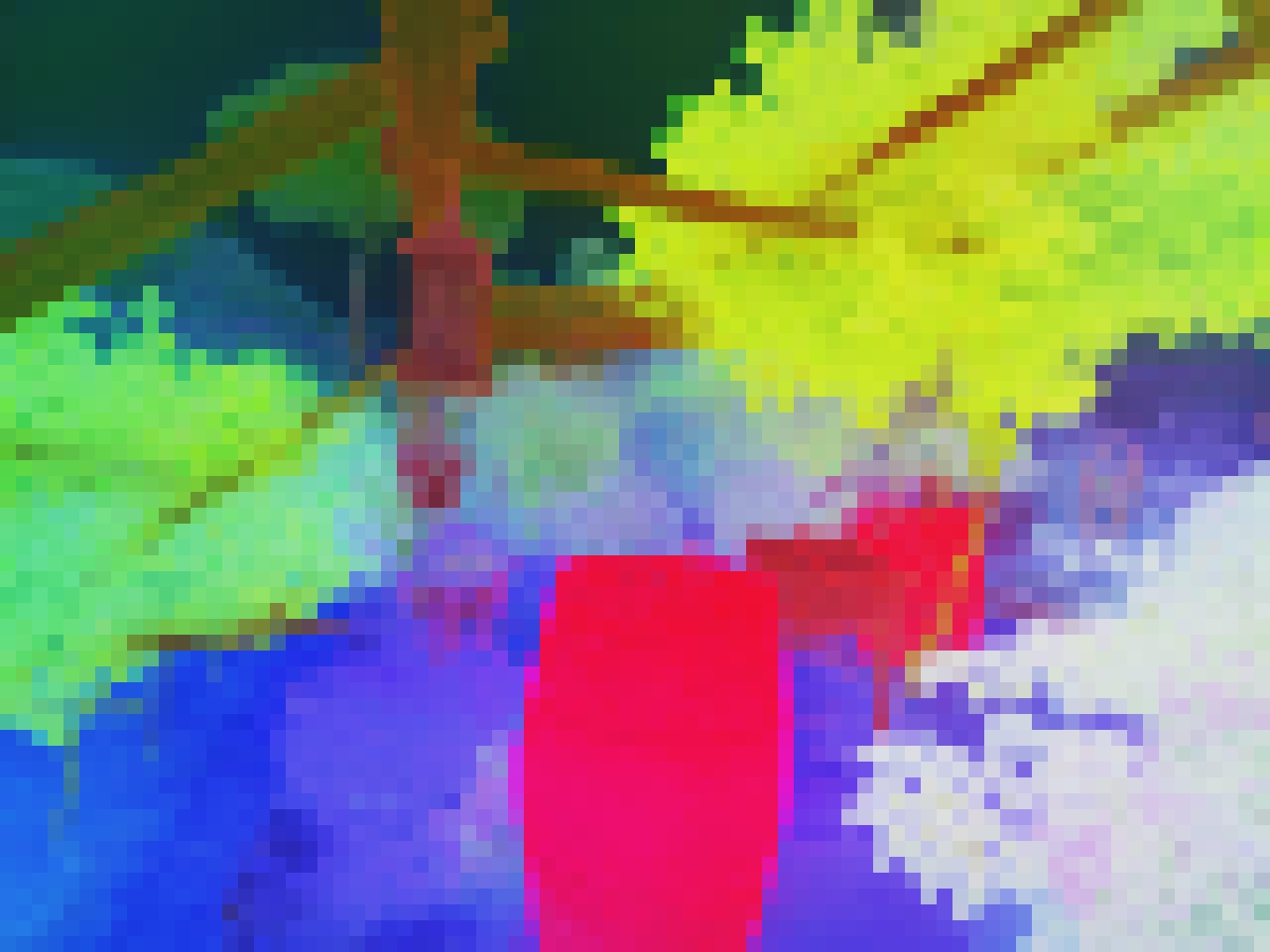}
    \end{subfigure}\vspace{0.25em}
    \begin{subfigure}{0.195\textwidth}
        \centering
        \includegraphics[width=\linewidth]{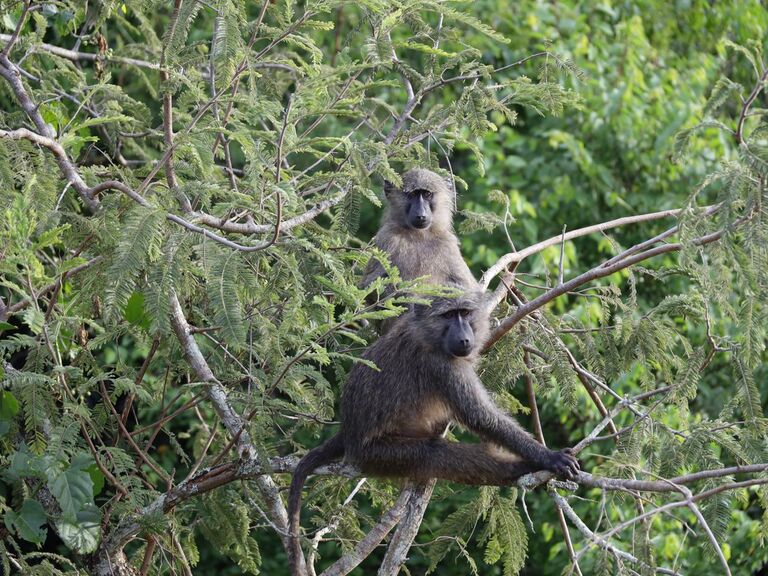}
    \end{subfigure}\hfill
    \begin{subfigure}{0.195\textwidth}
        \centering
        \includegraphics[width=\linewidth]{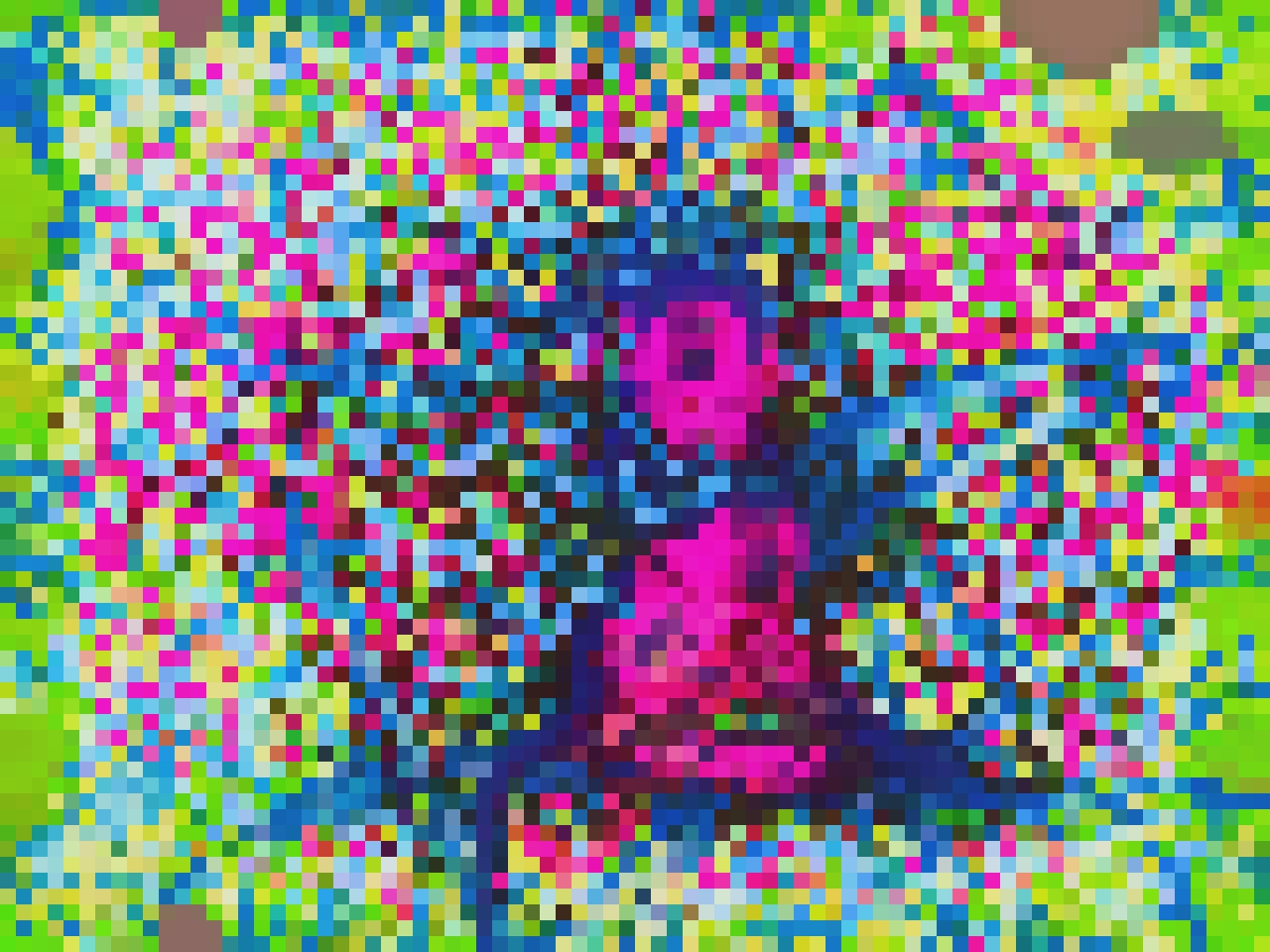}
    \end{subfigure}\hfill
    \begin{subfigure}{0.195\textwidth}
        \centering
        \includegraphics[width=\linewidth]{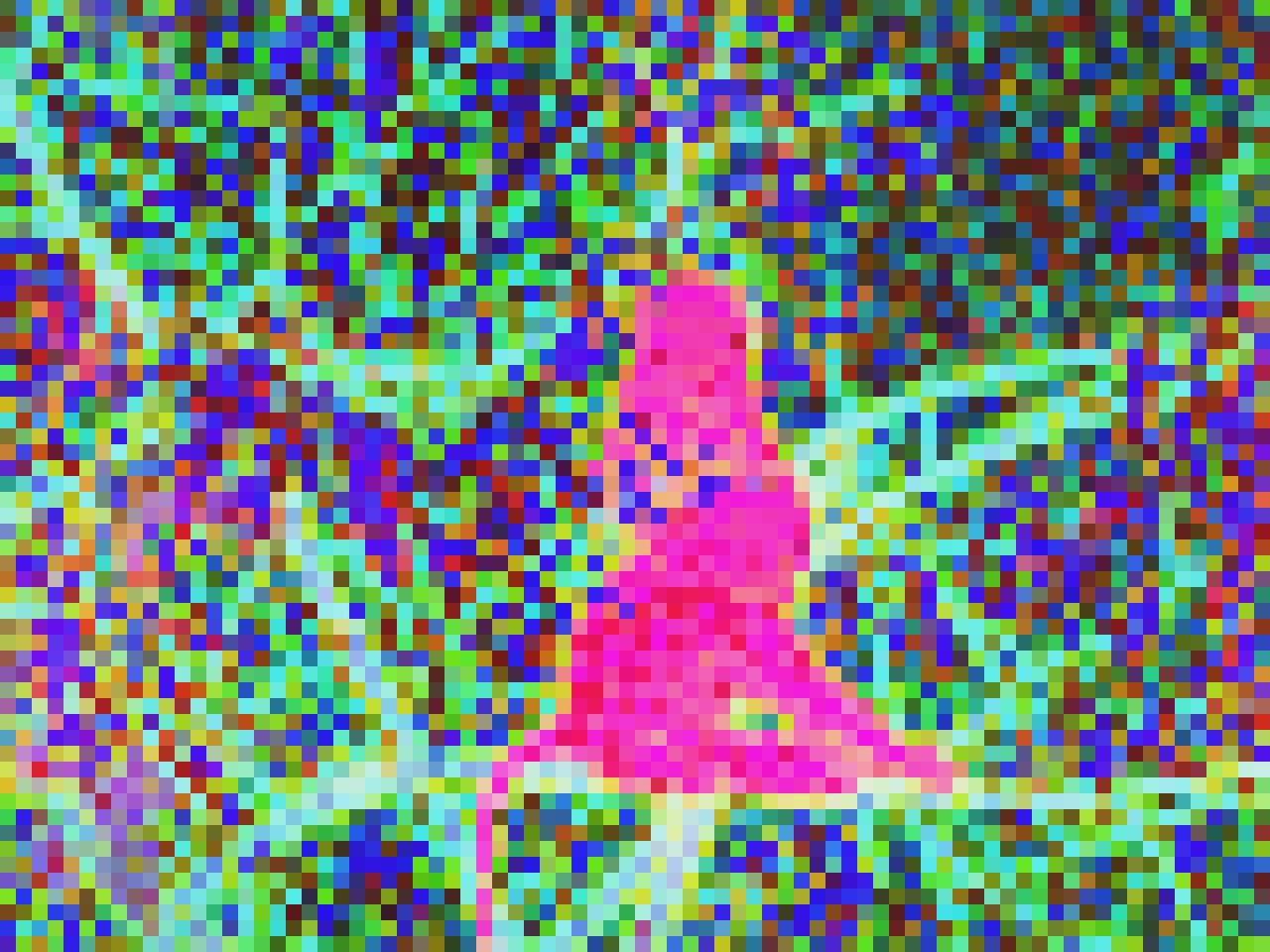}
    \end{subfigure}\hfill
    \begin{subfigure}{0.195\textwidth}
        \centering
        \includegraphics[width=\linewidth]{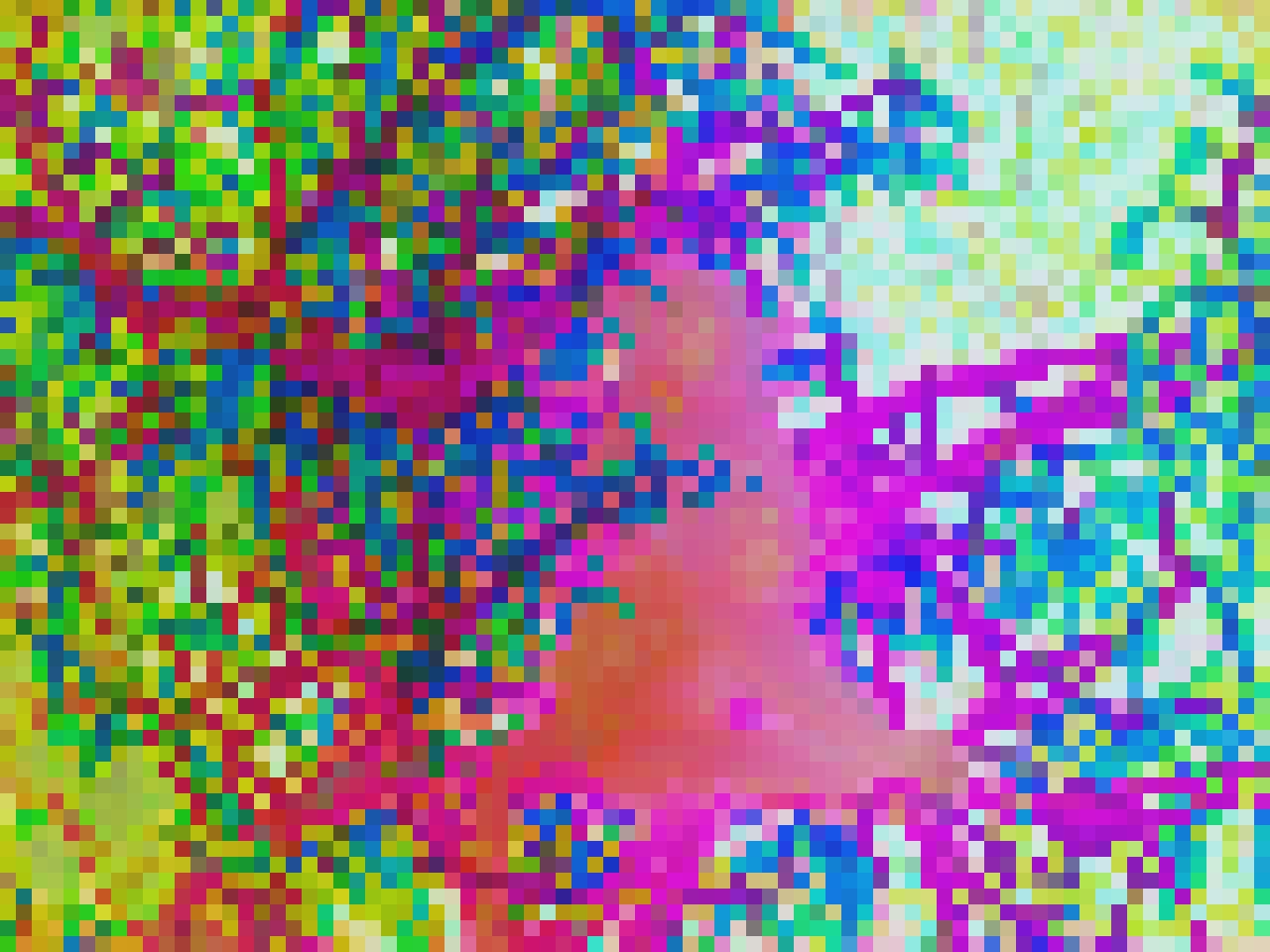}
    \end{subfigure}\hfill
    \begin{subfigure}{0.195\textwidth}
        \centering
        \includegraphics[width=\linewidth]{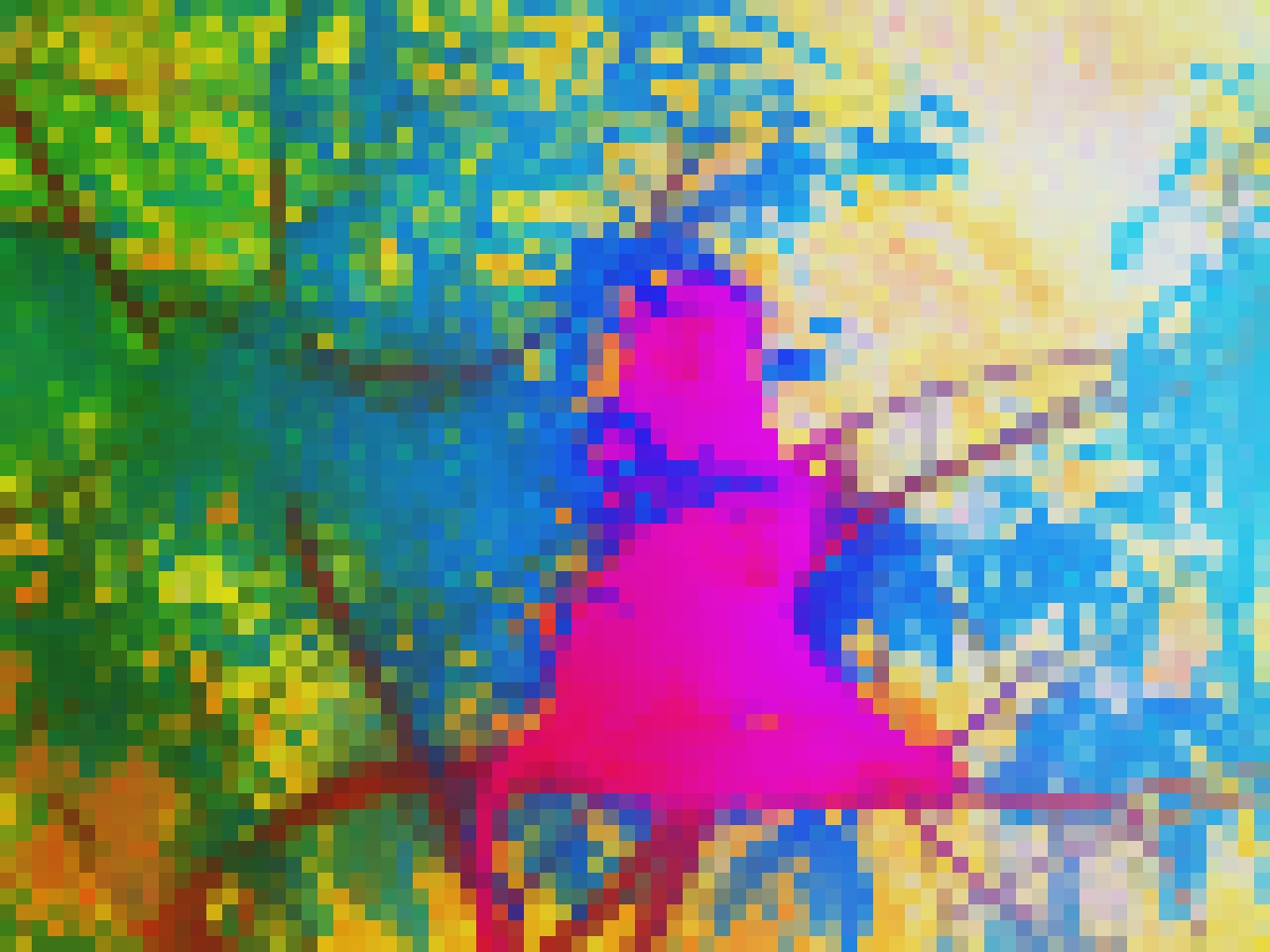}
    \end{subfigure}\vspace{0.25em}
    \begin{subfigure}{0.195\textwidth}
        \centering
        \includegraphics[width=\linewidth]{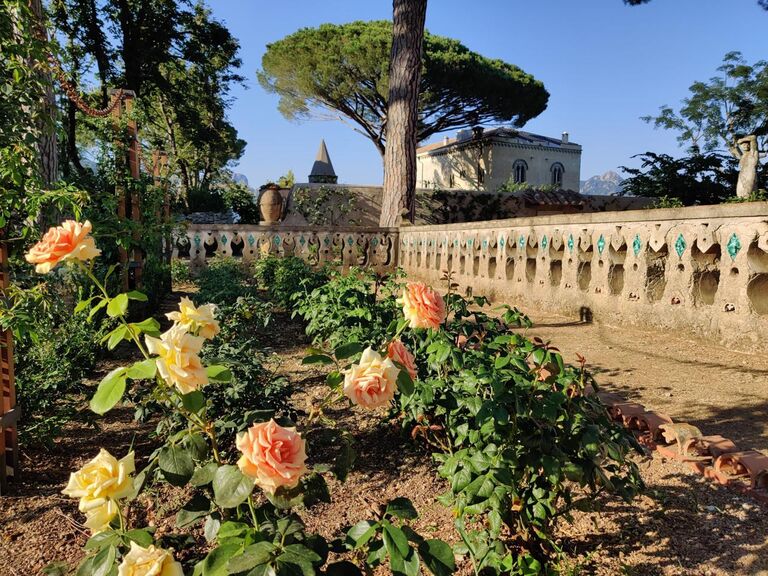}
    \end{subfigure}\hfill
    \begin{subfigure}{0.195\textwidth}
        \centering
        \includegraphics[width=\linewidth]{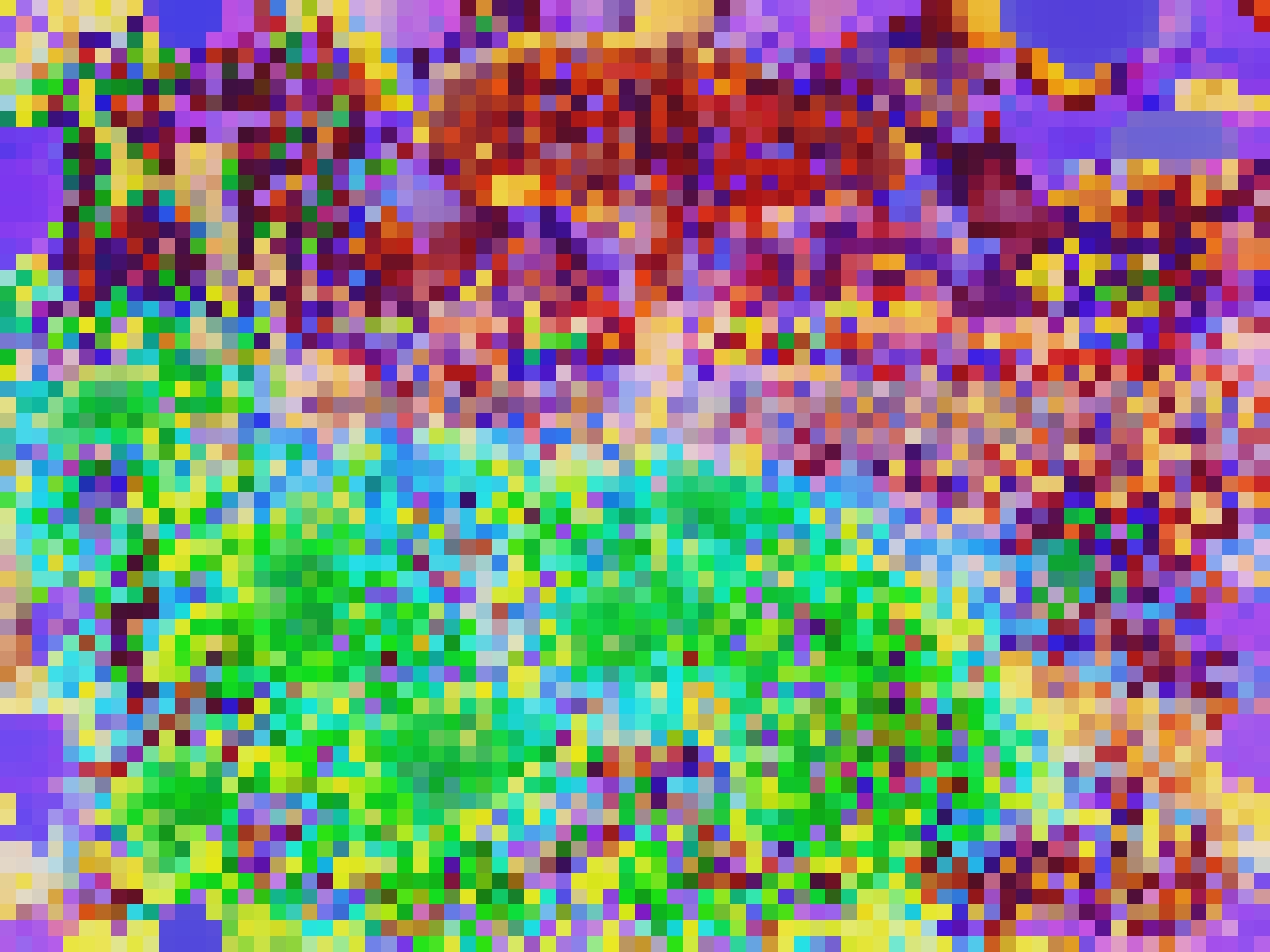}
    \end{subfigure}\hfill
    \begin{subfigure}{0.195\textwidth}
        \centering
        \includegraphics[width=\linewidth]{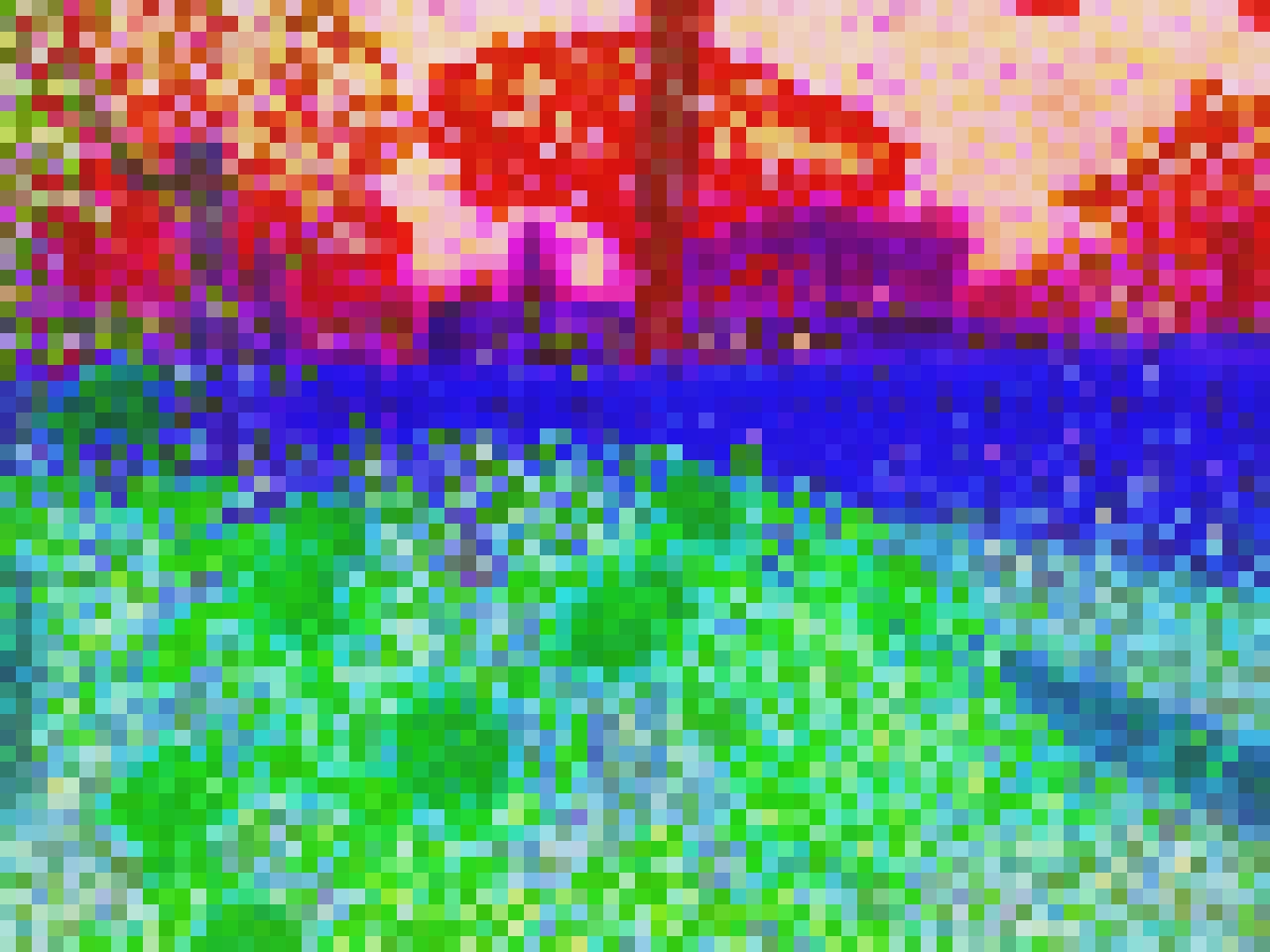}
    \end{subfigure}\hfill
    \begin{subfigure}{0.195\textwidth}
        \centering
        \includegraphics[width=\linewidth]{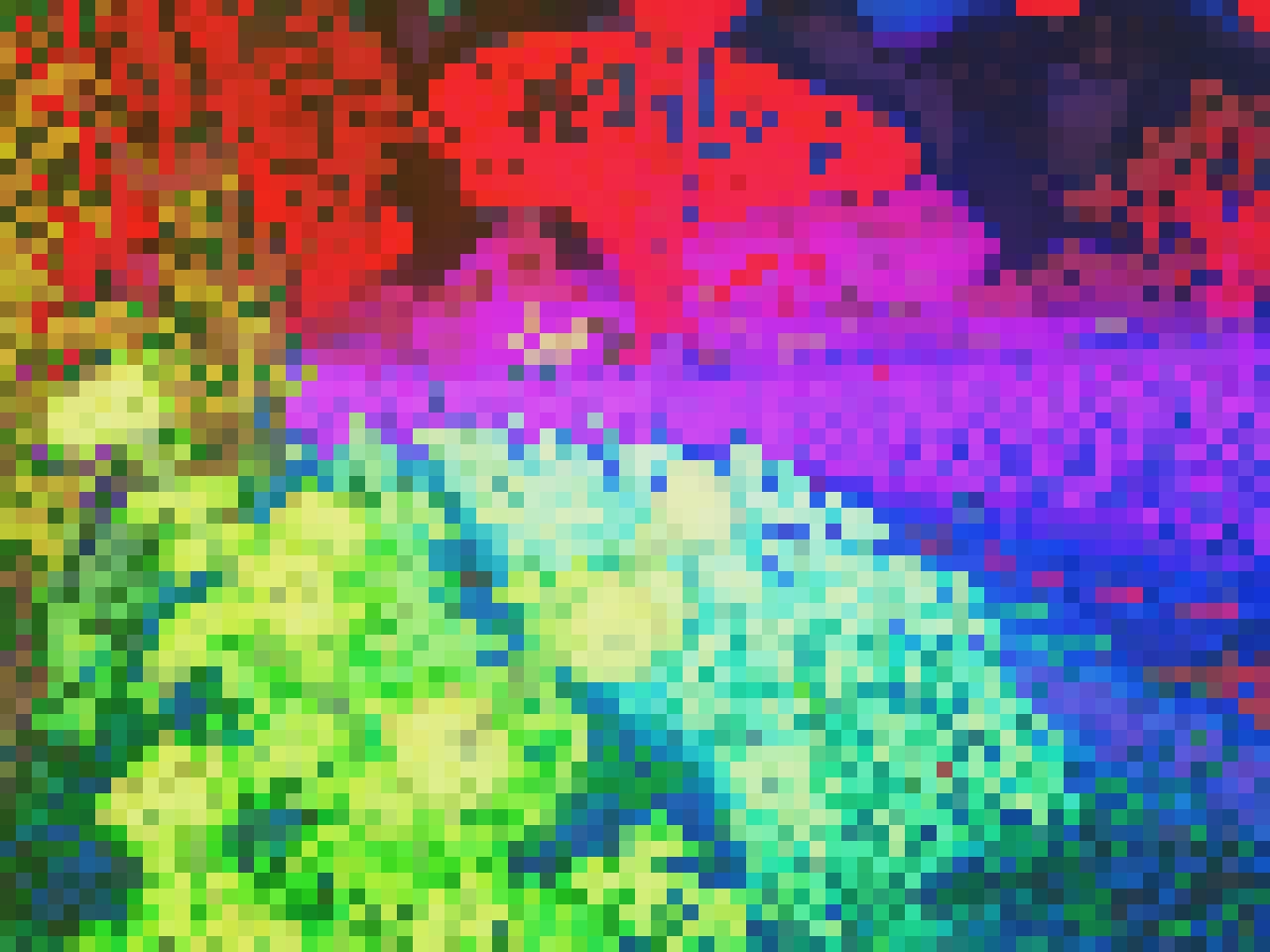}
    \end{subfigure}\hfill
    \begin{subfigure}{0.195\textwidth}
        \centering
        \includegraphics[width=\linewidth]{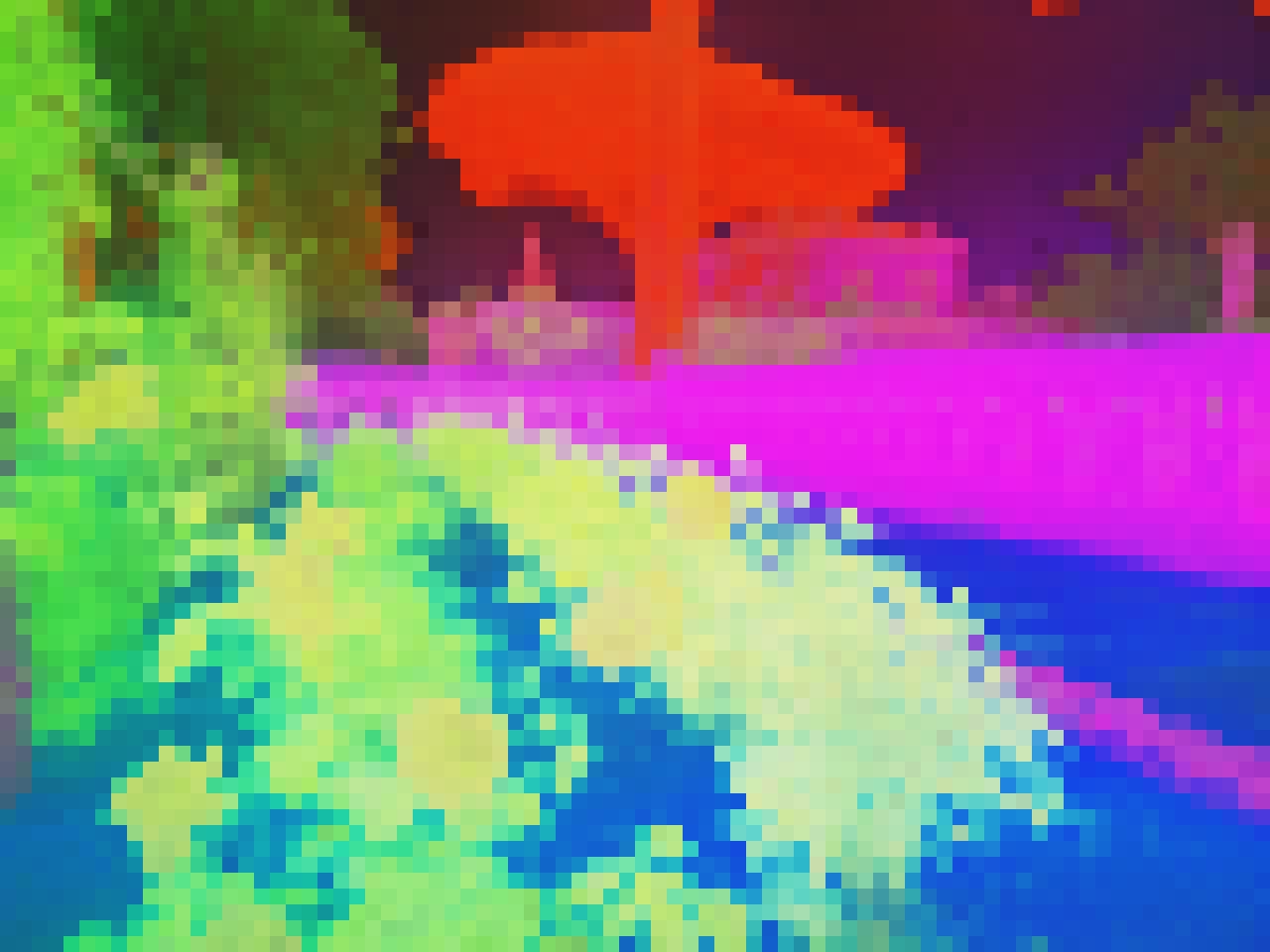}
    \end{subfigure}\vspace{0.25em}
    \begin{subfigure}{0.195\textwidth}
        \centering
        \includegraphics[width=\linewidth]{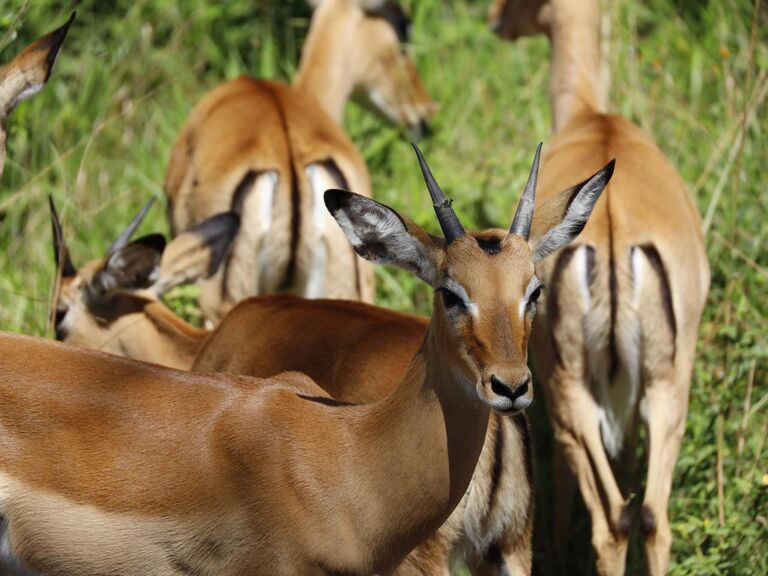}
        \caption*{Input}
    \end{subfigure}\hfill
    \begin{subfigure}{0.195\textwidth}
        \centering
        \includegraphics[width=\linewidth]{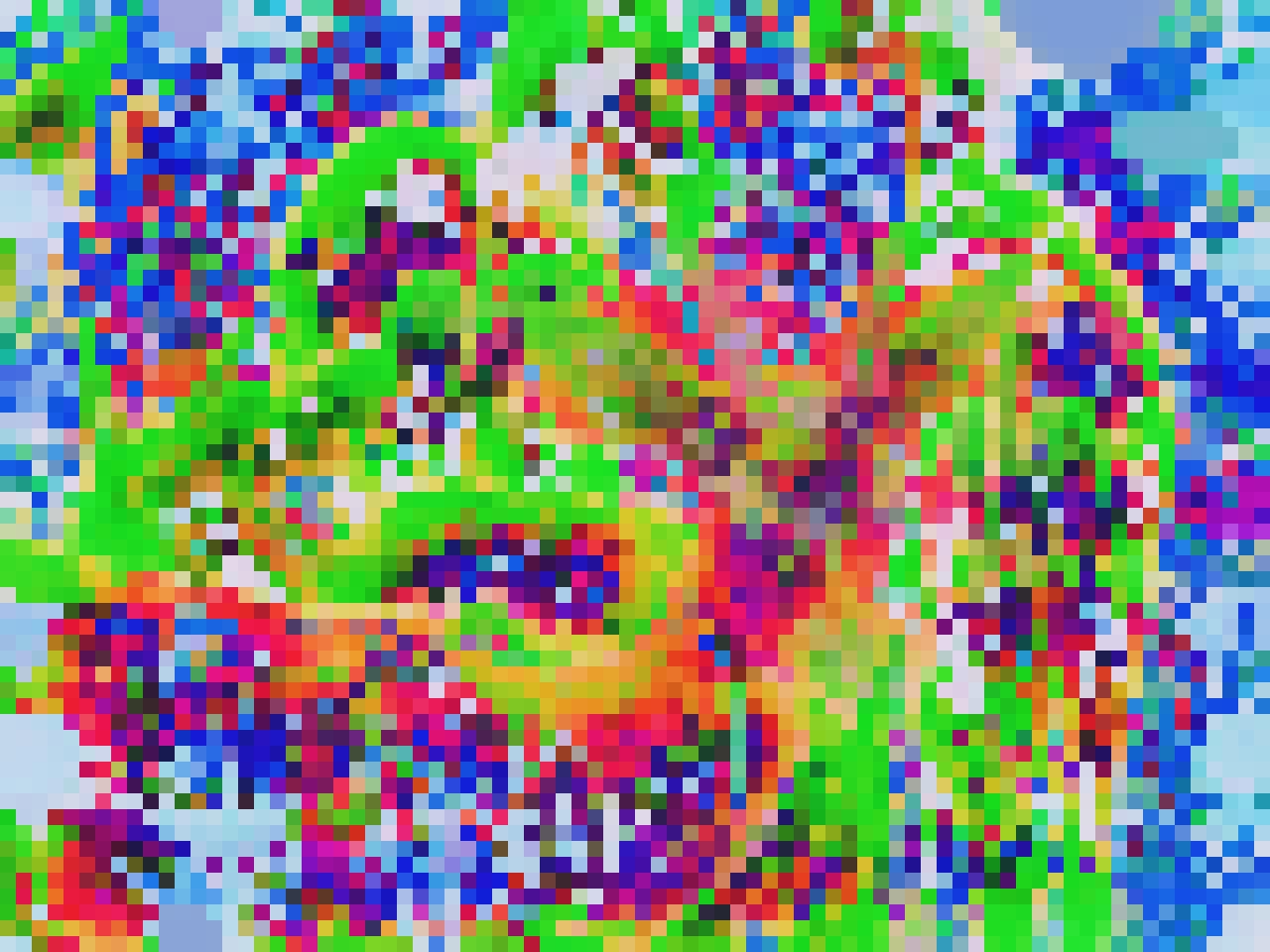}
        \caption*{SigLIP 2}
    \end{subfigure}\hfill
    \begin{subfigure}{0.195\textwidth}
        \centering
        \includegraphics[width=\linewidth]{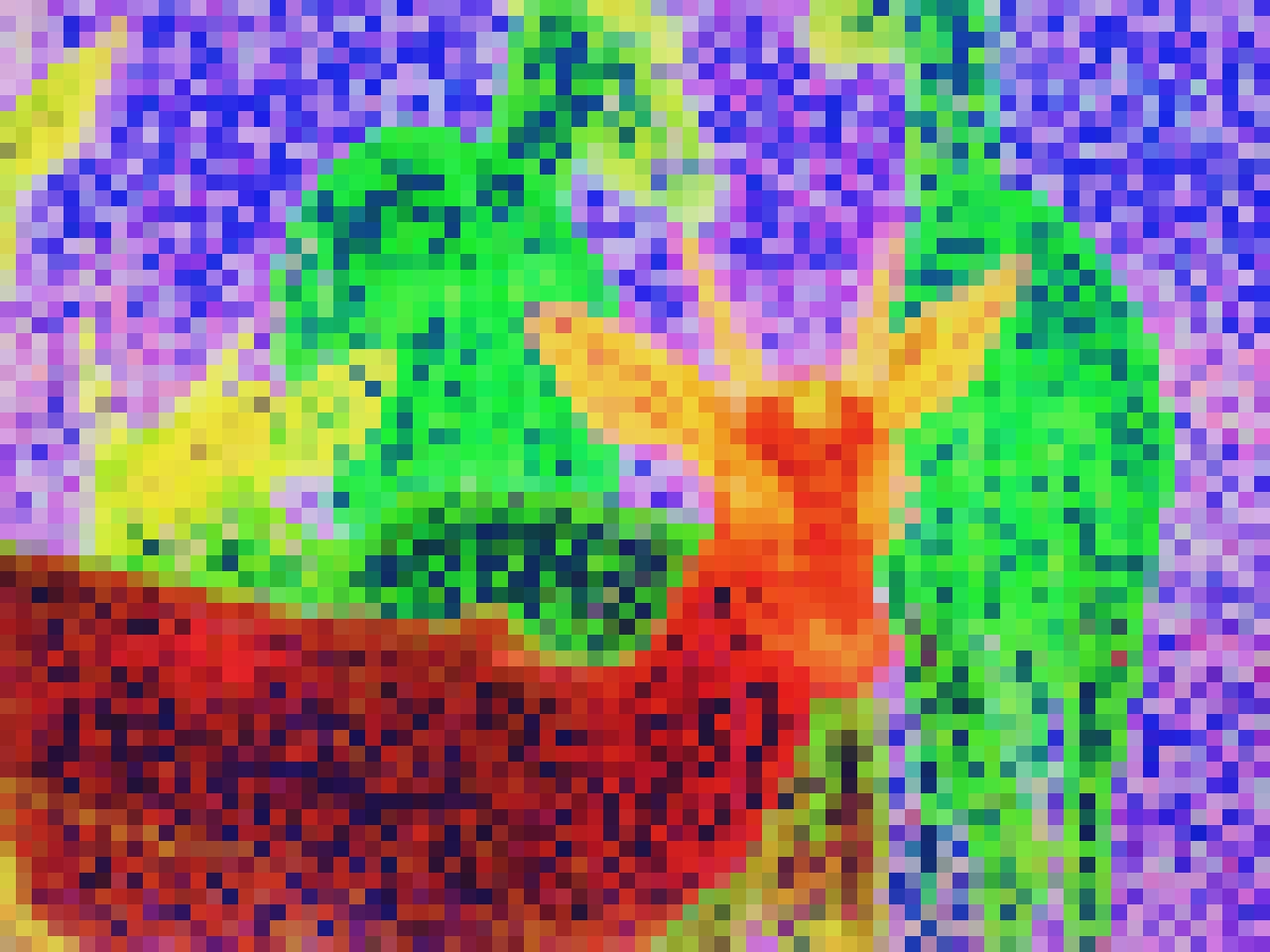}
        \caption*{PE Spatial}
    \end{subfigure}\hfill
    \begin{subfigure}{0.195\textwidth}
        \centering
        \includegraphics[width=\linewidth]{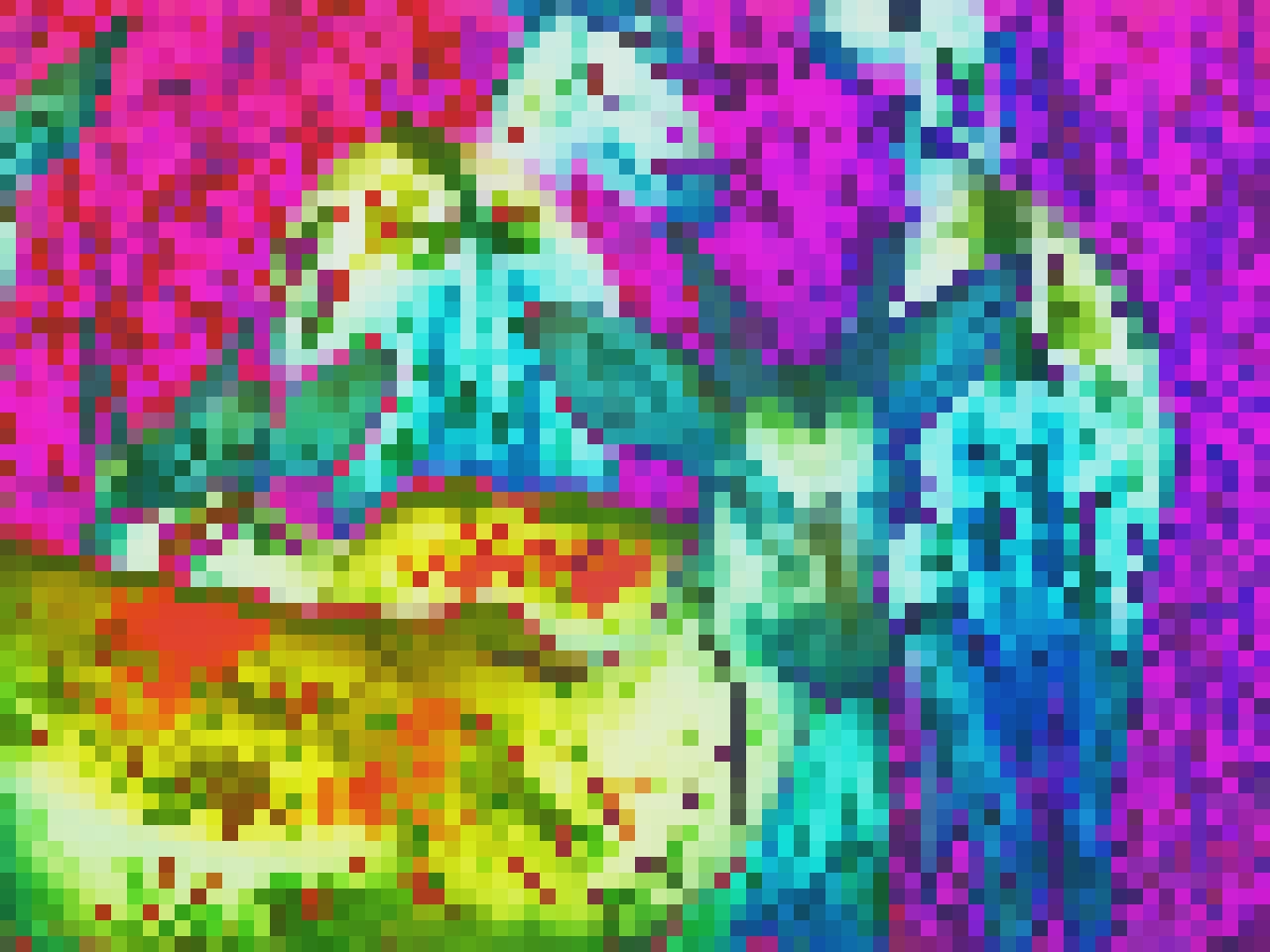}
        \caption*{DINOv2 w/reg}
    \end{subfigure}\hfill
    \begin{subfigure}{0.195\textwidth}
        \centering
        \includegraphics[width=\linewidth]{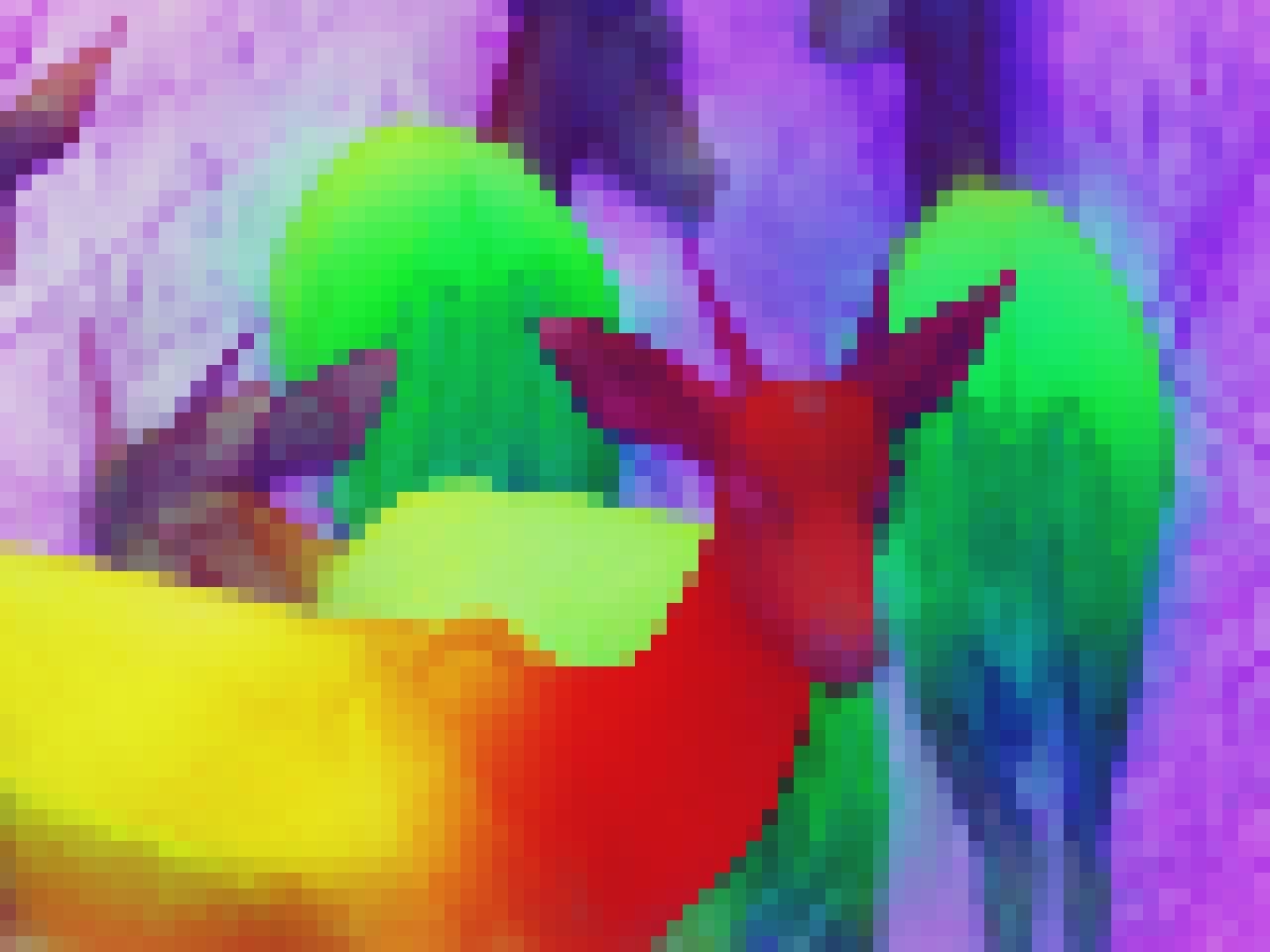}
        \caption*{DINOv3}
    \end{subfigure}
    \caption{
        Comparison of dense features.
        We compare several vision backbones by projecting their dense outputs using PCA and mapping them to RGB.
        From left to right:
        SigLIP 2 ViT-g/16,
        PEspatial ViT-G/14,
        DINOv2 ViT-g/14 with registers,
        DINOv3 ViT-7B/16.
        Images are forwarded at resolution \(1280{\times}960\) for models using patch 16 and \(1120{\times}840\) for patch 14, \ie all feature maps have size \(80{\times}60\).
    }
    \label{fig:results-pca}
\end{figure}

We start by analyzing DINOv3's dense feature maps qualitatively.
To this end, we project the dense feature space into 3 dimensions using principal component analysis (PCA), and map the resulting 3D space into RGB.
Because of the sign ambiguity in PCA (eight variants) and the arbitrary mapping between principal components and colors (six variants), we explore all combinations and report the visually most compelling one.
The resulting visualization is shown in \cref{fig:results-pca}.
Compared to other vision backbones, it can be seen that the features of DINOv3 are sharper, containing much less noise, and showing superior semantical coherence.

\subsubsection{Dense Linear Probing}
\label{sec:results-dense-linear-probing}

We perform linear probing on top of the dense features for two tasks: semantic segmentation and monocular depth estimation.
In both cases, we train a linear transform on top of the frozen patch outputs of DINOv3.
For semantic segmentation, we evaluate on the ADE20k~\citep{zhou2017scene}, Cityscapes~\citep{cordts2016cityscapes}, and PASCAL VOC 2012~\citep{pascal-voc-2012} datasets and report the mean intersection-over-union (mIoU) metric.
For depth estimation, we use the NYUv2~\citep{silberman2012indoor} and KITTI~\citep{geiger2013vision} datasets and report the root mean squared error (RMSE).

\begin{table}[t]
    \centering
    \small
    \caption{
        Dense linear probing results on semantic segmentation and monocular depth estimation with frozen backbones. 
        We report the mean Intersection-over-Union (mIoU) metric for the segmentation benchmarks ADE20k, Cityscapes, and VOC. 
        We report the Root Mean Squared Error (RMSE) metric for the depth benchmarks NYUv2 and KITTI. 
        For segmentation, all models are evaluated with input resolution adapted to 1024 patch tokens (\ie $448 \times 448$ for patch size 14, $512 \times 512$ for patch size 16). 
    }
    \label{tab:results-linear-dense}
    \begin{tabular}{@{}ll c ccc c c c @{}}
        \toprule
        & && \multicolumn{3}{c}{Segmentation} && \multicolumn{2}{c}{Depth} \\ 
        \cmidrule{4-6} \cmidrule{8-9}
        Method  & ViT && ADE20k & Citysc. & VOC && NYUv2 $\downarrow$ & KITTI $\downarrow$ \\
        \midrule
        \multicolumn{2}{@{}l}{\resultsTableHeaderAgg} && & && & \\
        AM-RADIOv2.5    & g/14     && 53.0 & 78.4 & 85.4 && 0.340 & 2.918 \\ 
        PEspatial      & G/14     && 49.3 & 73.2 & 82.7 && 0.362 & 3.082 \\
        \midrule
        \multicolumn{2}{@{}l}{\resultsTableHeaderWeakly} && & && & \\
        SigLIP~2        & g/16 && 42.7 & 64.8 & 72.7 && 0.494 & 3.273 \\
        PEcore         & G/14 && 38.9 & 61.1 & 69.2 && 0.590 & 4.119 \\
        \midrule
        \multicolumn{2}{@{}l}{\resultsTableHeaderSelf} && & && & \\
        Franca          & g/14     && 46.3 & 68.7 & 82.9 && 0.445 & 3.140 \\
        DINOv2          & g/14     && 49.5 & 75.6 & 83.1 && 0.372 & 2.624 \\
        Web-DINO        & 7B/14    && 42.7 &  68.3 & 76.1 &&   0.466  &   3.158  \\
        \midrule
        DINOv3          & 7B/16    && \bf 55.9 & \bf 81.1 & \bf 86.6 && \bf 0.309 & \bf 2.346 \\
    \bottomrule
    \end{tabular}
\end{table}

\paragraph[Results]{Results (\cref{tab:results-linear-dense})}
The segmentation results demonstrate the superior quality of our dense features. 
On the general ADE20k dataset, DINOv3 outperforms the self-supervised baselines by more than 6 mIoU points, and the weakly supervised baselines by more than 13 points.
Furthermore, DINOv3 surpasses PEspatial by more than 6 points, and AM-RADIOv2.5 by nearly 3 points. These results are remarkable as both are strong baselines, being distilled from the heavily supervised segmentation model SAM~\citep{kirillov2023segment}.
Similar results are observed on the self-driving benchmark Cityscapes, with DINOv3 achieving the best mIoU of $81.1$, surpassing AM-RADIOv2.5 by 2.5 points, and all other backbones by at least 5.5 points. 

On monocular depth estimation, DINOv3 again outperforms all other models by significant margins: the weakly-supervised models PEcore and SigLIP~2 are still lagging, with DINOv2 and the more advanced models derived from SAM are the closest competitors.
Interestingly, while PEspatial and AM-RADIO show strong performance on NYU, their performance is lower than DINOv2's on KITTI.
Even there, DINOv3 outperforms its predecessor DINOv2 by 0.278 RMSE.

Both sets of evaluations show the outstanding representation power of the dense features of DINOv3 and reflect the visual results from \cref{fig:results-pca}.
With only a linear predictor, DINOv3 allows robust prediction of object categories and masks, as well as physical measurements of the scene such as relative depth.
These results show that the features are not only visually sharp and properly localized, they also represent many important properties of the underlying observations in a linearly separable way.
Finally, the absolute performance obtained with a linear classifier on ADE20k (55.9 mIoU) is itself impressive, as it is not far from the absolute the state-of-the-art (63.0 mIoU) on this dataset.

\subsubsection{3D Correspondence Estimation} 
\label{sec:results-correspondence-estimation}

Understanding the 3D world has always been an important goal of computer vision
Image foundation models have recently fueled research in 3D understanding by offering \emph{3D-aware features}.
In this section, we evaluate the \emph{multi-view consistency} of DINOv3---that is, whether patch features of the same keypoint in different views of an object are similar---following the protocol defined in Probe3D~\citep{banani2024probing}.
We distinguish between \emph{geometric} and \emph{semantic} correspondence estimation. 
The former refers to matching keypoints for the \emph{same object instance} while the latter refers to matching keypoints for different instances of the \emph{same object class}.
We evaluate geometric correspondence on the NAVI dataset~\citep{jampani2023navi} and semantic correspondence on the SPair dataset~\citep{min2019spair71k}, and
measure performance with correspondence recall in both cases.
Please refer to \cref{app:exp-details:keypoint-matching} for more experimental details.

\begin{table}[t]
    \centering
    \small
    \caption{
        Evaluation of 3D consistency of dense representations.
        We estimate 3D keypoint correspondences across views following the evaluation protocol of Probe3D~\citep{banani2024probing}.
        To measure performance, we report the correspondence recall, \ie the percentage of correspondences falling into a specified distance.
    }
    \label{tab:results-correspondence-estimation}
    \begin{tabular}{@{}ll c ccc@{}}
        \toprule
        & && \multicolumn{1}{c}{Geometric} && \multicolumn{1}{c}{Semantic} \\ 
        \cmidrule{4-4} \cmidrule{6-6}
        Method  & ViT && NAVI && SPair \\
        \midrule
        \multicolumn{2}{@{}l}{\resultsTableHeaderAgg} && & \\
        AM-RADIOv2.5    & g/14   && 59.4 && 56.8 \\ 
        PEspatial      & G/14   && 53.8 && 49.6 \\
        \midrule
        \multicolumn{2}{@{}l}{\resultsTableHeaderWeakly} && & \\
        SigLIP~2        & g/16   && 49.4 && 42.6 \\
        PEcore         & G/14   && 39.9 && 23.1 \\
        \midrule
        \multicolumn{2}{@{}l}{\resultsTableHeaderSelf} && & \\
        Franca          & g/14   && 54.6 && 51.0 \\
        DINOv2          & g/14   && 60.1 && 56.1 \\
        Web-DINO        & 7B/14  && 55.0 && 32.2 \\        
        \midrule
        DINOv3          & 7B/16  && \bf 64.4 && \bf 58.7 \\
    \bottomrule
    \end{tabular}
\end{table}

\paragraph[Results]{Results (\cref{tab:results-correspondence-estimation})}
For geometric correspondences, DINOv3 outperforms all other models and improves over the second best model (DINOv2) by 4.3\% recall.
Other SSL scaling endeavors (Franca and WebSSL) lag behind DINOv2, showing that it is still a strong baseline.
Weakly-supervised models (PEcore and SigLIP~2) do not fare well on this task, indicating a lack of 3D awareness. 
For models with SAM distillation, AM-RADIO nearly reaches the performance of DINOv2, but PEspatial still lags behind it ($-11.6$\% recall), and even falls behind Franca ($-0.8$\% recall).
This suggests that self-supervised learning is a key component for strong performance on this task.
For semantic correspondences, the same conclusions apply.
DINOv3 performs best, outperforming both its predecessor ($+2.6$\% recall) and AM-RADIO ($+1.9$\% recall).
Overall, these impressive performance on keypoint matching are very promising signals for downstream use of DINOv3 in other 3D-heavy applications.

\subsubsection{Unsupervised Object Discovery}
\label{sec:results-object-discovery}

Powerful self-supervised features facilitate discovering object instances in images without requiring \emph{any} annotations~\citep{Vo21LOD,simeoni2021localizing,seitzer2023bridging,wang2023tokencut,simeoni2025unsupervised}. 
We test this capability for different vision encoders via the task of unsupervised object discovery, which requires class-agnostic segmentation of objects in images~\citep{Russell06ObjectDiscovery,tuytelaars2010bjectDiscovery,Cho2015ObjectDiscovery,Vo2019UnsupOptim}.
In particular, we use the non-parametric graph-based TokenCut algorithm~\citep{wang2023tokencut}, which has shown strong performance on a variety of backbones.
We run it on three widely used datasets: VOC 2007, VOC 2012~\citep{everingham2015pascal}, and COCO-20k~\citep{Lin2014cocodataset,Vo20rOSD}. 
We follow the evaluation protocol defined by \citet{simeoni2021localizing} and report the CorLoc metric. 
To properly compare backbones with different feature distributions, we perform a search over the main TokenCut hyperparameter, namely the cosine similarity threshold applied when constructing the patch graph used for partitioning. 
Originally, the best object discovery results were obtained with DINO~\citep{caron2021emerging} using the keys of the last attention layer.
However, this hand-crafted choice does not consistently generalize to other backbones.
For simplicity, we always employ the output features for all models. 

\begin{figure}[t]
    \centering
    \small
    \begin{minipage}{0.57\textwidth}
        \begin{tabular}{@{}ll c ccc@{}}
            \toprule
            Method & ViT & & VOC07 & VOC12 & COCO \\
            \midrule
            \multicolumn{2}{@{}l}{\resultsTableHeaderAgg} && & \\
            AM-RADIOv2.5    & g/14     && 55.0 & 59.7 & 45.9 \\ 
            PEspatial      & G/14     && 51.2 & 56.0 & 43.9 \\
            \midrule
            \multicolumn{2}{@{}l}{\resultsTableHeaderWeakly} && & \\
            SigLIPv2        & g/16     && 20.5 &  24.7 & 18.6 \\
            PEcore         & G/14     && 14.2 & 18.2 & 13.5 \\
            \midrule
            \multicolumn{2}{@{}l}{\resultsTableHeaderSelf} && & \\
            DINO            & S/16     && 61.1 & 66.0 & 48.7 \\
            DINO            & B/16     && 60.1 & 64.4 & 50.5 \\
            DINOv2          & g/14     && 55.6 & 60.4 & 45.4 \\
            Web-DINO        & 7B/14    && 26.1 & 29.7 & 20.9 \\
            \midrule
            DINOv3          & 7B/16    && \bf 66.1 & \bf 69.5 & \bf 55.1 \\
            \bottomrule
        \end{tabular}
    \end{minipage}
    \hfill
    \begin{minipage}{0.42\textwidth}
    \raisebox{-3em}{
    \renewcommand{\arraystretch}{0.2}
    \setlength{\tabcolsep}{0.3pt}
    \begin{tabular}{cc}
       \includegraphics[width=0.42\linewidth]{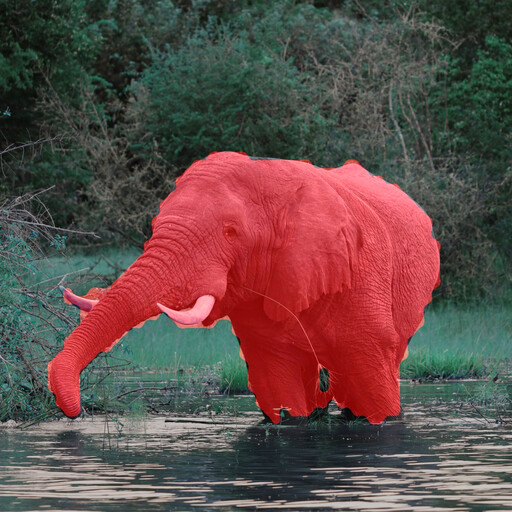}   &   \includegraphics[width=0.42\linewidth]{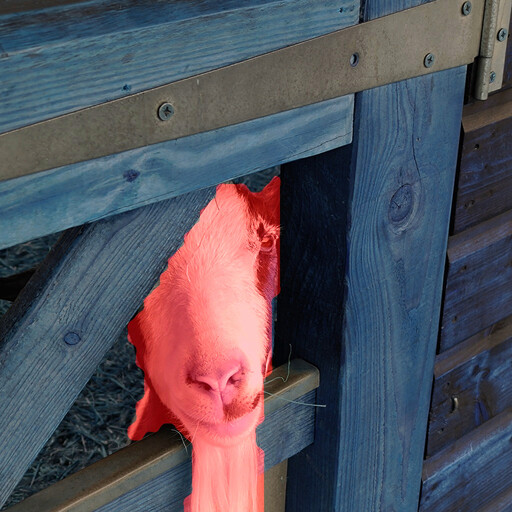} \\
        \includegraphics[width=0.42\linewidth]{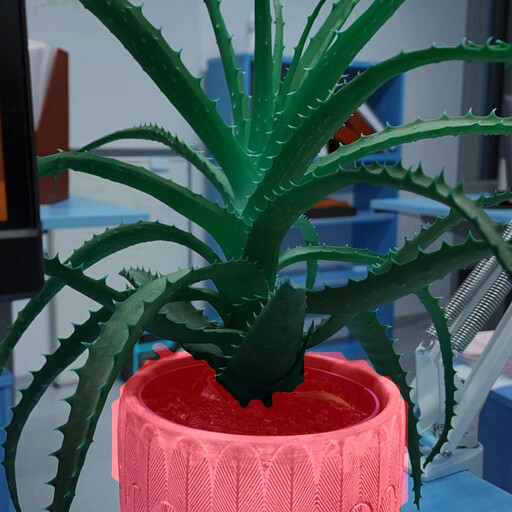} &  \includegraphics[width=0.42\linewidth]{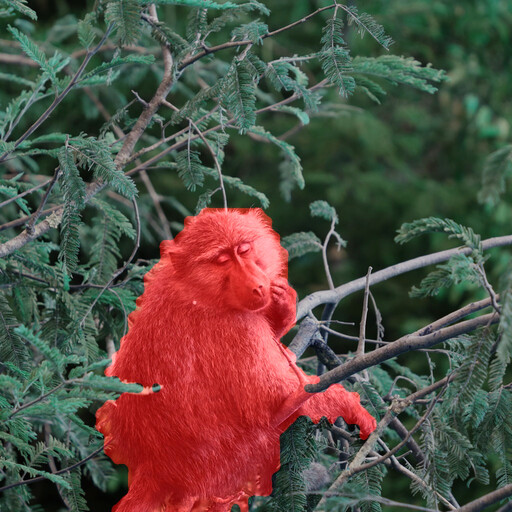}
    \end{tabular}
    }
    \end{minipage}
    \caption{
        Unsupervised object discovery.
        We apply TokenCut~\citep{wang2022self} on the output patch features of different backbones and report CorLoc metric. We also visualize predicted masks obtained with DINOv3 (red overlay on input images at res. $1024$), obtained \emph{with no annotation and no post-processing}.
    }
    \label{tab:results-object-discovery}
\end{figure}

\paragraph[Results]{Results (\cref{tab:results-object-discovery})}
The original DINO has set a very high bar for this task. 
Interestingly, while DINOv2 has shown very strong performance for pixel-wise dense tasks, it fails at object discovery.
This can in part be attributed to the artifacts present in the dense features (\cf \cref{fig:results-pca}).
DINOv3, with its clean and precise output feature maps outperforms both its predecessors, with a $5.9$ CorLoc improvement on VOC 2007, and all other backbones, whether self-, weakly-supervised or agglomerative.
This evaluation confirms that DINOv3's dense features are both semantically strong and well localized.
We believe that this will pave the way for more class-agnostic object detection approaches, especially in scenarios where annotations are costly or unavailable, and where the set of relevant classes is not confined to a predefined subset.

\subsubsection{Video Segmentation Tracking}
\label{sec:results-tracking}

Beyond static images, an important property of visual representations is their \emph{temporal consistency}, \ie whether the features evolve in a stable manner through time.
To test for this property, we evaluate DINOv3 on the task of video segmentation tracking: given ground-truth instance segmentation masks in the first frame of a video, the goal is to propagate these masks to subsequent frames.
We use the DAVIS~2017~\citep{pont20172017}, YouTube-VOS~\citep{xu2018youtubevos}, and MOSE~\citep{ding2023mose} datasets.
We evaluate performance using the standard \(\JnF\)-mean metric, which combines region similarity (\(\mathcal{J}\)) and contour accuracy (\(\mathcal{F}\))~\citep{perazzi2016benchmark}.
Following \citet{jabri2020space}, we use a non-parametric label propagation algorithm that considers the similarity between patch features across frames.
We evaluate at three input resolutions, using a short side length of 420/480 (S), 840/960 (M), and 1260/1440 (L) pixels for models with patch size 14/16 (matching the number of patch tokens).
The \(\JnF\) score is always computed at the native resolution of the videos.
See \cref{app:exp-details:tracking} for more detailed experimental settings.

\begin{table}[t]
    \centering
    \small
    \caption{
        Video segmentation tracking evaluation.
        We report the \(\JnF\)-mean on DAVIS, YouTube-VOS, and MOSE at multiple resolutions.
        For models with patch size 14/16, the small, medium and large resolutions correspond to a video short side of 420/480, 840/960, 1260/1140 pixels.
    }
    \label{tab:results-video-segmentation-tracking}
    \begin{tabular}{@{}ll ccc ccc ccc@{}}
        \toprule
        & & \multicolumn{3}{c}{DAVIS} & \multicolumn{3}{c}{YouTube-VOS} & \multicolumn{3}{c}{MOSE} \\
        \cmidrule(lr){3-5} \cmidrule(lr){6-8} \cmidrule(lr){9-11}
        Method & ViT & S & M & L & S & M & L & S & M & L \\
        \midrule
        \multicolumn{2}{@{}l}{\resultsTableHeaderAgg} & & & & & & & & & \\
        AM-RADIOv2.5    & g/14  & 66.5 & 77.3 & 81.4 & 70.1 & 78.1 & 79.2 & 44.0 & 52.6 & 54.3 \\
        PEspatial      & G/14  & 68.4 & 74.5 & 70.5 & 68.5 & 67.5 & 55.6 & 39.3 & 40.2 & 34.0 \\
        \midrule
        \multicolumn{2}{@{}l}{\resultsTableHeaderWeakly} & & & & & & & & & \\
        SigLIP 2        & g/16  & 56.1 & 62.3 & 62.9 & 52.0 & 57.3 & 55.1 & 28.0 & 30.3 & 29.2 \\
        PEcore         & G/14  & 48.2 & 53.1 & 49.8 & 34.7 & 33.0 & 25.3 & 17.8 & 19.0 & 15.4 \\
        \midrule
        \multicolumn{2}{@{}l}{\resultsTableHeaderSelf} & & & & & & & & & \\
        Franca          & g/14  & 61.8 & 66.9 & 66.5 & 67.3 & 70.5 & 67.9 & 40.3 & 42.6 & 41.9 \\
        DINOv2          & g/14  & 63.9 & 73.6 & 76.6 & 65.6 & 73.5 & 74.6 & 40.4 & 47.6 & 48.5 \\
        Web-DINO        & 7B/14 & 57.2 & 65.8 & 69.5 & 43.9 & 49.6 & 50.9 & 24.9 & 29.9 & 31.1 \\
        \midrule
        DINOv3          & 7B/16  & \bf 71.1 & \bf 79.7 & \bf 83.3 & \bf 74.1 & \bf 80.2 & \bf 80.7 & \bf 46.0 & \bf 53.9 & \bf 55.6 \\
        \bottomrule
    \end{tabular}
\end{table}

\begin{figure}[t]
    \centering
    \begin{subfigure}[b]{0.195\textwidth}
        \centering
        \includegraphics[width=\textwidth]{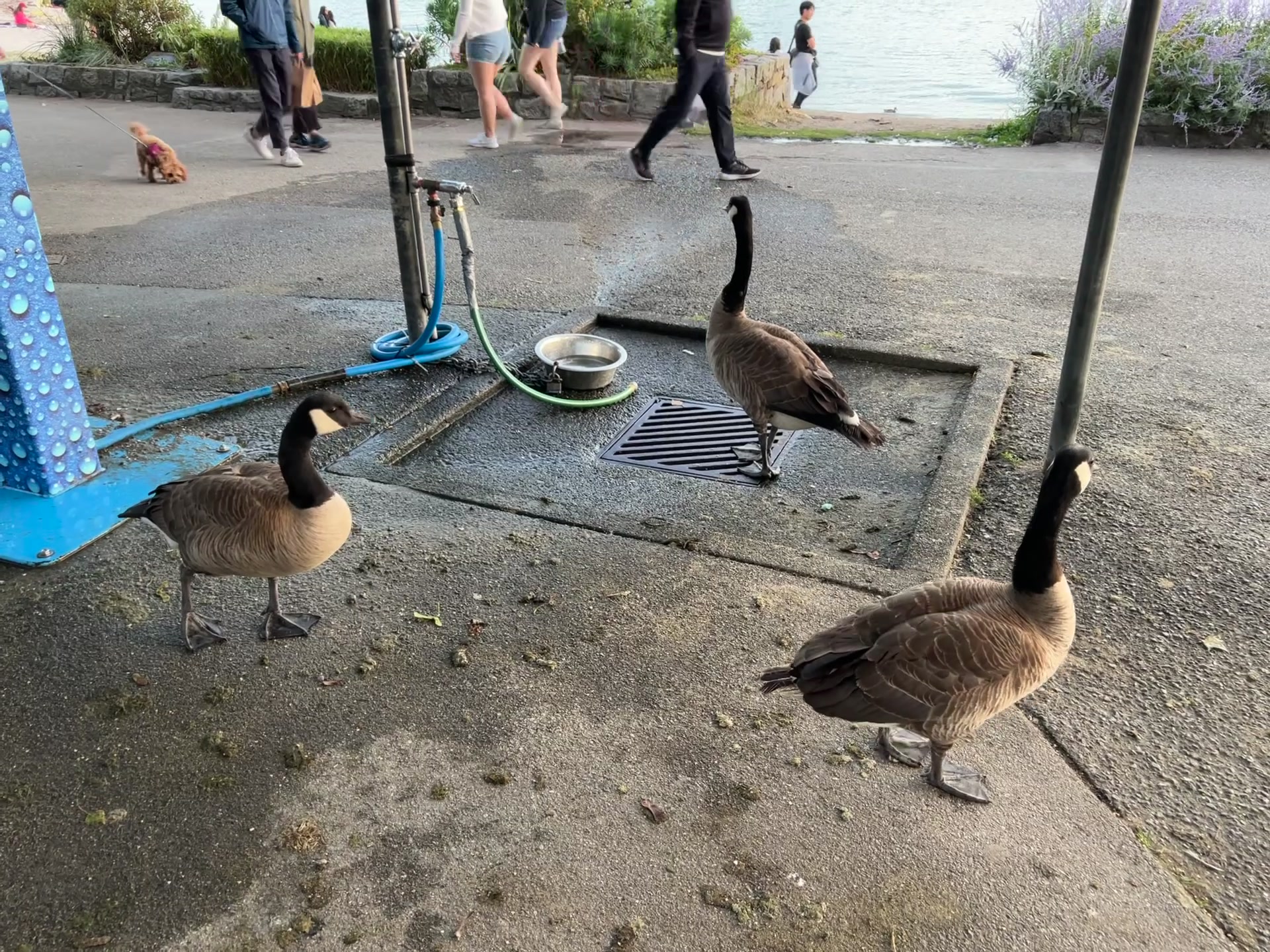}
    \end{subfigure}\hfill
    \begin{subfigure}[b]{0.195\textwidth}
        \centering
        \includegraphics[width=\textwidth]{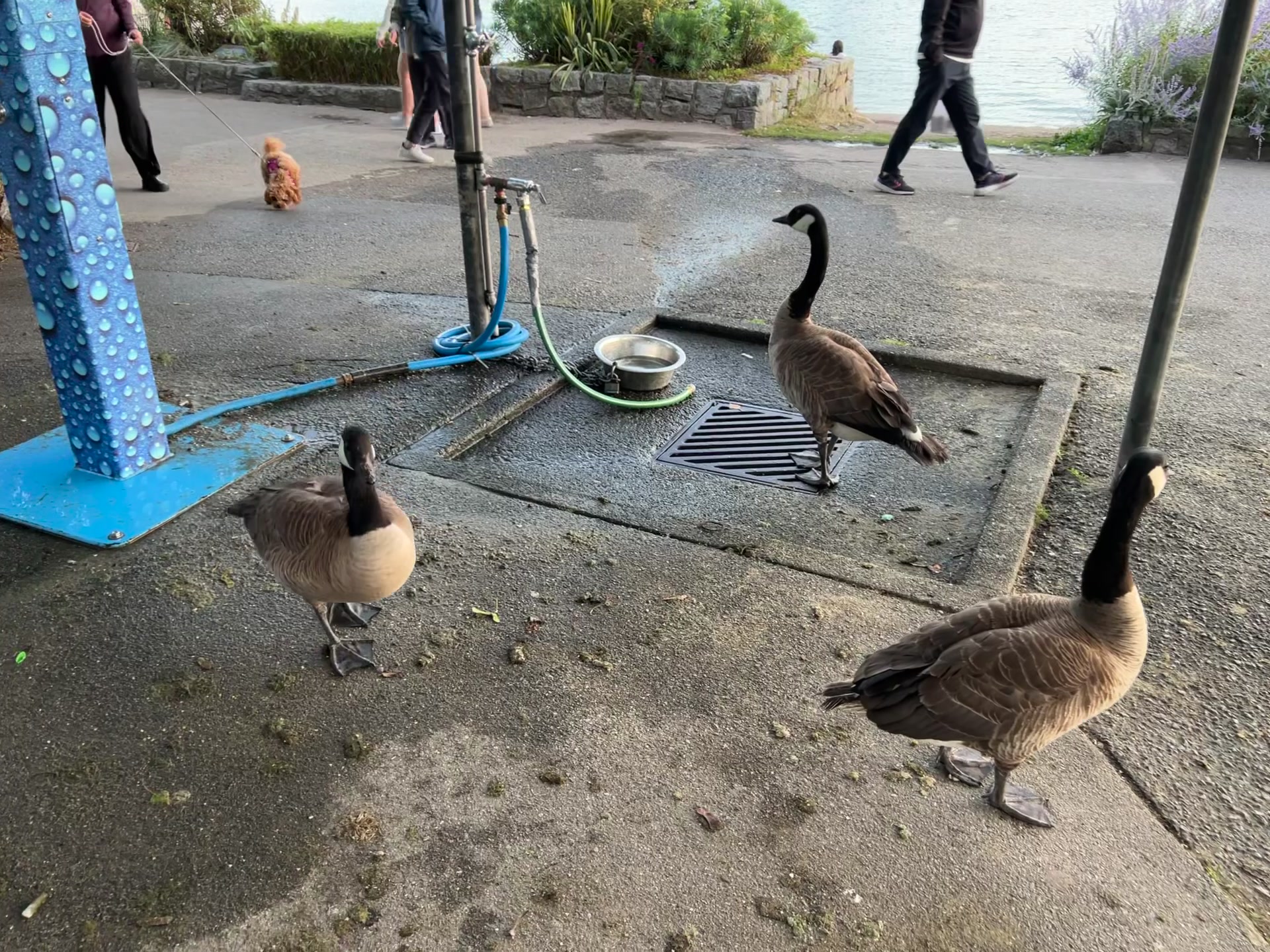}
    \end{subfigure}\hfill
    \begin{subfigure}[b]{0.195\textwidth}
        \centering
        \includegraphics[width=\textwidth]{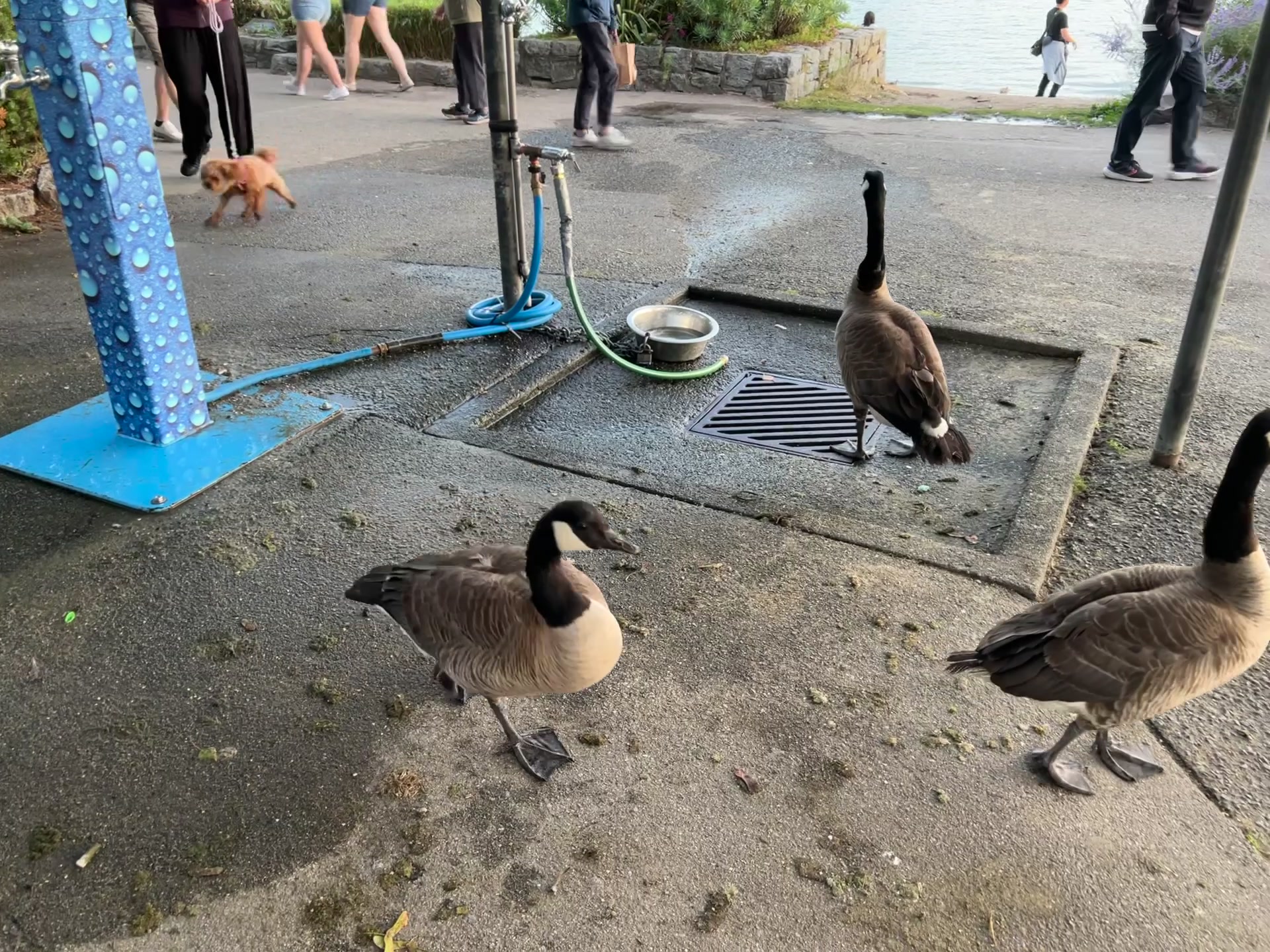}
    \end{subfigure}\hfill
    \begin{subfigure}[b]{0.195\textwidth}
        \centering
        \includegraphics[width=\textwidth]{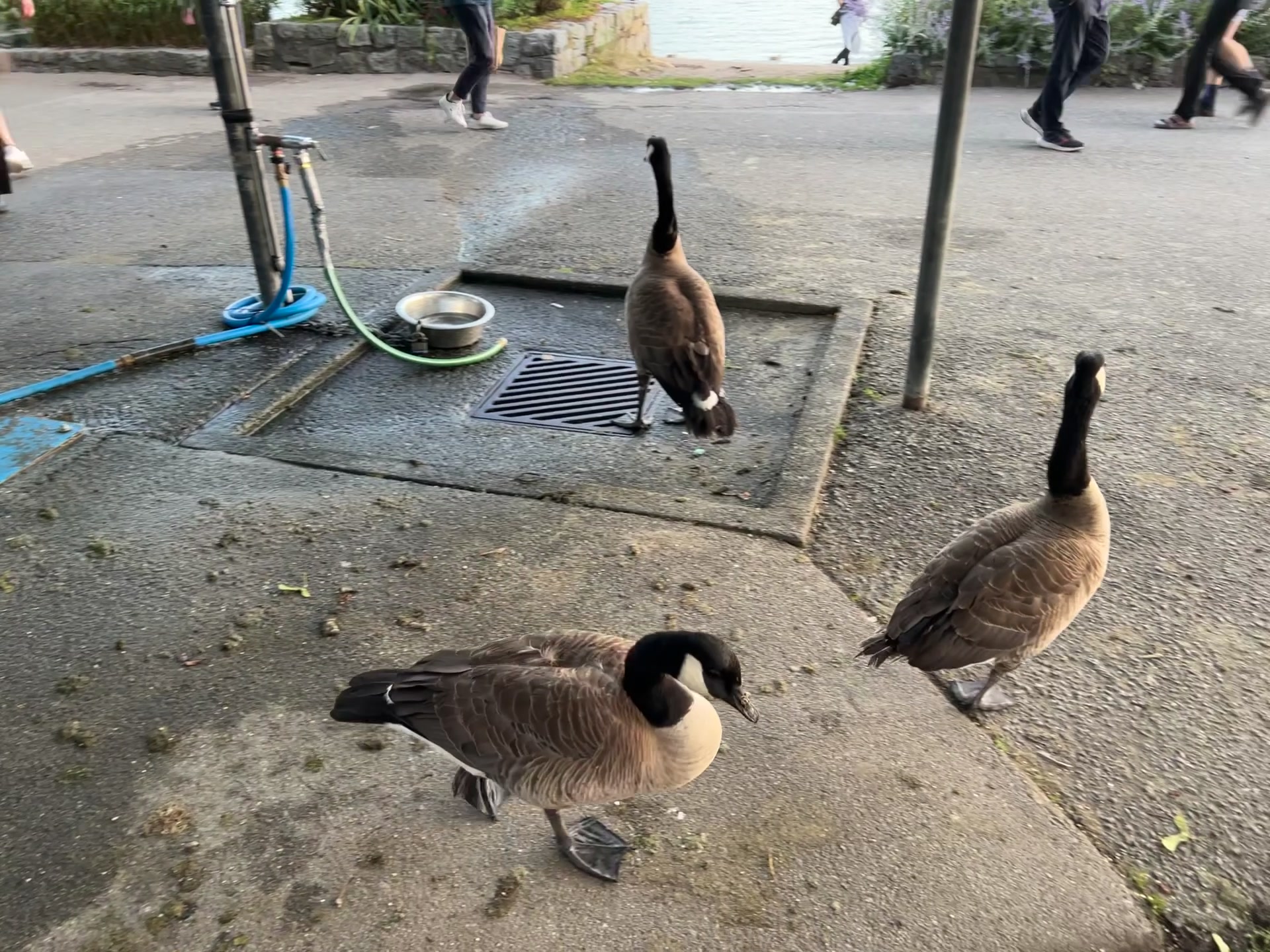}
    \end{subfigure}\hfill
    \begin{subfigure}[b]{0.195\textwidth}
        \centering
        \includegraphics[width=\textwidth]{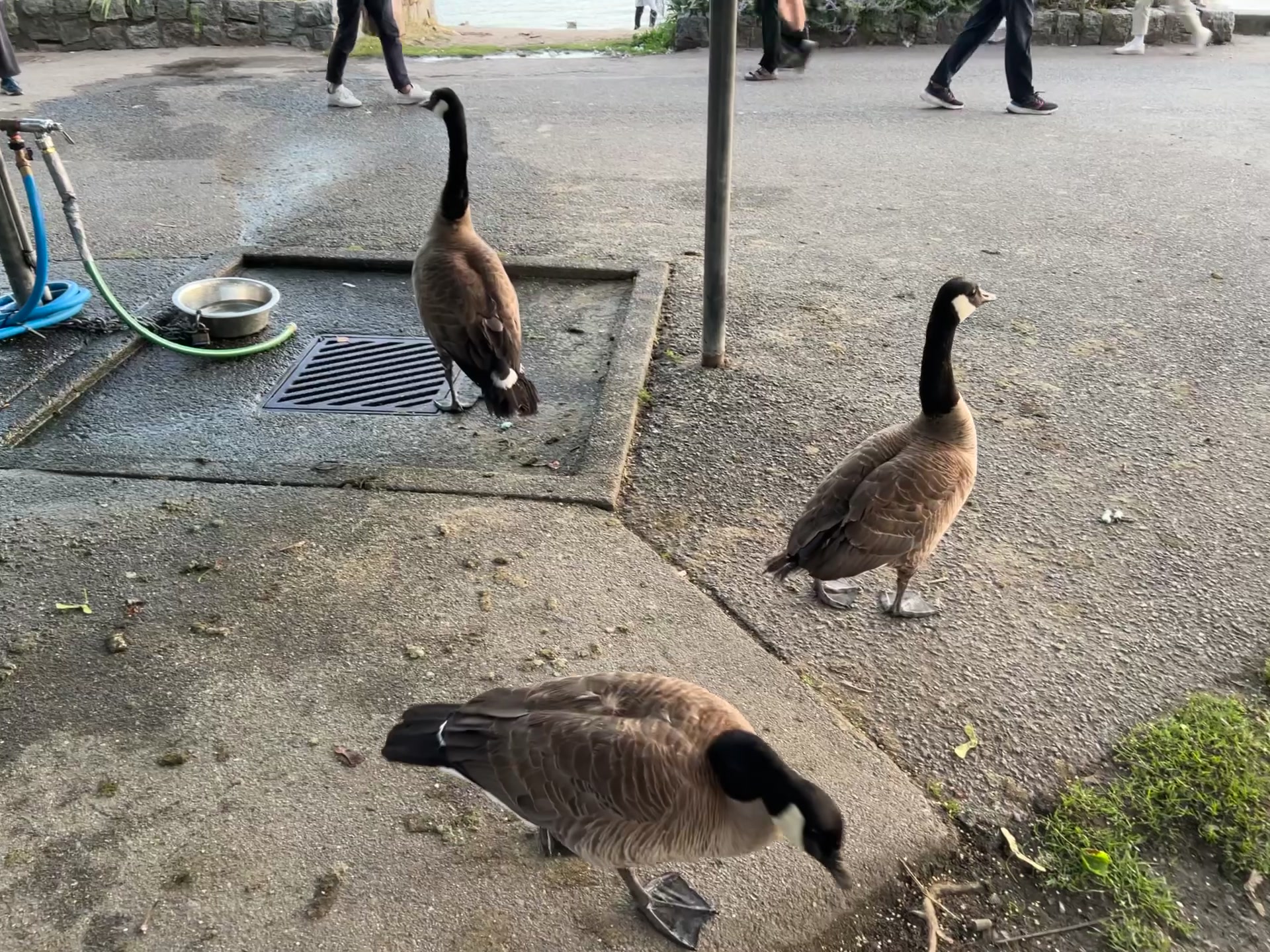}
    \end{subfigure}\vspace{0.25em}
    \begin{subfigure}[b]{0.195\textwidth}
        \centering
        \includegraphics[width=\textwidth]{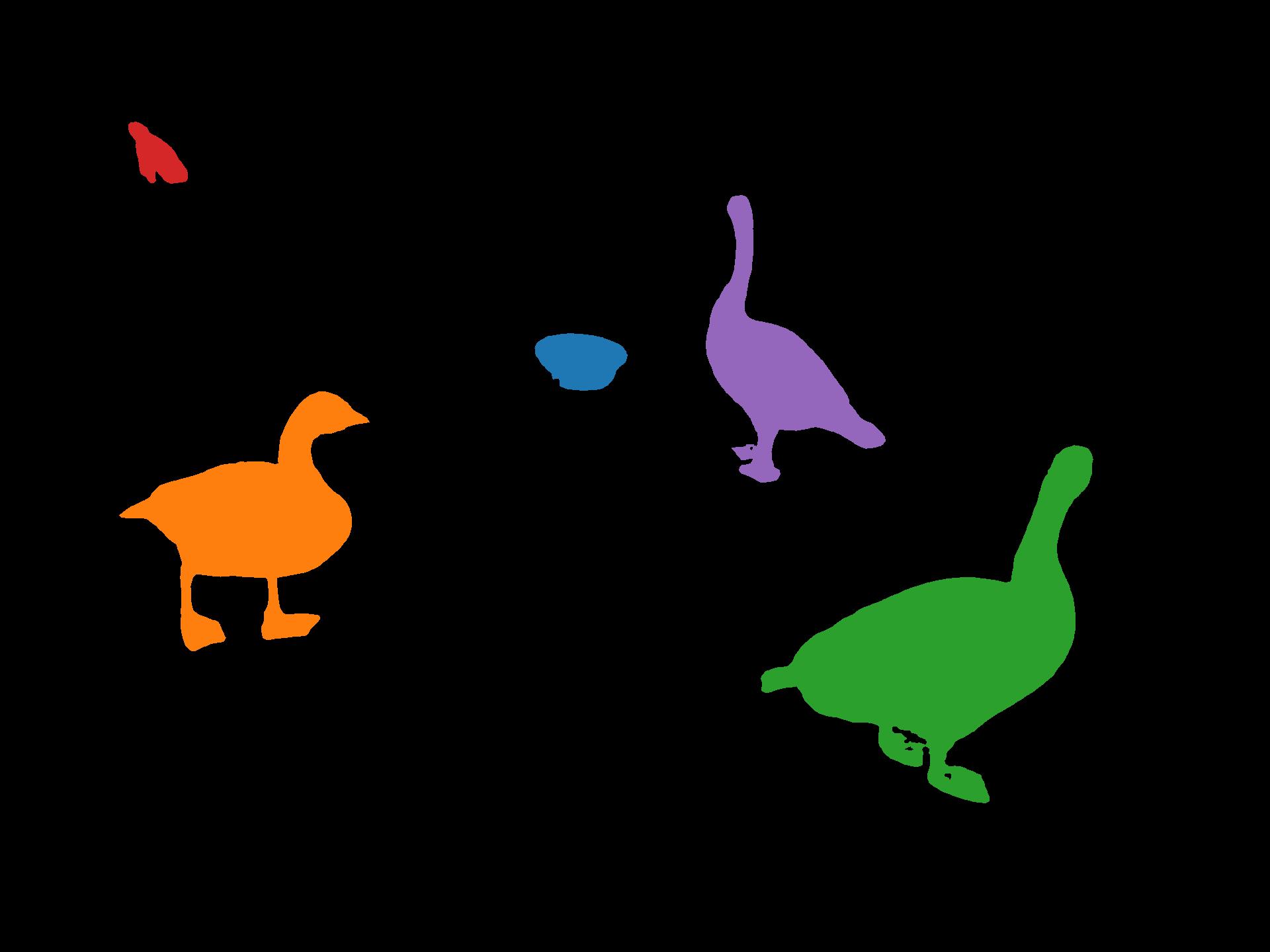}
        \caption*{Initial frame}
    \end{subfigure}\hfill
    \begin{subfigure}[b]{0.195\textwidth}
        \centering
        \includegraphics[width=\textwidth]{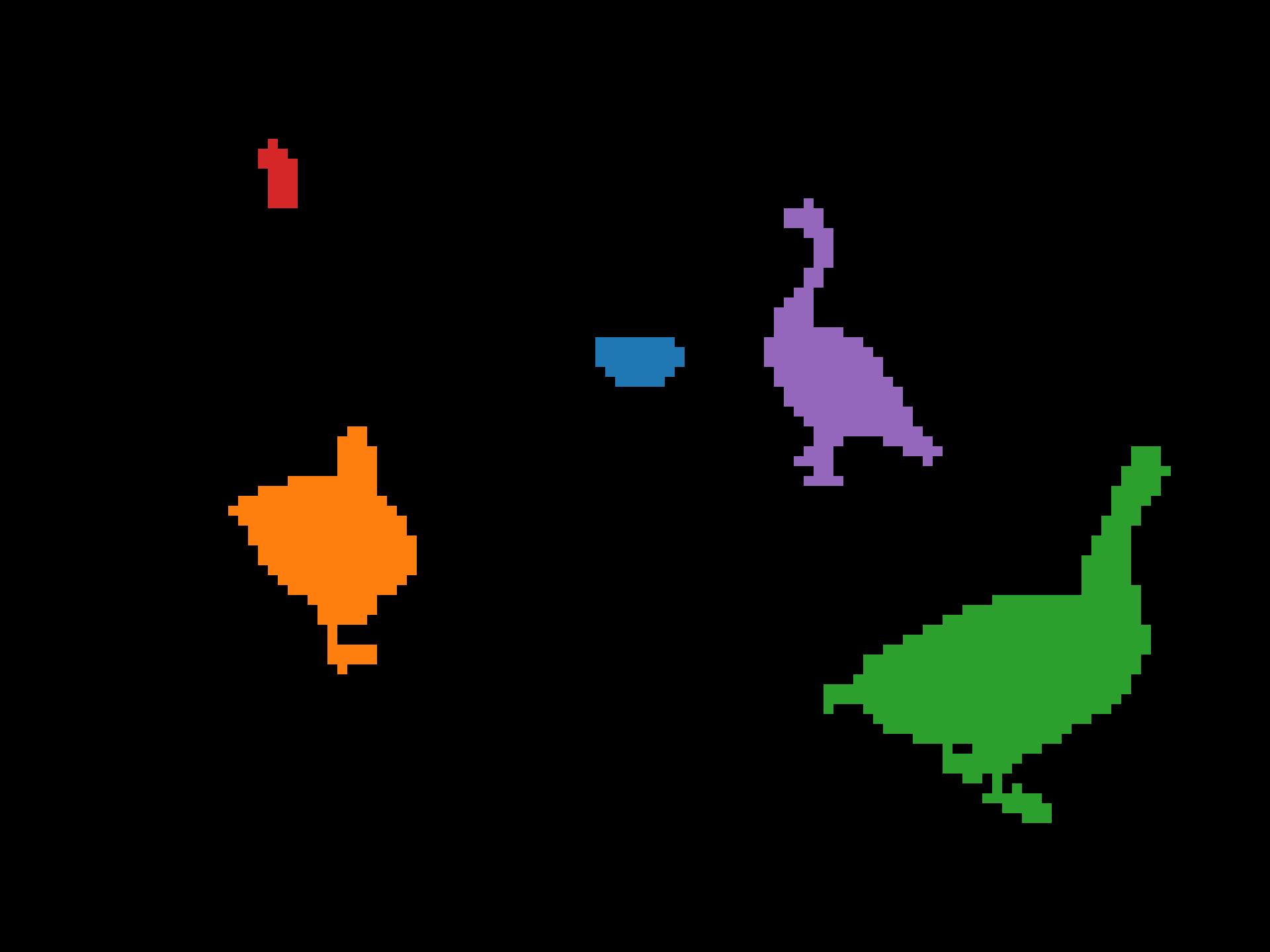}
        \caption*{Frame 30}
    \end{subfigure}\hfill
    \begin{subfigure}[b]{0.195\textwidth}
        \centering
        \includegraphics[width=\textwidth]{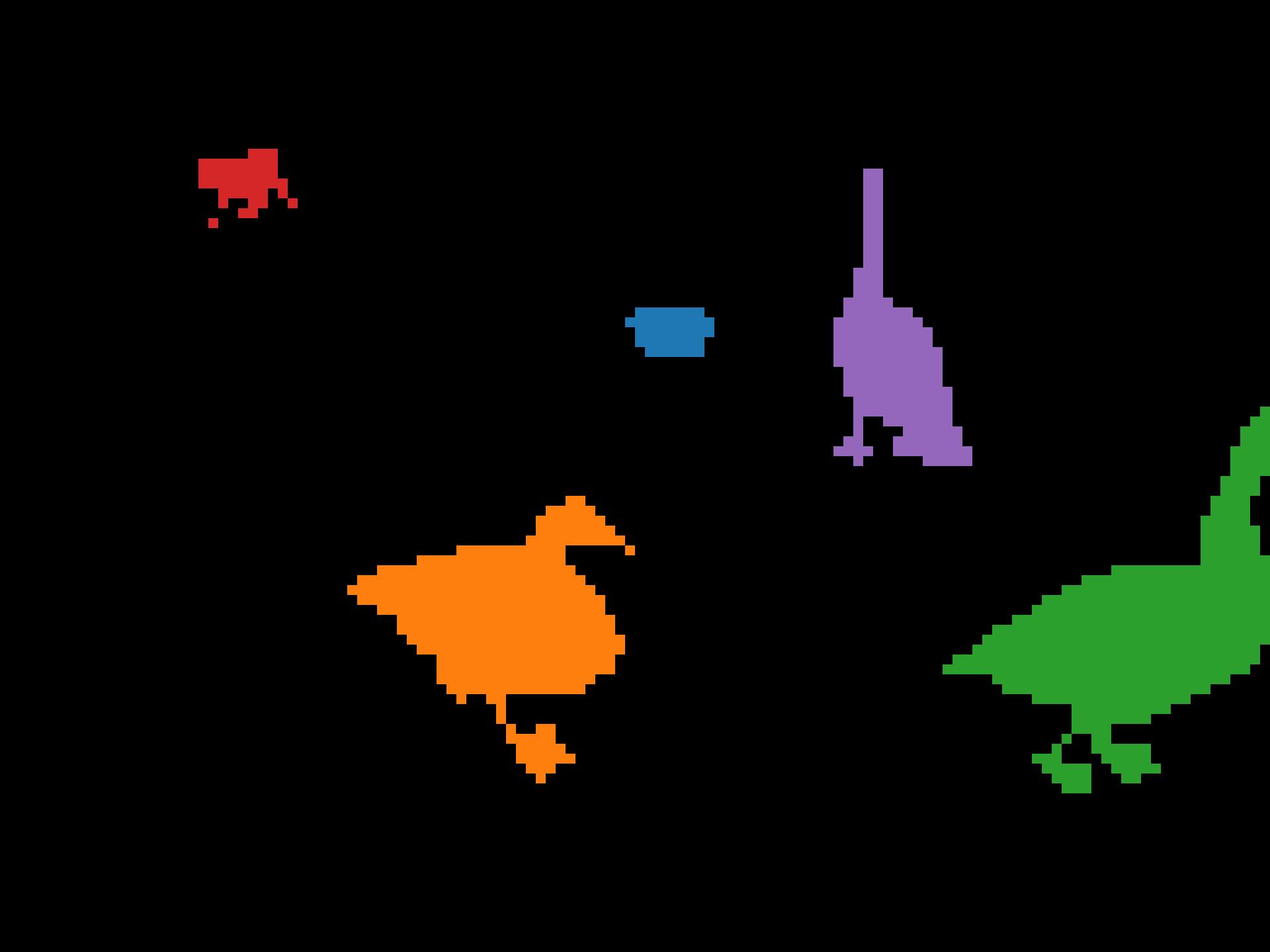}
        \caption*{Frame 60}
    \end{subfigure}\hfill
    \begin{subfigure}[b]{0.195\textwidth}
        \centering
        \includegraphics[width=\textwidth]{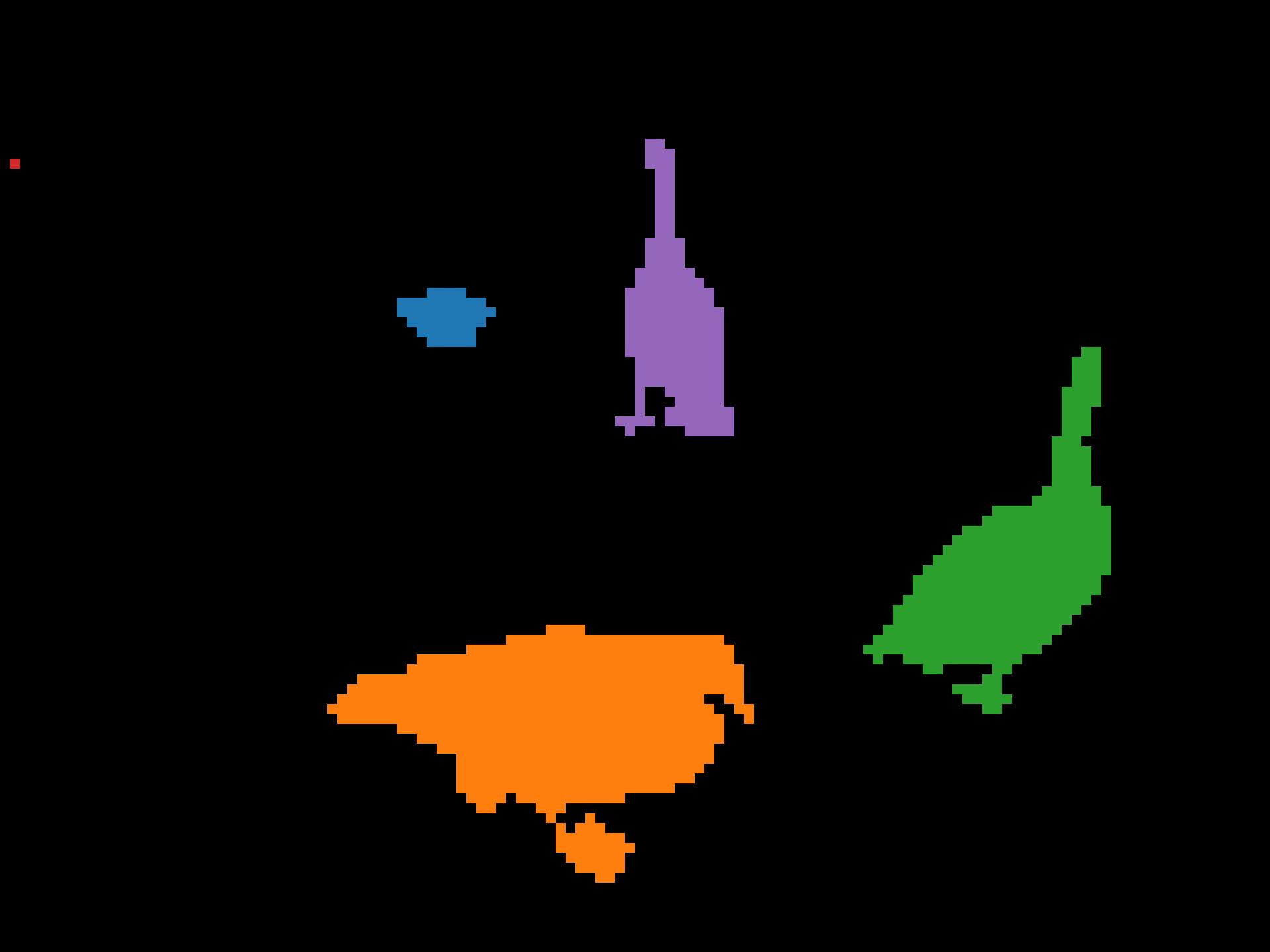}
        \caption*{Frame 90}
    \end{subfigure}\hfill
    \begin{subfigure}[b]{0.195\textwidth}
        \centering
        \includegraphics[width=\textwidth]{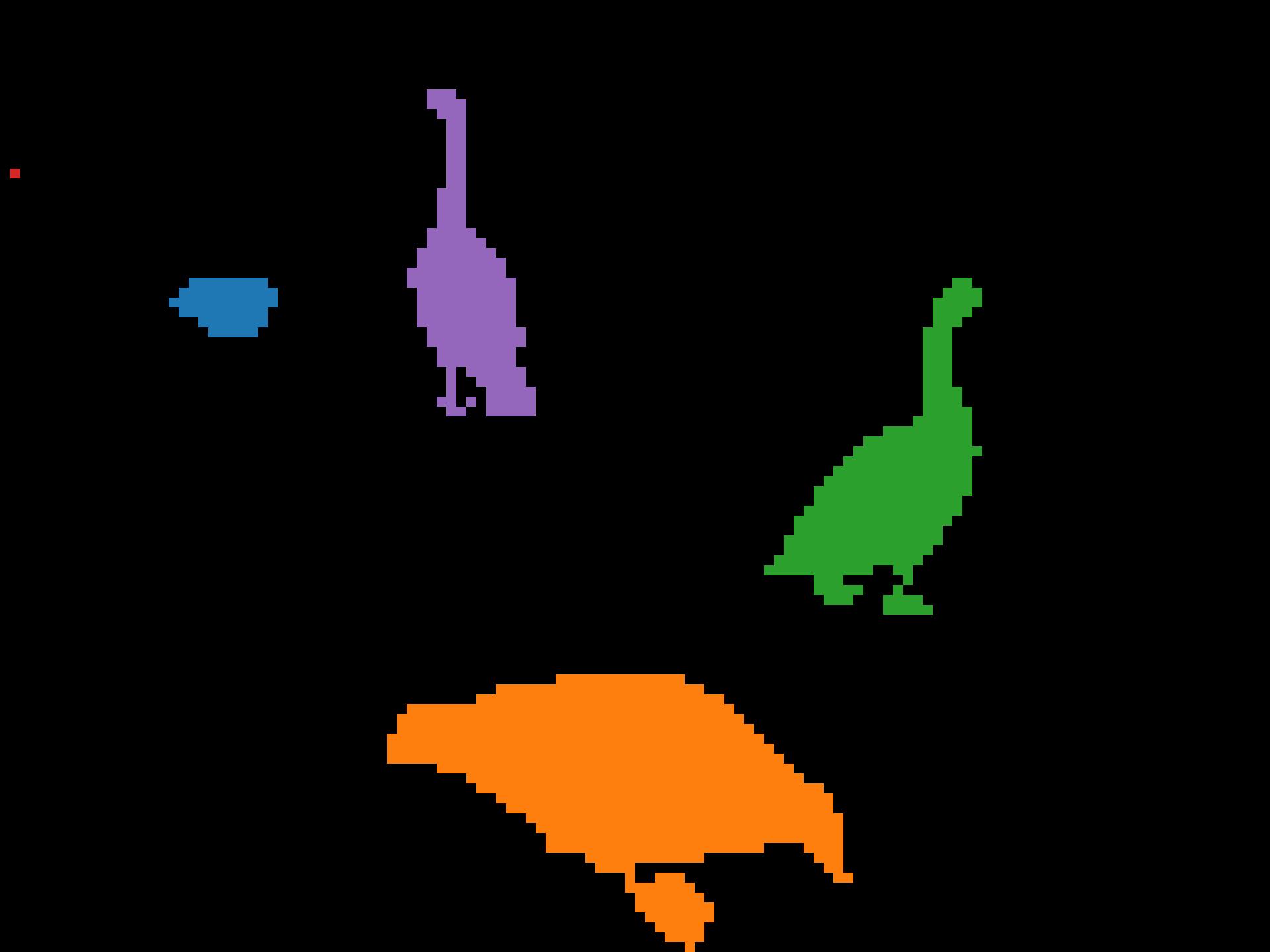}
        \caption*{Frame 120}
    \end{subfigure}
    \caption{
        Segmentation tracking example.
        Given the ground-truth instance segmentation masks for the initial frame, we propagate the instance labels to subsequent frames according to patch similarity in the feature space of DINOv3.
        The input resolution is \(2048{\times}1536\) pixels, resulting in \(128{\times}96\) patches.
    }
    \label{fig:segmentation-tracking-example}
\end{figure}

\paragraph[Results]{Results (\cref{tab:results-video-segmentation-tracking})} 
Aligned with all previous results, weakly-supervised backbones do not deliver convincing performance. 
PEspatial, distilled from the video model SAMv2, provides satisfactory performance, surpassing DINOv2 on smaller resolutions, but falling short on larger ones.
Across resolutions, DINOv3 outperforms all competitors, with a staggering $83.3$ $\JnF$ on DAVIS-L, $6.7$ points above DINOv2.
Furthermore, performance as a function of resolution follows a healthy trend, confirming that our model is able to make use of more input pixels to output precise, high-resolution feature maps (\cf \cref{fig:intro:dense-quality,fig:visualization-extreme-resolutions}).
In contrast, performance at higher resolutions stays almost flat for SigLIP~2 and PEcore, and degrades for PEspatial.
Interestingly, our image model, without any tuning on video, allows to properly track objects in time (see \cref{fig:segmentation-tracking-example}).
This makes it a great candidate to embed videos, allowing to build strong video models on top.

\subsubsection{Video Classification}
\label{sec:results-video-classification}

The previous results have shown the low-level temporal consistency of DINOv3's representations, allowing to accurately track objects in time. 
Going beyond, we evaluate in this section the suitability of its dense features for high-level video classification.
Similar to the setup of V-JEPA 2~\citep{assran2025vjepa2}, we train an \emph{attentive probe}---a shallow 4-layer transformer-based classifier---on top of patch features extracted from each frame.
This enables reasoning over temporal and spatial dimensions as the features are extracted independently per frame.
During evaluation, we either take a single clip per video, or use test-time augmentation (TTA) by averaging the predictions of 3 spatial and 2 temporal crops per video.
See \cref{app:exp-details:video-classification} for experimental details.
We run this evaluation on three datasets: UCF101~\citep{soomro2012ucf101}, Something-Something V2~\citep{goyal2017something}, and Kinetics-400~\citep{kay2017kinetics}, and report top-1 accuracy.
As an additional baseline, we report the performance of V-JEPA v2, a state-of-the-art SSL model for video understanding.

\begin{table}[t]
    \centering
    \small
    \caption{
        Video classification evaluation using attentive probes.
        We report top-1 accuracy on UCF101, Something-Something V2 (SSv2), and Kinetics-400 (K400).
        For each model, we report performance for evaluating a single clip per video, or applying test-time augmentation (TTA) by averaging the predicted probabilities from multiple clips.
    }
    \label{tab:results-video-classification}
    \begin{tabular}{@{}ll cc cc cc@{}}
        \toprule
        & & \multicolumn{2}{c}{UCF101} & \multicolumn{2}{c}{SSv2} & \multicolumn{2}{c}{K400} \\
        \cmidrule(lr){3-4} \cmidrule(lr){5-6} \cmidrule(lr){7-8}
        Method & ViT & Single & TTA & Single & TTA & Single & TTA \\
        \midrule
        \multicolumn{2}{@{}l}{\resultsTableHeaderAgg} & & & & & & \\
        AM-RADIOv2.5  & g/14  & 92.8 & 92.5 & 69.1 & 70.0 & 84.8 & 85.2 \\
        PEspatial    & G/14  & 92.7 & 92.8 & 66.4 & 68.4 & 83.5 & 84.8 \\
        \midrule
        \multicolumn{2}{@{}l}{\resultsTableHeaderWeakly} & & & & & & \\
        SigLIP~2    & g/16  & 93.6 & \bf 94.2 & 68.8 & 70.2 & 86.9 & 87.7 \\
        PEcore    & G/14  & 93.1 & 93.3 & 69.0 & 70.4 & \bf 87.9 & \bf 88.8 \\
        \midrule
        \multicolumn{2}{@{}l}{\resultsTableHeaderSelf} & & & & & & \\
        DINOv2     & g/14  & 93.5 & 93.8 & 67.4 & 68.4 & 84.4 & 85.6 \\
        V-JEPA 2   & g/16  & \bf 94.0 & 93.8 & \bf 73.8 & \bf 75.4 & 83.3 & 84.3 \\
        Web-DINO   & 7B/14 & 93.9 & 94.1 & 67.3 & 68.1 & 86.8 & 87.2 \\
        \midrule
        DINOv3     & 7B/16 & 93.5 & 93.5 & 70.1 & 70.8 & 87.8 & 88.2 \\
        \bottomrule
    \end{tabular}
\end{table}

\paragraph[Results]{Results (\cref{tab:results-video-classification})}
In line with the conclusion of the previous experiment, we find that DINOv3 can be successfully used for extracting strong video features.
As this evaluation involves training several layers of self-attention, the differences between models are less visible.
However, DINOv3 lands in the same range as PEcore and SigLIP~2, and clearly outperforms other models (DINOv2, AM-RADIO) across datasets.
UCF101 and K400 are appearance-focused, where strong category-level understanding of objects gives most of the performance.
SSv2 on the other hand, requires better understanding of motion---the dedicated video model V-JEPA v2 shines on this dataset.
Interestingly, the gap between DINOv3 and the weakly-supervised models is slightly bigger on this dataset.
This again confirms the suitability of DINOv3 to video tasks.

\subsection{DINOv3 has Robust and Versatile Global Image Descriptors}
\label{sec:results-global-features}

In this section, we evaluate DINOv3's ability to capture global image statistics. 
To this end, we consider classic classification benchmarks using linear probes (\cref{sec:results-classification-linear}) and instance retrieval benchmarks (\cref{sec:results-instance-rec}).
Again, we compare to the strongest publicly available image encoders.
In addition to the models from the previous section, we evaluate the two weakly supervised models AIMv2~\citep{fini2024multimodal}, trained using joint auto-regressive pixel and text prediction, and the massive EVA-CLIP-18B~\citep{sun2024eva}.

\subsubsection{Image Classification with Linear Probing}
\label{sec:results-classification-linear}

We train a linear classifier on top of DINOv3's output CLS token to evaluate the model on classification benchmarks.
We consider the ImageNet1k~\citep{deng2009imagenet} dataset and its variants to evaluate out-of-distribution robustness, and a suite of datasets from different domains to understand DINOv3's ability to distinguish fine-grained classes.
See \cref{app:exp-details:classification} for evaluation details.

\paragraph[Domain Generalization from ImageNet]{Domain Generalization from ImageNet (\cref{tab:results-imagenet-classification}) }
In this experiment, we train on ImageNet-\emph{train}, use ImageNet-\emph{val} as a \emph{validation set} to select hyperparameters, and transfer the best found classifier to different test datasets: ImageNet-\textbf{V2}~\citep{recht2019imagenet} and \textbf{ReaL}~\citep{beyer2020imagenetreal} are alternative sets of images and labels for ImageNet, used to test overfitting on the ImageNet validation set; \textbf{R}endition~\citep{hendrycks2021many} and \textbf{S}ketch~\citep{wang2019learning} show stylized and artificial versions of the ImageNet classes; \textbf{A}dversarial~\citep{hendrycks2021natural} and \textbf{Obj}ectNet~\citep{barbu2019objectnet} contain deliberately-chosen difficult examples; \textbf{C}orruptions~\citep{hendrycks2019benchmarking} measures robustness to common image corruptions.
For reference, we also list linear probing results from \citet{dehghani2023scaling} for ViTs trained using supervised classification on the massive JFT dataset (3B--4B images).
Note that these results follow a slightly different evaluation protocol and are not directly comparable to our results.

DINOv3 significantly surpasses all previous self-supervised backbones, with gains of +10\% on ImageNet-R, +6\% on -Sketch, +13\% on ObjectNet over the previously strongest SSL model DINOv2.
We note that the strongest weakly-supervised models, SigLIP~2 and PE, are now better than the strongest supervised ones (ViT-22B) on hard OOD tasks like ImageNet-A and ObjectNet. 
DINOv3 reaches comparable results on ImageNet-R and -Sketch, and, on the hard tasks ImageNet-A and ObjectNet, is closely behind PE, while exceeding SigLIPv2.
On ImageNet, while validation scores are 0.7--0.9 points behind SigLIPv2 and PE, the performance on the ``cleaner'' test sets -V2 and -ReaL is virtually the same.
Notably, DINOv3 achieves the best robustness to corruptions (ImageNet-C).
All in all, \emph{this is the first time that a SSL model has reached comparable results to weakly- and supervised models on image classification}---a domain which used to be the strong point of (weakly-)supervised training approaches.
This is a remarkable result, given that models like ViT-22B, SigLIP~2, and PE are trained using massive human-annotated datasets.
In contrast, DINOv3 learns purely from images, which makes it feasible to further scale/improve the approach in the future.

\begin{table}[t]
    \centering
    \small
    \caption{
        Classification accuracy of linear probes trained on ImageNet1k with frozen backbones. 
        Weakly- and self-supervised models are evaluated with image resolution adapted to 1024 patch tokens (\ie $448 \times 448$ for patch size 14, $512 \times 512$ for patch size 16).
        For reference, we also list results from \citet{dehghani2023scaling} using a different evaluation protocol (marked with $^\ast$).
    }
    \label{tab:results-imagenet-classification}
    \begin{tabular}{@{}ll c ccc c cc c ccc@{}}
        \toprule
        & && \multicolumn{3}{c}{ImageNet} && \multicolumn{2}{c}{Rendition} && \multicolumn{3}{c}{Hard} \\ 
        \cmidrule{4-6} \cmidrule{8-9} \cmidrule{11-13}
        Method  & ViT  && Val & V2 & ReaL && R & S && A & C $\downarrow$ & Obj. \\
        \midrule
        \multicolumn{2}{@{}l}{\resultsTableHeaderSuper} && & & && & && & & \\
        \citet{zhai2022scaling}$^\ast$ & G/14 && 89.0 & 81.3 & 90.6 && 91.7 & --- && 78.8 & --- & 69.6 \\
        \citet{chen2023pali}$^\ast$ & e/14 && 89.3 & 82.5 & 90.7 && 94.3 & --- && 81.6 & --- & 71.5 \\
        \citet{dehghani2023scaling}$^\ast$ & 22B/14 && 89.5 & 83.2 & 90.9 && 94.3 & --- && 83.8 & --- & 74.3 \\
        \midrule
        \multicolumn{2}{@{}l}{\resultsTableHeaderAgg} && & & && & && & & \\
        AM-RADIOv2.5 & g/14 && 88.0 & 80.2 & 90.3 && 83.8 & 67.1 && 81.3 & 27.1 & 68.4 \\
        \midrule
        \multicolumn{2}{@{}l}{\resultsTableHeaderWeakly} && & & && & && & & \\
        PEcore      & G/14 && \bf 89.3 & \bf 81.6 & 90.4 && \bf 92.2 & \bf 71.9 && \bf 89.0 & 22.7 & \bf 80.2 \\
        SigLIP~2     & g/16 && 89.1 & 81.6 & \bf 90.5 && \bf 92.2 & 71.8 && 84.6 & 30.0 & 78.6 \\
        AIMv2        & 3B/14 && 87.9 & 79.5 & 89.7 && 82.3 & 67.1 && 74.5 & 29.5 & 69.0 \\
        EVA-CLIP     & 18B/14 && 87.9 & 79.3 & 89.5 && 85.2 & 64.0 && 81.6 & 33.0 & 71.9 \\  
        \midrule
        \multicolumn{2}{@{}l}{\resultsTableHeaderSelf} && & & && & && & & \\
        Web-DINO   & 7B/14 && 85.9 & 77.1 & 88.6 && 75.6 & 64.0 && 71.6 & 31.2 & 69.7 \\
        Franca     & g/14   && 84.8 & 75.3 & 89.2 && 67.6 & 49.5 && 56.5 & 40.0 & 54.5 \\
        DINOv2     & g/14   && 87.3 & 79.5 & 89.9 && 81.1 & 65.4 && 81.7 & 24.1 & 66.4 \\
        \midrule
        DINOv3     & 7B/16  && 88.4 & 81.4 & 90.4 && 91.1 & 71.3 && 86.9 & \bf 19.6 & 79.0 \\ 
        \bottomrule
    \end{tabular}
\end{table}

\paragraph[Finegrained Classification]{Finegrained Classification (\cref{tab:results-classification-fine})}

\begin{table}[t]
    \begin{minipage}[b][][b]{.53\linewidth}
        \centering
        \small
        \caption{Finegrained classification benchmarks. 
        Fine-S averages over 12 datasets, see \cref{app:tab:results-classification-vtab} for full results.}
        \label{tab:results-classification-fine}
        \begin{tabular}{@{}l@{\hspace{6pt}}l@{\hspace{8pt}}c@{\hspace{6pt}}c@{\hspace{6pt}}c@{\hspace{6pt}}c@{}}
        \toprule
        Method & ViT & Fine-S & Places & iNat18 & iNat21 \\
        \midrule
        \multicolumn{2}{@{}l}{\resultsTableHeaderAgg} &&&& \\
        AM-RADIOv2.5 & g/14 & 93.9 & 70.2 & 79.0 & 83.7 \\
        \midrule
        \multicolumn{2}{@{}l}{\resultsTableHeaderWeakly} &&&& \\        
        SigLIP~2      & g/16 & 93.7 & 70.5 & 80.7 & 82.7 \\
        PEcore       & G/14 & \bf 94.5 & \bf 71.3 & \bf 86.6 & 87.0 \\
        AIMv2         & 3B/14 & 92.9 & 70.7 & 80.8 & 83.2 \\
        EVA CLIP      & 18B/14 & 92.9 & 71.1 & 80.7 & 83.5 \\
        \midrule
        \multicolumn{2}{@{}l}{\resultsTableHeaderSelf} &&&& \\
        Franca   & g/14  & 87.7 & 64.6 & 61.4 & 70.6 \\
        DINOv2   & g/14  & 92.6 & 68.2 & 80.7 & 86.1 \\
        Web-DINO & 7B/14 & 90.2 & 69.6 & 65.3 & 74.1 \\
        \midrule
        DINOv3   & 7B/16 & 93.0 & 70.0 & 85.6 & \bf 89.8 \\
        \bottomrule
    \end{tabular}
    \end{minipage}%
    \hfill
    \begin{minipage}[b][][b]{.44\linewidth}
        \centering
        \small
        \caption{
            Instance recognition benchmarks. 
            See \cref{app:tab:results-instance-retrieval} for additional metrics.
        }
        \label{tab:results-instance-recognition}
        \setlength{\tabcolsep}{4pt}
        \begin{tabular}{@{}cccc@{}}
        \toprule
        Oxford-H & Paris-H & Met (GAP) & AmsterTime \\  %
        \midrule
        \\
        47.5 & 85.7 & 30.5 & 23.1 \\  %
        \midrule
        \\       
        25.1 & 60.9 & 13.9 & 15.5 \\  %
        32.7 & 68.9 & 10.6 & 23.1 \\  %
        28.8 & 71.4 & 29.5 & 14.6 \\  %
        27.1 & 65.6 &  0.5 & 18.9 \\  %
        \midrule
        \\
        14.3 & 51.6 & 27.2 & 21.1 \\  %
        58.2 & 84.6 & 44.6 & 48.9 \\  %
        31.2 & 80.3 & 35.2 & 30.6 \\  %
        \midrule
        \bf 60.7 & \bf 87.1 & \bf 55.4 & \bf 56.5 \\  %
        \bottomrule
        \end{tabular}
    \end{minipage}
\end{table}

We also measure DINOv3's performance when training linear probes on several datasets for fine-grained classification.
In particular, we report the accuracy on 3 large datasets, namely Places205~\citep{zhou2014learning} for scene recognition, and iNaturalist 2018~\citep{van2018inaturalist} and iNaturalist 2021~\citep{van2021benchmarking}) for detailed plant and animal-species recognition, as well as the average over 12 smaller datasets covering scenes, objects, and textures (as in \citet{oquab2024dinov2}, here termed Fine-S).
See also \cref{app:tab:results-classification-vtab} for individual results on those datasets.

We find that, again, DINOv3 surpasses all previous SSL methods.
It also shows competitive results compared to the weakly-supervised methods, indicating its robustness and generalization capability across diverse finegrained classification tasks.
Notably, DINOv3 attains the highest accuracy on the difficult iNaturalist21 dataset at 89.8\%, outperforming even the best weakly-supervised model PEcore with 87.0\%.

\subsubsection{Instance Recognition}
\label{sec:results-instance-rec}

To evaluate the instance-level recognition capabilities of our model, we adopted a non-parametric retrieval approach. Here, database images are ranked by their cosine similarity to a given query image, using the output CLS token.
We benchmark performance across several datasets: the Oxford and Paris datasets for landmark recognition~\citep{radenovic2018revisiting}, the Met dataset featuring artworks from the Metropolitan Museum~\citep{ypsilantis2021met}, and AmsterTime, which consists of modern street view images matched to historical archival images of Amsterdam~\citep{yildiz2022amstertime}.
Retrieval effectiveness is quantified using mean average precision for Oxford, Paris, and AmsterTime, and global average precision for Met.
See \cref{app:exp-details:recognition} for more evaluation details.

\paragraph[Results]{Results (\cref{tab:results-instance-recognition,app:tab:results-instance-retrieval})}
Across all evaluated benchmarks, DINOv3 achieves the strongest performance by large margins, \eg improving over the second best model DINOv2 by +10.8 points on Met and +7.6 points on AmsterTime.
On this benchmark, weakly-supervised models are lagging far behind DINOv3, with the exception of AM-RADIO, which is distilled from DINOv2 features.
These findings highlight the robustness and versatility of DINOv3 for instance-level retrieval tasks, spanning both traditional landmark datasets and more challenging domains such as art and historical image retrieval.

\subsection{DINOv3 is a Foundation for Complex Computer Vision Systems}
\label{sec:results-system-level}

The previous two sections already provided solid signal for the quality of DINOv3 in both dense and global tasks. 
However, these results were obtained under ``model probing'' experimental protocols, using lightweight linear adapters or even non-parametric algorithms to assess the quality of features.
While such simple evaluations allowed to remove confounding factors from involved experimental protocols, they are not enough to evaluate the full potential of DINOv3 as a foundational component in a larger computer vision system.
Thus, in this section, we depart from the lightweight protocols, and instead train more involved downstream decoders and consider stronger, task-specific baselines.
In particular, we use DINOv3 as a basis for 
(1) object detection with Plain-DETR (\cref{sec:results-detection}), 
(2) semantic segmentation with Mask2Former (\cref{sec:results-segmentation}), 
(3) monocular depth estimation with Depth Anything (\cref{sec:results-depth-estimation}), 
and (4) 3D understanding with the Visual Geometry Grounded Transformer (\cref{sec:results-vggt}).
These tasks are only intended as explorations for what is possible with DINOv3.
Still, we find that building on DINOv3 unlocks competitive or even state-of-the-art results with little effort.

\subsubsection{Object Detection}
\label{sec:results-detection}
As a first task, we tackle the long-standing computer vision problem of object detection.
Given an image, the goal is to provide bounding boxes for all instances of objects of pre-defined categories.
This task requires both precise localization and good recognition, as boxes need to match the object boundaries and correspond to the correct category.
While performance on standard benchmarks like COCO~\citep{Lin2014cocodataset} is mostly saturated, we propose to tackle this task with a \emph{frozen} backbone, only training a small decoder on top.

\paragraph{Datasets and Metrics}
We evaluate DINOv3 on object detection capabilities with the COCO dataset~\citep{Lin2014cocodataset}, reporting results on the COCO-VAL2017 split.
Additionally, we evaluate out-of-distribution performance on the COCO-O evaluation dataset~\citep{mao2023cocoo}. 
This dataset contains the same classes but provides input images under six distribution shift settings.
For both datasets, we report mean Average Precision (mAP) with IoU thresholds in \([0.5:0.05:0.95]\).
For COCO-O, we additionally report the effective robustness (ER).
Since COCO is a small dataset, comprising only 118k training images, we leverage the larger Objects365 dataset~\citep{shao2019objects365} for pre-training the decoder, as is common practice.

\paragraph{Implementation}
We build upon the Plain-DETR~\citep{lin2023plaindetr}, but make the following modification:
We do not fuse the transformer encoder into the backbone, but keep it as a separate module, similar to the original DETR~\citep{carion2020detr}, which allows us to keep the DINOv3 backbone completely frozen during training and inference.
To the best of our knowledge, this makes it \emph{the first competitive detection model to use a frozen backbone}.
We train the Plain-DETR detector on Objects365 for 22 epochs at resolution 1536, then one epoch at resolution 2048, followed by 12 epochs on COCO at resolution 2048.
At inference time, we run at resolution 2048.
Optionally, we also apply test-time augmentation (TTA) by forwarding the image at multiple resolutions (from 1536 to 2880).
See \cref{app:exp-details:detection} for full experimental details.

\paragraph[Results]{Results (\cref{tab:results-object-detection})}
We compare our system with four models: 
EVA-02 with a Cascade detector~\citep{fang2024eva},
EVA-02 with Co-DETR~\citep{zong2023detrs},
InternImage-G with DINO~\citep{wang2022internimage},
and PEspatial with DETA~\citep{bolya2025perception}.
We find that our lightweight detector (100M parameters) trained on top of a frozen DINOv3 backbone manages to reach state-of-the-art performance.
For COCO-O, the gap is pronounced, showing that the detection model can effectively leverage the robustness of the DINOv3.
Interestingly, our model outperforms all previous models with much fewer trained parameters, with the smallest comparison point still using more than 300M trainable parameters.
We argue that achieving such strong performance without specializing the backbone is an enabler for various practical applications: A single backbone forward can provide features that support multiple tasks, reducing compute requirements.

\begin{table}[t]
    \centering
    \small
    \caption{
        Comparison with state-of-the-art systems on object detection.
        We train a detection adapter on top of a \emph{frozen} DINOv3 backbone.
        We show results on the validation set of the COCO and COCO-O datasets, and report the mAP across IoU thresholds, as well as the effective robustness (ER).
        Our detection system based on DINOv3 sets a new state of the art. 
        As the InternImage-G detection model has not been released, we were unable to reproduce their results or compute COCO-O scores.
    }
    \label{tab:results-object-detection}
    \begin{tabular}{@{}ll c ccc cc cc@{}}
    \toprule
    && & \multicolumn{3}{c}{Parameters} & \multicolumn{2}{c}{COCO} & \multicolumn{2}{c}{COCO-O} \\
    \cmidrule(lr){4-6} \cmidrule(lr){7-8} \cmidrule(lr){9-10}
    Model    & Detector & FT & Encoder & Decoder & Trainable & Simple & TTA &  mAP & ER \\
    \midrule
    EVA-02          & Cascade   & \orangefire & 300M    & ---   &  300M          & 64.1 & ---    & 63.6 & 34.7 \\
    InternImage-G   & DINO      & \orangefire & 6B      & ---   & 6B            & 65.1 & 65.3   & --- & --- \\
    EVA-02          & Co-DETR   & \orangefire & 300M    & ---   & 300M            & 65.4 & 65.9   & 63.7 & 34.3 \\
    PEspatial      & DETA      & \orangefire & 1.9B    & \phantom{0}50M & 2B    & 65.3 & 66.0   & 64.0 & 34.7 \\
    \midrule
    DINOv3          & Plain-DETR & \bluesnow  & \phantom{00}7B    & 100M & 100M & \textbf{65.6} & \textbf{66.1} & \textbf{66.4} &  \textbf{36.8} \\
    \bottomrule
\end{tabular}
\end{table}

\subsubsection{Semantic Segmentation}
\label{sec:results-segmentation}
Following the previous experiment, we now evaluate on semantic segmentation, another long-standing computer vision problem.
This task also requires strong, well localized representations, and expects a dense per-pixel prediction.
However, opposed to object detection, the model does not need to differentiate instances of the same object.
Similar to detection, we train a decoder on top of a \emph{frozen} DINOv3 model.

\paragraph{Datasets and Metrics}
We focus our evaluation on the ADE20k dataset~\citep{zhou2017scene}, which contains 150 semantic categories across 20k training images and 2k validation images.
We measure performance using the mean Intersection over Union (mIoU).
To train the segmentation model, we additionally use the COCO-Stuff \citep{caesar2018coco} and Hypersim \citep{roberts2021hypersim} datasets.
Those contain 164k images with 171 semantic categories, and 77k images with 40 categories respectively. 

\paragraph{Implementation}
To build a decoder that maps DINOv3 features to semantic categories, we combine ViT-Adapter \citep{chen2022vitadapter} and Mask2Former \citep{cheng2021mask2former}, similar to prior work~\citep{wang2022image,wang2022internimage,wang2023onepeace}.
However, in our case, the DINOv3 backbone remains frozen during training.
In order to avoid altering the backbone features, we further modify the original ViT-Adapter architecture by removing the injector component.
Compared to baselines, we also increase the embedding dimensions from $1024$ to $2048$, to support processing the $4096$-dimensional output of the DINOv3 backbone.
We start by pre-training the segmentation decoder on COCO-Stuff for $80$k iterations, followed by $10$k iterations on Hypersim \citep{roberts2021hypersim}.
Finally, we train for 20k iterations on the training split of ADE20k and report results on the validation split. 
All training is done at an input resolution of 896.
At inference time we consider two setups: single-scale, \ie we forward images at training resolution, or multi-scale, \ie we average predictions at multiple image ratios between \({\times}0.9\) and \(1.1\) the original training resolution.
We refer to \cref{app:exp-details:segmentation-system} for more experimental details.

\paragraph[Results]{Results (\cref{tab:results-for-ade20k})}
We compare our model's performance with several state-of-the-art baselines, including 
BEIT-3 \citep{wang2022image}, 
InternImage-H \citep{wang2022internimage} and 
ONE-PEACE \citep{wang2023onepeace}, and report results on additional datasets in \cref{app:tab:semantic-segmentation-other-datasets}.
Our segmentation model based on the frozen DINOv3 backbone reaches state-of-the-art performance, equaling that of ONE-PEACE ($63.0$ mIoU).
It also improves over all prior models on the COCO-Stuff~\citep{caesar2018coco} and VOC 2012~\citep{pascal-voc-2012} datasets.
As semantic segmentation requires accurate per-pixel predictions, vision transformer backbones pose a fundamental problem.
Indeed, the 16 pixel-wide input patches make the granularity of the prediction relatively coarse---encouraging solutions like ViT-Adapter. 
On the other hand, we have shown that we can obtain high-quality feature maps, even at very high resolutions up to 4096 (\cf \cref{fig:intro:dense-quality,fig:visualization-extreme-resolutions}); this corresponds to dense feature maps 512-tokens wide.
We hope that future work will be able to leverage these high-resolution features to reach state-of-the-art performance without having to rely on heavy decoders like ViT-Adapter with Mask2Former.

\begin{table}[t]
    \centering
    \small
    \caption{
        Comparison with state-of-the-art systems for semantic segmentation on ADE20k.
        We evaluate the model in a single- or multi-scale setup (respectively Simple and TTA).
        Following common practice, we run this evaluation at resolution 896 and report mIoU scores.
        BEIT3, ONE-PEACE and DINOv3 use a Mask2Former with ViT-Adapter architecture, and the decoder parameters take into account both.
        We report results on further datasets in \cref{app:tab:semantic-segmentation-other-datasets}
    }
    \begin{tabular}{lc ccc cc}
    \toprule
    & & \multicolumn{3}{c}{Parameters} &  \multicolumn{2}{c}{mIoU} \\
    \cmidrule(lr){3-5} \cmidrule(lr){6-7} 
    Model   &   FT       & Encoder & Decoder & Trainable &  Simple & TTA \\
    \midrule
    BEIT3           & \orangefire & 1.0B  & 550M & 1.6B &  62.0 & 62.8 \\  %
    InternImage-H   & \orangefire & 1.1B  & 230M & 1.3B &  62.5 & 62.9 \\  %
    ONE-PEACE       & \orangefire & 1.5B  & 710M & 2.2B &  62.0 & \textbf{63.0} \\  %
    \midrule
    DINOv3          & \bluesnow   & \phantom{0.}7B    & 927M & 927M &  \textbf{62.6} & \textbf{63.0} \\
    \bottomrule
    \end{tabular}
    \label{tab:results-for-ade20k}
\end{table}

\subsubsection{Monocular Depth Estimation}
\label{sec:results-depth-estimation}

We now consider building a system for monocular depth estimation.
To do so, we follow the setup of Depth Anything V2 (DAv2)~\citep{yang2024depthanythingv2}, a recent state-of-the-art method.
The key innovation of DAv2 is to use a large collection of synthetically generated images with ground truth depth annotations.
Critically, this relies on DINOv2 as a feature extractor that is able to bridge the \emph{sim-to-real} gap, a capability that other vision backbones like SAM~\citep{kirillov2023segment} do not show~\citep{yang2024depthanythingv2}.
Thus, we swap DINOv2 with DINOv3 in the DAv2 pipeline to see if we can achieve similar results.

\paragraph{Implementation}
Like DAv2, we use a Dense Prediction Transformer (DPT)~\citep{ranftl2021vision} to predict a pixelwise depth field, using features from four equally spaced layers of DINOv3 as input.
We train the model using the set of losses from DAv2 on DAv2's synthetic dataset, increasing the training resolution to $1024 \times 768$ to make use of DINOv3's high resolution capabilities.
In contrast to DAv2, we \emph{keep the backbone frozen} instead of finetuning it, testing the out-of-the-box capabilities of DINOv3.
We also found it beneficial to scale up the DPT head to obtain the full potential DINOv3 7B's larger features.
See \cref{app:exp-details:depth-system} for details.

\paragraph{Datasets and Metrics}
We evaluate our model on 5 real-world datasets (NYUv2~\citep{silberman2012indoor}, KITTI~\citep{geiger2013vision}, ETH3D~\citep{schoeps2017multiview}, ScanNet (from \citet{ke2025marigold}) and DIODE~\citep{vasiljevic2019diode}) in the zero-shot scale-invariant depth setup, similar to \citet{ranftl2020towards,ke2025marigold,yang2024depthanythingv2}.
We report the standard metrics absolute relative error (ARel) (lower is better) and $\delta_1$ (higher is better).
We refer to~\citet{yang2024depthanythingv1} for a description of those metrics.

\paragraph[Results]{Results (\cref{tab:results-depth-estimation-system-level})}
We compare to the state of the art for relative depth estimation: MiDaS~\citep{ranftl2020towards}, LeReS~\citep{yin2021learning}, Omnidata~\citep{eftekhar2021omnidata}, DPT~\citep{ranftl2021vision}, Marigold in the ensemble version~\citep{ke2025marigold} and DAv2.
Our depth estimation model reaches a new state-of-the-art on all datasets, only lacking behind in ARel on DIODE compared to DPT.
Remarkably, this is possible using a \emph{frozen backbone}, whereas all other baselines need to finetune the backbone for depth estimation.
In addition, this validates that DINOv3 inherits DINOv2's \emph{strong sim-to-real capabilities}, a desirable property that opens up the possibility for downstream tasks to use synthetically generated training data.

\begin{table}[t]
  \centering
  \small
  \caption{
    Comparison with state-of-the-art systems for relative monocular depth estimation.
    By combining DINOv3 with Depth Anything V2~\citep{yang2024depthanythingv2}, we obtain a SotA model for relative depth estimation.
  }
  \label{tab:results-depth-estimation-system-level}
  \setlength{\tabcolsep}{3.8pt}
  \begin{tabular}{@{}l c c cc c cc c cc c cc c cc@{}}
    \toprule
    & && \multicolumn{2}{c}{NYUv2} && \multicolumn{2}{c}{KITTI} && \multicolumn{2}{c}{ETH3D} && \multicolumn{2}{c}{ScanNet} && \multicolumn{2}{c}{DIODE}\\
    \cmidrule{4-5} \cmidrule{7-8} \cmidrule{10-11} \cmidrule{13-14} \cmidrule{16-17}
    Method & FT && ARel $\downarrow$ & $\delta_1 \uparrow$ && ARel $\downarrow$ & $\delta_1 \uparrow$ && ARel $\downarrow$ & $\delta_1 \uparrow$ && ARel $\downarrow$ & $\delta_1 \uparrow$ && ARel $\downarrow$ & $\delta_1 \uparrow$ \\
    \midrule
    MiDaS     & \orangefire && 11.1 & 88.5 && 23.6 & 63.0 && 18.4 & 75.2 && 12.1 & 84.6 && 33.2 & 71.5 \\
    LeReS     & \orangefire &&  9.0 & 91.6 && 14.9 & 78.4 && 17.1 & 77.7 &&  9.1 & 91.7 && 27.1 & 76.6 \\
    Omnidata  & \orangefire &&  7.4 & 94.5 && 14.9 & 83.5 && 16.6 & 77.8 &&  7.5 & 93.6 && 33.9 & 74.2 \\
    DPT       & \orangefire &&  9.8 & 90.3 && 10.0 & 90.1 &&  7.8 & 94.6 &&  8.2 & 93.4 && \textbf{18.2} & 75.8 \\
    Marigold  & \orangefire &&  5.5 & 96.4 &&  9.9 & 91.6 &&  6.5 & 96.0 &&  6.4 & 95.1 && 30.8 & 77.3 \\
    DAv2 (ViT-g) & \orangefire  &&  4.4 & 97.9 &&  7.5 & 94.7 && 13.1 & 86.5 && ---   & ---   && ---   & ---   \\
    \midrule
    DINOv3   & \bluesnow  &&  \textbf{4.3} & \textbf{98.0} &&  \textbf{7.3} & \textbf{96.7} &&  \textbf{5.4} & \textbf{97.5} &&  \textbf{4.4} & \textbf{98.1} && 25.6 & \textbf{82.2} \\
  \bottomrule
  \end{tabular}
\end{table}

\subsubsection{Visual Geometry Grounded Transformer with DINOv3}
\label{sec:results-vggt}

Finally, we consider 3D understanding with the recent Visual Geometry Grounded Transformer (VGGT)~\citep{wang2025vggt}.
Trained on a large set of 3D-annotated data, VGGT learns to estimate all important 3D attributes of a scene, such as camera intrinsics and extrinsics, point maps, or depth maps, in a single forward pass.
Using a simple, unified pipeline, it reaches state-of-the-art results on many 3D tasks while being more efficient than specialized methods---constituting a major advance in 3D understanding.

\paragraph{Implementation}
VGGT uses a DINOv2-pretrained backbone to obtain representations for different views of a scene, before fusing them with a transformer.
Here, we simply swap the DINOv2 backbone with DINOv3, using our ViT-L variant (see \cref{sec:family}) to match DINOv2 ViT-L/14 in the original work.
We run the same training pipeline as VGGT, including finetuning of the image backbone.
We switch the image resolution from $518 \times 518$ to $592 \times 592$ to accommodate DINOv3's patch size 16 and keep the the results comparable to VGGT. We additionally adopt a small number of hyperparameter changes detailed in \cref{app:exp-details:vggt}.

\begin{table}[t]
    \caption{
        3D understanding using Visual Geometry Grounded Transformer (VGGT)~\citep{wang2025vggt}. 
        Simply by swapping DINOv2 for DINOv3 ViT-L as the image feature extractor in the VGGT pipeline, we are able to obtain state-of-the-art results on various 3D geometry tasks.
        We reproduce baseline results from~\citet{wang2025vggt}. %
        We also report methods using ground truth camera information, marked with $^\ast$. 
        Camera pose estimation results are reported with AUC@30.  
    }
    \label{tab:results-vggt}
    \begin{subtable}{0.3\linewidth}
        \centering
        \small
        \caption{Camera pose estimation.}
        \label{tab:results-vggt-camera-pose}
        \setlength{\tabcolsep}{3pt}
        \begin{tabular}{@{}lcc@{}}
        \toprule
        Method & Re10K & CO3Dv2 \\
        \midrule
        DUSt3R & 67.7 & 76.7 \\
        MASt3R & 76.4 & 81.8 \\
        VG GSfM v2 & 78.9 & 83.4 \\
        CUT3R & 75.3 & 82.8 \\
        FLARE & 78.8 & 83.3 \\
        VGGT & 85.3 & 88.2 \\
        \midrule
        DINOv3 & \bf 86.3 & \bf 89.6 \\
        \bottomrule
        \end{tabular}
    \end{subtable}%
    \hfill
    \begin{subtable}{0.35\linewidth}
        \centering
        \caption{Multi-view estimation on DTU.}
        \label{tab:results-vggt-mvs}
        \setlength{\tabcolsep}{2pt}
        \small
        \begin{tabular}{@{}lccc@{}}
        \toprule
        Method & Acc.$\downarrow$ & Comp.$\downarrow$ & Overall$\downarrow$ \\
        \midrule
        \textcolor{gray}{Gipuma$^\ast$} & \textcolor{gray}{0.283} & \textcolor{gray}{0.873} & \textcolor{gray}{0.578} \\
        \textcolor{gray}{CIDER$^\ast$} & \textcolor{gray}{0.417} & \textcolor{gray}{0.437} & \textcolor{gray}{0.427} \\
        \textcolor{gray}{MASt3R$^\ast$} & \textcolor{gray}{0.403} & \textcolor{gray}{0.344} & \textcolor{gray}{0.374} \\
        \textcolor{gray}{GeoMVSNet$^\ast$} & \textcolor{gray}{0.331} & \textcolor{gray}{0.259} & \textcolor{gray}{0.295} \\
        DUSt3R & 2.677 & 0.805 & 1.741 \\
        VGGT & 0.389 & 0.374 & 0.382 \\
        \midrule
        DINOv3  & \bf 0.375 & \bf 0.361 & \bf 0.368 \\
        \bottomrule
        \end{tabular}
    \end{subtable}%
    \hfill
    \begin{subtable}{0.315\linewidth}
        \centering
        \small
        \caption{View matching on ScanNet-1500.}
        \label{tab:results-vggt-view-matching}
        \setlength{\tabcolsep}{3pt}
        \begin{tabular}{@{}lcc@{}}
        \toprule
        Method & AUC@5 & AUC@10 \\
        \midrule
        SuperGlue & 16.2 & 33.8 \\
        LoFTR & 22.1 & 40.8 \\
        DKM & 29.4 & 50.7 \\
        CasMTR & 27.1 & 47.0 \\
        Roma & 31.8 & 53.4 \\
        VGGT & 33.9 & 55.2 \\
        \midrule
        DINOv3 & \bf 35.2 & \bf 56.1 \\
        \bottomrule
        \end{tabular}
    \end{subtable}
\end{table}

\paragraph{Datasets and Metrics}
Following \citet{wang2025vggt}, we evaluate on camera pose estimation on the Re10K~\citep{zhou2018stereo} and CO3Dv2~\citep{reizenstein2021common} datasets, dense multi-view estimation on DTU~\citep{jensen2014large}, and two-view matching on ScanNet-1500~\citep{dai2017scannet}.
For camera pose estimation and two-view matching, we report the standard area-under-curve (AUC) metric.  %
For multi-view estimation, we report the smallest L2-distance between prediction to ground truth as ``Accuracy'', the smallest L2-distance from ground truth to prediction as ``Completeness'' and their average as `Overall''.
We refer to \citet{wang2025vggt} for details about method and evaluation.

\paragraph[Results]{Results (\cref{tab:results-vggt})}
We find that VGGT equipped with DINOv3 \emph{further improves over the previous state-of-the-art} set by VGGT on all three considered tasks---using DINOv3 leads to clear and consistent gains.
This is encouraging, given that we only applied minimal tuning for DINOv3.
These tasks span different levels of visual understanding: high-level abstraction of scene content (camera pose estimation), dense geometric prediction (multi-view depth estimation), and fine-grained pixel-level correspondence (view matching).
Together with the previous results on correspondence estimation (\cref{sec:results-correspondence-estimation}) and depth estimation (\cref{sec:results-depth-estimation}), we take this as further empirical evidence for the strong suitability of DINOv3 as a basis for 3D tasks.
Additionally, we anticipate further improvements from using the larger DINOv3 7B model. 

\section{Evaluating the Full Family of DINOv3 Models}
\label{sec:family}
\begin{figure}[t]
    \centering
    \begin{subfigure}[b]{0.5\textwidth}
        \centering
        \begin{tabular}{l r rr}
            \toprule
             &  & \multicolumn{2}{c}{Inference GFLOPs} \\
             \cmidrule{3-4}
            Model & \#Params & Res.~256 & Res.~512 \\
             \midrule
            CNX-Tiny & 29M & 5 & 20  \\
            CNX-Small & 50M & 11 & 46 \\
            CNX-Base & 89M  & 20 & 81 \\
            CNX-Large & 198M & 38 & 152  \\
             \midrule
            ViT-S &  21M & 12 & 63 \\
            ViT-S+ & 29M & 16 & 79 \\
            ViT-B &  86M  & 47 & 216 \\
            ViT-L &  300M & 163 & 721 \\
            ViT-H+ & 840M  & 450 & 1903 \\
             \midrule
            ViT-7B & 6716M & 3550 & 14515 \\
            \bottomrule
        \end{tabular}
        \caption{DINOv3 family of models.}
        \label{tab:family-models-flops}
    \end{subfigure}
    \hfill
    \begin{subfigure}[b]{0.48\textwidth}    
        \centering 
        \includegraphics{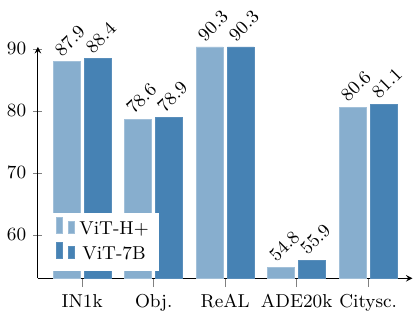} 
        \caption{ViT-H+ v.s. ViT-7B.}
        \label{fig:family:vith-7b}
    \end{subfigure}
    \caption{
        (a) Presentation of the distilled models' characteristics. 
        CNX stands for ConvNeXT. We present per model the number of parameters and the GFLOPs estimated on images of size $256\times256$ and $512\times512$. 
        (b) We compare DINOv3 ViT-H+ to its 7B-sized teacher; despite having almost 10$\times$ less parameters, the ViT-H+ is close to DINOv3 7B in performance.
    }
\end{figure}

In this section, we provide quantitative evaluations on the family of models distilled from our 7B-parameters model (See \cref{sec:distillation}). 
This family includes variants based on the Vision Transformer (ViT) and the ConvNeXt (CNX) architectures.
We provide the detailed parameter counts and inference FLOPs for all models in \cref{tab:family-models-flops}. 
These models cover a wide range of computational budgets to accommodate a broad spectrum of users and deployment scenarios. 
We conduct a thorough evaluation of all ViT (\cref{sec:family:vits}) and ConvNeXt variants to assess their performance across tasks.

\Cref{fig:intro:family} provides an overview comparison of the DINOv3 family versus other model collections.
The DINOv3 family significantly outperforms all others on dense prediction tasks.
This includes specialized models distilled from supervised backbones like AM-RADIO and PEspatial.
At the same time, our models achieve similar results on classification tasks, making them the optimal choice across compute budgets.

In \cref{sec:family:vits} detail our ViT models and compare them to other open-source alternatives.
Then, in \cref{sec:family:convnext}, we discuss the ConvNeXt models.
Finally, following \cref{sec:alignementtxt}, we trained a text encoder aligned with the output of our ViT-L model.
We present multi-modal alignment results for this model in \cref{sec:family:text}.

\subsection{A Vision Transformer for Every Use Case}
\label{sec:family:vits}
Our ViT family spans architectures from the compact ViT-S to the larger 840 million parameter ViT-H+ models.
The former is designed to run efficiently on resource-constrained devices such as laptops, the latter delivers state-of-the-art performance for more demanding applications.
We compare our ViT models to the best open-source image encoders of corresponding size, namely DINOv2~\citep{oquab2024dinov2}, SigLIP~2~\citep{tschannen2025siglip} and Perception Encoder~\citep{bolya2025perception}. 
For a fair comparison, we ensure that the input sequence length is equivalent across models. 
Specifically, for model with a patch size of 16 we input images of size $512 \times 512$ versus $448 \times 448$ when models are using patch size 14. 

\begin{table}[t]
  \centering
  \small
  \caption{
    Comparison of our family of models against open-source alternatives of comparable size.
    We showcase our ViT-\{S, S+, B, L, H+\} models on a representative set of global and dense benchmarks: classification (IN-ReAL, IN-R, ObjectNet), retrieval (Oxford-H), segmentation (ADE20k), depth (NYU), tracking (DAVIS at 960px), and keypoint matching (NAVI, SPair).
    We match the number of patch tokens for a fair comparison across models of different patch size.
  }
  \label{tab:family:distillation}
  \begin{tabular}{@{}llc cccc c ccccc@{}}
    \toprule
    & && \multicolumn{4}{c}{Global Tasks} && \multicolumn{5}{c}{Dense Tasks} \\
    \cmidrule{4-7} \cmidrule{9-13}
    Size & Model && IN-ReaL & IN-R & Obj. & Ox.-H && ADE20k  & NYU$\downarrow$ & DAVIS & NAVI & SPair \\
    \midrule
    S & DINOv2  && 87.3 & 54.0 & 47.8 & 39.5 && 45.5 & 0.446 & 73.6 & 53.4 & 51.6 \\
    S & DINOv3  && 87.0 & 60.4 & 50.9 & 49.5 && 47.0 & 0.403 & 72.7 & 56.3 & 50.4 \\
    S+ & DINOv3 && 88.0 & 68.8 & 54.6 & 50.0 && 48.8 & 0.399 & 75.5 & 57.1 & 55.2 \\  
    \midrule
    B & PEcore && 87.5 & 68.4 & 57.9 & 20.2 && 37.4 & 0.641 & 44.5 & 41.8 & 13.7 \\ 
    B & SigLIP 2 && 89.3 & 80.6 & 66.9 & 20.2 && 41.6 & 0.512 & 63.2 & 45.4 & 32.8\\
    B & DINOv2  && 89.0 & 68.4 & 57.3 & 51.0 && 48.4 & 0.416 & 72.9 & 56.9 & 57.1 \\
    B & DINOv3  && 89.3 & 76.7 & 64.1 & 58.5 && 51.8 & 0.373 & 77.2 & 58.8 & 57.2 \\ 
    \midrule
    L & PEcore && 90.1 & 87.7 & 74.9 & 25.6 && 39.7 & 0.650 & 48.2 & 42.1 & 19.2 \\
    L & SigLIP 2 && 90.1 & 89.2 & 75.0 & 21.4 && 43.6 & 0.484 & 66.3 & 47.8 & 41.9 \\
    L & DINOv2  && 89.7 & 79.1 & 64.7 & 55.7 && 48.8 & 0.394 & 73.4 & 59.9 & 57.0 \\
    L & DINOv3  && 90.2 & 88.1 & 74.8 & 63.1 && 54.9 & 0.352 & 79.9 & 62.3 & 61.2 \\
    \midrule
    SO400m & SigLIP 2 && 90.3 & 90.4 & 76.2 & 23.0 && 44.0 & 0.402 & 64.8 & 48.8 & 38.7 \\
    \midrule
    H+ & DINOv3 && 90.3 & 90.0 & 78.6 & 64.5 && 54.8 & 0.352 & 79.3 & 63.3 & 56.3 \\ 
    \bottomrule
  \end{tabular}
\end{table}

Our empirical study clearly demonstrates that DINOv3 models consistently outperform their counterparts on dense prediction tasks. 
Most notably, on the ADE20k benchmark, the DINOv3 ViT-L model achieves an improvement of over 6 mIoU points compared to the best competitor DINOv2.
The ViT-B variant shows a gain of approximately 3 mIoU points against the next best competitor. 
These substantial improvements highlight the effectiveness of DINOv3's local features in capturing fine-grained spatial details. 
Furthermore, evaluations on depth estimation tasks also reveal consistent performance gains over competing approaches.
This underscores the versatility of the DINOv3 family across different dense vision problems. 
Importantly, our models achieve competitive results on global recognition benchmarks such as ObjectNet and ImageNet-1k.
This indicates that the enhanced dense task performance does not come at the expense of global task accuracy. 
This balance confirms that DINOv3 models provide a robust and well-rounded solution, excelling across both dense and global vision tasks without compromise.

On another note, we want to also validate if the largest models that we distill capture all the information from the teacher.
To this end, we run a comparison of our largest ViT-H+ with the 7B teacher.
As shown in \cref{fig:family:vith-7b}, the largest student achieves performance that is on par with the 8 times larger ViT-7B model. 
This result not only validates the effectiveness of our distillation process but also demonstrates that, when guided by a high-quality teacher, smaller models can learn to deliver comparable levels of performance. 
This finding reinforces our belief that \emph{training very large models benefits the broader community}.
The strength of larger models can be successfully distilled into more efficient, smaller models with little or no loss of quality.

\begin{figure}[t]
    \centering
    \begin{tikzpicture}[every node/.style={inner sep=0, outer sep=0}]
        \newcommand{\imgHSpace}{0.15cm} %
        \newcommand{\imgVSpace}{0.15cm} %
        \newcommand{\imgSize}{0.21\textwidth}
        \node (img11) {\includegraphics[width=\imgSize]{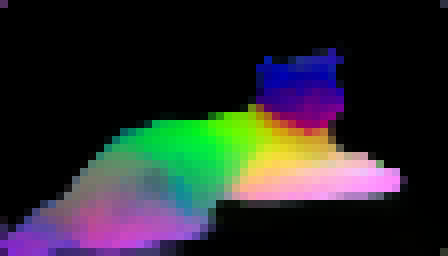}};
        \node[right=\imgHSpace of img11] (img12) {\includegraphics[width=\imgSize]{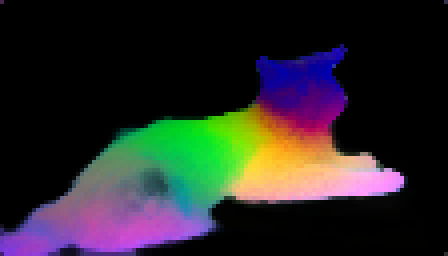}};
        \node[right=\imgHSpace of img12] (img13) {\includegraphics[width=\imgSize]{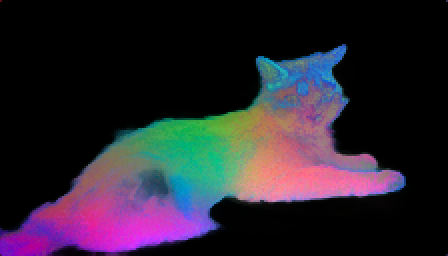}};
        \node[right=\imgHSpace of img13] (img14) {\includegraphics[width=\imgSize]{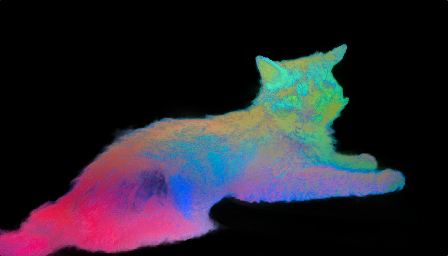}};
        \node[below=\imgVSpace of img11] (img21) {\includegraphics[width=\imgSize]{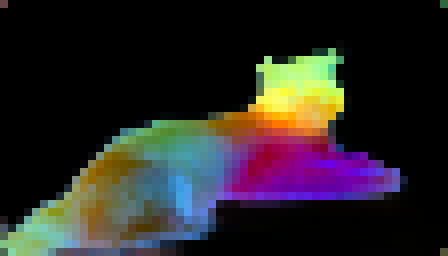}};
        \node[right=\imgHSpace of img21] (img22) {\includegraphics[width=\imgSize]{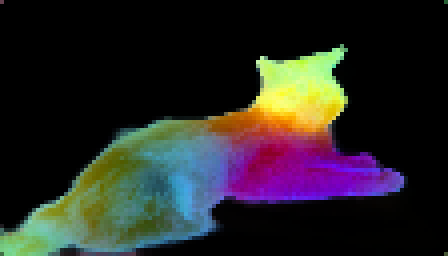}};
        \node[right=\imgHSpace of img22] (img23) {\includegraphics[width=\imgSize]{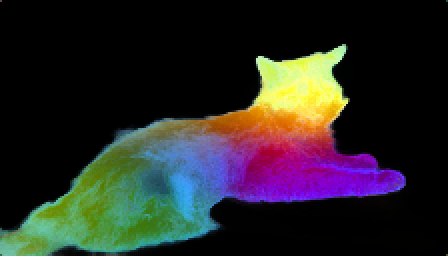}};
        \node[right=\imgHSpace of img23] (img24) {\includegraphics[width=\imgSize]{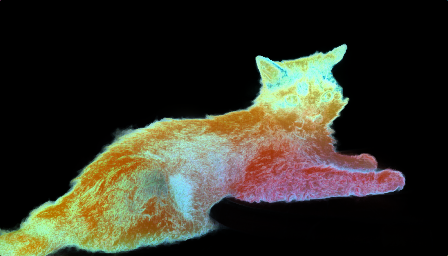}};
        \node[below=\imgVSpace of img21] (img31) {\includegraphics[width=\imgSize]{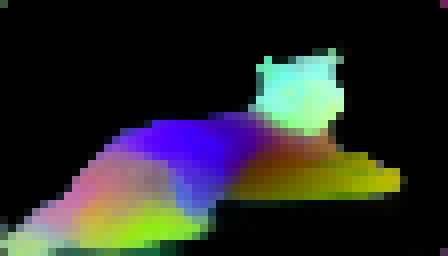}};
        \node[right=\imgHSpace of img31] (img32) {\includegraphics[width=\imgSize]{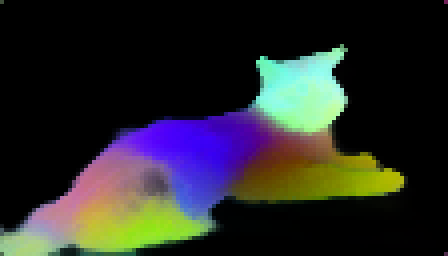}};
        \node[right=\imgHSpace of img32] (img33) {\includegraphics[width=\imgSize]{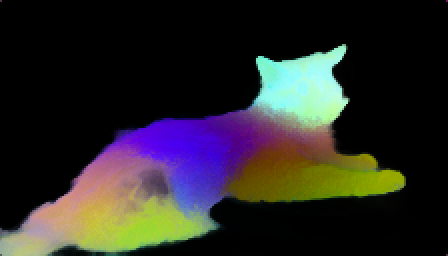}};
        \node[right=\imgHSpace of img33] (img34) {\includegraphics[width=\imgSize]{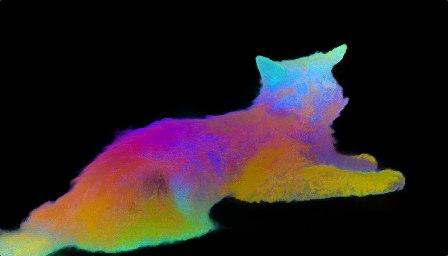}};
        \node[below=\imgVSpace of img31] (img41) {\includegraphics[width=\imgSize]{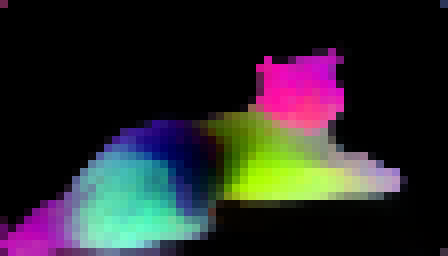}};
        \node[right=\imgHSpace of img41] (img42) {\includegraphics[width=\imgSize]{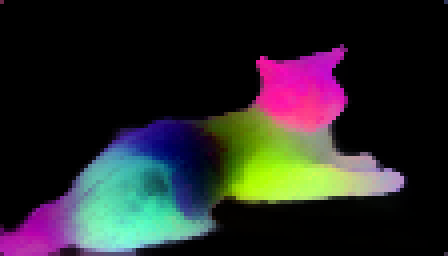}};
        \node[right=\imgHSpace of img42] (img43) {\includegraphics[width=\imgSize]{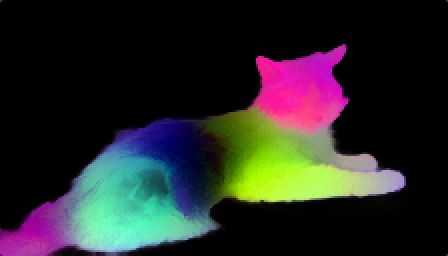}};
        \node[right=\imgHSpace of img43] (img44) {\includegraphics[width=\imgSize]{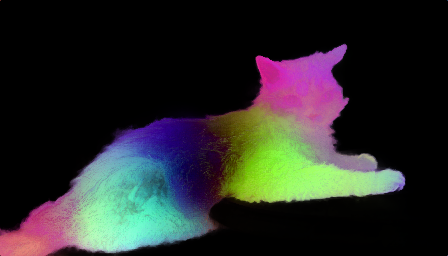}};
        \node[below=\imgVSpace of img41] (img51) {\includegraphics[width=\imgSize]{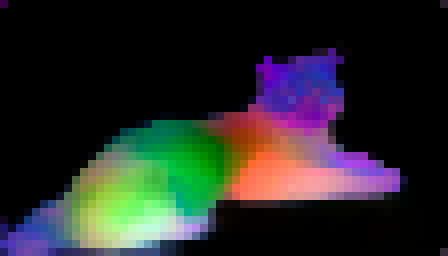}};
        \node[right=\imgHSpace of img51] (img52) {\includegraphics[width=\imgSize]{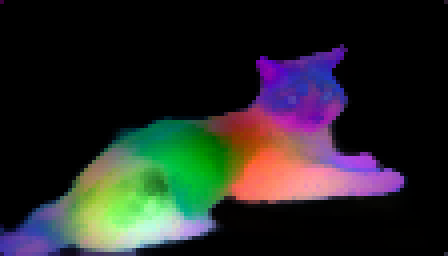}};
        \node[right=\imgHSpace of img52] (img53) {\includegraphics[width=\imgSize]{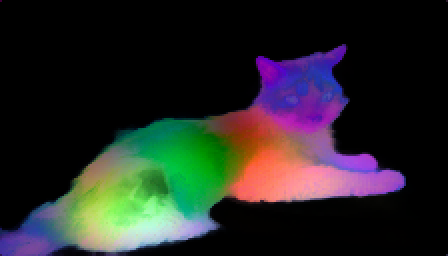}};
        \node[right=\imgHSpace of img53] (img54) {\includegraphics[width=\imgSize]{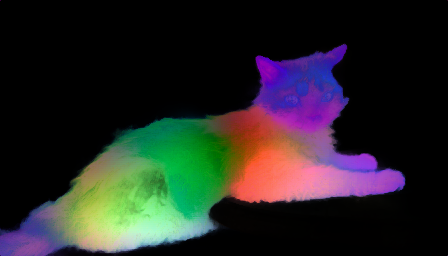}};
        \node[anchor=center] (label1) [left=0.3cm of img11.west] {\rotatebox{90}{ViT-S}};
        \node[anchor=center] (label2) [left=0.3cm of img21.west] {\rotatebox{90}{ViT-S+}};
        \node[anchor=center] (label3) [left=0.3cm of img31.west] {\rotatebox{90}{ViT-B}};
        \node[anchor=center] (label4) [left=0.3cm of img41.west] {\rotatebox{90}{ViT-L}};
        \node[anchor=center] (label5) [left=0.3cm of img51.west] {\rotatebox{90}{ViT-H+}};
        \node[below=0.2cm of img51.south] {\small $896\times512$};
        \node[below=0.2cm of img52.south] {\small $1792\times1024$};
        \node[below=0.2cm of img53.south] {\small $3584\times2048$};
        \node[below=0.2cm of img54.south] {\small $7168\times4096$};
    \end{tikzpicture}
    \caption{Stability of the features at multiple resolutions for the DINOv3 ViT family of models. 
    Top-to-bottom: ViT-S, S+, B, L, H+. 
    We run inference on an image at multiple resolutions, then perform principal component analysis on the features computed on a $1792\times1024$ image ($112\times64$ image tokens).
    We then project features at all resolutions onto the principal components 5--7 that we map to the RGB space for visualization. 
    While the models are functional at all resolutions, we observe that the features remain consistent across a large range of resolutions before drifting: for example, ViT-S+ features are stable between $896\times512$ and $3584\times2048$ inputs, while ViT-L barely starts drifting on the largest resolution $7168\times4096$. 
    ViT-H+ remains stable throughout the whole tested range.}
    \label{fig:pca:family}
\end{figure}
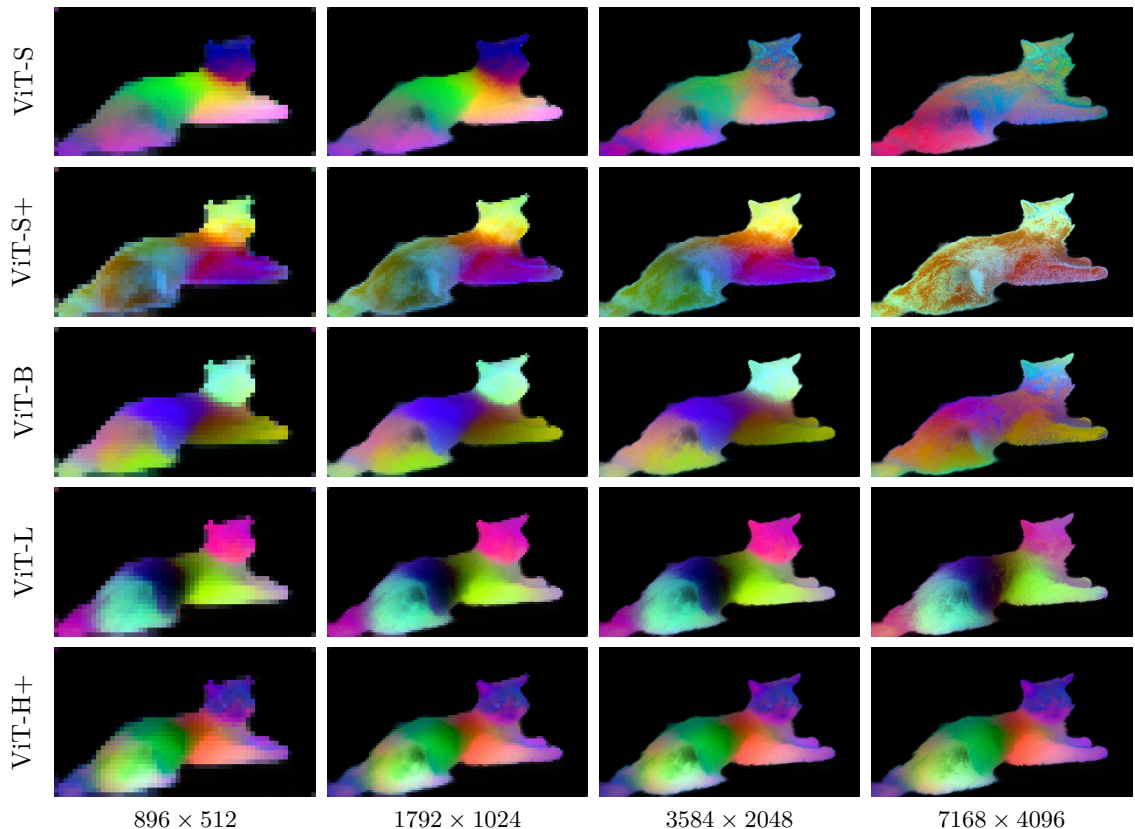

\subsection{Efficient ConvNeXts for Resource-Constrained Environments}
\label{sec:family:convnext}

In this section, we evaluate the quality of our ConvNeXt (CNX) models distilled from the 7B teacher. 
ConvNeXt models are highly efficient in terms of FLOPs and are well-suited for deployment on devices optimized for convolutional computations. 
Furthermore, transformer models often do not lend themselves well to quantization~\citep{bondarenko2021understanding}, whereas quantization of convolutional nets is a well explored subject.
We distill CNX architectures of size T, S, B, and L (see \cref{tab:family-models-flops}) and compare them to the original ConvNeXt models~\citep{liu2022convnet}.
These baselines achieve high performance on ImageNet-1k as they were trained in a supervised fashion using ImageNet-22k labels, and thus represent a strong competitor.
For this experiment, we provide results for global tasks at input resolutions 256 and 512, for ADE20k at resolution 512, and for NYU at resolution 640.

\begin{table}[t]
    \centering
    \caption{
      Evaluation of our distilled DINOv3 ConvNeXt models.
      We compare our models to off-the-shelf ConvNeXts trained supervised on ImageNet-22k~\citep{liu2022convnet}.
      For global tasks, we give results at input resolutions 256 and 512, as we found the supervised models to significantly degrade at resolution 512.
    }
    \begin{tabular}{@{}ll c  cc c cc c cc  c cc@{}}
    \toprule
    & && \multicolumn{8}{c}{Global Tasks} && \multicolumn{2}{c}{Dense Tasks} \\
    \cmidrule{4-11} \cmidrule{13-14}
    Size & Model && \multicolumn{2}{c}{IN-ReAL} && \multicolumn{2}{c}{IN-R} && \multicolumn{2}{c}{Obj.} && ADE20k & NYU$\downarrow$ \\
    \cmidrule{4-5} \cmidrule{7-8} \cmidrule{10-11}
    &&& 256 & 512 && 256 & 512 && 256 & 512 && & \\
    \midrule
    T & Sup.  && 87.3 & 83.0 && 45.0 & 33.0 && 44.5 & 27.1 && 24.8 & 0.666 \\
    T & DINOv3 && 86.6 & 87.7 && 73.7 & 74.1 && 52.6 & 58.7 && 42.7 & 0.448 \\
    \midrule
    S & Sup.  && 88.9 & 86.8 && 52.8 & 39.1 && 50.8 & 40.0 && 22.6 & 0.630 \\
    S & DINOv3  && 87.9 & 88.7 && 73.7 & 74.1 && 52.6 & 58.7 && 44.8 & 0.432\\
    \midrule
    B & Sup.  && 89.3 & 87.8 && 57.3 & 46.2 && 53.6 & 46.5 && 26.5 & 0.596 \\
    B & DINOv3  && 88.5 & 89.2 && 77.2 & 78.2 && 56.2 & 61.3 && 46.3 & 0.420 \\
    \midrule
    L & Sup.  && 89.6 & 88.1 && 58.4 & 46.6 && 55.0 & 47.7 && 33.3 & 0.567 \\
    L & DINOv3  && 88.9 & 89.4 && 81.3 & 82.4 && 59.3 & 65.2 && 47.8 & 0.403 \\
    \bottomrule
\end{tabular}
    \label{tab:family:convnext}
\end{table}

\paragraph[Results]{Results (\cref{tab:family:convnext})}
We find that on in-distribution image classification, our models slightly lag behind the supervised ones at resolution 256 (\eg $-0.7$ IN-ReAL for CNX-T).
However, the trend is reversed at resolution 512, with the supervised ConvNeXts significantly degrading, whereas our models scale with increased input resolution.
For out-of-distribution classification (IN-R, ObjectNet), there are significant gaps between the two model families for all sizes---a testament to the robustness of the DINOv3 CNX models. 
Furthermore, the DINOv3 models offer very large improvement on dense tasks. 
Indeed, for CNX-T, our model yields a $+17.9$ mIoU (42.7 versus 24.8) improvement, and for CNX-L, our model gets $+14.5$ mIoU (47.8 versus 33.3).
The combination of high performance and computational efficiency makes the distilled ConvNeXt models especially promising for real-world applications where resource constraints are critical.
Aside from that, the distillation of the ViT-7B model into smaller ConvNeXt models is particularly exciting, as it bridges two fundamentally different architectures. 
While ViT-7B is based on transformer blocks with a CLS token, ConvNeXt relies on convolutional operations without a CLS token, making this transfer of knowledge non-trivial. This achievement highlights the versatility and effectiveness of our distillation process.

\subsection[Zero-shot Inference with DINOv3-based dino.txt]{Zero-shot Inference with DINOv3-based \texttt{dino.txt}}
\label{sec:family:text}

As detailed in \cref{sec:alignementtxt}, we train a text encoder to align both the CLS token and the output patches of the distilled DINOv3 ViT-L model to text, following the recipe of dino.txt~\citet{jose2025dinov2}.
We evaluate the quality of the alignment both at the global- and patch-level on standard benchmarks. 
We report the zero-shot classification accuracy using the CLIP protocol~\citep{radford2021learning} on the ImageNet-1k, ImageNet-Adversarial, ImageNet-Rendition and ObjectNet benchmarks. 
For image-text retrieval, we evaluate on the COCO2017 dataset~\citep{coco2017} and report Recall@1 on both image-to-text (I $\rightarrow$ T) and text-to-image (T $\rightarrow$ I) tasks. 
To probe the quality of patch-level alignment, we evaluate our model on the open-vocabulary segmentation task using the common benchmarks ADE20k and Cityscapes, for which we report the mIoU metric.

\paragraph[Results]{Results (\cref{tab:dino_txt})} 
We compare our text-aligned DINOv3 ViT-L with competitors in the same size class. 
Compared to \citet{jose2025dinov2}, which aligns DINOv2 to text, DINOv3 leads to significantly better performance on all benchmarks. 
On global alignment tasks, we compare favorably to the original CLIP~\citep{radford2021learning} and strong baselines such as EVA-02-CLIP~\citep{sun2023eva} but slightly behind SigLIP2~\citep{tschannen2025siglip} and Perception Encoder~\citep{bolya2025perception}. 
On dense alignment tasks, our text-aligned model shows excellent performance on two challenging benchmarks ADE20K and Cityscapes thanks to clean feature maps of DINOv3. \omitme{to comment on dense task, depending on PE/SigLIP 2 results}

\begin{table}[t]
    \small
    \centering
    \caption{Comparing our text-aligned DINOv3 ViT-L to the state-of-the-art. 
    Our model achieves excellent dense alignment performance while staying competitive in global alignment tasks. 
    All compared models are of ViT-L size and operate on the same sequence length of 576.}
    \label{tab:dino_txt}
    \setlength{\tabcolsep}{4pt}
    \begin{tabular}{@{}l c cccc c cc c cc@{}}
    \toprule
     && \multicolumn{4}{c}{Classification} && \multicolumn{2}{c}{Retrieval} && \multicolumn{2}{c}{Segmentation} \\
    \cmidrule(lr){3-6} \cmidrule(lr){8-9} \cmidrule(lr){11-12}
    Method &&      IN1k & A & R & Obj. & & I $\rightarrow$ T & T $\rightarrow$ I & & ADE20k & Cityscapes \\
    \midrule
    CLIP && 76.6 & 77.5 & 89.0 & 72.3 && 57.9 & 37.1 && 6.0 & 11.5 \\
    EVA-02-CLIP && 80.4 & 82.9 & 93.2 & 78.5 && 64.1 & 47.9 && 10.9 & 14.1 \\
    \texttt{dino.txt}  && 81.6 & 83.2 & 88.8 & 74.5 && 62.5 & 45.0 && 19.2 & 27.4 \\
    SigLIP 2 && 83.1 & 84.3 & \textbf{95.7} & 84.4 && 71.4 & 55.3 && 10.8 & 16.3 \\
    PE && \textbf{83.5} & \textbf{89.0} & 95.2 & \textbf{84.7} && \textbf{75.9} & \textbf{57.1} && 17.6 & 21.4 \\
    \midrule
    DINOv3 \texttt{dino.txt} && 82.3 & 85.4 & 93.0 & 80.5 && 63.7 & 45.6 && \textbf{24.7} & \textbf{36.9} \\
    \bottomrule
    \end{tabular}
\end{table}

\section{DINOv3 on Geospatial Data}
\label{sec:satellite}
Our self-supervised learning recipe is generic and can be applied to any image domain. In this section, we showcase this universality by building a DINOv3 7B model for satellite images, which have very different characteristics (\eg object texture, sensor noise, and focal views) than the web images on which DINOv3 was initially developed.

\subsection{Pre-Training Data and Benchmarks}

Our satellite DINOv3 7B model is pre-trained on \SATdataset, a dataset of 493 millions of $512\times512$ images sampled randomly from Maxar RGB ortho-rectified imagery at 0.6 meter resolution.
We use the exact same set of hyper-parameters that are used for the web DINOv3 7B model, except for the RGB mean and std normalization that are adapted for satellite images, and the training length. 
Similar to the web model, our training pipeline for the satellite model consists of 100k iterations of initial pre-training with global crops ($256\times 256$), followed by 10k iterations using Gram regularization, and finalized with 8k steps of high resolution fine-tuning at resolution $512$. 
Also similar to the web model, we distill our 7B satellite model into a more manageable ViT-Large model to facilitate its use in low-budget regime.

We evaluate DINOv3 satellite and web models on multiple earth observation tasks. For the task of global canopy height mapping, we use the Satlidar dataset described in \cref{app:exp-details:geospatial}, which consists of one million $512\times 512$ images with LiDAR ground truths split into train/val/test splits with ratios 8/1/1. The splits include the Neon and S\~{a}o Paulo dataset used by \citet{tolan2024very}.
For national-scale canopy height mapping, we evaluate on Open-Canopy \citep{fogel2025open}, which combines SPOT 6-7 satellite imagery and aerial LiDAR data over 87,000 km$^2$ across France.
Since images in this dataset have 4 channels including the additional infra-red (IR) channel, we adapt our backbone by taking the average of the three channels in the weights of the patch embed module and adding it to the weights as the fourth channel.
We trained a DPT decoder on $512\times 512$ crops of images resized to 1667 to match the Maxar ground sample resolution. %

\begin{table}[t]
  \centering
  \small
  \caption{Evaluation of different backbones for high-resolution canopy height prediction. All models are trained with a DPT decoder. Results are presented either for experiments with the decoder trained on SatLidar and evaluated on IID samples (SatLidar Val) and OOD test sets (SatLidar Test, Neon and S\~{a}o Paulo), or for experiments with the decoder trained and evaluated on the Open-Canopy dataset.
  We list mean absolute error (MAE) and the block $R^2$ metric from \citet{tolan2024very}.
  For completeness, we additionally evaluate the original decoder of \citet{tolan2024very} that was trained on Neon dataset (denoted by $^\ast$).}
  \label{tab:canopy}
  \setlength{\tabcolsep}{4pt}
  \begin{tabular}{@{}ll c cc c cc cc cc c c@{}}
    \toprule
    \multirow{3}{*}{Method} & \multirow{3}{*}{Arch.} & & \multicolumn{9}{c}{SatLidar} & & \multirow{2}{*}{Open Canopy} \\
    \cmidrule{4-5} \cmidrule{7-12}
    & & & \multicolumn{2}{c}{SatLidar Val} 
       & & \multicolumn{2}{c}{SatLidar Test}  
          & \multicolumn{2}{c}{Neon Test}    
          & \multicolumn{2}{c}{S\~{a}o Paulo} & & \\
        \cmidrule{4-5} \cmidrule{7-8} \cmidrule{9-10} \cmidrule{11-12} \cmidrule{14-14} 
     & & & MAE↓ & $R^2$↑ & & MAE↓   & $R^2$↑ & MAE↓   & $R^2$↑ & MAE↓   & $R^2$↑ & & MAE↓ \\
    \midrule
    \citet{tolan2024very}$^\ast$  & ViT-L 
       & & 2.8    & 0.86   
       & & 4.0    & 0.61   
         & 2.7    & 0.73   
         & \textbf{5.4}    & 0.42   
         & & ---  \\
    \citet{tolan2024very}  & ViT-L 
       & & 2.4    & 0.90   
       & & 3.4    & 0.81   
         & 2.9    & 0.69   
         & \textbf{5.4} & 0.48  
         & & 2.42 \\
    DINOv3 Web  & ViT-7B 
       & & 2.4    & 0.90   
       & & 3.6    & 0.74   
         & 2.7    & 0.75
         & 5.9    & 0.34   
         & & 2.17  \\
    DINOv3 Sat  & ViT-L
       & & \textbf{2.2} & 0.91 
       & & \textbf{3.2} & 0.81 
         & \textbf{2.4} & \textbf{0.81}  
         & 5.8    & 0.42 
         & & 2.07 \\
    DINOv3 Sat  & ViT-7B 
       & & \textbf{2.2} & \textbf{0.92} 
       & & \textbf{3.2} & \textbf{0.82} 
         & 2.6 & 0.74  
         & 5.5    & \textbf{0.51} 
         & & \textbf{2.02} \\
    \bottomrule
  \end{tabular} %
\end{table}

Semantic geospatial tasks are assessed with GEO-Bench \citep{lacoste2023geo}, which comprises six classification and six segmentation tasks spanning various spatial resolutions and optical bands. The GEO-Bench tasks are diverse, including the detection of rooftop-mounted photovoltaic systems, classifying local climate zones, measuring drivers of deforestation, and detecting tree crowns. For high-resolution semantic tasks, we consider the land cover segmentation dataset LoveDA \citep{wang2022lovedaremotesensinglandcover}, the object segmentation dataset iSAID \citep{zamir2019isaidlargescaledatasetinstance}, and the horizontal detection dataset DIOR \citep{li2020object}. 

\subsection{Canopy Height Estimation}

Estimating canopy height from satellite imagery is a challenging metric task, requiring accurate recovery of continuous spatial structure despite random variations in slope, viewing geometry, sun angle, atmospheric scattering, and quantization artifacts.
This task is critical for global carbon monitoring and for forest and agriculture management \citep{harris2021}.
Following \citet{tolan2024very}, the first work to leverage a SSL backbone trained on satellite images for this task, we train a DPT head on top of frozen DINOv3 on the SatLidar1M training set, then evaluate it on i.i.d.~samples on SatLidar1M validation set as well as out-of-distribution test sets including SatLidar1M test, Neon and Sao Paulo.
We additionally train and evaluate on the Open-Canopy dataset. 

\omitme{revise text if removing Sao Paulo}
\paragraph[Results]{Results (\cref{tab:canopy})}
We compare different SSL backbones, denoting with ``DINOv3 Sat'' the model trained the \SATdataset dataset, and with ``DINOv3 Web'' the model trained on \LVDdataset (see \cref{sec:data}).
It can be seen that DINOv3 satellite models yield state-of-the-art performance on most benchmarks.
Our 7B satellite model sets the new state of the art on SatLidar1M val, SatLidar1M test and Open-Canopy, reducing MAE from $2.4$ to $2.2$, from $3.4$ to $3.2$ and from $2.42$ to $2.02$ respectively. These results show that DINOv3 training recipe is generic and can be effectively applied out-of-the-box to other domains. 
Interestingly, our distilled ViT-L satellite model performs comparably to its 7B counterpart, achieving comparable results on SatLidar1M and Open-Canopy while faring surprisingly better on Neon test set, reaching the lowest MAE of $2.4$ compared to $2.6$ of the 7B model and $2.9$ of \citet{tolan2024very}.
Our DINOv3 7B web model reaches decent performance on the benchmarks, outperforming \citet{tolan2024very} on SatLidar1M val, Neon and Open-Canopy but stays behind the satellite model.
This highlights the strength of domain-specific pretraining for physically grounded tasks like canopy height estimation, where sensor-specific priors and radiometric consistency are important.

\subsection{Comparison to the Earth Observation State of the Art}

\begin{table}[t]
    \centering
    \footnotesize
    \caption{
    Comparison of our DINOv3 models against strong baselines DOFA~\citep{xiong2024neural}, Prithvi-v2~\citep{szwarcman2024prithvi}, and \citet{tolan2024very} in Geo-Bench tasks. While Privthi-v2 and DOFA leverage all available optical bands, our models achieve significantly better performance with only RGB inputs.\looseness-1
    }
    \label{tab:Geobench}
    \begin{subtable}{\linewidth}
        \centering
        \caption{Classification tasks.}
        \vspace{-0.4em}
        \setlength{\tabcolsep}{4pt}
        \begin{tabular}{@{}lcccccccccc@{}}
            \toprule
            Method & Arch. & FT & Bands & m-BEnet & m-brick-kiln & m-eurosat & m-forestnet & m-pv4ger & m-so2sat & Mean \\
             \midrule
           DOFA  & ViT-L &\orangefire & all & 68.7 & 98.4 & 96.6 & 55.7 & 98.2 & 61.6 &  79.9 \\
           Best of Prithvi-v2  & ViT-L/H& \orangefire & all & 71.2  & {\bf98.8} & 96.4 & 54.1 & {98.1} & 59.1 & 79.6\\
           \citet{tolan2024very} & ViT-L & \bluesnow & RGB & 66.0 & 97.1 & 95.2 & 56.3 & 94.3  & 58.1 & 77.8\\
           DINOv3 Sat & ViT-L &\bluesnow & RGB & 73.0 & 96.5 & 94.1 & 60.6 & 96.0 & 57.4 & 79.6 \\
           DINOv3 Sat & 7B &\bluesnow & RGB &  74.0 & 97.2 & 94.8 & {\bf 62.3} & 96.1 & 62.1 & 81.1 \\ %
           DINOv3 Web & 7B & \bluesnow & RGB & {\bf 74.6} & 97.7& {\bf 97.0} & 57.9 & {\bf 98.3} & {\bf 63.8} & {\bf 81.6}  \\
            \bottomrule
        \end{tabular} %
     \end{subtable}
     
     \vspace{1em}
     \begin{subtable}{\linewidth}
        \centering
        \caption{Segmentation tasks.}
        \vspace{-0.4em}
        \setlength{\tabcolsep}{2pt}
         \begin{tabular}{@{}lcccccccccc@{}}
            \toprule
            Method & Arch. & FT & Bands & m-cashew$^{*}$ & m-chesapeake & m-NeonTree & m-nz-cattle & m-pv4ger-seg & m-SA-crop & Mean  \\
            \midrule
           DOFA & ViT-L&\orangefire& all & 81.2 &61.6  & 58.5 & 77.4 & 95.1 & 35.7 & 68.3\\
           Best of Prithvi-v2  & ViT-L/H& \orangefire & all& 90.2 & 69.4 & 59.1 & 81.0 & 95.3 & {\bf41.9} & 72.8 \\
           \citet{tolan2024very} & ViT-L & \bluesnow& RGB& 92.8 &73.7 & 58.1& 83.1& 94.7& 35.1 & 72.9\\
           DINOv3 Sat & ViT-L & \bluesnow & RGB & 94.2 & 75.6	& 61.8 & \textbf{83.7} & 95.2 & 36.8 & 74.5 \\ 
           DINOv3 Sat & 7B & \bluesnow & RGB & 94.1 & {\bf76.6} & 62.6 &  83.4 & 95.5 & 37.6 & 75.0 \\
           DINOv3 Web & 7B & \bluesnow & RGB & {\bf 96.0} & 76.5& {\bf 66.4}& {\bf 83.7} & {\bf 95.9}& 36.8& {\bf 75.9}  \\
            \bottomrule
        \end{tabular} %
    \end{subtable}
    \begin{flushleft}
        \vspace{-0.2em}
        {\footnotesize $^{*}$\textit{Conversion to 6 classes following \citet{szwarcman2024prithvi}.}}
    \end{flushleft}
\end{table}

We compare the performance of different methods for Earth observation tasks in~\cref{tab:Geobench} and~\cref{tab:sat_highres}.
The frozen DINOv3 satellite and web models set new state-of-the-art results on 12 out of 15 classification, segmentation, and horizontal object detection tasks. Our Geo-Bench results surpass prior models, including Prithvi-v2~\citep{szwarcman2024prithvi} and DOFA~\citep{xiong2024neural}, which use 6+ bands for Sentinel-2 and Landsat tasks, as well as task-specific fine-tuning (\cref{tab:Geobench}).
Despite using a frozen backbone with RGB-only input, the DINOv3 satellite model outperforms previous methods on the three unsaturated classification tasks and on five of six segmentation tasks. Interestingly, the DINOv3 7B web model is very competitve on these benchmarks.
It achieves comparable or stronger performance on many GEO-Bench tasks as well as on large-scale, high-resolution remote sensing benchmarks for segmentation and detection.
As shown in \cref{tab:Geobench} and \cref{tab:sat_highres}, the frozen DINOv3 web model establishes new leading results Geo-Bench tasks as well as for segmentation and detection tasks on the LoveDA and DIOR datasets.

These findings have broader implications for the design of geospatial foundation models.
Those have recently emphasized heuristic techniques such as multitemporal aggregation, multisensor fusion, or incorporating satellite-specific metadata \citep{brown2025alphaearthfoundationsembeddingfield, feng2025tesseratemporalembeddingssurface}.
Our results show that general-purpose SSL can match or exceed satellite-specific approaches for tasks that depend on precise object boundaries (segmentation or object detection). 
This supports emerging evidence finding that domain-agnostic pretraining can offer strong generalization even in specialized downstream domains \citep{lahrichi2025selfsupervisedpretrainingsatelliteimagery}.  

Collectively, our results suggest task-dependent benefits of domain-specific pretraining.
The DINOv3 satellite model excels in metric tasks like depth estimation, leveraging satellite-specific priors.
In contrast, the DINOv3 web model achieves state-of-the-art results on semantic geospatial tasks through diverse, universal representations.
The complementary strengths of both models illustrate the broad applicability and effectiveness of the DINOv3 SSL paradigm. %

\omitme{
}

\begin{figure}[t]
    {\small%
        \begin{tabular}{@{}c@{\hspace{0.64em}}c@{\hspace{0.64em}}c@{\hspace{0.64em}}c@{\hspace{0.64em}}c@{}}
        \includegraphics[width=0.19\linewidth]{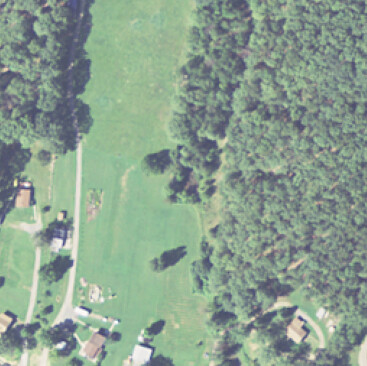}&
        \includegraphics[width=0.19\linewidth]{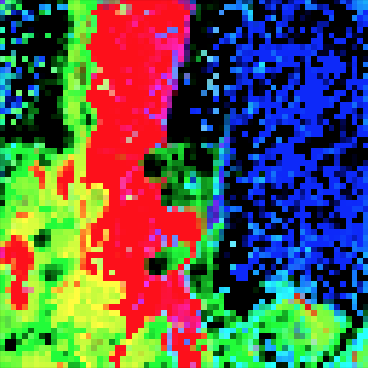}&
        \includegraphics[width=0.19\linewidth]{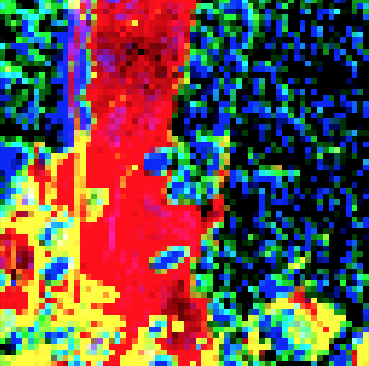}&
        \includegraphics[width=0.19\linewidth]{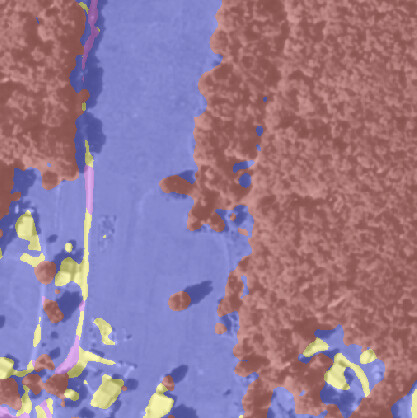}&
        \includegraphics[width=0.19\linewidth]{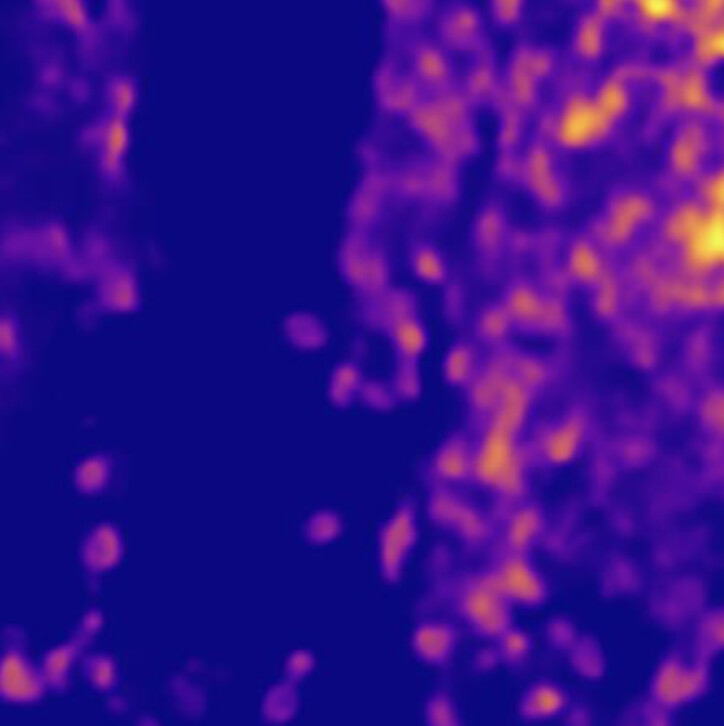}\\
        Img in Chesapeake & PCA DINOv2 & PCA DINOv3 & Segmentation v3 & Canopy height v3 \\
        \end{tabular}
    }
    \caption{
        Illustration of versatile applications in remote sensing made possible by a single DINOv3 model. The PCA on DINOv3 features shows finer details than DINOv2. The segmentation map was computed using only GEO-Bench chesapeake labels. The canopy height model decoder was trained on the Open-Canopy dataset using 4 channels (RGB + InfraRed), while inference was performed on RGB channels only. 
    }
    \omitme{
        \hl{[to do: update with the new model]}
    } 
    \omitme{using /home/coupriec/segmentation_viz.ipynb or /home/coupriec/lavida/segmentation_viz.ipynb on CW}
    \label{fig:satfig}
\end{figure}

 \begin{table}[t]
     \centering
     \small
     \caption{We compare the performance of DINOv3 to state-of-the-art models Privthi-v2~\citep{szwarcman2024prithvi}, BillionFM~\citep{Cha_2024} and SkySense V2~\citep{zhang2025skysense} for high resolution semantic geospatial tasks. We report mIoU for the segmentation datasets LoveDA (1024$\times$) and iSAID (896$\times$), and mAP for the detection dataset DIOR ($800\times$).}
     \label{tab:sat_highres}
      \begin{tabular}{@{}llc c c c c@{}}
        \toprule
          Method & Arch. & FT && LoveDA & iSAID & DIOR \\%Depth on Open-Canopy \\
          \midrule
           \multirow{2}{*}{Prev. SotA} & & \multirow{2}{*}{\orangefire} && BillionFM, ViT-G  & SkySense V2, Swin-G$^\ast$ & SkySense V2, Swin-G$^\ast$ \\%PVTv2 ViT-L\\ 
            & & && 54.4  & \textbf{71.9} & 79.5 \\%& 2.52 \\ 
          \midrule
          Decoder Arch. & & && UPerNet & UPerNet & Faster-RCNN \\%& DPT\\
          Privthi-v2 & ViT-H &  \orangefire && 52.2 &  62.8  & --- \\%& - \\
          DINOv3 Sat  & ViT-L & \bluesnow  &&   54.4 &  62.9  & 72.7 \\
          DINOv3 Sat  & ViT-7B & \bluesnow  &&  55.3  &  64.8  & 76.6 \\%&  {\bf 2.02} \\ %
          DINOv3 Web & ViT-7B & \bluesnow && \textbf{56.2} & 71.4 \omitme{71.2-71.6}  & {\bf 80.5} \\%& 2.17 \\ 
          \bottomrule
      \end{tabular}
    \begin{flushleft}
        {\footnotesize $^\ast$\textit{Uses modified DINOv2 SSL with supervised pretraining alignment on OpenStreetMap, reporting +0.8 mIoU on iSAID.}}
    \end{flushleft}
 \end{table}

\begin{figure}[t]
    \centering
    \begin{tabular}{cccc}
         \includegraphics[width=0.22\linewidth]{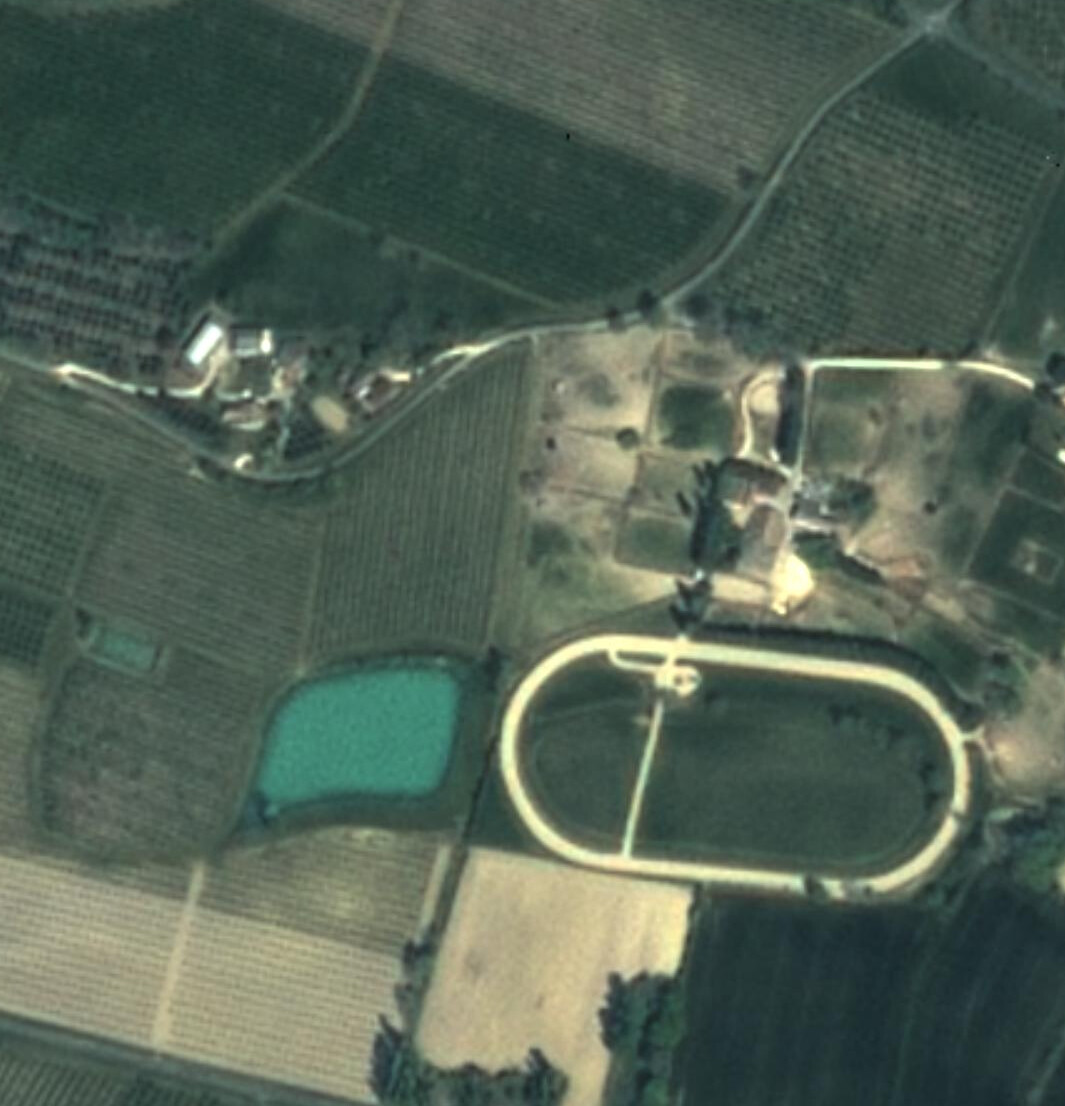}&
         \includegraphics[width=0.22\linewidth]{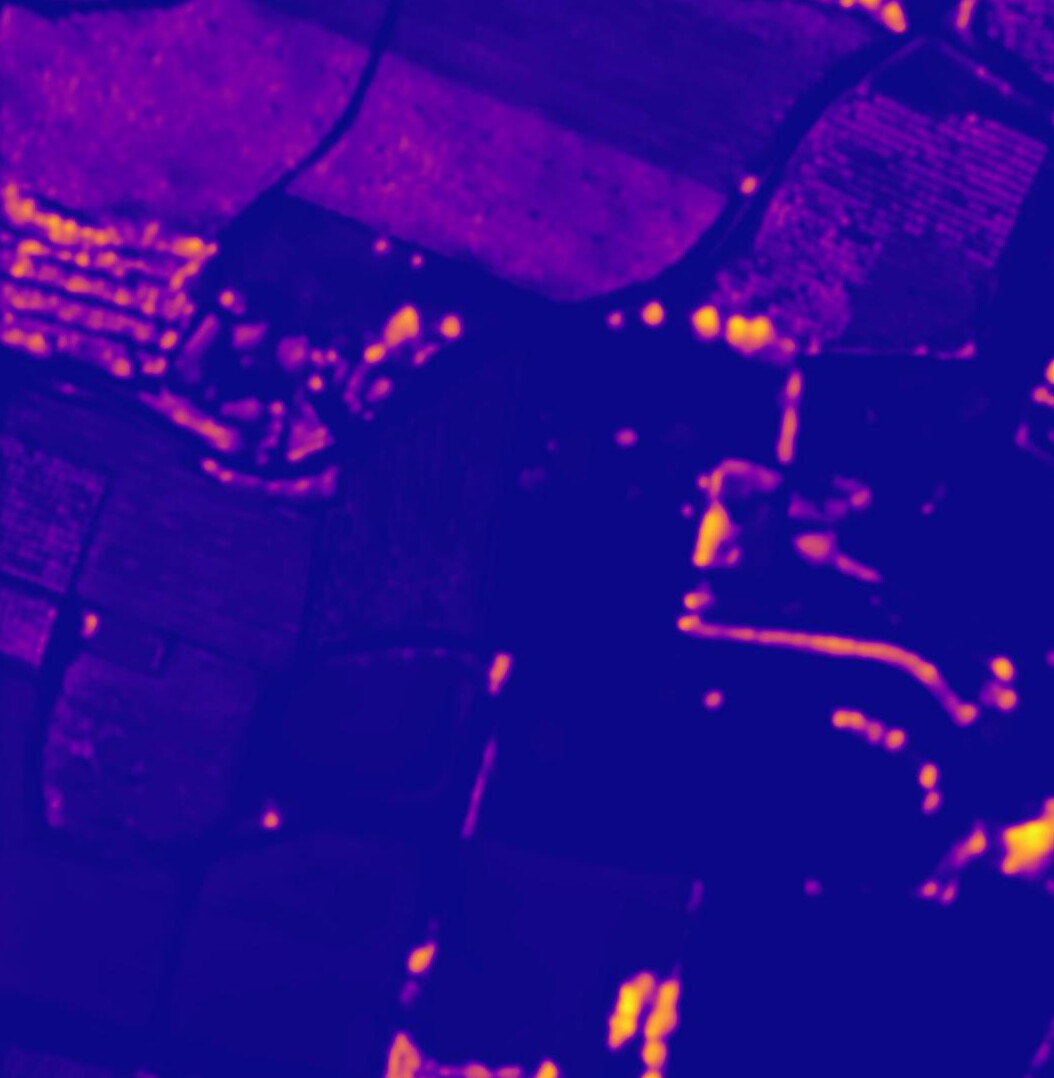}&
         \includegraphics[width=0.22\linewidth]{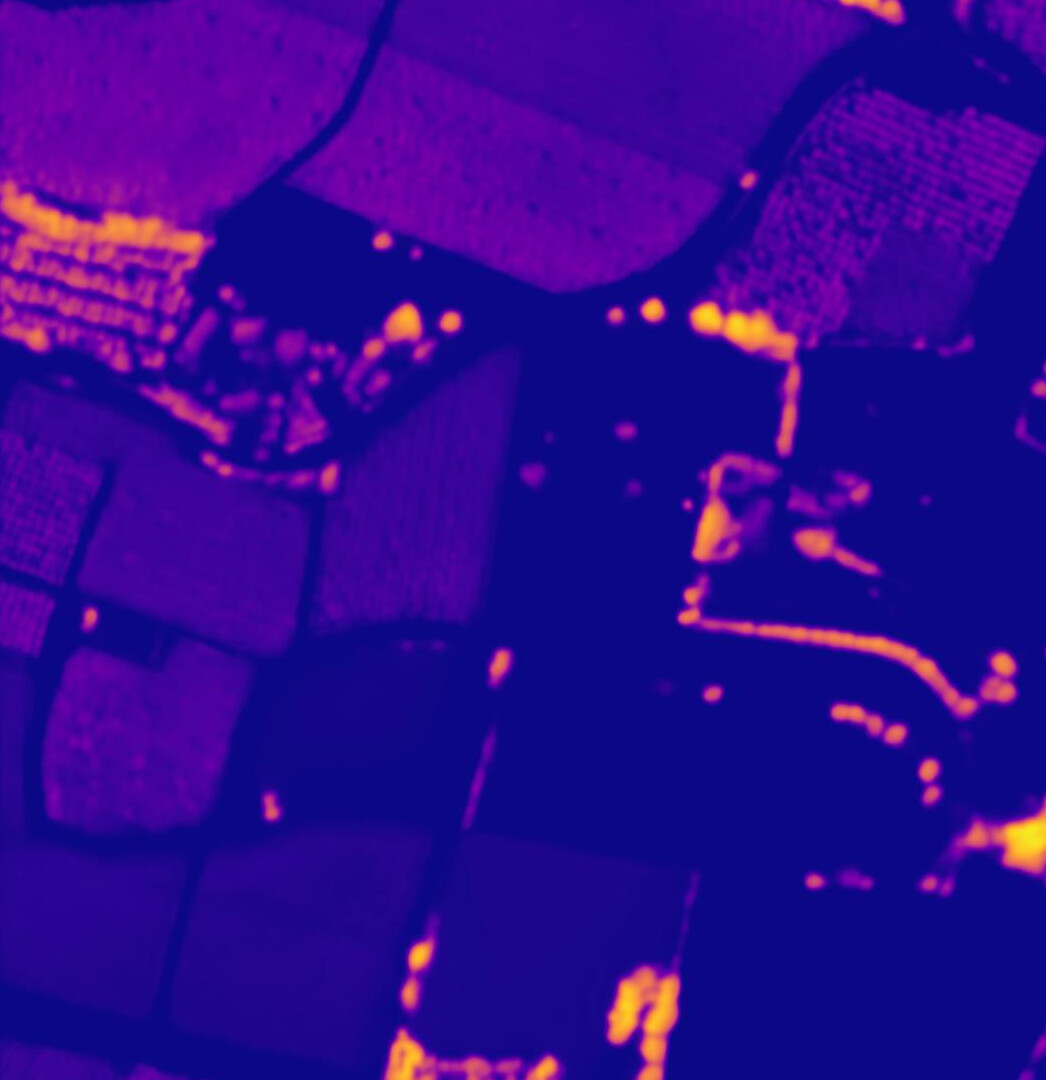}&
         \includegraphics[width=0.22\linewidth]{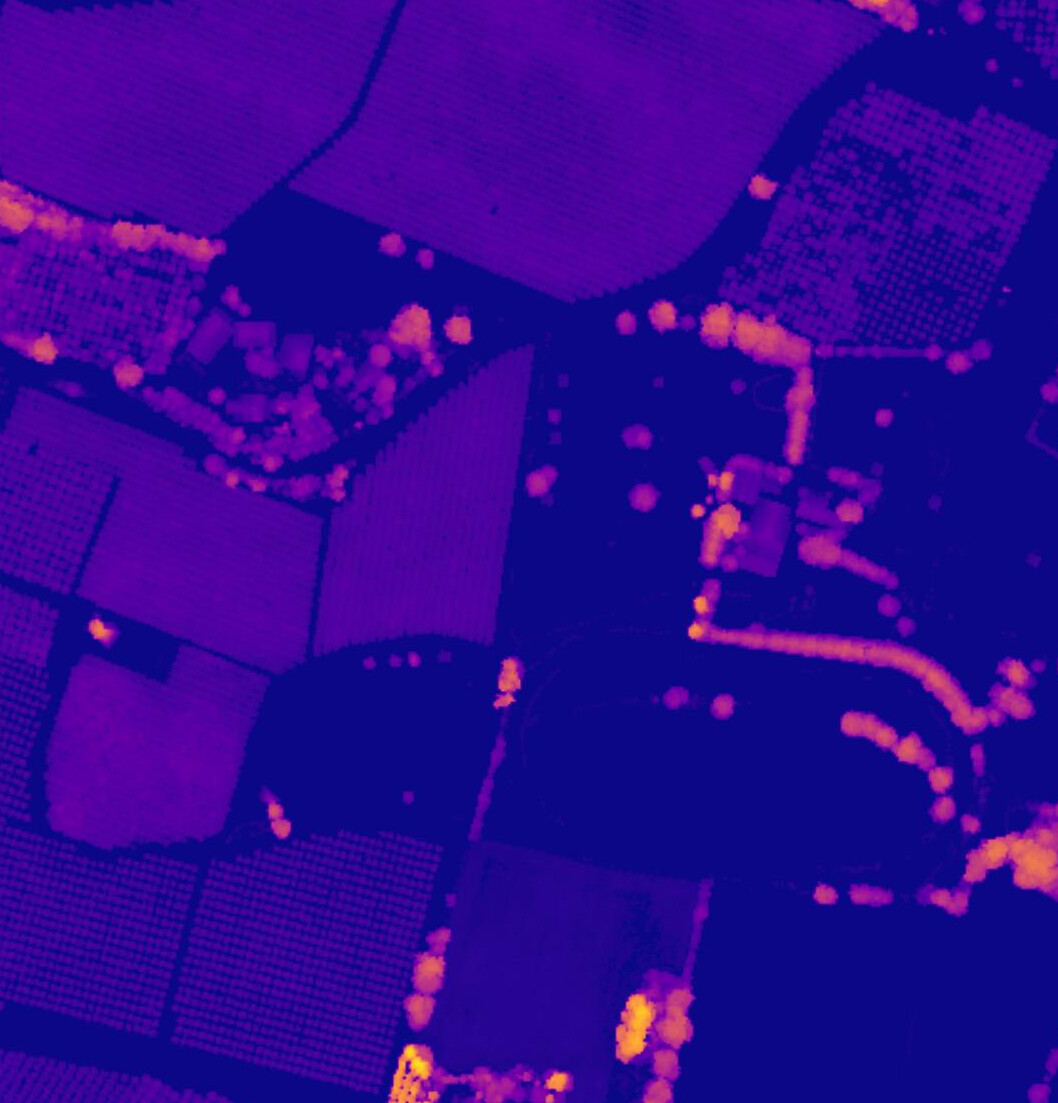}\\
         Image from OpenCanopy& Improved Tolan et al.& DINOv3 SAT-493M & Ground Truth\\ 
    \end{tabular}
    \caption{A qualitative comparison of the DINOv3 7B satellite model to \citet{tolan2024very} on the Open Canopy dataset. For both models, the decoder is trained on 448$\times$448 input images. It can be seen that DINOv3 produces more accurate maps, for example the accurate height for the trees on the field.}
    \label{fig:placeholder}
\end{figure}

\section{Environmental Impact}

\begin{table}[tb]
  \centering
  \caption{
    Carbon footprint of model training.
    We report the potential carbon emission of reproducing a full model pre-training, computed using a PUE of 1.1 and carbon intensity factor of 0.385kg CO$_2$eq/KWh.
  }
  \label{tab:carbon}
  \begin{tabular}{@{}ll ccccccc@{}}
    \toprule
    Model    & Arch.   & GPU type  & Power & Steps   & GPU hours   & PUE & Total power & Emission \\
             &         &           & (W)   &         &         &     & (MWh)       & (tCO$_2$eq)\\
    \midrule
    MetaCLIP & ViT-G  & A100-40GB & 400W  & 390k    & 368,640 & 1.1 & 160         & 62 \\
    DINOv2   & ViT-g    & A100-40GB & 400W  & 625k    & 22,016  & 1.1 & 9.7         & 3.7 \\
    DINOv3   & ViT-7B   & H100-SXM5 & 700W  & 1,000k  & 61,440  & 1.1 & 47          & 18 \\
    \bottomrule
  \end{tabular}
\end{table}

To estimate the carbon emission of our pre-training, we follow the methodology used in previous work in natural language processing~\citep{strubell2019energy,touvron2023llama} and SSL~\citep{oquab2024dinov2}.
We fix the value of all exogenous variables, \ie the Power Usage Effectiveness (PUE) and carbon intensity factor of a power grid to the same value as used by \citet{touvron2023llama}, \ie we assume a PUE of 1.1 and a carbon intensity factor of the US average of 0.385 kg CO$_2$eq/KWh.
For the power consumption of GPUs, we take their thermal design power: 400W for A100 GPUs and 700W for H100 GPUs.
We report the details of the computation for the pre-training of our ViT-7B in \cref{tab:carbon}.
For reference, we provide the analogous data for DINOv2 and MetaCLIP.
As another point of comparison, the energy required to train one DINOv3 model (47 MWh) is roughly equivalent to that required for 240,000 km of driving with an average electric vehicle. 

\paragraph{Carbon Footprint of the Whole Project}
In order to compute the carbon footprint of the whole project, we use a rough estimate of a total 9M GPU hours.
Using the same grid parameters as presented above, we estimate the total footprint to be roughly 2600 tCO$_2$eq.
For comparison, a full Boeing 777 return flight between Paris and New York corresponds to approximately 560 tCO$_2$eq.
Supposing 12 such flights per day, the environmental impact of our project represents half of all flights between these two cities for one day.
This estimate only considers the electricity for powering the GPUs and ignores other emissions, such as cooling, manufacturing, and disposal.

\section{Conclusion}

DINOv3 represents a significant advancement in the field of self-supervised learning, demonstrating the potential to revolutionize the way visual representations are learned across various domains. 
By scaling dataset and model size through meticulous data preparation, design, and optimization, DINOv3 showcases the power of self-supervised learning to eliminate the dependency on manual annotations. 
The introduction of the \gramname method effectively mitigates the degradation of dense feature maps over extended training periods, ensuring robust and reliable performance.

Together with the implementation of post-hoc polishing strategies, such as high-resolution post-training and distillation, we achieve state-of-the-art performance across a wide range of visual tasks with no fine-tuning of the image encoder. 
The DINOv3 suite of vision models not only sets new benchmarks but also offers a versatile solution across various resource constraints, deployment scenarios, and application use cases.
The progress made with DINOv3 is a testament to the promise of self-supervised learning in advancing the state of the art in computer vision and beyond.

\clearpage

\bibliographystyle{plainnat}
\bibliography{main}

\clearpage
\appendix

{\Large\textbf{Appendix}}
\FloatBarrier

\section{Artifacts and Outliers in Large-Scale Training}
\label{app:sec:artifacts}

This section provides a discussion about the emergence of artifacts and outliers that has recently been observed in the training of large models in both the LLM~\citep{an2025systematic} and the visual domains~\citep{darcet2024vision}. 
Borrowing the definition from \citet{an2025systematic}, outliers are typically characterized as network's activations whose values deviate significantly from the average of their distribution. %
During the training of DINOv3, we identified such outliers at different levels: some occurring at the patch level and others at the feature dimension level. 
We discuss bellow the different types of outlier observed, their impact on the training and results. 
We also discuss our different attempts at fixing them and our first conclusions. 

\subsection{High-Norm Patch Outliers}
 
\begin{figure}
    \centering
    \begin{subfigure}{.66\linewidth}
        \centering
        \small
        \setlength{\tabcolsep}{0.2pt}
        \renewcommand{\arraystretch}{0.2} %
        \begin{tabular}{ccccc}
            \centering
            & \input{figures/outliers/plot_norm_patches_registers_0_att_bias_False_value_gating_False} &
            \input{figures/outliers/plot_norm_patches_u1_untie_cp_False_gl_True_noLN} &  \input{figures/outliers/plot_norm_patches_registers_0_att_bias_True_value_gating_False} & 
             \input{figures/outliers/plot_norm_patches_registers_0_att_bias_False_value_gating_True}
             \\
            \includegraphics[height=2.1cm, width=2.1cm]{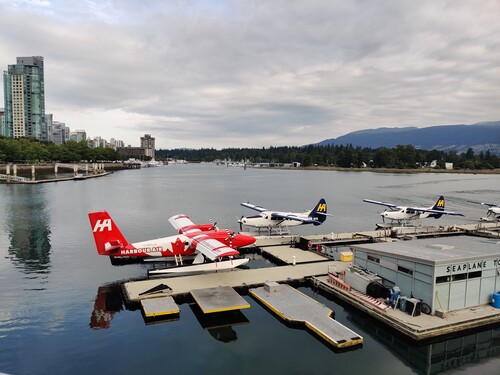} & 
            \includegraphics[height=2.1cm]{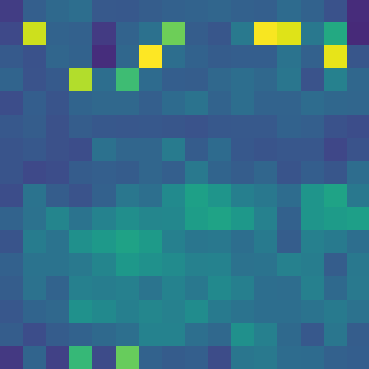} & 
            \includegraphics[height=2.1cm]{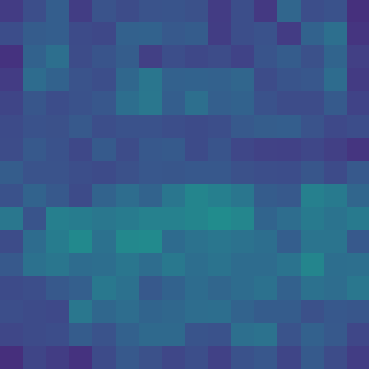} & 
            \includegraphics[height=2.1cm]{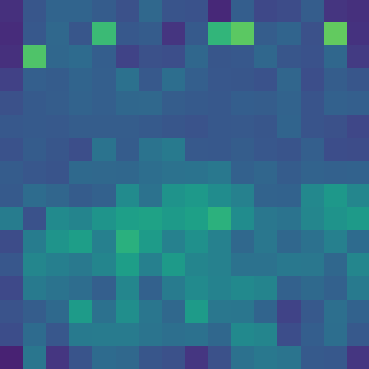} & 
            \includegraphics[height=2.1cm]{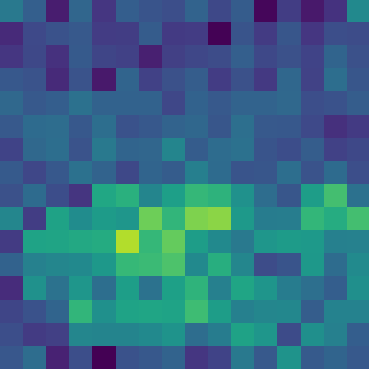}
            \\
            \includegraphics[height=2.1cm, width=2.1cm]{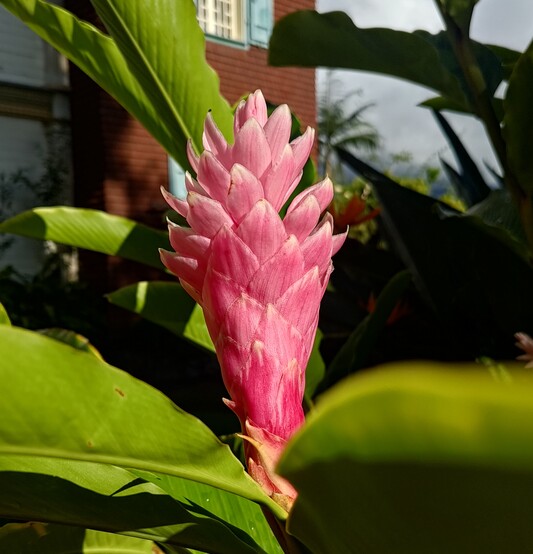} &  
            \includegraphics[height=2.1cm]{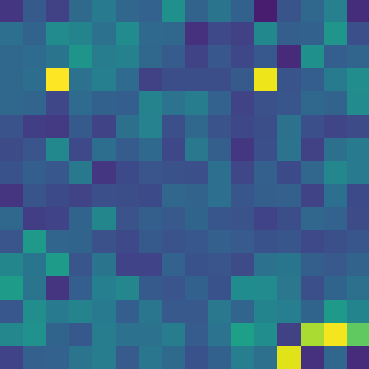} &  
            \includegraphics[height=2.1cm]{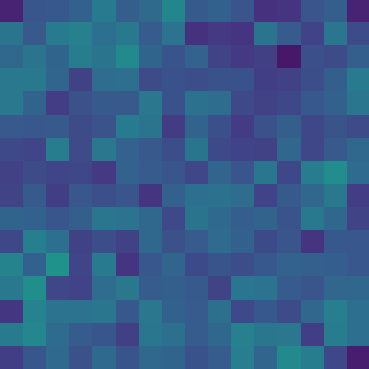} &  
            \includegraphics[height=2.1cm]{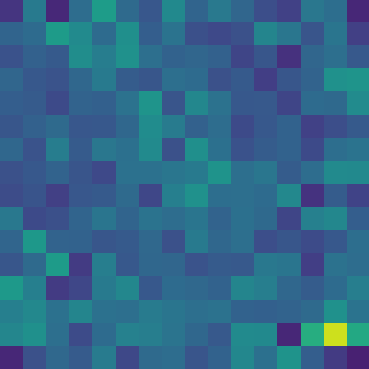} &  
            \includegraphics[height=2.1cm]{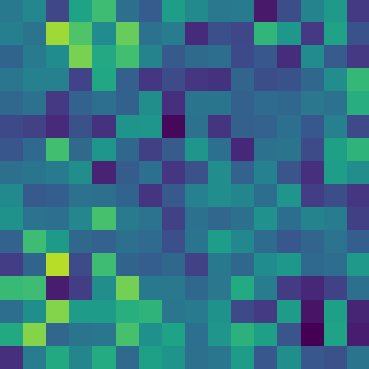}
            \\
            \\
            Image & $\varnothing$ & 4 Registers & Attention  & Value  \\
             & & &  Bias &  Gating \\
        \end{tabular}
        \centering
        \caption{Visualization of patch norms by outlier strategy. The bottom two rows share a colormap per row, from dark blue (low) to yellow (high).}
        \label{fig:outliers-regis-attention}
    \end{subfigure}
    \hfill
    \begin{subfigure}{.33\linewidth}
        \centering
        \small
        \captionsetup{justification=centering}
        \caption{Quantitative ablation.}
        \label{tab:outliers-regis-attention}
        \addtolength{\tabcolsep}{-0.3em}
        \begin{tabular}{l c cc}
            Outlier &  & IN1k & ADE20k \\
            Strategy &  & (Linear) & mIoU \\
            \toprule
            $\varnothing$ & & 86.4 & \bf 53.2 \\
            4 Registers & & \bf 86.6 & 53.0 \\
            Attention Bias & & 86.5 & 52.7 \\
            Value Gating & & 86.3 & 52.2 \\
        \end{tabular}
        \vspace{5em}    
    \end{subfigure}
    \caption{Impact of the different strategies to mitigate the presence of high-norm patch outliers, evaluated both (a) qualitatively and (b) quantitatively. 
    We produce results with a 7B model trained with our recipe for 150k iterations, without any high-norm handling strategy `$\varnothing$', when using four register tokens \citep{darcet2024vision}, or the attention bias and value gating strategies \citep{an2025systematic}. 
    In (a, first row), we plot the distribution of the output patch norms (sorted by ascending values) computed for three images. We also visualize the output patch norms per image (bottom two rows), with the same colormap--min and max values are computed per image over the different outlier strategies. }
    \label{fig:outliers}
\end{figure}

\citet{darcet2024vision} discovered that patch outliers negatively affect performance in DINOv2. 
These outliers are primarily characterized as high-norm tokens, often located in low-information background regions of an image. 
These tokens are observed to play a key role in the internal communication between patches and the CLS token. 
Additionally, this phenomenon affects other models as well, whether trained with supervision or not, such as CLIP~\citep{radford2021learning}.
When scaling to a 7B model, we observe the emergence of such high-norm patches, predominantly in the background area. In this section, we present results from 7B models trained for 150k iterations, which, although limited, provide us with initial signals to guide our decisions.
We plot the output patch norms (before the layer norm) in \cref{fig:outliers-regis-attention}, in the column `$\varnothing$', with high-norm patches in yellow appearing in the sky and other low-information areas.

\paragraph{Token Registers} 
In order to mitigate the appearance of such token outliers, \citep{darcet2024vision} proposes a simple yet effective solution: introducing additional tokens, called registers, into the input sequence of the ViT. 
Their role is to take over the internal communication between patches and the CLS. 
Following the conclusions, we use 4 registers and do not ablate further due to the high experimental cost.
\Cref{fig:outliers-regis-attention} illustrates examples of this strategy in action, where we observe the elimination of high-norm outliers, as further confirmed by the corresponding histogram of the norm distribution. 
Moreover, we quantitatively observe in \cref{tab:outliers-regis-attention} the benefit of incorporating additional register tokens on the ImageNet-1k (IN1k) benchmark.

\paragraph{Integrating Biases in the Attention Mechanism} 
Recent work by \citet{an2025systematic} investigates the appearance of outliers in the LLM realm across different models and architectures. The authors analyze different types of outliers which are observed to be intrinsically linked to the attention mechanism. They propose to mitigate the problem with several solutions from which we select two promising solutions which seem relevant and require minimal changes to the attention, specifically the explicit fixed bias, which we call `value gating', and the attention bias strategies.
The value gating strategy amounts to adding a learnable value bias $\mathrm{\bf v'} \in \mathbb{R}^d$ to the output of the attention, specifically by redefining the attention mechanism as 
\begin{equation}
    \mathrm{Attn}(Q,K,V; \mathrm{\bf k'}, \mathrm{\bf v'}) = \mathrm{softmax}(\frac{Q[K^T]}{\sqrt{d}})V + \mathrm{\bf v'}, 
    \label{eq:value-gatting}
\end{equation}
with $Q, K, V \in \mathbb{R}^{T\times d}$ the query, key, and value matrices and $d$ the dimensionality of the
hidden space. 
Alternatively, the attention bias, defined in \cref{eq:attn-bias}, consists in integrating two learnable bias terms $\mathrm{\bf k'}, \mathrm{\bf v'} \in \mathbb{R}^d$ in the keys and values matrices, respectively. It is defined as follows: 
\begin{equation}
    \mathrm{Attn}(Q,K,V; \mathrm{\bf k'}, \mathrm{\bf v'}) = \mathrm{softmax}(\frac{Q[K^T\mathrm{\bf k'}]}{\sqrt{d}}) \left[ \genfrac{}{}{0pt}{}{V}{\bf v'} \right]. 
    \label{eq:attn-bias}
\end{equation}
We observe in \cref{fig:outliers-regis-attention} that the value gating strategy substantially modifies the distribution of patch norms, resulting in generally higher norm values and the elimination of clear outliers. While the attention mechanism mitigates the presence of high-norm tokens, it does not completely resolve the issue, as some high-norm patches persist--as visible in the top row image--when compared to our results using register tokens. Notably, the best performance is achieved with the incorporation of the register tokens, which is why we adopt this strategy for all experiments reported is the paper.

\subsection{Feature Dimension Outliers}

The introduction of additional registers into the model architecture effectively resolves the issue of high-norm patch outliers. However, during the training of 7B models, we observe a distinct type of outlier that emerges not across patches, but within the feature (channel) dimension of the learned representations. Specifically, analysis of patch activations across transformer layers and training iterations reveals that a small subset of feature dimensions attain exceptionally large magnitudes, even as the norms across patches remain stable. Interestingly, these feature dimension outliers exhibit consistently high values across different patches and images, a behavior that contrasts with observations reported in \citep{an2025systematic}. Moreover, these outlier dimensions consistently persist across the layers of a given model, increasing in magnitude with depth and reaching their maximum values in the output layer. They also progressively increase in magnitude throughout the course of training.

We conduct experiments attempting to neutralize these dimensions during both training and inference. 
Our findings indicate that these dimensions play a significant role during training, as applying L2-regularization to suppress them results in a performance drop. 
However, removing these dimensions at inference time does not lead to significant performance changes, suggesting that they primarily carry trivial or non-informative signals. 
Additionally, we observe that the final layer normalization is trained to substantially scale down these outlier dimensions. 
Thus, we recommend to apply the final layer norm to the features of the final layer for downstream use.
Alternatively, applying batch normalization can also suppress these feature dimension outliers, as their elevated values are consistent across patches and images.

A word of caution applies to using features from earlier layers. 
As discussed above, these earlier layers are also affected by feature dimension outliers which can lead to ill-conditioned features.
While the final layer normalization is well-suited to normalize the distribution of the final features, its learned parameters may be suboptimal for applying it to the features of earlier layers. 
Indeed, we observe performance decreases for some tasks from doing so. 
In these cases, we found standard feature scaling techniques (\eg normalization with batch norm or principal component analysis) to be effective in dealing with the feature dimension outliers.
For example, for our semantic segmentation (\cref{sec:results-segmentation}) and depth estimation experiments (\cref{sec:results-depth-estimation}) using features from intermediate layers, we apply batch normalization.

\section{Additional Results}

\subsection{Evolution Over Years}
In \cref{fig:pcas}, we provide a rough evolution of state-of-the-art performance along years. 
Here, we provide the precise references and performances that we reported in the figure.
Please find it in \cref{app:fig:refs}.

\begin{table}[t]
    \centering
    \small
  \caption{
    Details of year of publication, performance, and reference of the numbers used in \cref{fig:pcas}.
    For all papers, we report top-1 accuracy of this algorithm with the largest model on ImageNet.
    For weakly- and self-supervised models, we provide linear probing performance.
    For dates, we use the year of first appearance on arXiv.
  }
  \label{app:fig:refs}
  \begin{tabular}{@{}l c ll c ll c ll@{}}
        \toprule
        && \multicolumn{2}{c}{Supervised} && \multicolumn{2}{c}{Weakly-Supervised} && \multicolumn{2}{c}{Self-Supervised} \\
        \cmidrule{3-4} \cmidrule{6-7} \cmidrule{9-10}
        Year && Top-1 & Reference && Top-1 & Reference && Top-1 & Reference \\
        \midrule
        2012 && 59.3 & \citet{krizhevsky2012imagenet} && \\
        2013 \\ 
        2014 \\
        2015 && 78.6 & \citet{he2016deep} && 34.9 & \citet{joulin2016learning} \\
        2016 \\
        2017 && 80.9 & \citet{xie2017aggregated} && \\
        2018 &&  & && 83.6 & \citet{mahajan2018exploring} && 38.2 & \citet{caron2018deep} \\
        2019 && 84.3 & \citet{tan2019efficientnet} && & && 68.6 & \citet{he2020momentum} \\
        2020 && 87.5 & \citet{kolesnikov2020big} &&  & && 75.3 & \citet{caron2020unsupervised} \\
        2021 && 88.6 & \citet{dosovitskiy2020image} && 88.4 & \citet{radford2021learning} && 82.3 & \citet{zhou2021ibot} \\
        2022 && \\
        2023 && 89.5 & \citet{dehghani2023scaling} && & && 86.5 & \citet{oquab2024dinov2} \\
        2024 && \\
        2025 &&  & && 89.3 & \citet{bolya2025perception} && 88.4 & This work \\
        \bottomrule
  \end{tabular}
  \vspace{3em}
\end{table}

\subsection{Per-Layer Analysis}

In this section, we evaluate the quality of our features across the various layers of the DINOv3 7B model. 
Specifically, we present results from five representative tasks: classification (IN-1k val, ImageNet-ReAL and ObjectNet), segmentation (ADE20k), depth estimation (NYU), tracking (DAVIS), and 3D correspondence estimation (NAVI).
For the first 3 benchmarks, a linear layer is trained on the outputs of each backbone layer to assess feature performance as in \cref{sec:results-dense-linear-probing,sec:results-classification-linear}.
For tracking and correspondence estimation, we use non-parametric approaches as in \cref{sec:results-correspondence-estimation,sec:results-tracking}.

\begin{figure}[t]
    \centering
    \begin{subfigure}{0.32\textwidth}
        \caption{Classification}
        \label{app:fig:exp:layer-IN}
        \includegraphics{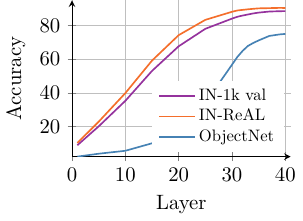}
    \end{subfigure}
    \hspace{0.1\textwidth}
    \begin{subfigure}{0.32\textwidth}
        \caption{Segmentation (ADE20k)}
        \label{app:fig:exp:layer-ade20k}
        \includegraphics{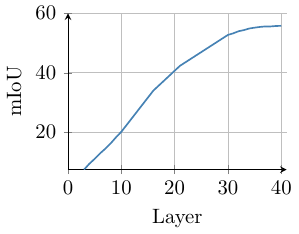}
    \end{subfigure}

    \begin{subfigure}{0.32\textwidth}
        \caption{Depth (NYU) $\downarrow$}
        \label{app:fig:exp:layer-nyu}
        \includegraphics{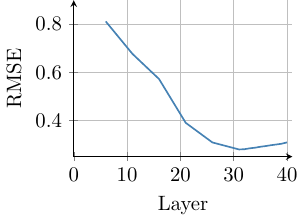}
    \end{subfigure}%
    \hfill
    \begin{subfigure}{0.32\textwidth}
        \caption{Tracking (DAVIS)}
        \label{app:fig:exp:layer-davis}
        \includegraphics{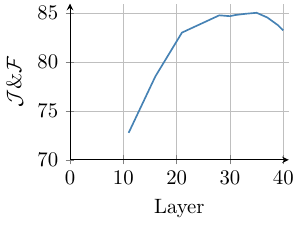}
    \end{subfigure}%
    \hfill
    \begin{subfigure}{0.32\textwidth}
        \caption{3D Corr. Estimation (NAVI)}
        \label{app:fig:exp:layer-navi}
        \includegraphics{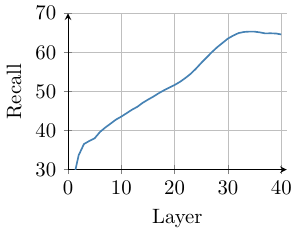}
    \end{subfigure}
    \caption{
        Results on five benchmarks using features from intermediate layers of DINOv3 7B.
        Evaluations (a-c) use a linear layer (see \cref{sec:results-dense-linear-probing,sec:results-classification-linear}), while (d, e) use a non-parametric approach (see \cref{sec:results-correspondence-estimation,sec:results-tracking}).
    }
    \label{app:fig:exp:layers}
\end{figure}

The results are shown in \cref{app:fig:exp:layers}.
We find that for classification and dense tasks, performance increases smoothly over the layers.
Depth estimation, tracking, and 3D correspondence estimation peak around layer 32, indicating that, for tasks where geometry plays a significant role, the downstream performance of DINOv3 can be improved by considering earlier layers.
On the other hand, the performance of intermediate layers only slightly improves compared to the last one, making it a good default choice.

\subsection{Additional Results to Main Results Section}

We give additional experimental results complementing the main results in \cref{sec:results}. 
In \cref{app:tab:results-classification-vtab}, we show per-dataset results for finegrained classification on small datasets with linear probing (Fine-S, see \cref{sec:results-classification-linear}).
In \cref{app:tab:results-instance-retrieval}, we give full results for the instance recognition evaluation (\cref{sec:results-instance-rec}), adding more metrics.
Finally, in \cref{app:tab:semantic-segmentation-other-datasets}, we give complementary results for our state-of-the-art semantic segmentation model (\cref{sec:results-segmentation}) on the COCO-Stuff~\citep{caesar2018coco}, PASCAL VOC 2012~\citep{pascal-voc-2012} and Cityscapes~\citep{geiger2013vision} datasets.

\begin{table}[tb]
    \centering
    \footnotesize
    \caption{Per-dataset results for finegrained classification on small datasets with linear probing (Fine-S, see \cref{sec:results-classification-linear}), following \citet{oquab2024dinov2}.
    }
    \label{app:tab:results-classification-vtab}
    \setlength{\tabcolsep}{4pt}
    \begin{tabular}{@{}ll c ccccccccccccc@{}}
    \toprule
    Method & ViT && Food & C10 & C100 & SUN & Cars & Aircr. & VOC & DTD & Pets & Cal101 & Flowers & CUB & Avg \\
    \midrule
    \multicolumn{2}{@{}l}{\resultsTableHeaderAgg} && &&&&&&&&&&&& \\
    AM-RADIOv2.5 & g/14    && 96.5 & 99.5 & 95.0 & 82.8 & 95.4 & 91.7 & 90.3 & 88.6 & 96.7 & 98.8 & 99.7 & 91.5 & 93.9 \\
    \midrule
    \multicolumn{2}{@{}l}{\resultsTableHeaderWeakly} && &&&&&&&&&&&& \\
    SigLIP & g/16          && 97.7 & 99.3 & 92.7 & 85.1 & 96.5 & 88.7 & 91.0 & 87.7 & 98.7 & 90.3 & 99.7 & 90.3 & 93.7 \\
    PE-core & G/14         && 97.8 & 99.5 & 95.3 & 85.2 & 96.5 & 92.0 & 90.5 & 88.2 & 98.7 & 93.3 & 99.5 & 93.3 & 94.5 \\
    AIMv2 & 3B/14          && 96.6 & 99.3 & 93.3 & 83.4 & 95.6 & 84.2 & 90.5 & 87.4 & 96.8 & 90.7 & 99.7 & 90.7 & 92.9 \\
    EVA CLIP & 18B/14      && 96.9 & 99.5 & 95.4 & 85.0 & 95.4 & 81.6 & 90.2 & 87.1 & 98.4 & 90.6 & 99.6 & 90.6 & 92.9 \\
    \midrule
    \multicolumn{2}{@{}l}{\resultsTableHeaderSelf} && &&&&&&&&&&&& \\
    Franca & g/14          && 89.2 & 98.6 & 90.4 & 73.7 & 89.7 & 74.1 & 89.4 & 80.6 & 93.2 & 97.6 & 97.8 & 78.4 & 87.7 \\
    DINOv2 & g/14          && 95.6 & 99.5 & 94.5 & 78.9 & 94.6 & 88.5 & 88.4 & 86.8 & 96.8 & 95.9 & 99.7 & 91.6 & 92.6 \\
    Web-DINO & 7B/14       && 96.1 & 99.5 & 93.4 & 77.5 & 95.0 & 88.8 & 87.0 & 79.9 & 92.9 & 93.1 & 99.6 & 78.9 & 90.2 \\
    \midrule
    DINOv3 & 7B/16         && 96.9 & 99.6 & 96.0 & 81.1 & 95.0 & 88.2 & 88.2 & 87.2 & 97.0 & 94.8 & 99.7 & 92.4 & 93.0 \\
    \bottomrule
    \end{tabular}
\end{table}

\begin{table}[ht]
    \centering
    \small
    \caption{
        Full results for instance recognition, presenting additional metrics for \cref{sec:results-instance-rec}.
    }
    \label{app:tab:results-instance-retrieval}
    \begin{tabular}{ll c cc c cc c ccc c c}
        \toprule
        & && \multicolumn{2}{c}{Oxford} && \multicolumn{2}{c}{Paris} && \multicolumn{3}{c}{Met} && AmsterTime \\
        \cmidrule{4-5} \cmidrule{7-8} \cmidrule{10-12} \cmidrule{14-14}
        Method & ViT && M & H && M & H && GAP & GAP- & ACC && mAP \\
        \midrule
        \multicolumn{2}{@{}l}{\resultsTableHeaderAgg} && \\
        AM-RADIOv2.5 & g/16 && 72.8 & 50.7 && 93.3 & 85.3 && 30.5 & 65.9 & 69.0 && 46.7 \\
        \midrule
        \multicolumn{2}{@{}l}{\resultsTableHeaderWeakly} &&  \\
        SigLIPv2 & g/16     && 49.3 & 25.1 && 79.3 & 60.9 &&  0.0 &  0.0 &  0.2 && 15.5 \\
        PE-core  & G/14     && 57.4 & 32.7 && 83.6 & 68.9 && 10.6 & 34.8 & 44.9 && 23.1 \\
        AIMv2    & 3B/14    && 55.0 & 28.8 && 85.6 & 71.4 && 29.5 & 67.3 & 69.9 && 23.1 \\
        EvaCLIP  & 18B/14   && 55.2 & 27.1 && 81.8 & 65.6 &&  0.5 &  4.3 & 11.0 && 18.9 \\
        \midrule 
        \multicolumn{2}{@{}l}{\resultsTableHeaderSelf} & \\
        Franca   & g/14     && 44.6 & 14.3 && 73.8 & 51.6 && 27.2 & 54.3 & 57.7 && 21.1 \\
        DINOv2   & g/14     && 78.2 & 58.2 && 92.7 & 84.6 && 44.6 & 73.0 & 75.2 && 48.9 \\
        Web-DINO   & 7B/14    && 64.1 & 31.2 && 89.8 & 80.3 && 35.2 & 67.3 & 71.3 && 30.6 \\
        \midrule
        DINOv3   & 7B/16    && 81.1 & 60.7 && 93.3 & 87.1 && 55.4 & 77.7 & 80.7 && 56.5 \\
        \bottomrule
    \end{tabular}
\end{table}

\begin{table}[b]
\centering
    \small
    \caption{
        Comparison with state-of-the-art systems on semantic segmentation on other datasets, complementary to the ADE20k results in \cref{tab:results-for-ade20k}.
        We report mIoU scores when evaluating the model in a single- or multi-scale (TTA) setup and compare against the best previously published result for each dataset:
        \citet{fang2022eva} for COCO-Stuff 164k, \citet{wang2022internimage} for Cityscapes, and \citet{zoph2020rethinking} for VOC 2012.
        We use input resolutions of 1280 for COCO-Stuff, 1024 for VOC 2012, and 1280 for Cityscapes.
        All baselines require finetuning of the backbone, while we keep the DINOv3 backbone frozen.
    }
   \begin{tabular}{c c cc cc cc}
        \toprule
        Method & FT & \multicolumn{2}{c}{COCO-Stuff 164k} & \multicolumn{2}{c}{Cityscapes} & \multicolumn{2}{c}{VOC 2012} \\
        \cmidrule(lr){3-4} \cmidrule(lr){5-6} \cmidrule(lr){7-8} 
           &  & Single & TTA & Single & TTA  & Single & TTA \\ 
        \midrule
        Previous Best & \orangefire & 53.7 & 53.7 & \textbf{86.3} & \textbf{87.0} & --- & 90.0 \\
        \midrule
        DINOv3          & \bluesnow  & \textbf{53.8} & \textbf{54.0} &
        86.1 & 86.7 &
        \textbf{90.1} & \textbf{90.4} \\
        \bottomrule
    \end{tabular}
    \label{app:tab:semantic-segmentation-other-datasets}
\end{table}

\subsection{Classification on OCR-Heavy Datasets}

In this experiment, we evaluate DINOv3 on classification tasks that require some form of character recognition.
These tasks include street-sign, logo, and product classification.
We compare our model to the best self-supervised model (DINOv2 g) and the best weakly-supervised one (PE-core G).
We run this evaluation on images of resolution 512 for our model, and adjust for patch size to match the sequence length for the others.
We report the result of this experiment in \cref{tab:appendix:logos}.

\begin{table}[t]
    \centering
    \small
    \caption{
        Comparison of DINOv3 classification performance on OCR-heavy datasets.
        These are notoriously hard datasets for SSL.
        We compare DINOv3 with the best DINOv2 model (g), along with the best weakly-supervised PE-core model (G).
    }
    \label{tab:appendix:logos}
    \begin{tabular}{ll c cc ccc}
        \toprule
        Model && GTSRB & Logo-2K+ & FlickrLogos-32 & RP2K & Products-10K & SOProducts \\
        \midrule
        DINOv2 & ViT-g          & 78.2 & 52.9 & 83.6 & 91.4 & 70.8 & 57.6 \\
        PE-core & ViT-G              & 94.8 & 93.2 & 99.0 & 93.1 & 80.6 & 80.7 \\
        \midrule
        DINOv3-7B & ViT-7B         & 87.5 & 86.0 & 86.3 & 94.7 & 74.5 & 65.2 \\
        \bottomrule
    \end{tabular}
\end{table}

We see that our new model DINOv3 drastically outperforms its predecessor DINOv2.
However, the gap with weakly-superpvised models remains large.
Since our model does not leverage pair image-text data during training, it has a much harder time learning glyph associations.
Recent work from \citet{fan2025scaling} hints at the impact of training data on the performance on this type of tasks.
Since the main focus of our work is on improving dense features, we leave closing this gap for future work.

\subsection{Fairness Analysis}

We evaluate the geographical fairness and diversity of the DINOv3 7B model across different income buckets and regions, following the protocol of \citet{goyal2022fairness}.
For reference, we include the results obtained with DINOv2 and SEERv2. The results indicate that DINOv3 delivers somewhat consistent performance across income categories, although there is a notable performance drop of 23\% in the low-income bucket compared to the highest-income bucket. The medium and high-income buckets exhibit comparable performance. Regionally, DINOv3 achieves relatively good scores across different regions; however, a relative difference of over 14\% is observed between Europe and Africa, which is an improvement over the relative difference of more than 17\% seen with DINOv2.

\begin{table}[t]
    \centering
    \caption{Geographical fairness and diversity analysis across income buckets and regions, following the protocol of \citet{goyal2022fairness}.
    }
    \label{tab:results-geographical-fairness}
    \begin{tabular}{llc ccc cccc}
        \toprule
        & & \multicolumn{3}{c}{Income Buckets} & \multicolumn{4}{c}{Regions} \\
        \cmidrule(lr){3-5} \cmidrule(lr){6-9}
        Method & Arch. & low & medium & high & Africa & Asia & Americas & Europe \\
        \midrule
        SEERv2 & RG-10B & 59.7 & 78.5 & 86.6 & 65.9 & 76.3 & 81.1 & 85.6 \\
        \midrule
        DINOv2 & ViT-g/14 & 67.4 & 83.3 & 90.5 & 74.0 & 81.6 & 86.2 & 89.7  \\
        DINOv3 & ViT-7B & 69.6 & 85.7 & 90.9 & 76.7 & 83.0 & 88.0 & 90.7  \\
        \bottomrule
    \end{tabular}
\end{table}

\section{Implementation Details}
\label{app:implem-details}

We use multi-crop~\citep{caron2020unsupervised} with $2$ global crops ($256\times256$ px) and $8$ local crops ($112\times112$ px) seen by the student model, resulting in a total sequence length of $3.7$M tokens. 
The teacher EMA (exponential moving average of the student) processes the global crops only. We apply the $\mathcal{L_{\mathrm{DINO}}}$ loss on the class token of all student local crops and both teacher global crops, and between pairs of different global crops for both models.
A random proportion in $[0.1, 0.5]$ of the global crops patch tokens seen by the student are masked with 50\% probability, and we apply the $\mathcal{L}_{\mathrm{iBOT}}$ loss between these and the visible tokens seen by the teacher EMA.
We apply the $\mathcal{L}_{\mathrm{DKoleo}}$ loss to small batches of 16 class tokens of the first global crop seen by the student.
We train for 1M iterations using a fully-sharded data-parallel setup in Pytorch, using bfloat16 and 8-bit floating-point matrix multiplications.
We use a constant learning rate of $0.0004$ with a warmup of 100k iterations, a weight decay of $0.04$, a learning rate decay factor of $0.98$ per layer, a stochastic depth (layer dropout) value of $0.4$ and an EMA factor of $0.999$ for the teacher. 
Remaining hyperparameters can be found in the configuration files in the code release.

For the \gramname step, we use a loss weight of $w_{\mathrm{Gram}} = 2$ and update the Gram teacher every 10k steps for a maximum of three updates.
For high-resolution adaptation (\cref{sec:algo:hrft}), we sample from the following pairs of global/local/Gram teacher crop resolutions with the following probabilities: $(512, 112, 768)$ with $p=0.3$, $(768, 112, 1152)$ with $p=0.3$, $(768, 168, 1152)$ with $p=0.3$, $(768, 224, 1152)$ with $p=0.05$, and $(768, 336, 1152)$ with $p=0.05$. 
These values were obtained empirically.

\section{Experimental Details}

In this section, we provide detailed descriptions of the datasets and evaluation metrics used across all benchmarks in this paper.

\subsection{Semantic Segmentation: Linear Probing}
\label{app:exp-details:segmentation-probing}

\paragraph{Datasets and Metrics} 
We evaluate semantic segmentation performance of DINOv3 obtained via linear probing on three benchmark datasets: ADE20k~\citep{zhou2017scene}, VOC12~\citep{pascal-voc-2012}, and Cityscapes~\citep{cordts2016cityscapes}. The evaluation metric reported is the standard mean Intersection-over-Union (mIoU).

\paragraph{Evaluation Protocol} 
To assess the quality of the dense features, we train a linear classifier on the training set of each benchmark. This linear layer is applied on top of the patch output features (after layer normalization) of the frozen backbone, with the features further normalized using a trained batch normalization layer. For all backbones, we perform a hyperparameter sweep using the AdamW optimizer, varying the learning rate over \(\{1 \times 10^{-4}, 3 \times 10^{-4}, 1 \times 10^{-3}\}\) and weight decay over \(\{1 \times 10^{-4}, 1 \times 10^{-3}\}\).

\subsection{Depth Estimation: Linear Probing}
\label{app:exp-details:depth-probing}

\paragraph{Datasets and Metrics} 
We evaluate the quality of DINOv3 features for geometric tasks on the depth benchmarks NYUv2~\citep{silberman2012indoor} and KITTI~\citep{geiger2013vision} datasets. 
Results are reported using the Root Mean Squared Error (RMSE) metric.

\paragraph{Evaluation Protocol}
To assess the quality of the dense features, we train a linear classifier on the training set of each benchmark. This linear layer is applied on top of the patch output features (after layer normalization) of the frozen backbone, with the features further normalized using a trained batch normalization layer.
For all backbones, we perform a hyperparameter sweep using the AdamW optimizer, varying the learning rate over \([1 \times 10^{-4}, 3 \times 10^{-4}, 1 \times 10^{-3}]\) and weight decay over \([1 \times 10^{-4}, 1 \times 10^{-3}]\).

\subsection{3D Keypoint Matching}
\label{app:exp-details:keypoint-matching}

\paragraph{Datasets and Metrics}
Geometric correspondence is evaluated on the NAVI dataset~\citep{jampani2023navi}, and semantic correspondence on the SPair dataset~\citep{min2019spair71k}.
For NAVI, we use images resized to a side length of 448/512 pixels  for models with patch size 14/16.
For SPAir, we use images resized to a side length of 896/1024 pixels for models with patch size 14/16.
To measure performance, we report the correspondence recall, \ie the percentage of correspondences falling into a specified distance.

\paragraph{Evaluation Protocol}
For NAVI, we follow the protocol defined in Probe3D~\citep{banani2024probing}.
Specifically, we subsample 1/4 of the object views, and for each source view select another dest.~view within a maximum rotation of 120 degrees to create an image pair (source, dest.) to perform patch matching on. 
For each image pair, each patch of source (within the object) is matched to a patch in dest. 
The top-1000 matches with highest cosine similarity are kept for evaluation, and a 3D distance error is computed for each match based on the known camera pose and depth maps of both images. 
This allows to compute recall errors with varying thresholds, for which we use thresholds of 1cm, 2cm, and 5cm.
We then compute the average recall across thresholds as the correspondence recall.

For each evaluated backbone, we use the features of the final layer, and evaluate them with and without the final layer norm applied. 
This is because we noticed bad performance for some models when applying the final layer norm.
We report the maximum of both results.

\subsection{Unsupervised Object Discovery}
\label{app:exp-details:object-discovery}

\paragraph{Datasets and Metrics}
For this task, the objective is to generate a single bounding box per image that highlights any object depicted in the scene. We follow the protocol of \citet{simeoni2021localizing} for unsupervised object discovery and evaluate all backbones on the detection benchmarks VOC07~\citep{pascal-voc-2007}, VOC12~\citep{pascal-voc-2012}, and COCO20K~\citep{Lin2014cocodataset,Vo20rOSD}. COCO20K is a subset of the COCO2014 trainval dataset~\citep{Lin2014cocodataset}, consisting of 19,817 randomly selected images, as proposed in \citep{Vo20rOSD} and commonly used for this task. For each image, a single bounding box is generated. For evaluation, we use the \textit{Correct Localization} (CorLoc) metric, which computes the percentage of correctly localized boxes. A predicted box is considered correct if its \textit{intersection over union} (IoU) with any ground-truth bounding box exceeds 0.5.

\paragraph{Evaluation Protocol}
To evaluate the quality of the image encoders, we employ the TokenCut strategy~\citep{wang2023tokencut}. This method organizes image patches into a fully connected graph, where the edges represent similarity scores between pairs of patches, computed using backbone features learned by the transformer. The salient object patches are identified by applying the Normalized Cut algorithm, which solves a graph-cut problem. A bounding box is then fitted to the resulting salient object mask.
All images are input to the encoders at their full resolution, and we use the patch output features for all image encoder. To account for differences in feature distributions among models, we perform a sweep over TokenCut’s unique hyperparameter: the similarity threshold used in graph construction. Specifically, we vary the threshold between 0 and 0.4 in increments of 0.05.

\subsection{Video Segmentation Tracking}
\label{app:exp-details:tracking}

\paragraph{Datasets and Metrics}
For this task, we use the DAVIS~2017 \citep{pont20172017}, YouTube-VOS \citep{xu2018youtubevos} and MOSE~\citep{ding2023mose} datasets.
DAVIS defines a training set of 60 videos and a validation set of 30 videos for which all frames are annotated with ground-truth instance segmentation masks.
For YouTube-VOS, only the training set is annotated and publicly available, while the validation set is gated behind an evaluation server.
To mimic the DAVIS setup, we take a random subset of 2758 annotated videos (80\%) as the training set and the remaining 690 videos (20\%) as the validation set.
In a similar fashion, we split the MOSE dataset into 1206 videos for validation and 301 for testing.
For all datasets, we evaluate performance using the standard \(\JnF\)-mean metric \citep{perazzi2016benchmark}, which combines the region similarity (\(\mathcal{J}\)) and contour accuracy (\(\mathcal{F}\)) scores.
Only the objects annotated in the first frame are tracked and evaluated, while objects that appear later in the video are ignored, even if their ground-truth masks are annotated.

\paragraph{Evaluation Protocol}
Similar to \citet{rajasegaran2025empirical}, we implement a non-parametric protocol for label propagation based on patch similarity, which is computed as a cosine similarity between features extracted from a frozen DINOv3 backbone.
We assume that the first frame of the video is labeled with instance segmentation masks, which we represent as a one-hot vector per patch.
For each frame, we compute the cosine similarity between all its patch features, all patches of the first frame, and all patches of a small number of past frames.
Focusing on a single patch in the current frame, we consider the \(k\) most similar patches within a spatial neighborhood, and compute a weighted average of their labels to obtain a prediction for the current patch.
After processing one frame, we move to the next one, treating the previous predictions as soft instance segmentation labels.
When forwarding individual frames through the backbone, we resize the image such that the shortest side matches a certain size, preserving aspect ratio up to the nearest multiple of the patch size.\footnote{
For example, DAVIS videos are natively \(480{\times}854\) and we want to process them at resolution 960.
For a model with patch size 16, we resize the frames to \(960{\times}1712\) with a slight horizontal stretch, resulting in a \(60{\times}107\) feature map.
Instead, for a model with patch size 14, we resize the frames to \(966{\times}1708\) with a slight vertical stretch, resulting in a \(69{\times}122\) feature map.
}
Patch similarity and label propagation are computed at the resolution of the resulting features, then the mask probabilities are bilinearly resized to the native resolution for computing \(\JnF\).
We consider several hyperparameter combinations, \eg the number of past frames to use as context, the number of neighbors \(k\), and the size of the spatial neighborhood, as summarized in \cref{tab:exp-details-tracking-hyperparameters}.
We perform hyperparameter selection on the training set of DAVIS, and then apply the best combination to the test splits of all datasets.

\begin{table}[tb]
    \centering
    \caption{
        List of hyperparameters evaluated for video segmentation tracking on the training split of DAVIS 2017 \citep{pont20172017}.
        The best hyperparameters, highlighted, are applied to all datasets.
    }
    \label{tab:exp-details-tracking-hyperparameters}
    \begin{tabular}{@{}ccccc@{}}
        \toprule
        Max context length & Neighborhood mask size & Neighborhood mask shape & Top-K & Temperature \\
        \midrule
        \phantom{0}7 & 12 & Square & \phantom{0}5 & 0.2\phantom{0} \\
        \phantom{0}7 & 12 & Circle & \phantom{0}5 & 0.2\phantom{0} \\
        \phantom{0}7 & \phantom{0}5 & Square & \phantom{0}5 & 0.2\phantom{0} \\
        \phantom{0}7 & 24 & Square & \phantom{0}5 & 0.2\phantom{0} \\
        \rowcolor{lightgray} \phantom{0}7 & \(\infty\) & --- & \phantom{0}5 & 0.2\phantom{0} \\
        \phantom{0}7 & 12 & Square & \phantom{0}5 & 0.01  \\
        \phantom{0}7 & 12 & Square & \phantom{0}5 & 0.1\phantom{0}  \\
        \phantom{0}7 & 12 & Square & \phantom{0}5 & 0.7\phantom{0}  \\
        \phantom{0}4 & 12 & Square & \phantom{0}5 & 0.2\phantom{0}  \\
        10 & 12 & Square & \phantom{0}5 & 0.2\phantom{0}  \\
        15 & 12 & Square & \phantom{0}5 & 0.2\phantom{0}  \\
        \phantom{0}7 & 12 & Square & \phantom{0}3 & 0.2\phantom{0}  \\
        \phantom{0}7 & 12 & Square & 10 & 0.2\phantom{0}  \\
        \phantom{0}7 & 12 & Square & 15 & 0.2\phantom{0}  \\
        15 & 12 & Circle & 10 & 0.1\phantom{0} \\
        15 & 24 & Circle & 10 & 0.1\phantom{0} \\
        15 & 36 & Circle & 10 & 0.1\phantom{0} \\
        15 & \(\infty\) & --- & 10 & 0.1\phantom{0} \\
        \bottomrule
    \end{tabular}
\end{table}

\subsection{Video Classification}
\label{app:exp-details:video-classification}

\paragraph{Datasets and Metrics}
We evaluate DINOv3 on video classification using the UCF101 \citep{soomro2012ucf101}, Something-Something V2 \citep{goyal2017something}, and Kinetics-400 \citep{kay2017kinetics} datasets.
At a high level, we extract a fixed number of frames from each video, encode them with a frozen backbone, collect all patch features into a flat sequence, which we then feed to a shallow transformer-based classifier trained with regular supervision on a set of labeled videos.
In previous work, \eg \citep{assran2025vjepa2}, this protocol is referred to as an \emph{attentive probe}, a hint to the \emph{linear probes} used for image classification.
In the following paragraphs, we describe our implementation of the protocol.

\paragraph{Training}
At training time, we select 16 frames at random temporal locations from each video, keeping track of the corresponding timestamps.
We also sample the parameters of a spatial crop that covers between 40\% and 100\% of the area--these parameters will be shared across all frames of the video to avoid jittering.
We then process each frame with DINOv3 as an independent \(256{\times}256\) image, extracting \(16{\times}16\) patch features and discarding the CLS token.
For each patch, we keep track of its spatial coordinates defined on a \([0,1]^2\) box.
The patch features from all frames are linearly projected to \(1024\) dimensions, concatenated into a flat sequence of length \(16{\times}16{\times}16=4096\), and then fed to four self-attention blocks that model the spatial and temporal relationships between the patches.
To ensure the model has access to positional information, we inject the timestamp and spatial coordinates of each patch both as an additive sin-cos embedding before the blocks \citep{vaswani2017attention}, and as a 3D factorized RoPE with random spatial rotations in each attention head~\citep{heo2024rotary}.
After the four blocks, we apply a cross-attention block with a single position-less learnable query to aggregate the information from all patches into a single vector, which is then linearly projected to obtain the final classification logits.
The stack of self-attention blocks, the cross-attention block, the positional embeddings, and the final projection constitute the video classifier, which we train for 20 epochs with batch size 64 with a standard cross-entropy loss.
In practice, we train a set of classifiers in parallel, one for each combination of learning rate
\(\{1\cdot 10^{-4}, 2\cdot 10^{-4}, 5\cdot 10^{-4}, 1\cdot 10^{-3}, 2\cdot 10^{-3}, 5\cdot 10^{-3}\}\)
and weight decay
\(\{10^{-3}, 10^{-2}, 10^{-1}\}\).
For each dataset, we use 90\% of the training set to update the model parameters, 10\% of the training set to choose the best combination of learning rate and weight decay, and finally report performance of the chosen model on the validation split.

\begin{figure}[t]
    \centering
    \includegraphics[width=\linewidth]{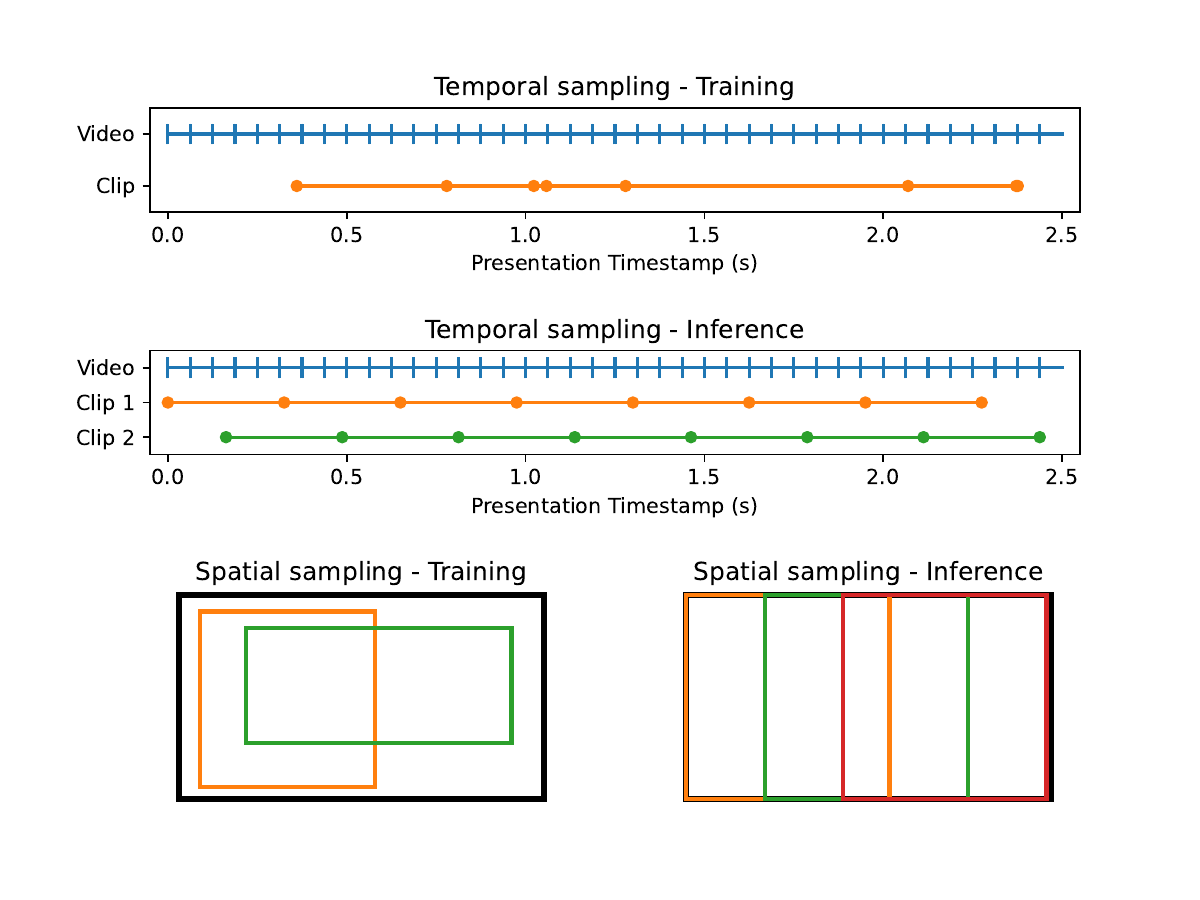}
    \caption{
        Sampling clips for video classification.
        Choosing a clip for training or inference means determining the coordinates of a spatial crop and which frames/timestamps to sample.
        At training time, we sample clips at random by choosing random frames from the whole video and by applying a spatial crop that covers \(\ge40\%\) of the area.
        At inference time, we select clips in a deterministic way.
        Spatially, we take the three largest square crops aligned to the left, middle and right.
        Temporally, we take two overlapping sets of frames such that they cover as much of the video as possible and their timestamps interleave.
    }
    \label{fig:video_attn_pool_clip_sampling}
\end{figure}

\paragraph{Inference}
At inference time, we follow a deterministic strategy to sample a single clip per video: we take the first frame, the last frame, and uniformly-spaced frames in between, for a total of 16.
From each frame, we crop the largest center square and resize it to \(256{\times}256\) pixels, possibly losing information from the sides of rectangular videos.
We then feed these frames to DINOv3 and to the classifier to obtain a prediction for the video.
Alternatively, we follow \citet{assran2025vjepa2} and perform test-time augmentation (TTA) by selecting multiple frame sequences and multiple spatial crops, processing them independently and then averaging the class probabilities to obtain the final prediction.
Clip sampling is exemplified in \cref{fig:video_attn_pool_clip_sampling}.

\paragraph{Baselines}
For the chosen baseline models, we use the same evaluation protocol, \ie feature extraction, classifier architecture, training procedure and inference protocol, with a few differences.
The input resolution is \(256{\times}256\) pixels for models that use patch size 16, and \(224{\times}224\) pixels for patch size 14.
This way, all backbones yield an identical number of tokens, and therefore afford the same amount of computation in the classifier.
All models process videos frame by frame independently, since they are trained on images.
The only exception is V-JEPA 2, to which we feed whole clips to extract time-aware features.
Since V-JEPA 2 reduces the temporal axis by a factor of two, \eg yielding 8 time steps given 16 frames as input, we duplicate each patch token to match the sequence length of other models.

\subsection{Image Classification with Linear Probing}
\label{app:exp-details:classification}

\paragraph{Datasets and Metrics}
We evaluate the global quality of the DINOv3 model using the widely adopted linear probing evaluation. 
We train a linear transform on the training set of ImageNet-1k~\citep{deng2009imagenet} and evaluate results on the val set.
We assess the generalization quality of the model by evaluating the transfer to classification test sets: ImageNet-\textbf{V2}~\citep{recht2019imagenet} and \textbf{ReaL}~\citep{beyer2020imagenetreal}, which provide alternative sets of images and labels for ImageNet, designed to test for overfitting on the original ImageNet validation set. Additionally, we consider the \textbf{R}endition~\citep{hendrycks2021many} and \textbf{S}ketch~\citep{wang2019learning} datasets, which present stylized and artificial versions of ImageNet classes; the \textbf{A}dversarial~\citep{hendrycks2021natural} and \textbf{Obj}ectNet~\citep{barbu2019objectnet} datasets, which contain deliberately challenging examples; and the \textbf{C}orruptions~\citep{hendrycks2019benchmarking} dataset, which measures robustness to common image corruptions. 
We report top-1 classification accuracy as the evaluation metric for all datasets but ImageNet-C, for which we report the mean corruption error (mCE, see \citep{hendrycks2019benchmarking}).

For fine-grained datasets, we consider the same collection of 12 datasets from \citet{oquab2024dinov2}, which we call Fine-S here: Food-101~\citep{dataset-food101}, CIFAR-10~\citep{krizhevsky2009learning}, CIFAR-100~\citep{krizhevsky2009learning}, SUN397~\citep{dataset-sun397}, StanfordCars~\citep{dataset-stanfordcars}, FGVC-Aircraft~\citep{dataset-aircraft}, VOC 2007~\citep{pascal-voc-2007}, DTD~\citep{dataset-dtd}, Oxford Pets~\citep{dataset-pets}, Caltech101~\citep{dataset-caltech101}, Flowers~\citep{dataset-flowers}, and CUB200~\citep{dataset-cub}, as well as the larger datasets Places205~\citep{zhou2014learning}, iNaturalist 2018~\citep{van2018inaturalist}, and iNaturalist 2021~\citep{van2021benchmarking}.

\paragraph{Evaluation Protocol}
For the larger datasets ImageNet, Places205, iNaturalist 2018 and iNaturalist 2021, we use the following procedure.
For each baseline, we train a linear layer on the final features of the CLS token (after the layer norm) using the ImageNet-1k training set~\citep{deng2009imagenet}. 
Specifically, we use SGD with a momentum of 0.9, and train for 10 epochs with a batch size of 1024.
We sweep the learning rates $\{1 \times 10^{-4}, 2 \times 10^{-4}, 5 \times 10^{-4}, 1 \times 10^{-3}, 2 \times 10^{-3}, 5 \times 10^{-3}, 1 \times 10^{-2}, 2 \times 10^{-2}, 5 \times 10^{-2}, 1 \times 10^{-1}, 2 \times 10^{-1}, 5 \times 10^{-1}, 1 \times 10^{0}, 2 \times 10^{0}, 5 \times 10^{0}\}$ and weight decay values $\{0, 1e-5\}$ and use the validation set of ImageNet-1k to select the best combination.
During training, we use random resize crop augmentation with standard Inception-crop parameters.
For the datasets in Fine-S, following \citet{oquab2024dinov2}, we use a lighter weight evaluation using scitkit-learn's  LogisticRegression implementation with the L-BFGS solver.

In both cases, we evaluate models at resolutions resulting in 1024 patch tokens, that is, $448 \times 448$ for patch size 14, and $512 \times 512$ for patch size 16.
The images are resized such that the shorter side matches the chosen side length, then take the central square crop.

\subsection{Instance Recognition}
\label{app:exp-details:recognition}

\paragraph{Datasets and Metrics}
We use the Oxford and Paris datasets for landmark recognition~\citep{radenovic2018revisiting}, the Met dataset featuring artworks from the Metropolitan Museum~\citep{ypsilantis2021met}, and AmsterTime, which consists of modern street view images matched to historical archival images of Amsterdam~\citep{yildiz2022amstertime}.
In \cref{tab:results-instance-recognition}, we report mean average precision (mAP) for Oxford-Hard, Paris-Hard, and AmsterTime, and global average precision (GAP) for Met.
In \cref{app:tab:results-instance-retrieval}, we additional give mAP for Oxford-Medium and Paris-Medium, as well as the additional metrics GAP- and accuracy (see \citep{ypsilantis2021met}).
For Oxford and Paris, we resize all images such that the larger side is 224 pixels long, keeping the aspect ratio, then take a full center crop, yielding an image of resolution of $224 \times 224$.
For AmsterTime, we resize all images such that the shorter side is 256 pixels long, keeping the aspect ratio, then take a center crop of size $224 \times 224$.
For Met, we evaluate all images close to their original resolution, resizing both to the nearest multiple of patch size (resulting in a long side 508/512 for patch size 14/16).

\paragraph{Evaluation Protocol}
The image similarity is computed using cosine distance between the CLS tokens computed for query and target images.
We follow the evaluation protocols of \citep{radenovic2018revisiting} for Oxford and Paris, of \citep{yildiz2022amstertime} for AmsterTime, and of \citet{ypsilantis2021met} for Met.
For Met, this includes tuning the hyperparameters k and $\tau$ with a grid search, optimizing GAP on the validation set of Met, and whitening the features using a PCA estimated on the training set of Met.

\subsection{Object Detection}
\label{app:exp-details:detection}

\paragraph{Datasets and Metrics}
We evaluate DINOv3 on object detection on the COCO \citep{Lin2014cocodataset} and COCO-O \citep{mao2023cocoo} datasets.
COCO is a standard benchmark for object detection, covering 80 object categories, and containing 118k training images and 5k validation images.
COCO-O is an evaluation-only dataset with the same categories as COCO, but in more challenging visual conditions, such as scenes with significant occlusions, cluttered backgrounds, and varying lighting conditions.
For training the object detection model, we also leverage the Objects365 \citep{shao2019objects365} dataset, which contains 2.5M images and covers 365 object categories, a subset of which maps directly to COCO classes.
For both COCO and COCO-O, we report mean Average Precision (mAP) computed at IoU thresholds of \([0.5:0.05:0.95]\).

\paragraph{Architecture}
Our approach builds upon the Plain-DETR~\citep{lin2023plaindetr} implementation, with several modifications.
We do not fuse the transformer encoder into the \emph{backbone}, but keep it as a separate module, similar to the original DETR \citep{carion2020detr}.
This allows us to keep the DINOv3 backbone completely frozen during training and inference, making it the first competitive detection model to do so.
From a DINOv3 ViT-7B/16 backbone, we extract features from four intermediate layers, namely \([10, 20, 30, 40]\).
For each patch, we concatenate intermediate features channel-wise, giving a feature dimension of \(4 \cdot 4096 = 16384\), which is further increased by the windowing strategy described below.
The backbone features feed into the \emph{encoder}, which is a stack of 6 self attention blocks with embedding dimension of 768.
The \emph{decoder} is a stack of 6 cross attention blocks with the same embedding dimension, where 1500 ``one-to-one'' queries and 1500 ``one-to-many'' queries attend to the patch tokens of the encoders to predict bounding boxes and class labels.

\paragraph{Image Pre-Processing}
Training is performed in three stages as described below, one with a base image resolution of 1536 pixels and two with a base resolution of 2048.
Following DETR, we apply random horizontal flipping ($p=0.5$), followed by either
(i) random resizing, where the shortest side is uniformly sampled between 920 pixels (resp. 1228) and the base resolution of the stage (1536 or 2048), or
(ii) a random crop retaining 60--100\% of the original image area, followed by resizing as in (i).
At evaluation time, images are resized so that the shortest side is 2048 without additional augmentation, and both sides are rounded up to the nearest multiple of the patch size.

\paragraph{Windowing strategy}
We then apply a windowing strategy that combines a global view of the image with smaller views, to allow the backbone to process objects at all scales. The number of windows is fixed to $3 \times 3$, and their sizes vary according to the input resolution.
As an example, for an image of size \(1536 \times 2304\):
\begin{enumerate}
    \item
    The image is divided into \(3 \times 3\) non-overlapping windows of size \(512 \times 768\).
    Each window is forwarded through the backbone, resulting in \(32 \times 48\) patch tokens of dimension 16384.
    The features of all windows are spatially reassembled into a \((3 \cdot 32) \times (3 \cdot 48)\) feature map.
    \item
    The whole image is resized to \(512{\times}768\) and forwarded through the backbone, resulting in a feature map of \(32{\times}48\) patch tokens of dimension 16384.
    These features are then bilinearly upsampled to \(96{\times}144\), matching the size of the windowed feature map.
    \item
    Finally, the features maps from steps 1 and 2 are concatenated channel-wise, resulting in a \(96{\times}144\) feature map of dimension \(2 \cdot 16384 = 32768\).
    This feature map is then flattened as a sequence of \(96*144\) tokens and fed to the encoder.
\end{enumerate}

\paragraph{Training}

We follow a training curriculum in three stages, using the Objects365 dataset~\citep{shao2019objects365} and the COCO dataset~\citep{Lin2014cocodataset} at increasing resolutions.
Throughout training, we use the AdamW optimizer~\citep{loshchilov2017decoupled} with a weight decay of 0.05.
Following DETR, we use the Focal Loss~\citep{lin2018focallossdenseobject} as classification loss, with a weight of 2, L1 loss as bounding box loss with a weight of 1, complemented by the GIoU~\citep{rezatofighi2019generalizedintersectionunionmetric} loss with a weight of 2.
The stages are as follows:
\begin{enumerate}
    \item 
    We begin training on Objects365 at base resolution 1536 pixels.
    We train for 22 epochs with global batch size of 32, which we distribute over 32 GPUs.
    After an initial warmup of 1000 steps, the learning rate is set to \(5\cdot10^{-5}\) and is divided by 10 after the 20th epoch.
    \item
    We then continue training on Objects365 at base resolution 2048 pixels.
    We train for for 4 epochs with learning rate \(2.5\cdot10^{-5}\).
    \item 
    We finish training by doing 12 epochs on COCO at base resolution 2048.
    After a linear warmup of 2000 iterations, the learning rate follows a cosine decay schedule, starting at \(2.5\cdot10^{-5}\) and reaching \(2.5\cdot10^{-6}\) at the 8th epoch.
    In this part we use the IA-BCE classification loss~\citep{cai2024aligndetrenhancingendtoendobject} instead of the simple Focal Loss from DETR.
    We observed this loss to increase the model performance at transfer time, but not at pretraining time.
    As this loss mixes class and box information, it brings its full potential if the box predictions are already well initialized.
    The GIoU loss weight is set to 4 in this part to encourage better box alignment.
\end{enumerate}

\paragraph{Test-Time Augmentation}
At test time, we follow the inference procedure described above, resizing images such that the short side is 1536 or 2048.
At those resolutions, the COCO mAP is 65.4 and 65.6, respectively.
Alternatively, we can apply the test-time augmentation (TTA) strategy from \citet{bolya2025perception}, which consists in flipping and resizing the image to multiple resolutions, and merging the predictions with SoftNMS~\citep{Bodla_2017_ICCV}.
Specifically, each image is processed at resolutions of \([1536, 1728, 1920, 2112, 2304, 2496, 2688, 2880]\), yielding an mAP of 66.1.

\subsection{Semantic Segmentation}
\label{app:exp-details:segmentation-system}

\paragraph{Datasets and Metrics}
We evaluate DINOv3 on semantic segmentation on the ADE20k~\citep{zhou2017scene}, Cityscapes~\citep{cordts2016cityscapes}, COCO-Stuff~\citep{caesar2018coco}, and VOC 2012 \citep{pascal-voc-2012} datasets. 
ADE20k is a widely used benchmark for semantic segmentation with 150 semantic categories, varying from outdoor scenery to images of people and objects inside a house. 
In addition, COCO-Stuff and Hypersim \citep{roberts2021hypersim} datasets are used for pre-training the model.
COCO-Stuff is a larger dataset (118k training images) than ADE20k containing 80 thing classes and 91 stuff classes, while Hypersim is a photorealistic synthetic dataset presenting indoor scenes with 40 semantic categories, with sharper and more accurate annotations. 
More than half of the Hypersim images contain 21 or more objects, making it a good candidate for helping the model learn rich information of the scenes. 
The evaluation metric reported is mIoU for all datasets.

\paragraph{Architecture}
We adapt the ViT-Adapter and Mask2former configurations that other baselines use~\citep{wang2023onepeace}, with several differences.
First, to ensure that our backbone remains frozen and its activations are not altered, we remove the injector component of the ViT-Adapter. 
This makes our backbone output features to be directly used in the extractor module. 
Second, the embedding dimensions in the Mask2former decoder are scaled to 2048 instead of the default 1024 to adapt to our backbone output dimension of 4096, while other baselines' backbones usually present an output dimension of 1024 or 1536.
As inputs to the decoder, we extract features from four intermediate layers of the DINOv3 7B/16 backbone, namely layers \([10, 20, 30, 40]\).
We apply the final layer norm to the features of all layers and add a learned batch normalization.

\paragraph{Training Protocol}
For generating results on COCO-Stuff, we train the model using a cosine scheduler, with a 6k linear warmup and a maximum learning rate of 1.5e-5. 
The model is trained for 80k iterations, at resolution 1280 pixels.
As for training on the other datasets---ADE20k, Cityscapes and VOC 2012---we first pre-train the decoder on COCO-Stuff for 80k iterations, with a 6k linear warmup and a learning rate of 1.5e-5, following a cosine scheduler. 
This helps the model learn diverse semantic categories (171 categories) on a larger dataset than ADE20k. 
The model is then trained on Hypersim for 10k iterations at a learning rate of 2.5e-5 following a cosine scheduler with a 1.5k linear warmup. 
Corresponding to roughly 2 epochs, this step helps our model learn high-quality image-to-mask correspondence due to their photorealistic synthetic nature.
Finally, our model is trained on ADE20k for 20k iterations with a learning rate of 3e-5, again with a 1.5k linear warmup and a cosine schedule. 
We report our final result on the validation set. 
For Cityscapes and VOC 2012, learning rates of 1.5e-5 and 1e-5 are used respectively. 
For all training, a batch size of 16 and the AdamW optimizer is used.

\paragraph{Inference}
For single-scale evaluation, sliding inference is used for evaluating the models---the image is first resized at the training resolution (\eg a $400 \times 500$ image will be resized to $896 \times 1120$ pixels for ADE20k, since the model was trained at resolution 896). Then, a sliding window method is used with a stride (\eg stride of 596 pixels for ADE20k) on square crops (\eg $896 \times 896$ pixels) to generate a prediction for each crop, sliding through the image. 
These results are then aggregated and rescaled to the original image size to generate a final prediction.
For test-time augmentation, both ADE20k and VOC 2012 images were rescaled to ratios of [0.9, 0.95, 1.0, 1.05, 1.1] of the evaluation resolution, and each image was also flipped horizontally to generate a total of 10 predictions per sample. 
After sliding inference on each image, they were rescaled to the original image shape and averaged. 
COCO-Stuff 164K's TTA mIoU was obtained by simply using an additional horizontally flipped image per sample, and for Cityscapes, ratios of [1.0, 1.5, 2.0] of the evaluation resolution were applied.

\subsection{Monocular Depth Estimation}
\label{app:exp-details:depth-system}

\paragraph{Implementation Details}
Our approach differs from Depth Anything v2 (DAv2)~\citep{yang2024depthanythingv2} primarily in the configuration of image resolution, which is set to $768 \times 1024 pixels$, and the network architecture. 
The backbone is kept frozen throughout training, while a dropout rate of 0.05 is applied at the end of the DPT head~\citep{ranftl2021vision}. 
As input to the decoder, we extract features from four intermediate layers of the DINOv3 7B/16 backbone, namely layers \([10, 20, 30, 40]\).
We apply the final layer norm to the features of all layers and add a learned batch normalization.
The depth estimation output is discretized into 256 uniformly distributed bins covering the range from 0.001m to 100m. 
Training employs a base learning rate of 1e-3, scheduled using PolyLR with a power of 3.5 and an initial linear warmup phase lasting 12k iterations. 
To enhance robustness and generalization, we apply a suite of augmentations: Gaussian blur, Gaussian noise, AutoContrast, AutoEqualise, ColorJitter, rotation, and left-right flip.

\paragraph{Datasets and Metrics}
We train the model on the dataset of DAv2, which consists of synthetically generated images from the IRS, TartanAir, BlendedMVS, Hypersim, and VKITTI2 datasets.
We evaluate on five datasets: NYUv2~\citep{silberman2012indoor}, KITTI~\citep{geiger2013vision}, ETH3D~\citep{schoeps2017multiview}, ScanNet from \citep{ke2025marigold}, and DIODE~\citep{vasiljevic2019diode}.
We adopt the zero-shot scale-invariant depth setup, and report the standard metrics absolute relative error and $\delta_1$ (see~\citep{yang2024depthanythingv1}).

\subsection{Visual Geometry Grounded Transformer with DINOv3}
\label{app:exp-details:vggt}

\paragraph{Implementation Details}
Compared to the original VGGT~\citep{wang2025vggt}, we adopt the following changes:
(1) we use an image size of 592 instead of 518; this is to match the number of patch tokens that DINOv2 produces,
(2) adopting a smaller learning rate, specifically from 0.0002 to 0.0001, and 
(3) using a concatenation of the four intermediate layers of DINOv3 ViT-L rather than just the last layer as input to the downstream modules.
Interestingly, we found that using four intermediate layers brings a benefit for DINOv3, whereas doing the same for DINOv2 brings no additional performance gains.
We also experimented with a version closer to the original VGGT setup (image size 512, same learning rate, final layer), and already found this untuned version to improve over the original VGGT work across all tested benchmarks.

\subsection{Geospatial}
\label{app:exp-details:geospatial}
\paragraph{Evaluation details} %

In all of the evaluations, we keep the backbone frozen and only train lightweight classifiers or decoders that are specialized for the tasks.
For GEO-Bench classification, we train a linear classifier for 2400 iterations with a batch size of 32. We use SGD optimizer, cosine learning rate annealing, and select the best learning rate between $1\mathrm{e}{-5}$ and $1$. Unless otherwise specified, segmentation evaluations use a DPT decoder~\citep{ranftl2021vision}, with a learning rate selected based on performance on the validation set with a grid search of four values in $[3\mathrm{e}{-5}, 1\mathrm{e}{-4}, 3\mathrm{e}{-4}, 1\mathrm{e}{-3}]$.   

On LoveDA and iSAID datasets, we train an UPerNet decoder \citep{xiao2018unifiedperceptualparsingscene} for 80k iterations, with a batch size of 8, and a linear warm-up of 1500 iterations in line with \citep{wang2024mtpadvancingremotesensing}.
All other hyper-parameters such as crop size and weight decay are the same as in \citep{wang2024mtpadvancingremotesensing}.
Following previous work~\citep{tolan2024very,wang2022lovedaremotesensinglandcover}, we use a DPT head for canopy height prediction evaluations and train a Faster RCNN~\citep{ren2015faster} detector for 12 epochs for object detection tasks.

\paragraph{Satlidar dataset}

The Satlidar dataset consists in one Million of $512\times 512$ Maxar images and corresponding dense lidar measurements collected from different locations as described in Table \cref{tab:satlidar}.
The images were extracted from larger tiles, the numbers of tiles for each sub-dataset are specified in the table.       
\begin{table}[tb]
    \centering
    \caption{Description of the Satlidar dataset.}
    \label{tab:satlidar}
    \begin{tabular}{cccc}
    \toprule
    Subdataset & Path & Amount of tiles & Purpose \\  
    \midrule
Kalimantan &\tiny{\url{https://daac.ornl.gov/cgi-bin/dsviewer.pl?ds_id=1540}}&86&	train/val/test\\
OpenDC	& \tiny{\url{https://opendata.dc.gov/datasets/2020-lidar-classified-las/about}}&68&	train/val/test\\
Brazil	&	\tiny{\url{https://daac.ornl.gov/cgi-bin/dsviewer.pl?ds_id=1644}}&37&	train/val/test\\
Mozambique	&	\tiny{\url{https://daac.ornl.gov/cgi-bin/dsviewer.pl?ds_id=1521}}&144&	train/val/test\\
Neon		&	\tiny{\url{https://data.neonscience.org/data-products/DP3.30015.001}} &5366&train/val/test\\
CA20Graup	&	\tiny{\url{https://portal.opentopography.org/datasetMetadata?otCollectionID=OT.092021.6339.1}}&99&	train/val/test\\
CA17Duvall &	\tiny{\url{https://portal.opentopography.org/datasetMetadata?otCollectionID=OT.042020.6339.2}}&56&	train/val/test\\
Netherlands	&	\tiny{\url{https://geotiles.citg.tudelft.nl/}}&13& train/val/test\\
Sao Paulo	& \tiny{\url{https://daac.ornl.gov/CMS/guides/LiDAR\_Forest\_Inventory\_Brazil.html}} &4&	test \\
CA brande	& \tiny{\url{https://doi.org/10.5069/G9C53J18}} &1& test\\
\bottomrule
    \end{tabular}
\end{table}%

\end{document}